\documentclass{article}

\usepackage{microtype}
\usepackage{graphicx}
\usepackage{booktabs}
\usepackage{multirow}
\usepackage{tabularx}
\usepackage{placeins}
\usepackage{xcolor}
\usepackage{amsmath,amssymb}
\usepackage{tikz}
\usepackage{pgfplots}
\usepackage{pgfplotstable}
\pgfplotsset{compat=1.18}
\usepackage{figures/style/castor}
\usepackage{hyperref}
\usepackage[preprint]{icml2026}
\usepackage[no-math]{fontspec}
\newcommand{\todo}[1]{\textcolor{red}{\textbf{[TODO #1]}}}
\newcommand{\rMZ}{\ensuremath{r_{\mathrm{MZ}}}}
\newcommand{\rU}{\ensuremath{r_{\mathrm{S}}}}
\newcommand{\rR}{\ensuremath{r_{\mathrm{X}}}}

\newcommand{\HtwoH}{H200-h}
\newcommand{\repoURL}{\url{https://github.com/cerulean-works/castor}}
\newcommand{\hfURL}{\url{https://huggingface.co/collections/cerulean-works/castor}}

\newcommand{\sixNpanel}{6{,}226}
\newcommand{\sixNadv}{3{,}212}
\newcommand{\sixNplain}{1{,}240}
\newcommand{\sixNharmful}{4{,}452}
\newcommand{\sixNbenign}{1{,}774}
\newcommand{\sixK}{20}
\newcommand{\sixMaxTokens}{512}
\newcommand{\sixTemperature}{1.0}
\newcommand{\sixNboot}{1{,}000}

\newcommand{\sixNmz}{18}
\newcommand{\sixNu}{48}

\newcommand{\sixRmzRaw}{0.930}

\newcommand{\sixRuRaw}{0.767}
\newcommand{\sixGap}{0.173}

\newcommand{\sixGapCI}{[0.162, 0.185]}
\newcommand{\sixGapCIbases}{[0.155, 0.195]}
\newcommand{\sixGapJackSE}{0.023}

\newcommand{\sixRmzCIbases}{[0.983, 0.986]}
\newcommand{\sixRuCIbases}{[0.790, 0.830]}
\newcommand{\sixRuJackSE}{0.023}

\newcommand{\sixNbaseDraws}{883}
\newcommand{\sixVulnMin}{0.463}
\newcommand{\sixVulnMax}{0.512}
\newcommand{\sixRelMin}{0.935}
\newcommand{\sixRelMax}{0.953}
\newcommand{\sixPlainVulnMin}{0.221}
\newcommand{\sixPlainVulnMax}{0.295}
\newcommand{\sixBenignVulnMin}{0.566}
\newcommand{\sixBenignVulnMax}{0.741}
\newcommand{\sixPlainRefusalMin}{70\,\%}
\newcommand{\sixPlainRefusalMax}{78\,\%}
\newcommand{\sixPlainGap}{0.099}
\newcommand{\sixPlainGapCI}{[0.086, 0.114]}
\newcommand{\sixHarmfulGap}{0.141}
\newcommand{\sixHarmfulGapCI}{[0.133, 0.149]}
\newcommand{\sixBenignGap}{0.136}
\newcommand{\sixBenignGapCI}{[0.125, 0.149]}

\newcommand{\sixBpDef}{0.984}
\newcommand{\sixBpAuxTwo}{0.986}
\newcommand{\sixBpAuxThree}{0.983}
\newcommand{\sixBpDefAuxTwo}{0.830}
\newcommand{\sixBpDefAuxThree}{0.814}
\newcommand{\sixBpAuxTwoAuxThree}{0.790}

\newcommand{\cciNtemplates}{91}
\newcommand{\cciNbehaviours}{300}
\newcommand{\cciHjbN}{1{,}212}

\newcommand{\fourteenKappaRefusal}{0.757}
\newcommand{\fourteenKappaFour}{0.727}
\newcommand{\fifteenN}{1{,}200}

\newcommand{\fifteenAccW}{0.880}
\newcommand{\fifteenAccWCI}{[0.853, 0.905]}
\newcommand{\fifteenAccU}{0.758}
\newcommand{\fifteenWorstAccW}{0.782}

\newcommand{\fifteenKappaRefusal}{0.766}
\newcommand{\fifteenKappaFour}{0.673}
\newcommand{\fifteenJudgeHumanKappa}{0.517}

\newcommand{\fifteenQthreeGAccW}{0.896}
\newcommand{\fifteenFlipsEightyFive}{7}

\newcommand{\fifteenMinFlipsPerModel}{1}

\newcommand{\fifteenTwinsMinusRefs}{+0.029}
\newcommand{\fifteenTwinsMinusRefsCI}{[\ensuremath{-}0.036, 0.109]}
\newcommand{\fifteenPrecWj}{0.580}
\newcommand{\fifteenPrecHjb}{0.602}
\newcommand{\fifteenGapCorr}{0.175}
\newcommand{\fifteenGapCorrCI}{[0.163, 0.187]}
\newcommand{\fifteenGapNone}{0.173}
\newcommand{\fifteenLevelCorrMin}{0.273}
\newcommand{\fifteenLevelCorrMax}{0.301}
\newcommand{\fifteenLevelRawMin}{0.46}
\newcommand{\fifteenLevelRawMax}{0.51}
\newcommand{\fifteenRecallAdvMin}{0.907}
\newcommand{\fifteenBySourceWj}{0.840}
\newcommand{\fifteenBySourceHjb}{0.844}
\newcommand{\fifteenBySourceOthersMin}{0.905}
\newcommand{\fifteenBySourceOthersMax}{0.936}
\newcommand{\fifteenDisagreeAccMin}{0.395}
\newcommand{\fifteenDisagreeAccMax}{0.399}
\newcommand{\fifteenAgreeAccMin}{0.914}
\newcommand{\fifteenAgreeAccMax}{0.945}
\newcommand{\fifteenQthreeGDisagreeAccMin}{0.601}
\newcommand{\fifteenQthreeGDisagreeAccMax}{0.605}
\newcommand{\sevenSiteLayer}{11}
\newcommand{\sevenSitePos}{\ensuremath{-}2}

\newcommand{\sevenGap}{0.401}
\newcommand{\sevenGapCI}{[0.384, 0.418]}
\newcommand{\sevenGapCIbases}{[0.303, 0.548]}

\newcommand{\sevenProjRelMin}{0.993}
\newcommand{\sevenProjRelMax}{0.999}
\newcommand{\sevenBpDef}{0.932}
\newcommand{\sevenBpAuxTwo}{0.980}
\newcommand{\sevenBpAuxThree}{0.976}
\newcommand{\sevenBpDefAuxTwo}{0.416}
\newcommand{\sevenBpDefAuxThree}{0.594}
\newcommand{\sevenBpAuxTwoAuxThree}{0.674}
\newcommand{\sevenLink}{0.362}

\newcommand{\sevenLinkCI}{[0.339, 0.382]}
\newcommand{\sevenLinkCIbases}{[0.267, 0.494]}

\newcommand{\sevenHarmGap}{0.591}
\newcommand{\sevenHarmGapCI}{[0.584, 0.598]}
\newcommand{\sevenHarmRmz}{1.000}
\newcommand{\sevenHarmRu}{0.409}

\newcommand{\sevenCosMZ}{0.947}

\newcommand{\sevenCosU}{\ensuremath{-}0.021}
\newcommand{\sevenCosUabsMax}{0.046}
\newcommand{\sevenOwnSiteGap}{0.101}
\newcommand{\sevenOwnSiteGapCI}{[0.094, 0.108]}
\newcommand{\sevenOwnSiteLink}{0.102}
\newcommand{\sevenOwnSiteLinkCI}{[0.087, 0.118]}
\newcommand{\sevenCosReadoutGap}{0.359}
\newcommand{\sevenCosReadoutLink}{0.373}
\newcommand{\sevenQthreeGLink}{0.315}
\newcommand{\sevenNpassing}{5}
\newcommand{\sevenNanPlainMax}{0.003}
\newcommand{\sevenLinkBpDefAuxTwo}{0.324}
\newcommand{\sevenLinkBpDefAuxThree}{0.267}
\newcommand{\sevenLinkBpAuxTwoAuxThree}{0.494}
\newcommand{\sevenPlainGap}{0.152}
\newcommand{\sevenPlainGapCI}{[0.140, 0.164]}
\newcommand{\sevenHarmfulGap}{0.212}
\newcommand{\sevenHarmfulGapCI}{[0.203, 0.222]}
\newcommand{\sevenBenignGap}{0.179}
\newcommand{\sevenBenignGapCI}{[0.167, 0.191]}
\newcommand{\sevenValidityMin}{\ensuremath{-}0.78}
\newcommand{\sevenValidityMax}{\ensuremath{-}0.55}

\newcommand{\sevenSiteMapLastMin}{0.58}
\newcommand{\sevenSiteMapLastMax}{0.62}

\newcommand{\sevenSiteMapGapMin}{0.02}
\newcommand{\sevenSiteMapGapMax}{0.86}
\newcommand{\sevenCausalNdouble}{10}
\newcommand{\sevenCausalN}{12}
\newcommand{\sevenCausalDefOneX}{1.92$\times$}
\newcommand{\sevenCausalDefFourX}{1.79$\times$}

\newcommand{\twentytwoDefOneX}{4.08$\times$}
\newcommand{\twentytwoDefFourX}{3.99$\times$}

\newcommand{\twentytwoRandomMax}{0.025}

\newcommand{\twentytwoNanPlainMax}{0.005}
\newcommand{\twentytwoNanAdvMax}{0.014}
\newcommand{\twentytwoNanBenignMax}{0.003}
\newcommand{\eighteenRotNpass}{11}
\newcommand{\eighteenRotMean}{0.903}
\newcommand{\eighteenRotMedian}{0.986}
\newcommand{\eighteenMismatchMean}{0.043}
\newcommand{\eighteenUnrotMin}{\ensuremath{-}0.018}
\newcommand{\eighteenUnrotMax}{0.025}

\newcommand{\eighteenSameNpass}{12}
\newcommand{\eighteenSameMin}{0.810}
\newcommand{\eighteenRotFailT}{0.427}
\newcommand{\eighteenStdNpass}{11}
\newcommand{\eighteenInvMin}{\ensuremath{-}0.013}
\newcommand{\eighteenInvMax}{0.018}
\newcommand{\eighteenOwnPMin}{0.799}
\newcommand{\eighteenOwnPMax}{0.982}
\newcommand{\nineBoth}{0.985}

\newcommand{\nineBothCI}{[0.983, 0.987]}

\newcommand{\nineInit}{0.822}

\newcommand{\nineInitCI}{[0.811, 0.833]}

\newcommand{\nineOrder}{0.822}

\newcommand{\nineOrderCI}{[0.810, 0.833]}

\newcommand{\nineNeither}{0.823}

\newcommand{\nineNeitherCI}{[0.811, 0.834]}

\newcommand{\nineInitMinusOrder}{+0.000}
\newcommand{\nineInitMinusOrderCI}{[\ensuremath{-}0.008, 0.009]}
\newcommand{\nineInitMinusOrderJackSE}{0.062}
\newcommand{\nineInitMinusNeither}{\ensuremath{-}0.001}
\newcommand{\nineInitMinusNeitherCI}{[\ensuremath{-}0.008, 0.007]}
\newcommand{\nineInitMinusNeitherJackSE}{0.051}
\newcommand{\nineOrderMinusNeither}{\ensuremath{-}0.001}
\newcommand{\nineOrderMinusNeitherCI}{[\ensuremath{-}0.008, 0.006]}
\newcommand{\nineOrderMinusNeitherJackSE}{0.040}
\newcommand{\nineBothMinusNeither}{+0.162}
\newcommand{\nineBothMinusNeitherCI}{[0.151, 0.174]}

\newcommand{\nineBpDX}{0.851}
\newcommand{\nineBpAX}{0.843}
\newcommand{\nineBpDA}{0.830}
\newcommand{\nineBpXY}{0.815}
\newcommand{\nineBpDY}{0.800}
\newcommand{\nineBpAY}{0.793}

\newcommand{\nineBenignOrderMinusNeither}{\ensuremath{-}0.011}
\newcommand{\nineBenignOrderMinusNeitherCI}{[\ensuremath{-}0.019, \ensuremath{-}0.001]}
\newcommand{\nineBenignInitMinusOrder}{+0.012}
\newcommand{\nineBenignInitMinusOrderCI}{[0.004, 0.021]}

\newcommand{\nineFifthRX}{0.834}
\newcommand{\nineFifthRXCI}{[0.820, 0.846]}
\newcommand{\nineFifthRY}{0.818}
\newcommand{\nineFifthRYCI}{[0.803, 0.832]}
\newcommand{\nineFifthXYref}{0.815}
\newcommand{\nineFifthXYrefCI}{[0.801, 0.828]}
\newcommand{\nineFifthInitMinusOrder}{+0.016}
\newcommand{\nineFifthInitMinusOrderCI}{[0.002, 0.029]}
\newcommand{\nineFifthInitMinusNeither}{+0.019}
\newcommand{\nineFifthInitMinusNeitherCI}{[0.006, 0.032]}
\newcommand{\nineFifthOrderMinusNeither}{+0.003}
\newcommand{\nineFifthOrderMinusNeitherCI}{[\ensuremath{-}0.010, 0.017]}

\newcommand{\nineSigmaCE}{0.0125}
\newcommand{\nineSigmaOlmes}{0.0058}
\newcommand{\nineMbarCE}{2.6551}
\newcommand{\nineMbarOlmes}{0.5789}

\newcommand{\nineZLargest}{1.60}
\newcommand{\nineTrajDevX}{0.0101}
\newcommand{\nineTrajDevY}{0.0097}
\newcommand{\nineTrajDevR}{0.0082}
\newcommand{\nineTrajDevReleased}{0.0083}
\newcommand{\elevenRungTwo}{0.840}
\newcommand{\elevenRungTwoCI}{[0.828, 0.851]}

\newcommand{\elevenRu}{0.811}
\newcommand{\elevenRmz}{0.984}

\newcommand{\elevenBenignRungTwo}{0.791}
\newcommand{\elevenBenignRu}{0.849}

\newcommand{\elevenDivRatioEarly}{86\,\%}
\newcommand{\elevenDivRatioMid}{94\,\%}
\newcommand{\elevenDivRatioEnd}{94.4\,\%}
\newcommand{\elevenDivCosEnd}{0.110}
\newcommand{\elevenDivRelLtwoEnd}{1.34}

\newcommand{\elevenDivSeedRelLtwoEnd}{1.42}
\newcommand{\elevenDivSeedCosEnd}{0.0004}
\newcommand{\elevenDivStepEarly}{2{,}500}
\newcommand{\elevenDivStepMid}{35{,}000}
\newcommand{\elevenDivStepEnd}{69{,}369}
\newcommand{\thirtyfiveDivRatioRoneEarly}{85\,\%}
\newcommand{\thirtyfiveDivRatioRoneEnd}{93.5\,\%}
\newcommand{\thirtyfiveDivRelLtwoRoneEnd}{1.32}

\newcommand{\thirtyfiveDivRatioDEarly}{85\,\%}
\newcommand{\thirtyfiveDivRatioDEnd}{94.4\,\%}
\newcommand{\thirtyfiveDivRelLtwoDEnd}{1.34}

\newcommand{\thirtyfiveDivSeedRelLtwoEnd}{1.42}
\newcommand{\thirtyfiveTrajDevRtwo}{0.0110}
\newcommand{\thirtyfiveTrajDevRone}{0.0082}
\newcommand{\thirtyfiveTrajDevReleased}{0.0083}
\newcommand{\twelveNwithin}{7}
\newcommand{\twelveNsources}{7}
\newcommand{\twelveMaxAminusRef}{1.0}
\newcommand{\twelveWmin}{2.8}
\newcommand{\twelveWmax}{7.6}
\newcommand{\twelveFmax}{1.6}

\newcommand{\twelveNharmful}{4{,}452}

\newcommand{\twelveFloorRcorr}{0.997}
\newcommand{\twelveAvsRefRcorr}{0.996}

\newcommand{\twelveSeedChangeRcorrMin}{0.981}
\newcommand{\twelveSeedChangeRcorrMax}{0.989}
\newcommand{\twelveLtwoAvsAiTwo}{1.72}
\newcommand{\twelveLtwoFloor}{1.67}

\newcommand{\twelveLtwoSeedMin}{3.04}
\newcommand{\twelveLtwoSeedMax}{3.06}

\newcommand{\twentythreeNitems}{1{,}200}
\newcommand{\twentythreeNprompts}{100}
\newcommand{\twentythreeNtwins}{4}
\newcommand{\twentythreeK}{3}
\newcommand{\twentythreeNboot}{1{,}000}
\newcommand{\twentythreeRmzHuman}{0.635}

\newcommand{\twentythreeRuHuman}{0.479}

\newcommand{\twentythreeGapHuman}{0.156}
\newcommand{\twentythreeGapHumanCI}{[0.036, 0.294]}
\newcommand{\twentythreeRelgapHuman}{0.246}
\newcommand{\twentythreeRelgapHumanCI}{[0.062, 0.451]}
\newcommand{\twentythreeRmzJudge}{0.773}

\newcommand{\twentythreeRuJudge}{0.570}

\newcommand{\twentythreeGapJudge}{0.203}
\newcommand{\twentythreeGapJudgeCI}{[0.101, 0.341]}
\newcommand{\twentythreeRelgapJudge}{0.263}
\newcommand{\twentythreeRelgapJudgeCI}{[0.132, 0.440]}
\newcommand{\twentythreeDiff}{+0.047}
\newcommand{\twentythreeDiffCI}{[\ensuremath{-}0.060, 0.157]}
\newcommand{\twentythreeDiffHi}{0.157}
\newcommand{\twentythreeRelgapDiff}{+0.017}
\newcommand{\twentythreeRelgapDiffCI}{[\ensuremath{-}0.136, 0.176]}

\newcommand{\twentythreeKappaFour}{0.52}
\newcommand{\twentythreeKappaRefusal}{0.65}
\newcommand{\twentythreeNdisagreements}{332}
\newcommand{\twentythreeJudgeN}{1{,}148}
\newcommand{\twentythreeJudgeAcc}{0.807}
\newcommand{\twentythreeJudgeKappa}{0.606}
\newcommand{\twentythreeJudgePrec}{0.613}
\newcommand{\twentythreeJudgeRecall}{0.971}
\newcommand{\twentythreeJudgeRate}{0.477}
\newcommand{\twentythreeHumanRate}{0.301}

\newcommand{\twentythreeGapHarry}{0.144}
\newcommand{\twentythreeGapHarryCI}{[0.045, 0.246]}
\newcommand{\twentythreeGapCoauthor}{0.124}
\newcommand{\twentythreeGapCoauthorCI}{[0.007, 0.259]}
\newcommand{\twentythreeAdjHarry}{214}
\newcommand{\twentythreeAdjCoauthor}{102}
\newcommand{\twentythreeAdjNeither}{16}

\newcommand{\fortyoneScatterN}{3{,}212}
\newcommand{\fortyoneScatterRMz}{0.985}
\newcommand{\fortyoneScatterRawMz}{0.933}
\newcommand{\fortyoneScatterRankMz}{9}

\newcommand{\fortyoneScatterRU}{0.813}
\newcommand{\fortyoneScatterRawU}{0.771}
\newcommand{\fortyoneScatterRankU}{24}

\newcommand{\fortyoneSourceN}{10}
\newcommand{\fortyoneSourceGapMin}{0.082}
\newcommand{\fortyoneSourceGapMax}{0.213}

\newcommand{\fortyoneSourceNboot}{1{,}000}
\newcommand{\fortyoneStableN}{3{,}212}

\newcommand{\fortyoneStableEvery}{4.7\,\%}
\newcommand{\fortyoneStableNone}{6.7\,\%}
\newcommand{\fortyoneStableNevery}{151}
\newcommand{\fortyoneStableNnone}{215}

\newcommand{\fortyoneAuditRefToComp}{212}
\newcommand{\fortyoneAuditCompToRef}{10}
\newcommand{\fortyoneInvalidMin}{8.1\,\%}
\newcommand{\fortyoneInvalidMax}{10.0\,\%}
\newcommand{\twentyfourSpeedup}{2.22$\times$}
\newcommand{\twentyfourSpeedupSft}{3.12$\times$}
\newcommand{\twentyfourSpeedupDpo}{1.00$\times$}
\newcommand{\twentyfourCostA}{29.8}
\newcommand{\twentyfourCostB}{66.0}
\newcommand{\twentyfourCostAsft}{17.0}

\newcommand{\twentyfourWfloor}{6.40}
\newcommand{\twentyfourWratioAB}{1.249}
\newcommand{\twentyfourWratioAR}{1.163}

\newcommand{\twentyfourWseed}{34.3}
\newcommand{\twentyfourWsftRatioAB}{1.254}

\newcommand{\twentyfourRstarDiff}{0.0012}
\newcommand{\twentyfourRstarDiffCI}{[\ensuremath{-}0.0037, 0.0065]}

\newcommand{\twentyfourLevelsWithin}{7}

\newcommand{\twentyfourGenJudge}{1.77}
\newcommand{\twentyfourTwinAllIn}{31.6}
\newcommand{\twentyfiveFirstStep}{1}
\newcommand{\twentyfiveRelLtwoStepOne}{\ensuremath{2.5\times 10^{-6}}}

\newcommand{\twentyfiveRelLtwoHundredSame}{\ensuremath{1.9\times 10^{-2}}}
\newcommand{\twentyfiveRelLtwoHundredCross}{\ensuremath{2.0\times 10^{-2}}}
\newcommand{\twentyfiveCrossOverSame}{1.07}
\newcommand{\twentyfiveCrossOverSameB}{1.02}
\newcommand{\twentyfiveUnequalStepOneMin}{93}
\newcommand{\twentyfiveUnequalStepOneMax}{96}
\newcommand{\twentyfiveNtensors}{99}
\newcommand{\twentyfiveLossDiffMin}{7}
\newcommand{\twentyfiveLossDiffMax}{26}
\newcommand{\twentyfiveGradDiffMin}{4}
\newcommand{\twentyfiveGradDiffMax}{6}
\newcommand{\twentyfiveMpsCur}{1.91$\times$}
\newcommand{\twentyfiveMpsZone}{1.44$\times$}
\newcommand{\twentyfiveTsZone}{0.81$\times$}
\newcommand{\twentyfiveScreenFail}{4}
\newcommand{\twentyfiveScreenN}{6}
\newcommand{\twentyfiveBound}{22.1}

\newcommand{\aZeroNmz}{18}
\newcommand{\aZeroNu}{48}
\newcommand{\aZeroRmz}{0.985}

\newcommand{\aZeroRu}{0.814}

\newcommand{\sixteenRmzOneFifty}{0.988}

\newcommand{\sixteenGapOneFifty}{0.204}
\newcommand{\sixteenGapOneFiftyCI}{[0.189, 0.218]}
\newcommand{\sixteenGapOneFiftyCIbases}{[0.187, 0.216]}

\newcommand{\sixteenRmzThreeHundred}{0.981}

\newcommand{\sixteenGapThreeHundred}{0.211}
\newcommand{\sixteenGapThreeHundredCI}{[0.198, 0.224]}
\newcommand{\sixteenGapThreeHundredCIbases}{[0.203, 0.219]}

\newcommand{\sixteenGapOneB}{0.171}
\newcommand{\sixteenGapOneBCI}{[0.160, 0.183]}
\newcommand{\sixteenGapOneBCIbases}{[0.154, 0.193]}

\newcommand{\sixteenDiffOneFiftyThreeHundred}{\ensuremath{-}0.008}
\newcommand{\sixteenDiffOneFiftyThreeHundredCI}{[\ensuremath{-}0.024, 0.010]}
\newcommand{\sixteenDiffOneFiftyOneB}{+0.032}
\newcommand{\sixteenDiffOneFiftyOneBCI}{[0.014, 0.050]}
\newcommand{\sixteenDiffThreeHundredOneB}{+0.040}
\newcommand{\sixteenDiffThreeHundredOneBCI}{[0.024, 0.055]}
\newcommand{\sixteenNboot}{1{,}000}

\newcommand{\sixteenVulnMeanOneFifty}{0.784}
\newcommand{\sixteenVulnMeanThreeHundred}{0.703}
\newcommand{\sixteenVulnMeanOneB}{0.491}
\newcommand{\sixteenGapHarmOneFifty}{0.064}

\newcommand{\sixteenGapHarmOneB}{0.165}
\newcommand{\sixteenGapQguardOneFifty}{0.112}

\newcommand{\sixteenGapQguardOneB}{0.147}
\newcommand{\sixteenCostTwinOneFifty}{15.97}

\newcommand{\sixteenCostTwinThreeHundred}{21.15}

\newcommand{\gridNseeds}{2}

\newcommand{\twentysevenRu}{0.814}

\newcommand{\twentysevenNboot}{2{,}000}

\newcommand{\twentysevenRepOther}{0.828}
\newcommand{\twentysevenRepOtherCI}{[0.818, 0.839]}
\newcommand{\twentysevenRungTwoMinusRepOther}{+0.011}
\newcommand{\twentysevenRungTwoMinusRepOtherCI}{[0.001, 0.022]}

\newcommand{\twentysevenBenignRungTwo}{0.793}
\newcommand{\twentysevenBenignRungTwoCI}{[0.775, 0.811]}
\newcommand{\twentysevenBenignRu}{0.851}
\newcommand{\twentysevenBenignRepOther}{0.826}
\newcommand{\twentysevenPlainRungTwo}{0.893}
\newcommand{\twentysevenPlainRu}{0.890}
\newcommand{\twentysevenBpDefRep}{0.839}

\newcommand{\twentysevenBpRepAuxTwo}{0.827}

\newcommand{\twentysevenBpRepAuxThree}{0.829}

\newcommand{\twentysevenBpAuxTwo}{0.989}

\newcommand{\twentysevenBpAuxThree}{0.981}

\newcommand{\twentysevenBpDefAuxTwo}{0.833}

\newcommand{\twentysevenBpDefAuxThree}{0.815}

\newcommand{\twentysevenBpAuxTwoAuxThree}{0.793}

\newcommand{\twentysevenReleasedPairMax}{0.833}
\newcommand{\twentysevenReleasedPairMin}{0.793}

\newcommand{\twentysevenRungTwoMinusEleven}{\ensuremath{-}0.0003}
\newcommand{\twentysevenRuMinusEleven}{+0.0024}
\newcommand{\twentysevenRmzMinusEleven}{+0.0006}
\newcommand{\twentysevenRepOtherMinusEleven}{+0.003}
\newcommand{\twentysevenBpMaxAbsShift}{0.006}

\newcommand{\twentysevenRelMin}{0.936}
\newcommand{\twentysevenRelMax}{0.951}

\newcommand{\thirtyfiveRungTwoMean}{0.844}
\newcommand{\thirtyfiveRungTwoMeanCI}{[0.835, 0.853]}
\newcommand{\thirtyfiveRungTwoMeanMinusU}{0.030}
\newcommand{\thirtyfiveRungTwoMeanMinusUCI}{[0.021, 0.040]}
\newcommand{\thirtyfiveRRminusRD}{0.011}
\newcommand{\thirtyfiveRRminusRDCI}{[\ensuremath{-}0.001, 0.023]}

\newcommand{\thirtyfiveBpRtwoD}{0.842}

\newcommand{\thirtyfiveBpRRtwo}{0.851}

\newcommand{\thirtyfiveBpRtwoAuxTwo}{0.826}

\newcommand{\thirtyfiveBpRtwoAuxThree}{0.814}

\newcommand{\thirtyfiveSamePairMin}{0.839}
\newcommand{\thirtyfiveSamePairMax}{0.851}
\newcommand{\thirtyfiveRmzRtwo}{0.979}
\newcommand{\thirtyfiveRmzRtwoCI}{[0.975, 0.982]}

\newcommand{\thirtyfiveGapFraction}{17.8\,\%}
\newcommand{\thirtyfiveGapFractionCI}{[12.6, 22.5]\,\%}
\newcommand{\thirtyfiveBenignRRtwo}{0.839}
\newcommand{\thirtyfiveBenignRRtwoCI}{[0.820, 0.857]}
\newcommand{\thirtyfiveExcessClearCells}{8}
\newcommand{\thirtyfiveSecondaryRows}{16}
\newcommand{\thirtyfiveInclMaxDiff}{0.0006}

\newcommand{\twentysevenConfigRstar}{0.9999}
\newcommand{\twentysevenConfigRstarCI}{[0.9988, 1.0012]}
\newcommand{\twentysevenConfigRstarRaw}{0.945}

\newcommand{\twentysevenConfigEffect}{+0.0003}
\newcommand{\twentysevenConfigEffectCI}{[\ensuremath{-}0.0003, 0.0008]}

\newcommand{\twentysevenConfigFactor}{1.25}

\newcommand{\twentysevenConfigWeightsSameCell}{8.97}

\newcommand{\twentysevenConfigWeightsMZwithin}{34.6}
\newcommand{\twentysevenConfigWeightsRatioSeed}{0.26}
\newcommand{\twentysevenConfigRatioMean}{1.40$\times$}
\newcommand{\twentysevenConfigRatioMin}{1.16}
\newcommand{\twentysevenConfigRatioMax}{1.61}
\newcommand{\twentysevenConfigNwithinBound}{2}
\newcommand{\twentysevenConfigNsameCell}{12}

\newcommand{\twentysevenConfigLevelsWithin}{84}
\newcommand{\twentysevenConfigLevelsTotal}{84}

\newcommand{\twentysevenConfigLevelsMaxAbs}{2.23}

\newcommand{\twentysevenConfigLevelsSourceMeanMin}{\ensuremath{-}0.41}
\newcommand{\twentysevenConfigLevelsSourceMeanMax}{\ensuremath{-}0.11}

\newcommand{\eightNmodels}{150}
\newcommand{\eightNrecipes}{25}

\newcommand{\eightNprompts}{3{,}212}
\newcommand{\eightNrPairs}{10{,}800}
\newcommand{\eightNmzPairsRecipe}{3}
\newcommand{\eightNuPairsRecipe}{12}
\newcommand{\eightNrPairsRecipe}{864}

\newcommand{\eightGapMean}{0.173}

\newcommand{\eightGapCIrecipes}{[0.167, 0.179]}
\newcommand{\eightGapCIprompts}{[0.166, 0.180]}
\newcommand{\eightRmz}{0.984}

\newcommand{\eightRu}{0.811}

\newcommand{\eightRr}{0.779}

\newcommand{\eightUminusR}{0.032}

\newcommand{\eightSeedOverRecipe}{5.4}

\newcommand{\eightNrecipesOrdered}{25}
\newcommand{\eightShareRecipe}{0.51}
\newcommand{\eightSharePretrain}{0.31}
\newcommand{\eightSharePosttrain}{0.18}

\newcommand{\eightGapRecipeMin}{0.145}
\newcommand{\eightGapRecipeMax}{0.209}
\newcommand{\eightNrecipesGapCIaboveZero}{25}
\newcommand{\eightGapMain}{0.175}

\newcommand{\eightVulnRecipeMax}{0.664}
\newcommand{\eightOutlierRr}{0.647}
\newcommand{\eightOutlierUminusR}{0.146}
\newcommand{\eightMinUminusR}{0.002}

\newcommand{\eightShareRecipeHarm}{0.64}

\newcommand{\eightShareRecipeQguard}{0.23}
\newcommand{\eightSharePretrainQguard}{0.49}

\newcommand{\eightShareRecipeAgree}{0.38}
\newcommand{\eightSharePretrainAgree}{0.39}

\newcommand{\eightGapPlain}{0.100}

\newcommand{\eightShareRecipePlain}{0.62}

\newcommand{\eightShareRecipeHarmful}{0.59}

\newcommand{\eightShareRecipeBenign}{0.12}
\newcommand{\eightSharePretrainBenign}{0.72}

\newcommand{\eightGapWjNonMain}{0.139}
\newcommand{\eightGapWjNonMainMin}{0.121}
\newcommand{\eightGapWjNonMainMax}{0.177}
\newcommand{\eightGapWjMain}{0.134}
\newcommand{\eightGapHjb}{0.210}

\newcommand{\eightGapHjbNonMain}{0.210}
\newcommand{\eightGapHjbNonMainMin}{0.169}
\newcommand{\eightGapHjbNonMainMax}{0.291}
\newcommand{\eightGapHjbMain}{0.214}
\newcommand{\eightGapResid}{0.525}

\newcommand{\eightRuResid}{0.421}
\newcommand{\eightRrResid}{0.365}
\newcommand{\eightGapResidNonMain}{0.524}
\newcommand{\eightGapResidNonMainMin}{0.472}
\newcommand{\eightGapResidNonMainMax}{0.598}
\newcommand{\eightGapResidMain}{0.553}
\newcommand{\eightResidNprompts}{1{,}212}
\newcommand{\eightResidNbehaviours}{300}
\newcommand{\eightResidNtemplates}{91}
\newcommand{\eightTransferK}{322}
\newcommand{\eightTransferMz}{0.994}
\newcommand{\eightTransferU}{0.931}
\newcommand{\eightTransferR}{0.919}

\newcommand{\eightTransferUShortfall}{7}

\newcommand{\eightRefMain}{0.671}
\newcommand{\eightRefAll}{0.636}
\newcommand{\eightRefMin}{0.443}
\newcommand{\eightRefMax}{0.704}
\newcommand{\eightOlmesPearsonRecipes}{\ensuremath{-}0.41}
\newcommand{\eightOlmesSpearmanRecipes}{\ensuremath{-}0.48}

\newcommand{\eightOlmesPearsonBases}{\ensuremath{-}0.34}

\newcommand{\eightCostTotal}{4{,}969.9}

\newcommand{\eightCostTrainCell}{29.9}

\newcommand{\eightAuditNitems}{400}
\newcommand{\eightAuditNpopulation}{383}
\newcommand{\eightAuditNgreedy}{17}

\newcommand{\eightAuditNadjudicated}{73}
\newcommand{\eightAuditPrecWj}{0.553}
\newcommand{\eightAuditPrecWjCI}{[0.428, 0.675]}
\newcommand{\eightAuditOneFlipWj}{0.020}

\newcommand{\eightAuditPrecWjMinusFifteenAbs}{0.03}
\newcommand{\eightAuditPrecHjb}{0.484}
\newcommand{\eightAuditPrecHjbCI}{[0.363, 0.601]}
\newcommand{\eightAuditOneFlipHjb}{0.024}

\newcommand{\eightAuditPrecHjbMinusFifteenAbs}{0.12}
\newcommand{\eightAuditPrecRecipeMin}{0.01}
\newcommand{\eightAuditPrecRecipeMax}{0.89}

\newcommand{\eightAuditHetP}{0.321}
\newcommand{\eightAuditHetT}{12.69}
\newcommand{\eightAuditHetTnull}{11.51}
\newcommand{\eightAuditHetNperm}{10{,}000}
\newcommand{\eightAuditHetAlpha}{0.01}

\newcommand{\eightAuditFlipsToCross}{14}
\newcommand{\eightAuditFlipCandidates}{33}

\newcommand{\eightAuditJudgeRate}{0.549}
\newcommand{\eightAuditJudgeRateCI}{[0.488, 0.614]}
\newcommand{\eightAuditHumanRate}{0.297}
\newcommand{\eightAuditHumanRateCI}{[0.237, 0.357]}
\newcommand{\eightAuditPrecision}{0.528}
\newcommand{\eightAuditPrecisionCI}{[0.436, 0.615]}
\newcommand{\eightAuditRecall}{0.977}
\newcommand{\eightAuditRecallCI}{[0.956, 0.994]}
\newcommand{\eightAuditFPmass}{0.259}

\newcommand{\eightAuditFNmass}{0.007}

\newcommand{\eightLevelRaw}{0.528}
\newcommand{\eightLevelCorr}{0.278}
\newcommand{\eightLevelCorrCI}{[0.230, 0.324]}
\newcommand{\eightLevelCorrRecipeMin}{0.246}

\newcommand{\eightLevelCorrRecipeMax}{0.347}
\newcommand{\eightLevelCorrRecipeMaxCI}{[0.289, 0.404]}
\newcommand{\eightLevelCorrModelMin}{0.211}
\newcommand{\eightLevelCorrModelMax}{0.363}

\newcommand{\eightAuditGap}{0.168}

\newcommand{\eightAuditGapCIrecipes}{[0.162, 0.174]}
\newcommand{\eightAuditGapCIprompts}{[0.161, 0.175]}
\newcommand{\eightAuditGapCIaudit}{[0.144, 0.174]}

\newcommand{\eightAuditGapNone}{0.173}

\newcommand{\eightAuditAccWj}{0.782}

\newcommand{\fifteenAccSourceWj}{0.840}
\newcommand{\eightAuditAccHjb}{0.844}

\newcommand{\fifteenAccSourceHjb}{0.844}
\newcommand{\eightAuditAccW}{0.804}

\newcommand{\eightAuditNscored}{356}
\newcommand{\eightAuditKappaRefusal}{0.768}
\newcommand{\eightAuditKappaFour}{0.740}
\newcommand{\eightShareRecipeCorr}{0.49}
\newcommand{\eightShareRecipeCorrCI}{[0.12, 0.70]}
\newcommand{\eightShareRecipeCorrPct}{49}
\newcommand{\eightShareRecipeCorrPctCI}{[12\,\%, 70\,\%]}

\newcommand{\eightSharePretrainCorr}{0.32}
\newcommand{\eightSharePretrainCorrCI}{[0.16, 0.59]}
\newcommand{\eightSharePretrainCorrPct}{32}
\newcommand{\eightSharePretrainCorrPctCI}{[16\,\%, 59\,\%]}

\newcommand{\eightSharePosttrainCorr}{0.19}
\newcommand{\eightSharePosttrainCorrCI}{[0.11, 0.37]}
\newcommand{\eightSharePosttrainCorrPct}{19}
\newcommand{\eightSharePosttrainCorrPctCI}{[11\,\%, 37\,\%]}

\newcommand{\eightSeedOverRecipeTimesCorr}{5.7$\times$}
\newcommand{\eightSeedOverRecipeTimesCorrCI}{[4.4, 7.2]}

\newcommand{\twentyoneNrecipes}{24}

\newcommand{\twentyoneLink}{0.575}
\newcommand{\twentyoneLinkCI}{[0.569, 0.581]}
\newcommand{\twentyoneLinkCommon}{0.306}

\newcommand{\twentyoneLinkOneFiftyLastToken}{0.34}

\newcommand{\twentyonePerRecipeSD}{0.065}
\newcommand{\twentyonePerRecipeSDCommon}{0.147}
\newcommand{\twentyoneNrecipesRunsThrough}{21}
\newcommand{\twentyoneNrecipesPartly}{3}
\newcommand{\twentyoneNrecipesDiffPositive}{24}

\newcommand{\twentyoneLinkThreeHundred}{0.492}
\newcommand{\twentyoneLinkThreeHundredCI}{[0.476, 0.515]}

\newcommand{\twentyoneHarmGapAZeroCommon}{0.591}
\newcommand{\twentyoneRefusalGapAZeroCommon}{0.403}

\newcommand{\twentyoneRefusalGapAZeroSite}{0.268}

\newcommand{\twentyoneHarmGapBCommon}{0.083}
\newcommand{\twentyoneRefusalGapBCommon}{0.323}

\newcommand{\twentyoneRefusalGapBSite}{0.201}

\newcommand{\twentyoneHarmGapThreeHundredCommon}{0.007}

\newcommand{\twentyoneHarmLinkMax}{0.14}

\newcommand{\fortysixNShared}{21}
\newcommand{\fortysixNSpecific}{4}

\newcommand{\fortysixSharedHarmRuMinThirteen}{0.989}
\newcommand{\fortysixSharedHarmRuMaxThirteen}{0.994}

\newcommand{\fortysixSpecificHarmRuMinThirteen}{0.414}
\newcommand{\fortysixSpecificHarmRuMaxThirteen}{0.454}

\newcommand{\fortysixSharedWithinTokenHarmRuThirteen}{0.74}
\newcommand{\fortysixSharedWithinTokenRefusalRuMinThirteen}{0.73}
\newcommand{\fortysixSharedWithinTokenRefusalRuMaxThirteen}{0.80}
\newcommand{\fortysixSharedWithinTokenHarmGapMinThirteen}{0.24}
\newcommand{\fortysixSharedWithinTokenHarmGapMaxThirteen}{0.30}
\newcommand{\fortysixSharedWithinTokenRefusalGapMinThirteen}{0.19}
\newcommand{\fortysixSharedWithinTokenRefusalGapMaxThirteen}{0.23}

\newcommand{\fortysixHarmfulEndsInWord}{84\,\%}
\newcommand{\fortysixHarmlessEndsInWord}{19\,\%}
\newcommand{\fortysixNRuns}{75}
\newcommand{\fortysixNRunsToken}{71}
\newcommand{\fortysixNRunsOdd}{4}
\newcommand{\fortysixMainRefusalUOddThirteen}{0.664}
\newcommand{\fortysixMainRefusalUTokenThirteen}{0.817}
\newcommand{\fortysixOtherRefusalUDiffMaxThirteen}{0.04}
\newcommand{\fortysixTokenEtaMinThirteen}{0.97}
\newcommand{\fortysixTokenEtaMaxThirteen}{0.99}
\newcommand{\fortysixOddEtaMinThirteen}{0.03}
\newcommand{\fortysixOddEtaMaxThirteen}{0.05}
\newcommand{\fortysixOddTokPairHarmRuMinThirteen}{0.99}

\newcommand{\fortysixOddPairHarmRuMinThirteen}{0.13}
\newcommand{\fortysixOddPairHarmRuMaxThirteen}{0.18}
\newcommand{\fortysixOddRunHarmGapMinThirteen}{0.85}

\newcommand{\fortysixOddTokRunHarmGapMinThirteen}{0.01}

\newcommand{\fortyfourNprompts}{3{,}211}

\newcommand{\fortyfourFlipMZ}{0.108}
\newcommand{\fortyfourFlipMZCI}{[0.104, 0.112]}

\newcommand{\fortyfourTransferMZ}{0.750}
\newcommand{\fortyfourTransferMZCI}{[0.740, 0.759]}
\newcommand{\fortyfourStrongFlipMZ}{\ensuremath{4.2\times 10^{-6}}}

\newcommand{\fortyfourFlipU}{0.170}
\newcommand{\fortyfourFlipUCI}{[0.164, 0.176]}

\newcommand{\fortyfourTransferU}{0.609}
\newcommand{\fortyfourTransferUCI}{[0.594, 0.622]}
\newcommand{\fortyfourStrongFlipU}{\ensuremath{2.1\times 10^{-3}}}

\newcommand{\fortyfourFlipR}{0.179}
\newcommand{\fortyfourFlipRCI}{[0.172, 0.185]}

\newcommand{\fortyfourTransferR}{0.587}
\newcommand{\fortyfourTransferRCI}{[0.571, 0.602]}
\newcommand{\fortyfourStrongFlipR}{\ensuremath{3.9\times 10^{-3}}}

\newcommand{\fortyfourStrongFlipMZCount}{1}
\newcommand{\fortyfourPairPromptsMZ}{240{,}825}

\newcommand{\fortyfourTransferUminusR}{0.022}
\newcommand{\fortyfourNtemplates}{91}
\newcommand{\fortyfourTemplateMZ}{0.980}
\newcommand{\fortyfourTemplateMZCI}{[0.973, 0.983]}

\newcommand{\fortyfourTemplateU}{0.741}
\newcommand{\fortyfourTemplateUCI}{[0.680, 0.788]}

\newcommand{\fortyfourTemplateR}{0.713}
\newcommand{\fortyfourTemplateRCI}{[0.641, 0.768]}

\newcommand{\fortyfourTemplateUminusR}{+0.028}
\newcommand{\fortyfourTemplateUminusRCI}{[0.015, 0.042]}
\newcommand{\fortyfourTemplateRelMin}{0.909}

\newcommand{\fortyfourTemplateRelMax}{0.990}

\newcommand{\fortyfiveSDRecipe}{3.5}

\newcommand{\fortyfiveSDPretrain}{2.7}

\newcommand{\fortyfiveSDPosttrain}{2.1}

\newcommand{\fortyfiveSDSampling}{0.15}

\newcommand{\fortyfiveShareRecipeCI}{[0.13, 0.72]}

\newcommand{\fortyfiveSharePretrainCI}{[0.15, 0.58]}

\newcommand{\fortyfivePreMinusPostPtSq}{3.1}
\newcommand{\fortyfivePreMinusPostPtSqCI}{[\ensuremath{-}0.6, 6.9]}

\newcommand{\fortyfiveSEOneOne}{3.44}
\newcommand{\fortyfiveSEOneOneCI}{[2.96, 3.88]}

\newcommand{\fortyfiveSEOneFour}{2.93}

\newcommand{\fortyfiveSETwoOne}{2.43}

\newcommand{\fortyfiveSEThreeOne}{1.99}
\newcommand{\fortyfiveSEThreeOneCI}{[1.71, 2.24]}

\newcommand{\fortyfiveSEOneOneRounded}{3.4}

\newcommand{\fortyfiveLeOnePtNpre}{12}
\newcommand{\fortyfiveLeOnePtNpost}{1}
\newcommand{\fortyfiveFloorNpre}{8}

\newcommand{\fortyfiveFlipOverall}{0.234}
\newcommand{\fortyfiveFlipOverallCI}{[0.176, 0.293]}
\newcommand{\fortyfiveFlipModelOverall}{0.257}
\newcommand{\fortyfiveFlipGeTwo}{0.148}
\newcommand{\fortyfiveFlipGeTwoCI}{[0.099, 0.197]}

\newcommand{\fortyfiveNrecipePairsGeTwo}{198}
\newcommand{\fortyfiveFlipBinZeroOne}{0.450}

\newcommand{\fortyfiveFlipBinEightUp}{0.006}

\newcommand{\fortyfiveRelOne}{0.759}
\newcommand{\fortyfiveRelOneCI}{[0.751, 0.766]}
\newcommand{\fortyfiveRelTwo}{0.863}
\newcommand{\fortyfiveRelTwoCI}{[0.858, 0.867]}
\newcommand{\fortyfiveRelThree}{0.904}
\newcommand{\fortyfiveRelThreeCI}{[0.900, 0.908]}
\newcommand{\fortyfiveRelSix}{0.919}
\newcommand{\fortyfiveRelSixCI}{[0.915, 0.922]}
\newcommand{\fortyfiveNoiseBase}{0.162}
\newcommand{\fortyfiveNoisePost}{0.015}
\newcommand{\fortyfiveNoiseSampling}{0.064}
\newcommand{\fortyfiveNoiseTotal}{0.241}
\newcommand{\fortyfivePreMinusPostBenignPtSq}{12.8}
\newcommand{\fortyfivePreMinusPostBenignPtSqCI}{[5.1, 22.4]}
\newcommand{\fortyfiveSEOneOneBenign}{4.51}

\newcommand{\fortyfivePreMinusPostQguardPtSqCI}{[\ensuremath{-}0.4, 5.1]}

\newcommand{\seventyfourBinLo}{10}
\newcommand{\seventyfourBinHi}{20}

\newcommand{\seventyfourTopOneKeepUPct}{27}
\newcommand{\seventyfourTopOneKeepMZPct}{67}

\newcommand{\seventysixNprompts}{3{,}211}

\newcommand{\seventysixRunCorrect}{75}
\newcommand{\seventysixRunN}{75}

\newcommand{\seventysixRunFifty}{0.981}

\newcommand{\seventysixRecipeCorrect}{24}
\newcommand{\seventysixRecipeN}{50}

\newcommand{\seventysixNsub}{200}
\newcommand{\seventysixFiftyN}{50}
\newcommand{\seventysixRunFiftyPct}{98}

\newcommand{\seventysixRecipeHeldOne}{0.40}
\newcommand{\seventysixRecipeHeldTwo}{0.46}
\newcommand{\seventysixRecipeHeldThree}{0.48}

\newcommand{\seventythreeGapCentred}{0.229}
\newcommand{\seventythreeGapCentredCI}{[0.209, 0.237]}

\newcommand{\seventythreeGapRaw}{0.268}

\newcommand{\seventythreeLinkCentred}{0.580}
\newcommand{\seventythreeLinkCentredCI}{[0.574, 0.589]}

\newcommand{\seventythreeLinkRaw}{0.575}

\newcommand{\seventythreeLogitLink}{0.349}
\newcommand{\seventythreeLogitLinkCI}{[0.343, 0.356]}

\newcommand{\seventythreePartial}{0.528}
\newcommand{\seventythreePartialCI}{[0.521, 0.534]}

\newcommand{\seventythreeLinkProjCentred}{0.506}
\newcommand{\seventythreeLinkProjCentredClusterCI}{[0.457, 0.539]}

\newcommand{\seventythreeNpairs}{72}

\newcommand{\seventythreeEtaProjMin}{0.19}
\newcommand{\seventythreeEtaProjMax}{0.57}
\newcommand{\seventythreeLensFinalMin}{0.710}
\newcommand{\seventythreeLensFinalMax}{0.940}
\newcommand{\eightynineDoseMeanPMinusTwo}{0.741}
\newcommand{\eightynineDoseRcMinusTwo}{0.881}

\newcommand{\eightynineDoseRcMinusHalf}{0.992}
\newcommand{\eightynineDoseMeanPZero}{0.484}

\newcommand{\eightynineDoseRcPlusHalf}{0.989}

\newcommand{\eightynineDoseMeanPPlusTwo}{0.214}
\newcommand{\eightynineDoseRcPlusTwo}{0.869}
\newcommand{\eightynineGateDiff}{0.140}
\newcommand{\eightynineGateDiffCI}{[0.136, 0.145]}
\newcommand{\eightynineIdentityRc}{1.001}
\newcommand{\eightynineIdentityRcCI}{[0.997, 1.005]}
\newcommand{\eightynineClosureTransplant}{0.483}
\newcommand{\eightynineClosureTransplantCI}{[0.451, 0.514]}
\newcommand{\eightynineClosureVsTwinTransplant}{0.433}
\newcommand{\eightynineClosureVsTwinTransplantCI}{[0.403, 0.465]}
\newcommand{\eightynineClosureTransplantPairMin}{0.426}
\newcommand{\eightynineClosureTransplantPairMax}{0.536}
\newcommand{\eightynineClosurePermuted}{\ensuremath{-}1.229}
\newcommand{\eightynineClosurePermutedCI}{[\ensuremath{-}1.392, \ensuremath{-}1.095]}

\newcommand{\eightynineClosureMzDonor}{0.498}
\newcommand{\eightynineClosureMzDonorCI}{[0.466, 0.532]}

\newcommand{\eightynineClosureSiteEleven}{0.029}
\newcommand{\eightynineClosureSiteElevenCI}{[\ensuremath{-}0.019, 0.071]}

\newcommand{\eightynineClosureSiteElevenPairMin}{\ensuremath{-}0.284}
\newcommand{\eightynineClosureSiteElevenPairMax}{0.228}
\newcommand{\eightynineClosureIdentity}{\ensuremath{-}0.020}
\newcommand{\eightynineClosureIdentityCI}{[\ensuremath{-}0.051, 0.012]}

\newcommand{\eightynineTransplantMinusSiteEleven}{0.454}
\newcommand{\eightynineTransplantMinusSiteElevenCI}{[0.401, 0.509]}
\newcommand{\eightynineMzDonorMinusTransplant}{+0.016}
\newcommand{\eightynineMzDonorMinusTransplantCI}{[\ensuremath{-}0.013, 0.042]}
\newcommand{\eightyninePairRcMin}{0.803}
\newcommand{\eightyninePairRcMax}{0.836}
\newcommand{\eightynineEditRcMin}{0.891}
\newcommand{\eightynineEditRcMax}{0.904}
\newcommand{\eightynineSelfTransplantMin}{0.925}
\newcommand{\eightynineSelfTransplantMax}{0.955}

\newcommand{\eightynineOffsetSDMin}{0.61}
\newcommand{\eightynineOffsetSDMax}{0.79}

\newcommand{\ninetyoneClosureNewNine}{0.495}
\newcommand{\ninetyoneClosureNewNineCI}{[0.472, 0.518]}
\newcommand{\ninetyoneClosureReversed}{0.504}
\newcommand{\ninetyoneClosureReversedCI}{[0.470, 0.535]}
\newcommand{\ninetyoneClosureSeedThree}{0.495}
\newcommand{\ninetyoneClosureSeedThreeCI}{[0.466, 0.526]}
\newcommand{\ninetyoneClosureSeedFour}{0.487}
\newcommand{\ninetyoneClosureSeedFourCI}{[0.455, 0.517]}
\newcommand{\ninetyoneClosureOtherSeed}{0.491}
\newcommand{\ninetyoneClosureOtherSeedCI}{[0.466, 0.516]}
\newcommand{\ninetyoneClosureAllTwelve}{0.492}
\newcommand{\ninetyoneClosureAllTwelveCI}{[0.469, 0.515]}

\newcommand{\ninetyoneNewNineMinusEightyNine}{+0.013}
\newcommand{\ninetyoneNewNineMinusEightyNineCI}{[\ensuremath{-}0.013, 0.039]}
\newcommand{\ninetyoneIdentityNewNine}{\ensuremath{-}0.023}
\newcommand{\ninetyoneIdentityNewNineCI}{[\ensuremath{-}0.041, \ensuremath{-}0.005]}
\newcommand{\ninetyoneClosurePairMin}{0.395}
\newcommand{\ninetyoneClosurePairMax}{0.622}
\newcommand{\ninetyoneIdentityRecordMin}{0.997}
\newcommand{\ninetyoneIdentityRecordMax}{1.004}
\newcommand{\ninetyoneOffsetSDMin}{0.59}
\newcommand{\ninetyoneOffsetSDMax}{0.88}

\newcommand{\ninetyoneNpairsAll}{12}
\newcommand{\eightyAdopted}{0.558}
\newcommand{\eightyAdoptedCI}{[0.537, 0.577]}
\newcommand{\eightyAdoptedNpairs}{24}
\newcommand{\eightySwapMZ}{0.966}
\newcommand{\eightySwapMZCI}{[0.962, 0.970]}
\newcommand{\eightySwapMZNpairs}{3}
\newcommand{\eightySwapU}{0.788}
\newcommand{\eightySwapUCI}{[0.771, 0.802]}
\newcommand{\eightySwapUNpairs}{12}
\newcommand{\eightyCross}{0.470}
\newcommand{\eightyCrossCI}{[0.446, 0.490]}
\newcommand{\eightyCrossNpairs}{48}
\newcommand{\eightySwapGap}{+0.178}
\newcommand{\eightySwapGapCI}{[0.165, 0.194]}
\newcommand{\eightyAdoptedMinusU}{\ensuremath{-}0.256}
\newcommand{\eightyAdoptedMinusUCI}{[\ensuremath{-}0.278, \ensuremath{-}0.235]}
\newcommand{\eightyAdoptedMinusCross}{+0.088}
\newcommand{\eightyAdoptedMinusCrossCI}{[0.081, 0.096]}
\newcommand{\eightyAdoptedJackSE}{0.003}
\newcommand{\eightySwapGapJackSE}{0.023}
\newcommand{\eightyAdoptedMinusCrossJackSE}{0.014}

\newcommand{\eightyRecordU}{0.811}
\newcommand{\eightyMeanVulnTulu}{0.49}
\newcommand{\eightyRelTulu}{0.94}
\newcommand{\eightyMeanVulnSwap}{0.25}
\newcommand{\eightyRelSwap}{0.96}
\newcommand{\eightyExploreTwentyFiveAdopted}{0.553}
\newcommand{\eightyExploreTwentyFiveU}{0.835}
\newcommand{\eightyExploreFiftyAdopted}{0.578}
\newcommand{\eightyExploreFiftyU}{0.839}
\newcommand{\eightyFpCorrect}{12}
\newcommand{\eightyFpN}{12}
\newcommand{\eightyFpAccCI}{[0.74, 1.00]}
\newcommand{\eightyFpTrueMean}{0.530}
\newcommand{\eightyFpImpostorMean}{0.501}
\newcommand{\eightyFpMarginMin}{0.001}
\newcommand{\eightyFpNprompts}{3{,}211}
\newcommand{\eightyFpThousand}{0.87}
\newcommand{\eightyFpThousandRange}{[0.67, 1.00]}

\newcommand{\eightyFpFifty}{0.14}
\newcommand{\eightyFpFiftyRange}{[0.00, 0.42]}

\newcommand{\eightyFpImpostorSameRecipe}{0}

\newcommand{\eightyFpImpostorFalconOrigTen}{5}
\newcommand{\eightyFpImpostorDolmaNoFlan}{4}
\newcommand{\eightyFpImpostorDolmaSixPlus}{3}
\newcommand{\eightyMeanVulnTuluTwentyFive}{0.523}
\newcommand{\eightyMeanVulnSwapTwentyFive}{0.364}
\newcommand{\eightyMeanVulnTuluFifty}{0.521}
\newcommand{\eightyMeanVulnSwapFifty}{0.382}
\newcommand{\eightyNprompts}{3{,}212}
\newcommand{\eightyPosthocWjN}{2{,}000}
\newcommand{\eightyPosthocWjAdopted}{0.390}
\newcommand{\eightyPosthocWjAdoptedCI}{[0.362, 0.418]}

\newcommand{\eightyPosthocWjSwapU}{0.778}

\newcommand{\eightyPosthocWjTuluU}{0.854}

\newcommand{\eightyPosthocWjAdoptedMinusU}{\ensuremath{-}0.464}

\newcommand{\eightyPosthocHjN}{1{,}212}
\newcommand{\eightyPosthocHjAdopted}{0.821}
\newcommand{\eightyPosthocHjAdoptedCI}{[0.806, 0.837]}

\newcommand{\eightyPosthocHjSwapU}{0.772}

\newcommand{\eightyPosthocHjTuluU}{0.773}

\newcommand{\eightyPosthocHjAdoptedMinusU}{+0.048}
\newcommand{\eightyPosthocHjAdoptedMinusUCI}{[0.029, 0.069]}
\newcommand{\eightyPosthocAdoptedHjMinusWj}{+0.431}
\newcommand{\eightyPosthocAdoptedHjMinusWjCI}{[0.400, 0.463]}

\newcommand{\eightyPosthocLadderSwapUMinusR}{+0.034}
\newcommand{\eightyPosthocLadderSwapUMinusRCI}{[0.029, 0.040]}
\newcommand{\eightyPosthocLadderSwapMZMinusU}{+0.178}
\newcommand{\eightyPosthocLadderSwapMZMinusUCI}{[0.165, 0.194]}
\newcommand{\eightyPosthocLadderSwapRatio}{5.2}
\newcommand{\eightyPosthocLadderSwapRatioCI}{[4.4, 6.3]}

\newcommand{\eightyPosthocLadderTuluUMinusR}{+0.009}
\newcommand{\eightyPosthocLadderTuluUMinusRCI}{[0.006, 0.013]}
\newcommand{\eightyPosthocLadderTuluMZMinusU}{+0.175}
\newcommand{\eightyPosthocLadderTuluMZMinusUCI}{[0.162, 0.187]}
\newcommand{\eightyPosthocLadderTuluRatio}{18.5}
\newcommand{\eightyPosthocLadderTuluRatioCI}{[13.3, 30.9]}
\newcommand{\eightyPosthocLadderUMinusRDiff}{+0.025}
\newcommand{\eightyPosthocLadderUMinusRDiffCI}{[0.019, 0.031]}
\newcommand{\eightyAuditPrecision}{0.759}
\newcommand{\eightyAuditPrecisionCI}{[0.583, 0.868]}
\newcommand{\eightyAuditRecall}{0.888}
\newcommand{\eightyAuditRecallCI}{[0.751, 0.980]}

\newcommand{\eightyAuditFpMass}{0.054}
\newcommand{\eightyAuditFpMassCI}{[0.030, 0.085]}
\newcommand{\eightyAuditFnMass}{0.021}
\newcommand{\eightyAuditFnMassCI}{[0.004, 0.044]}

\newcommand{\eightyAuditNjv}{96}
\newcommand{\eightyAuditKappaRefusal}{0.634}

\newcommand{\eightAuditPrecDclm}{0.748}
\newcommand{\eightAuditPrecDclmCI}{[0.282, 0.987]}
\newcommand{\eightAuditNjvDclm}{10}
\newcommand{\sixGapPlainMin}{0.082}
\newcommand{\sixGapPlainMax}{0.099}
\newcommand{\sixGapAdvMin}{0.147}
\newcommand{\sixGapAdvMax}{0.173}
\newcommand{\sixRmzJackSE}{0.0010}
\newcommand{\fortyoneSorryRu}{0.910}
\newcommand{\fiftythreeMinMZ}{0.966}
\newcommand{\fiftythreeMaxMZ}{0.994}
\newcommand{\fiftythreeMinU}{0.716}
\newcommand{\fiftythreeMaxU}{0.857}
\newcommand{\fiftythreeMinR}{0.456}
\newcommand{\fiftythreeMaxR}{0.863}
\newcommand{\fiftythreeNbelow}{925}
\newcommand{\fiftythreeNbelowOutlier}{735}
\newcommand{\fiftythreeMinNoOutlier}{0.520}
\newcommand{\eightUminusRColourPct}{6}
\newcommand{\eightOutlierRmz}{0.973}
\newcommand{\eightOutlierRu}{0.794}
\newcommand{\eightOutlierGap}{0.179}
\newcommand{\eightRmzRecipeMin}{0.973}
\newcommand{\eightRmzRecipeMax}{0.989}
\newcommand{\eightVulnRecipeSecond}{0.576}
\newcommand{\eightGapInclFour}{0.1731}
\newcommand{\eightGapFour}{0.1728}
\newcommand{\fortyfiveShareRecipeQguardCI}{[0.04, 0.37]}
\newcommand{\fortyfiveShareRecipeBenignCI}{[0.00, 0.30]}
\newcommand{\fortyfiveSharePretrainBenignCI}{[0.50, 0.89]}
\newcommand{\fortyfiveSDRecipeQguard}{1.7}
\newcommand{\fortyfiveSDPretrainQguard}{2.4}
\newcommand{\fortyfiveSDPosttrainQguard}{1.8}
\newcommand{\eightTransferMzNonMain}{0.994}
\newcommand{\eightTransferMzNonMainMin}{0.990}
\newcommand{\eightTransferMzNonMainMax}{0.998}

\newcommand{\eightTransferUNonMain}{0.931}
\newcommand{\eightTransferUNonMainMin}{0.908}
\newcommand{\eightTransferUNonMainMax}{0.944}

\newcommand{\eightTransferRNonMain}{0.918}
\newcommand{\eightTransferRNonMainMin}{0.823}
\newcommand{\eightTransferRNonMainMax}{0.950}

\newcommand{\sixteenVarPOneFifty}{0.042}
\newcommand{\sixteenNeverRefusedOneFifty}{0.154}
\newcommand{\sixteenAlwaysRefusedOneFifty}{0.000}
\newcommand{\sixteenVarPThreeHundred}{0.069}
\newcommand{\sixteenNeverRefusedThreeHundred}{0.134}
\newcommand{\sixteenAlwaysRefusedThreeHundred}{0.007}
\newcommand{\sixteenVarPOneB}{0.122}
\newcommand{\sixteenNeverRefusedOneB}{0.085}
\newcommand{\sixteenAlwaysRefusedOneB}{0.116}
\newcommand{\sixteenGapHarmOneFiftyCI}{[0.059, 0.069]}

\newcommand{\sixteenGapHarmOneBCI}{[0.152, 0.180]}
\newcommand{\sixteenGapQguardOneFiftyCI}{[0.104, 0.120]}

\newcommand{\sixteenGapQguardOneBCI}{[0.137, 0.157]}
\newcommand{\sixteenGapAgreeOneFifty}{0.169}
\newcommand{\sixteenGapAgreeOneFiftyCI}{[0.157, 0.182]}
\newcommand{\sixteenGapAgreeThreeHundred}{0.201}
\newcommand{\sixteenGapAgreeThreeHundredCI}{[0.188, 0.213]}
\newcommand{\sixteenGapAgreeOneB}{0.161}
\newcommand{\sixteenGapAgreeOneBCI}{[0.150, 0.172]}
\newcommand{\sixteenGapPlainOneFifty}{0.130}
\newcommand{\sixteenGapPlainOneFiftyCI}{[0.117, 0.145]}
\newcommand{\sixteenGapPlainThreeHundred}{0.094}

\newcommand{\sixteenGapPlainOneB}{0.100}

\newcommand{\sixteenGapHarmfulOneFifty}{0.159}

\newcommand{\sixteenGapHarmfulThreeHundred}{0.154}

\newcommand{\sixteenGapHarmfulOneB}{0.140}

\newcommand{\sixteenGapBenignOneFifty}{0.132}

\newcommand{\sixteenGapBenignThreeHundred}{0.127}

\newcommand{\sixteenGapBenignOneB}{0.138}

\newcommand{\sixteenRefOneFifty}{0.318}
\newcommand{\sixteenRefThreeHundred}{0.438}
\newcommand{\sixteenRefOneB}{0.672}
\newcommand{\sixteenLossDiffAuxTwo}{0.0006}
\newcommand{\twentyfourProfNccl}{74\,\%}
\newcommand{\twentyfourProfExposed}{63\,\%}
\newcommand{\twentyfourProfCompute}{17\,\%}
\newcommand{\twentyfourProfIdle}{21\,\%}
\newcommand{\twentyfourCandSpeedZeroTwo}{1.99$\times$}
\newcommand{\twentyfourCandSpeedZeroOne}{3.11$\times$}
\newcommand{\twentyfourCandSpeedZeroThreeTuned}{1.40$\times$}
\newcommand{\twentyfourCandSpeedEnv}{0.99$\times$}
\newcommand{\twentyfourCandZeroTwoDev}{1.22}
\newcommand{\twentyfourCandCheck}{0.81}
\newcommand{\sixtyoneCostPretrainRerunMax}{745}

\newcommand{\eightCostGenJudgeCellCompleted}{1.87}
\newcommand{\computeTotal}{11{,}396.6}

\newcommand{\computeSmallModelsTotal}{445.6}
\newcommand{\threeOneAMax}{4.7\,\%}
\newcommand{\threeOneAWorstStep}{10{,}000}
\newcommand{\threeOneAFinal}{2.0\,\%}
\newcommand{\threeRepTwoMin}{2.6}
\newcommand{\threeRepTwoMax}{15.6\,\%}
\newcommand{\threeOneASameStepsMin}{2.1}
\newcommand{\threeOneASameStepsMax}{4.5\,\%}
\newcommand{\threeRelLtwoRepTwo}{1.12}
\newcommand{\threeRelLtwoReleased}{1.15}
\newcommand{\threeRelLtwoOtherSeeds}{1.41}
\newcommand{\threeNewSeedZLoss}{0.05}
\newcommand{\threeNewSeedZBench}{0.53}
\newcommand{\threeReproBenchGapSD}{1.3}
\newcommand{\fourRpipe}{0.811}
\newcommand{\fourRpipeCI}{[0.798, 0.823]}
\newcommand{\fourRseed}{0.933}
\newcommand{\fourRseedCI}{[0.928, 0.937]}
\newcommand{\fourRatesWithin}{7}
\newcommand{\fourNsources}{7}
\newcommand{\tenRpipe}{0.949}
\newcommand{\tenRpipeCI}{[0.946, 0.953]}
\newcommand{\tenRseed}{0.922}
\newcommand{\tenRseedCI}{[0.916, 0.927]}
\newcommand{\tenRatesWithin}{6}
\newcommand{\tenNsources}{7}
\newcommand{\fourLtwoSftAiTwo}{4{,}955}
\newcommand{\fourLtwoSftSeeds}{42.6}
\newcommand{\tenLtwoSftAiTwo}{15.6}
\newcommand{\tenLtwoSftSeeds}{47.9}
\newcommand{\tenWjGap}{5.2}
\newcommand{\tenWjSeedDiff}{5.6}
\newcommand{\tenSftRatesMaxDiff}{0.8}
\newcommand{\tenSftRcorrAiTwo}{0.997}
\newcommand{\tenDpoRcorrAiTwo}{0.990}
\newcommand{\tenUpdSftPct}{0.013\,\%}
\newcommand{\tenUpdDpoPct}{0.3\,\%}
\newcommand{\twelveParentSwapWj}{4.2}
\newcommand{\twelveHarmfulWithin}{6}
\newcommand{\twelveHarmfulNsources}{6}
\newcommand{\sixRmzRawCI}{[0.927, 0.933]}
\newcommand{\sixRuRawCI}{[0.754, 0.779]}
\newcommand{\sixOneSeedHold}{16}
\newcommand{\sixOneSeedPairs}{18}
\newcommand{\sixOneSeedMissA}{0.0023}
\newcommand{\sixOneSeedMissB}{0.0006}
\newcommand{\fortyfourTransferMZPct}{75.0}

\newcommand{\fortyfourTransferUPct}{60.9}

\newcommand{\fortyfourTransferRPct}{58.7}

\newcommand{\fortyfiveFlipOverallPct}{23.4}

\newcommand{\seventyfourSeenMZTenTwentyPct}{2.2}
\newcommand{\seventyfourSeenMZTenTwentyPctCI}{[1.9\,\%, 2.6\,\%]}
\newcommand{\seventyfourSeenUTenTwentyPct}{25.3}
\newcommand{\seventyfourSeenUTenTwentyPctCI}{[24.5\,\%, 26.2\,\%]}
\newcommand{\seventyfourSeenRTenTwentyPct}{26.6}

\newcommand{\seventyfourSeenUTwentyPct}{8.9}

\newcommand{\seventyfourSeenMZTwentyPct}{0.05}

\newcommand{\seventysixRecipeAccPct}{48}

\newcommand{\seventyfourTopOneKeepMZPctCI}{[55\,\%, 79\,\%]}
\newcommand{\seventyfourTopOneKeepUPctCI}{[19\,\%, 35\,\%]}



\icmltitlerunning{Identically Trained Language Models Break on Different Jailbreaks}

\begin{document}

\twocolumn[
\icmltitle{Identically Trained Language Models Break on Different Jailbreaks}

\icmlsetsymbol{equal}{*}

\begin{icmlauthorlist}
  \icmlauthor{Chi-Wei Chang}{cerulean,ntu}
  \icmlauthor{Mu-Tien Kuo}{cerulean,nycu}
  \icmlauthor{Kuan-Wei Lee}{cerulean,nycu}
  \icmlauthor{Meng-Hsun Tsai}{nycu}
\end{icmlauthorlist}

\icmlaffiliation{cerulean}{Cerulean Labs}
\icmlaffiliation{ntu}{National Taiwan University, Taipei, Taiwan}
\icmlaffiliation{nycu}{National Yang Ming Chiao Tung University, Hsinchu, Taiwan}
\icmlcorrespondingauthor{Meng-Hsun Tsai}{tsaimh@nycu.edu.tw}
\icmlkeywords{safety, refusal, jailbreaks, random seeds, pretraining, twin study, reproducibility}

\vskip 0.3in
]

\printAffiliationsAndNotice{}

\begin{abstract}
Jailbreak benchmarks test one checkpoint and treat which jailbreaks break it as a property of its pretraining data and safety training.
We ask whether another pretraining run of the same data, given the same safety training, would break on the same jailbreaks.
For each of \eightNrecipes{} pretraining data recipes, we take 1B base models from 3 pretraining seeds and give each the same safety post-training with 2 post-training seeds, for \eightNmodels{} models in all.
We then collect \sixK{} responses from every model to a shared set of jailbreak prompts and correlate the per-prompt non-refusal rates between every pair of models.
On average, two post-trained copies of the same pretraining run have per-prompt non-refusal rates that correlate at \eightRmz{}.
A different pretraining seed lowers the correlation to \eightRu{}, and a different data recipe lowers it only slightly further, to \eightRr{}.
These results indicate that which jailbreaks break a checkpoint depends much more on its pretraining run than on its data, a pattern that holds in all \eightNrecipes{} data recipes and at 150M and 300M parameters.
In practice, the best of \fortyfourNtemplates{} attack templates found on one checkpoint stays the best after a new pretraining seed in only \seventyfourTopOneKeepUPct{}\,\% of model pairs, against \seventyfourTopOneKeepMZPct{}\,\% after a new post-training seed.
Each pretraining run also leaves a fingerprint: across \seventysixNprompts{} prompts, a model's per-prompt non-refusal rates identify its pretraining run among all \seventysixRunN{} runs every time, and a random subset of \seventysixFiftyN{} prompts still does so \seventysixRunFiftyPct{}\,\% of the time.
\end{abstract}

\section{Introduction}
\label{sec:intro}

\begin{figure*}[t]
  \centering
  \input{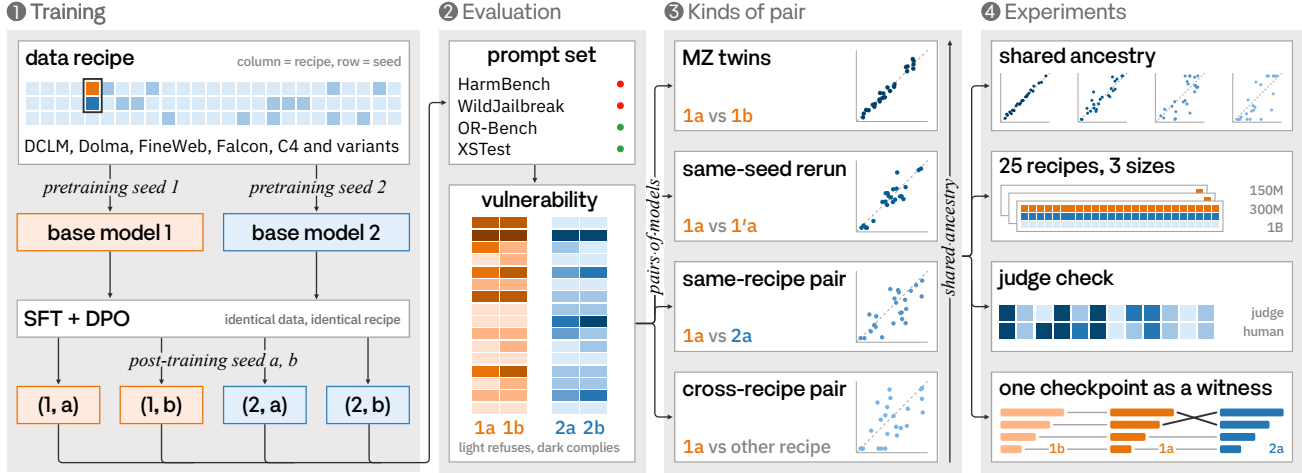}
  \caption{\textbf{Study design.}
  \emph{Training:} two pretraining seeds of one data recipe give two base models, and each base model is given the same safety post-training under two post-training seeds.
  \emph{Evaluation:} every model answers the same prompt set, and a strip shows one model's vulnerability, a square per prompt (light: the model refuses; dark: it complies).
  \emph{Kinds of pair,} ordered by the ancestry they share: MZ twins share a base model; a same-seed rerun repeats a released run's initialization and data order, so the two differ only in hardware and software; a same-recipe pair spans two runs of one data recipe, a cross-recipe pair two data recipes.
  \emph{Experiments:} we compare how closely each kind of pair agrees, prompt by prompt, across data recipes and model sizes, check the judge against human labels, and ask what the jailbreaks found on one checkpoint say about its pretraining run (Section~\ref{sec:setup}).}

\label{fig:design}
\end{figure*}

Deploying language models safely requires knowing whether their safeguards can withstand adversarial jailbreaks. Previous work approaches this on three fronts: benchmarks that track attack success rates \citep{chao2024jailbreakbench}, red teams that uncover safeguard-breaking prompts \citep{jiang2024wildteaming,hughes2024bon}, and mechanistic studies that identify refusal directions in model activations \citep{arditi2024refusaldirection}. These works each evaluate a single checkpoint yet treat which jailbreaks break it as a property of its pretraining data and safety training~\citep{mazeika2024harmbench,shen2025pandaguard}, a reading that overlooks the randomness of pretraining itself. Two training runs that differ only in their random seed can reach nearly the same aggregate accuracy yet disagree on individual predictions \citep{zhong2021instancelevel,damour2020underspecification} and on measured biases \citep{damour2020underspecification}, a concern MultiBERTs and PolyPythias raised for pretraining itself \citep{sellam2022multiberts,vanderwal2025polypythias}. Safety behavior also varies across seeds: fine-tuning seeds shift a model's harmfulness \citep{fraser2025finetuningsafety}, sampling seeds and temperatures flip refusal decisions \citep{larsen2025instability}. Furthermore, branches of one pretraining run that differ only in their last 1\,\% of tokens refuse differently after identical post-training \citep{lu2026finalwindow}. Each of these safety studies varies a seed after pretraining, or a final slice of one pretraining run, never a whole pretraining run. To our knowledge, no study has tested whether a second pretraining run of the same data, given the same safety training, would break on the same jailbreaks as the first.

To answer this, we start from DataDecide \citep{magnusson2025datadecide}, which pretrains a 1B-parameter model on each of \eightNrecipes{} data recipes with 3 pretraining seeds. We give each resulting base model the same safety post-training with 2 post-training seeds, yielding \eightNmodels{} models. We then collect \sixK{} responses from every model to each of \eightNprompts{} jailbreak prompts and compute the Pearson correlation of per-prompt non-refusal rates between every pair of models (\S\ref{sec:setup}). To separate a pretraining run from its data, we compare two kinds of pair (Figure~\ref{fig:design}): two models sharing one base model and differing only in their post-training seed are MZ twins, after monozygotic twins; two that share only a data recipe form a same-recipe pair. How much more closely MZ twins correlate than same-recipe pairs measures what sharing a pretraining run adds.

Our comparisons show that two post-trained copies of one pretraining run break on nearly the same jailbreaks, while two pretraining runs of the same data break on noticeably different ones. Changing only the post-training seed leaves per-prompt correlation between two models at \eightRmz{} on average over data recipes. Changing the pretraining seed first lowers it to \eightRu{}, and changing the data recipe as well lowers it only slightly further, to \eightRr{}. This ordering holds in all \eightNrecipesOrdered{} data recipes. We observe the same drop at 150M and 300M parameters and in author-labeled responses from four models (\S\ref{sec:grid}).
We also pretrain one of these runs twice more from its own seed on our hardware and software.
The new models correlate with the original and with each other barely more than a same-recipe pair does, which suggests that MZ twins agree because they share one trained run rather than its seed (\S\ref{sec:ladder}).
In practice, the attack template that works best on one checkpoint stays the best after a new pretraining seed in only \seventyfourTopOneKeepUPct{}\,\% of model pairs, against \seventyfourTopOneKeepMZPct{}\,\% after a new post-training seed. The prompts that break a model also identify its pretraining run: its per-prompt non-refusal rates pick out its MZ twin among the models of all \seventysixRunN{} pretraining runs every time, and a random subset of \seventysixFiftyN{} prompts still does so \seventysixRunFiftyPct{}\,\% of the time (\S\ref{sec:witness}). These results indicate that which jailbreaks break a checkpoint depends far more on its pretraining run than on its data.

Our contributions are as follows:
\begin{enumerate}
  \item We show that which jailbreaks break a model follows its pretraining run far more than its data, in every data recipe, at three sizes and under human labels.
  \item We show that what breaks one checkpoint is a signature of its pretraining run: the best attack template usually changes under a new pretraining seed, with a small set of prompts enough to identify the run even after new post-training (\S\ref{sec:witness}).
  \item We release our code on GitHub\footnote{\repoURL{}} and the weights of our safety-post-trained models at 150M, 300M and 1B, with per-prompt judge labels, on Hugging Face.\footnote{\hfURL{}}
\end{enumerate}

\section{Experimental setup}
\label{sec:setup}

This section defines the models, comparisons and measurements that the results rely on. We first describe the models: base models from DataDecide \citep{magnusson2025datadecide}, reruns of one of them on our hardware, and the same safety post-training for every base model. We then define which pairs of models we compare and on which prompts. Last, we explain how each model's responses are sampled and judged, and how two models' responses are correlated.

\paragraph{Base models.}
A \emph{data recipe} is a pretraining corpus as DataDecide releases it: a source dataset with its filtering, deduplication and mixing choices.
A \emph{pretraining seed} is the seed of one pretraining run, which in DataDecide sets initialization and data order together.
For each data recipe, DataDecide releases three 1B runs that are trained for the same 100B tokens and differ only in this seed (Table~\ref{tab:base-models}).
A \emph{base model} is the final checkpoint of one pretraining run.
We disable dropout, so the seed is the run's only controlled randomness. GPU nondeterminism can still affect training beyond the seed, and Section~\ref{sec:ladder} measures its effect.

Our main data recipe is DCLM-Baseline \citep{li2024datacomplm}, which makes up most of OLMo~2 1B's pretraining mix.
DataDecide's architecture is a Llama block, so we load every checkpoint as a Transformers \texttt{LlamaForCausalLM} \citep{wolf2020transformers} by renaming and splitting its tensors, and check at each size that the converted model's logits match OLMo's reference code (Appendix~\ref{app:pipeline}).
\paragraph{Our reruns.}
A pretraining seed fixes a run's initialization and data order, but not the hardware and software it runs on.
We pretrain base model 1 twice more with DataDecide's pretraining seed, on our H200 GPUs. Compared with each other, these two \emph{same-seed reruns} differ only by GPU nondeterminism, and compared with DataDecide's base model 1 they also differ in hardware and software.
To separate the seed's two variables, we also train two \emph{partial reruns}, each sharing only base model 1's initialization or only its data order (Table~\ref{tab:base-models}).
Each rerun uses DataDecide's code and data and starts from initial weights bit-identical to the released run it copies (Appendix~\ref{app:factorial}).
\paragraph{Post-training.}
Following OLMo~2 1B \citep{olmo2025olmo2}, we post-train every base model with T\"ulu~3 \citep{lambert2024tulu3}, using T\"ulu~3's published hyperparameters except for sequence handling (Appendix~\ref{app:pipeline}).
Safety data enters in supervised fine-tuning (SFT), through the WildGuardMix \citep{han2024wildguard}, WildJailbreak \citep{jiang2024wildteaming} and CoCoNot \citep{brahman2024coconot} parts of the T\"ulu~3 mixture.
We then run direct preference optimization \citep[DPO;][]{rafailov2023dpo} on the OLMo~2 1B preference mix and stop before OLMo~2's two reinforcement-learning stages, which add no safety data.
We post-train each base model with \gridNseeds{} \emph{post-training seeds}, or 4 on the main data recipe.
A post-training seed sets only the order of the training examples.
The same 4 seeds post-train our reruns and DataDecide's 150M and 300M runs of that data recipe.
\paragraph{Kinds of pair.}
We name each kind of pair by what the base models of its two models share.
\emph{MZ twins} share a pretraining run and its seed but differ in the post-training seed.
Like monozygotic twins, they share all they inherit.
A \emph{same-recipe pair} shares only a data recipe but differs in the pretraining seed, using released runs.

A \emph{cross-recipe pair} shares neither a data recipe nor a pretraining run.
Because MZ twins and same-recipe pairs differ in whether they share a pretraining run, their difference measures how much sharing a pretraining run raises the correlation.
Similarly, same-recipe and cross-recipe pairs differ in whether they share a data recipe, so their difference measures what the data recipe adds.
\paragraph{Prompt set.}
The prompt set draws from seven public sources: HarmBench's standard, contextual and human-jailbreak prompts \citep{mazeika2024harmbench}, SORRY-Bench \citep{xie2024sorrybench}, StrongREJECT \citep{souly2024strongreject}, XSTest \citep{rottger2024xstest}, OR-Bench-Hard-1K \citep{cui2024orbench} and WildJailbreak's evaluation set \citep{jiang2024wildteaming}.
Because T\"ulu~3's safety data includes jailbreaks collected in the wild, some benchmark prompts also appear in our post-training data.
We drop these under T\"ulu~3's 8-gram test, along with every prompt too long for the context window (Appendix~\ref{app:pipeline}).
The remaining \sixNpanel{} prompts hold \sixNplain{} plain harmful and \sixNbenign{} benign prompts beside our primary subset of \sixNadv{} adversarial-harmful prompts that wrap a harmful request in a jailbreak.
Among the adversarial-harmful, HarmBench's \cciHjbN{} human-jailbreak prompts each wrap one of \cciNbehaviours{} harmful requests in one of \cciNtemplates{} attack templates.
\paragraph{Sampling and judging.}
We sample \sixK{} responses from each model to every prompt with vLLM \citep{kwon2023vllm}, at temperature \sixTemperature{} and top-p 1, up to \sixMaxTokens{} new tokens.
Following T\"ulu~3 and OLMo~3 \citep{olmo2025olmo3}, we use WildGuard's refusal label as our judge \citep{han2024wildguard}, with Qwen3Guard-Gen-8B as a second judge \citep{zhao2025qwen3guard}.
A model's \emph{vulnerability} on a prompt is its non-refusal rate over coherent samples (Appendix~\ref{app:judge}), those that pass a rule-based check for repetition and prompt echoes \citep{antonelli2026tttsafety}.
A model with no coherent sample on a prompt has no vulnerability there; one adversarial-harmful prompt has none in one model, so analyses that need every model's full profile use the other \seventysixNprompts{}.

Its vulnerabilities on all prompts form its \emph{vulnerability profile}.
To check that the profiles do not depend on how samples are labeled, Appendix~\ref{app:judge} validates WildGuard against human labels, and Appendix~\ref{app:robustness} recomputes vulnerability four other ways.
\paragraph{Statistics.}
Since the question is whether two models break on the same prompts, we compare their vulnerability profiles by their Pearson correlation $r$.
Each vulnerability is estimated from only \sixK{} samples, so sampling noise keeps even two identical models from correlating at 1.
We estimate the correlation without this noise by Spearman's correction \citep{spearman1904attenuation}, which divides $r$ by $\sqrt{\rho_a\rho_b}$, and call the result the \emph{corrected correlation}.
Here each model's \emph{reliability} $\rho$ is the correlation between the vulnerability profiles from its odd- and even-indexed samples, stepped up to \sixK{} samples by the Spearman--Brown formula.
For each kind of pair, we report the mean corrected correlation: \rMZ{} for MZ twins, \rU{} for same-recipe pairs and \rR{} for cross-recipe pairs.
We call the difference \rMZ{}~$-$~\rU{} the \emph{ancestry gap}, since it measures what sharing a pretraining run adds.
Each mean and gap carries a 95\,\% bootstrap interval over prompts or over data recipes, reported in full in Table~\ref{tab:levels} and Appendices~\ref{app:judge} to~\ref{app:prereg}.
Each prediction and its reading rule were fixed before its experiment, separating tested predictions from exploratory findings (Appendix~\ref{app:prereg}).

\section{Results}\label{sec:results}
We ask whether two models that share a pretraining run break on the same jailbreak prompts more often than two models that share only its data.
Section~\ref{sec:grid} measures the ancestry gap in every data recipe and at three model sizes, then checks it against human labels.
Section~\ref{sec:ladder} reruns pretraining from one seed, in full and in part, to ask whether the gap follows the seed or the run, since GPU nondeterminism makes two runs from one seed differ.
Section~\ref{sec:witness} asks what the differences between pretraining runs mean for someone who evaluates a single checkpoint.

\subsection{Vulnerability across data recipes and model sizes}
\label{sec:grid}

\begin{table*}[t]
  \centering\scriptsize
  \caption{\label{tab:levels}Mean corrected correlation by kind of pair.
  The top row averages over all \eightNrecipes{} data recipes, with 95\,\% intervals over data recipes.
  The lower rows use the main data recipe's twelve models at each size, with 95\,\% intervals over prompts.}
  \setlength{\tabcolsep}{2.4pt}
\begin{tabular}{lllllll}
\toprule
\shortstack{Models}& \shortstack{MZ\\twins} & \shortstack{Same-recipe\\pairs} & \shortstack{Cross-recipe\\pairs} & \shortstack{Ancestry\\gap} & \shortstack{Same-recipe $-$\\cross-recipe} & \shortstack{Gap $\div$ (same-recipe\\$-$ cross-recipe)} \\
\midrule
All 25 data recipes, 1B & 0.984 [0.982, 0.985] & 0.811 [0.804, 0.817] & 0.779 [0.766, 0.788] & 0.173 [0.167, 0.179] & 0.032 [0.024, 0.043] & 5.4$\times$ [4.0, 7.2] \\
Main data recipe, 1B & 0.985 [0.983, 0.987] & 0.814 [0.801, 0.825] & --- & 0.171 [0.160, 0.183] & --- & --- \\
Main data recipe, 300M & 0.981 [0.978, 0.985] & 0.770 [0.756, 0.784] & --- & 0.211 [0.198, 0.224] & --- & --- \\
Main data recipe, 150M & 0.988 [0.983, 0.993] & 0.784 [0.769, 0.799] & --- & 0.204 [0.189, 0.218] & --- & --- \\
 
\bottomrule
\end{tabular}

\end{table*}

\begin{figure}[t]
  \centering
  \providecommand{\recipesunit}{288pt}%
\providecommand{\recipesrow}{7.6pt}%
\pgfplotstableread[col sep=tab]{figures/data/recipes_levels.tsv}\recipeslevels%
\pgfplotstablegetrowsof{\recipeslevels}%
\pgfmathtruncatemacro{\recipeslast}{\pgfplotsretval-1}%
\def\recipesmain{-10}%
\foreach \i in {0,...,\recipeslast}{%
  \pgfplotstablegetelem{\i}{main}\of\recipeslevels%
  \ifnum\pgfplotsretval=1 \xdef\recipesmain{\i}\fi%
}%
\input{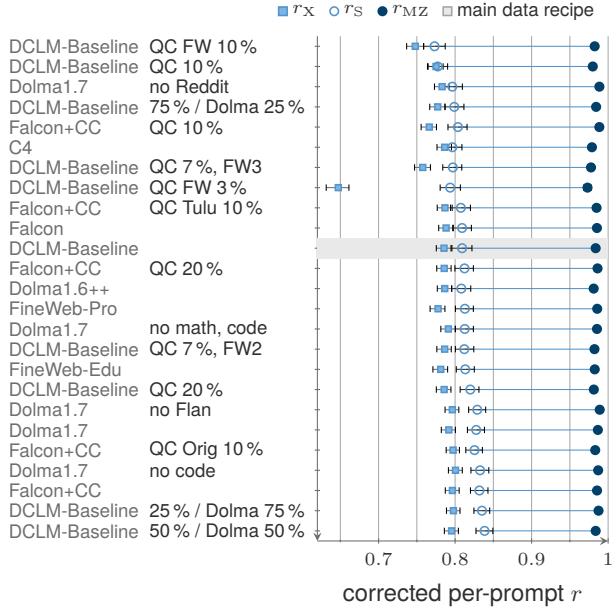}%
\begin{tikzpicture}
\begin{axis}[
  castor axis,
  name=recipes, scale only axis,
  width=0.38*\recipesunit, height=25*\recipesrow,
  y dir=reverse, ymin=-0.5, ymax=24.5,
  table/col sep=tab,
  ytick={0,...,24}, yticklabels={},
  ymajorgrids=false, xmajorgrids,
  xmin=0.62, xmax=1.0, xtick={0.7,0.8,0.9,1.0}, minor xtick={0.65,0.75,0.85,0.95}, xminorgrids,
  xticklabel style={/pgf/number format/fixed, /pgf/number format/precision=2},
  xlabel={corrected per-prompt $r$},
  legend style={at={(1,1.01)}, anchor=south east, legend columns=4,
    /tikz/every even column/.append style={column sep=3pt}},
]
\fill[band] (axis cs:0.62,\recipesmain-0.5) rectangle (axis cs:1.0,\recipesmain+0.5);
\castorrecipelabels{\recipeslevels}{y}
\addplot[quiver={u=\thisrow{rmz}-\thisrow{ru}, v=0}, draw=pair-unrelated, line width=0.4pt,
  forget plot] table[x=ru, y=y] {figures/data/recipes_levels.tsv};
\addplot[only marks, mark=square*, mark size=1.1pt, draw=pair-unrelated, fill=pair-cross,
  error bars/x dir=both, error bars/x explicit, error bars/error bar style={draw=muted}]
  table[x=rr, y=y, x error plus expr=\thisrow{rr_hi}-\thisrow{rr},
        x error minus expr=\thisrow{rr}-\thisrow{rr_lo}] {figures/data/recipes_levels.tsv};
\addlegendentry{$\rR$}
\addplot[only marks, mark=o, draw=pair-unrelated, mark options={fill=box},
  error bars/x dir=both, error bars/x explicit, error bars/error bar style={draw=text}]
  table[x=ru, y=y, x error plus expr=\thisrow{ru_hi}-\thisrow{ru},
        x error minus expr=\thisrow{ru}-\thisrow{ru_lo}] {figures/data/recipes_levels.tsv};
\addlegendentry{$\rU$}
\addplot[only marks, mark=*, draw=pair-mz, fill=pair-mz,
  error bars/x dir=both, error bars/x explicit, error bars/error bar style={draw=text}]
  table[x=rmz, y=y, x error plus expr=\thisrow{rmz_hi}-\thisrow{rmz},
        x error minus expr=\thisrow{rmz}-\thisrow{rmz_lo}] {figures/data/recipes_levels.tsv};
\addlegendentry{$\rMZ$}
\addlegendimage{only marks, mark=square*, mark size=3pt, draw=grid, fill=band}
\addlegendentry{main data recipe}
\end{axis}
\end{tikzpicture}
  \caption{\textbf{In each of the \eightNrecipes{} data recipes, the point estimates order MZ twins above same-recipe pairs and same-recipe pairs above cross-recipe pairs.}
  Each row is one data recipe, sorted by the ancestry gap with the largest at the top, and the main data recipe's row is on a band.
  A filled circle marks the mean corrected correlation of MZ twins, an open circle that of same-recipe pairs and a square that of the cross-recipe pairs with one model in the data recipe.
  The line joining the two circles is the data recipe's ancestry gap.
  Bars are 95\,\% intervals from a bootstrap over prompts.}
  \label{fig:grid}
\end{figure}

We first test whether sharing a pretraining run makes two models break on the same prompts beyond what sharing its data recipe already does.
Our grid holds \eightNmodels{} models: \eightNrecipes{} data recipes, 3 base models each, and post-training seeds 1 and 2 on every base model (Figure~\ref{fig:grid}, Table~\ref{tab:levels}).
Within each data recipe the ancestry gap is the mean corrected correlation of its \eightNmzPairsRecipe{} MZ twins minus that of its \eightNuPairsRecipe{} same-recipe pairs.
A gap of zero would mean that a pretraining run adds nothing beyond its data recipe.
MZ twins correlate at \eightRmz{} on average and same-recipe pairs at \eightRu{}, so sharing a base model raises the corrected correlation by \eightGapMean{}, with a 95\,\% interval over data recipes of \eightGapCIrecipes{}.
Taken one data recipe at a time, the ancestry gap lies between \eightGapRecipeMin{} and \eightGapRecipeMax{}, and its interval over prompts stays above zero in all \eightNrecipesGapCIaboveZero{}.
Cross-recipe pairs correlate at \eightRr{}, so a different data recipe on top of a different pretraining run lowers the correlation only slightly more.
Not sharing a base model lowers the correlation by \eightSeedOverRecipe{} times as much as not sharing a data recipe does.
This suggests that over these data recipes, whether two models break on the same prompts depends mainly on whether they share a pretraining run, and much less on whether they share a data recipe.

How often a model breaks, unlike which prompts break it, depends strongly on the data recipe.
The correlations above compare which prompts break two models, not how often each model is broken.
To see what makes that rate differ between models, we take each model's mean vulnerability on adversarial-harmful prompts, correct it for the judge's precision (Appendix~\ref{app:judge}), and split its variance across models into three percentages.
The data recipe accounts for \eightShareRecipeCorrPct{}\,\% \eightShareRecipeCorrPctCI{} of the variance, and the base model within its data recipe for \eightSharePretrainCorrPct{}\,\% \eightSharePretrainCorrPctCI{}.
The post-training seed accounts for the remaining \eightSharePosttrainCorrPct{}\,\% \eightSharePosttrainCorrPctCI{}; all three intervals resample data recipes.
This suggests that the data recipe, which adds little to which prompts break a model, matters for how often it breaks.

We next ask whether the prompts that break one model most often also break its partner in each kind of pair.
We find that the prompts that break one model most also break its MZ twin almost as often.
For each pair of models we take the first model's most effective tenth of prompts, the \eightTransferK{} prompts on which it breaks most often.
A prompt can land in this tenth partly by a lucky draw of samples, which would inflate the first model's own score on it.
We therefore choose the tenth on half of the first model's samples and score that model on the other half, a fresh draw in which the same luck cannot repeat \citep{kriegeskorte2009circular}.
We then divide the second model's mean vulnerability on these prompts by the first model's own, so a ratio of 1 means the second model breaks there just as often.
The ratio is \eightTransferMz{} for MZ twins, \eightTransferU{} for same-recipe pairs and \eightTransferR{} for cross-recipe pairs.
A model from another pretraining run of the same data recipe is thus broken there about \eightTransferUShortfall{}\% less often.
Changing the data recipe too lowers the ratio only slightly further, so most of the loss is there once the pretraining run changes.

All models tend to refuse the plainest harmful requests and break under the strongest jailbreak templates, so any two vulnerability profiles would agree to some degree whatever the models' ancestry.
HarmBench's \eightResidNprompts{} human-jailbreak prompts let us remove this shared part of the agreement, because every prompt wraps one of \eightResidNbehaviours{} harmful requests in one of \eightResidNtemplates{} jailbreak templates, and templates and requests alike recur across many prompts, so the effect of each component can be estimated on its own.
For every model, one least-squares fit estimates how strong a given template is and how hard a given request is to refuse.
Subtracting both from a prompt's vulnerability leaves the part that comes from its particular pairing of template and request.
Removing the two effects raises the ancestry gap on these prompts from \eightGapHjb{} to \eightGapResid{}.
This suggests that what MZ twins share goes beyond which templates and requests work (Appendix~\ref{app:robustness}).

Every model so far has 1B parameters, which leaves open whether the gap holds at other model sizes.
We repeat the comparison with the main data recipe's three 150M and three 300M pretraining runs, each post-trained with the four post-training seeds that the main data recipe's 1B base models have, where every other data recipe has two (Section~\ref{sec:setup}).
The ancestry gap is \sixteenGapOneFifty{} at 150M and \sixteenGapThreeHundred{} at 300M, against \sixteenGapOneB{} for the main data recipe's twelve 1B models (Table~\ref{tab:levels}, Figure~\ref{fig:grid-sizes}).
At every size MZ twins correlate between \sixteenRmzThreeHundred{} and \sixteenRmzOneFifty{}.
Including Qwen3Guard as the judge, the gap holds at every size we test, not only at 1B, with Appendix~\ref{app:robustness} reporting the intervals.
Appendix~\ref{app:prereg} lists the pre-registered predictions for the 25-recipe grid and these sizes, with what was observed.

Every model so far also learned to refuse from the same safety data, namely Tülu~3's, so the gap could be specific to that data.
To test this, we post-train the main data recipe's three 1B base models again with post-training seeds 1 and 2, replacing Tülu~3's safety data in SFT with as many rows of other public safety data and leaving DPO unchanged.
The replacement rows come from Nemotron-SFT-Safety-v2 \citep{nvidia2025nemotron3nano}, Llama-Nemotron's safety subset \citep{bercovich2025llamanemotron}, PKU-SafeRLHF \citep{ji2024pkusaferlhf}, BeaverTails \citep{ji2023beavertails} and the harmless split of HH-RLHF \citep{bai2022hhrlhf} (Appendix~\ref{app:robustness-safety-data}).
Under the second safety data the ancestry gap is \eightySwapGap{}, against \eightGapMain{} for the same base models and seeds under Tülu~3's safety data (Appendix~\ref{app:robustness}).
We observe the gap under other public safety data as well as under Tülu~3's.

We then audit the judge with labels the authors wrote on four models, since a judge error repeated in the near-identical responses of MZ twins would inflate their correlation.
The four models are two base models of the main data recipe at post-training seeds 1 and 2, so they form 2 MZ pairs and 4 same-recipe pairs.
Two authors labeled \twentythreeK{} samples from each model on \twentythreeNprompts{} adversarial-harmful prompts, for \twentythreeNitems{} responses in all.
Each author labeled every response alone, blind to the judge, the model and the other author's labels.
Their Cohen's $\kappa$ on refusal was \twentythreeKappaRefusal{}, close to the Fleiss $\kappa$ of 0.72 that WildGuard's three annotators reached on refusal \citep{han2024wildguard}.
With two annotators there is no majority vote, so the authors resolved each disagreement by discussion, blind to judge and model, before any comparison with the judge.
On both label sets we take the mean Pearson correlation of MZ twins minus that of same-recipe pairs, left uncorrected because every model has the same three samples.
The author labels give a difference of \twentythreeGapHuman{} \twentythreeGapHumanCI{} over prompts, and the judge gives \twentythreeGapJudge{} on the same responses.
Each author's own labels, before adjudication, give a positive difference as well, \twentythreeGapHarry{} and \twentythreeGapCoauthor{} (Appendix~\ref{app:judge}).
This suggests that on these four models the difference comes from the responses rather than the judge.

\subsection{Same-seed and partial reruns}
\label{sec:ladder}

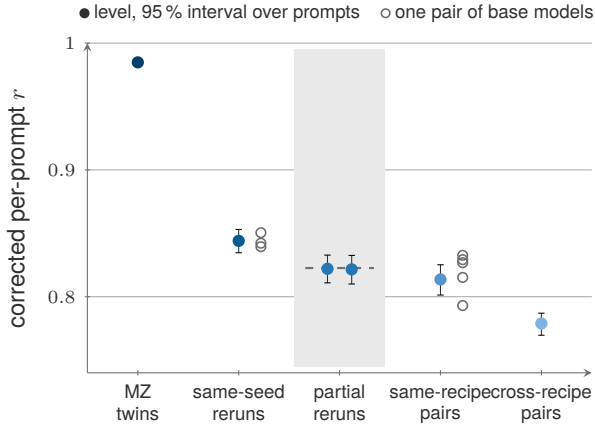
\begin{figure}[t]
  \centering
  \begin{tikzpicture}
\begin{axis}[
  castor axis,
  height=0.72\columnwidth,
  ymin=0.74, ymax=1.0,
  xmin=0.5, xmax=5.5,
  xtick={1,2,3,4,5},
  xticklabels={MZ\\twins, same-seed\\reruns, partial\\reruns, same-recipe\\pairs, cross-recipe\\pairs},
  xticklabel style={align=center},
  ylabel={corrected per-prompt $r$},
  legend style={at={(0.5,1.03)}, anchor=south, legend columns=2, /tikz/every even column/.append style={column sep=6pt}},
]
\addlegendimage{castor level, fill=text}
\addlegendentry{level, 95\,\% interval over prompts}
\addlegendimage{castor basepair}
\addlegendentry{one pair of base models}
\fill[band] (axis cs:2.55,0.745) rectangle (axis cs:3.45,0.995);
\addplot[muted, dashed, thick, no markers, forget plot]
  table[x=x, y=r, col sep=tab] {figures/data/ladder_strip.tsv};
\foreach \rung/\role in {MZ/pair-mz, rung2/pair-rerun, init/pair-partial, order/pair-partial, U/pair-unrelated, R/pair-cross}{
  \edef\plotcmd{\noexpand\addplot[castor level, fill=\role, draw=\role, castor keep={rung}{\rung}, forget plot]}
  \plotcmd
    table[x=x, y=r, col sep=tab,
          y error plus expr=\thisrow{hi}-\thisrow{r},
          y error minus expr=\thisrow{r}-\thisrow{lo}] {figures/data/ladder_levels.tsv};
}
\addplot[castor basepair, forget plot]
  table[x expr=\thisrow{x}+0.22, y=r, col sep=tab] {figures/data/ladder_pairs.tsv};
\end{axis}
\end{tikzpicture}
  \caption{\textbf{Same-seed reruns sit nearer same-recipe pairs than MZ twins.}
  Filled marks are mean corrected correlations on the \eightNprompts{} adversarial-harmful prompts, with 95\,\% intervals over prompts. MZ twins and same-recipe pairs come from the twelve 1B models of the main data recipe, over \aZeroNmz{} and \aZeroNu{} pairs. The same-seed mark averages three pairs of base models, each rerun with base model 1 and the two reruns together, sixteen model pairs each.
  The cross-recipe mark averages all \eightNrPairs{} cross-recipe pairs of the 25-recipe grid, whose models include six of the twelve. The gray band holds the partial reruns, sharing only the initialization on the left and only the data order on the right. Read them against the band's dashed segment, its own pairs that share neither, not against the same-recipe pairs beside it (Appendix~\ref{app:factorial}). Each open mark is one pair of base models. At same-recipe pairs, three pair released base models, and two pair the first rerun with base model 2 or 3, so those two are not same-recipe pairs.}
  \label{fig:ladder}
\end{figure}

Every pair of runs in Section~\ref{sec:grid} differs in its pretraining seed, so the gap there cannot tell us whether the seed alone makes MZ twins break on the same prompts.
We refer to a run as one pretraining of a base model from start to finish, so two runs from one seed share its initialization and data order but not the GPU nondeterminism along the way.
Our two same-seed reruns test this. Each is a new run from base model 1's pretraining seed, trained on our hardware and software. Figure~\ref{fig:ladder} compares the reruns with twelve 1B models, the main data recipe's three released base models with four post-training seeds each.
Among these twelve models, MZ twins reach a mean corrected correlation of \aZeroRmz{} and same-recipe pairs reach \aZeroRu{}. Together with base model 1, the two reruns give three runs from one pretraining seed and so three pairs of base models. Pairing each of one base model's four post-training seeds with each of the other's gives 16 pairs of models per pair of base models, 48 in all.
The same-seed mean over these 48 is \thirtyfiveRungTwoMean{} \thirtyfiveRungTwoMeanCI{}, which is \thirtyfiveRungTwoMeanMinusU{} \thirtyfiveRungTwoMeanMinusUCI{} above the same-recipe pairs. It lies much nearer the same-recipe pairs than the MZ twins, and its interval covers neither. Pair by pair, the first rerun reaches \twentysevenBpDefRep{} with base model 1, the second rerun reaches \thirtyfiveBpRtwoD{} and the two reruns reach \thirtyfiveBpRRtwo{} with each other. All three lie above the three pairs of released base models, which range from \twentysevenReleasedPairMin{} to \twentysevenReleasedPairMax{}. Pretraining a run again from its own seed thus gives models that correlate with the original run's models, and with each other, only slightly more than a same-recipe pair does.

A DataDecide pretraining seed sets both the initialization and the data order, so we next ask whether either one alone makes two runs' models correlate more closely.
We post-train base models 1 and 2 and the two partial reruns with four post-training seeds each, giving the sixteen models in the gray band of Figure~\ref{fig:ladder}. Each partial rerun takes one factor from base model 1 and the other from base model 2. Pairs that share neither reach \nineNeither{}, against \nineInit{} for pairs that share only the initialization and \nineOrder{} for pairs that share only the data order. Measured against sharing neither, the initialization changes the correlation by \nineInitMinusNeither{} \nineInitMinusNeitherCI{} and the data order by \nineOrderMinusNeither{} \nineOrderMinusNeitherCI{}, and the two differ from each other by \nineInitMinusOrder{} \nineInitMinusOrderCI{}. All three intervals cover zero, so in these sixteen models sharing only the initialization or only the data order adds no correlation we can detect.

Every pair of partial reruns that shares one factor compares a run pretrained on DataDecide's hardware and software with one pretrained on ours, so a drop from that switch alone could hide a small share of either factor (Appendix~\ref{app:factorial}). Our two same-seed reruns test that switch directly, as sharing our hardware and software lets their pair separate the switch from GPU nondeterminism. We compare how closely the two reruns correlate with each other against how closely the first rerun correlates with base model 1, which DataDecide pretrained. The two reruns correlate with each other \thirtyfiveRRminusRD{} \thirtyfiveRRminusRDCI{} more closely than the first rerun does with base model 1. That interval covers zero, so moving hardware and software costs no correlation we can detect.

Together, the reruns suggest that what a new pretraining seed changes in Section~\ref{sec:grid} is the particular run trained from it.
Rerunning the same seed gives models that sit much nearer same-recipe pairs than MZ twins. Sharing only the initialization or only the data order adds no correlation we can detect, and neither does sharing hardware and software. The ancestry gap thus measures what sharing one particular run adds, and a new seed is one way to train a different run.
Every interval in this section bootstraps over prompts, so it shows how far a number would move with other prompts, not with other base models.

\subsection{One checkpoint as a witness}
\label{sec:witness}

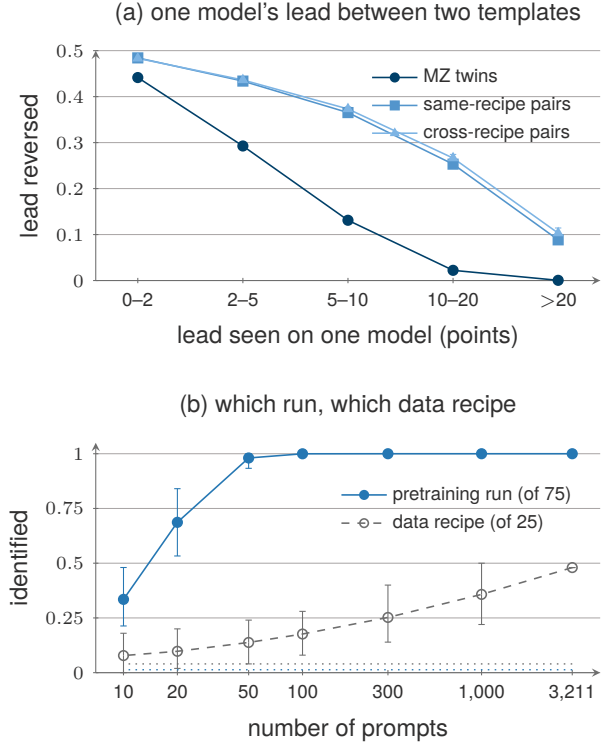
\begin{figure}[t]
  \centering
  \begin{tikzpicture}
\begin{axis}[castor axis, name=a, width=\columnwidth, height=0.56\columnwidth,
  xmin=0.6, xmax=5.4, xtick={1,2,3,4,5},
  xticklabels={0--2, 2--5, 5--10, 10--20, $>$20},
  ymin=0, ymax=0.5, ytick={0,0.1,0.2,0.3,0.4,0.5},
  yticklabel style={/pgf/number format/fixed, /pgf/number format/precision=1},
  xlabel={lead seen on one model (points)},
  ylabel={lead reversed},
  title={(a) one model's lead between two templates}, title style={font=\castorlabel, text=text},
  legend style={at={(0.97,0.97)}, anchor=north east}]
\foreach \pair/\role/\mk in {MZ/pair-mz/*, U/pair-unrelated/square*, R/pair-cross/triangle*}{
  \edef\plotcmd{\noexpand\addplot[\role, mark=\mk, mark size=1.8pt, castor keep={pair}{\pair},
    error bars/y dir=both, error bars/y explicit, error bars/error bar style={draw=\role}]}
  \plotcmd table[x=x, y=value, col sep=tab,
    y error plus expr=\thisrow{hi}-\thisrow{value},
    y error minus expr=\thisrow{value}-\thisrow{lo}] {figures/data/leaderboard_observed.tsv};
}
\legend{MZ twins, same-recipe pairs, cross-recipe pairs}
\end{axis}
\begin{axis}[castor axis, at={(a.south west)}, anchor=north west, yshift=-0.26\columnwidth,
  width=\columnwidth, height=0.56\columnwidth,
  xmode=log, xmin=7, xmax=4600, ymin=0, ymax=1.05, ytick={0,0.25,0.5,0.75,1},
  yticklabel style={/pgf/number format/fixed},
  xtick={10,20,50,100,300,1000,3211},
  xticklabels={10, 20, 50, 100, 300, 1{,}000, 3{,}211},
  log ticks with fixed point,
  xlabel={number of prompts},
  ylabel={identified},
  title={(b) which run, which data recipe}, title style={font=\castorlabel, text=text},
  legend style={at={(0.97,0.86)}, anchor=north east}]
\addplot[series-main, mark=*, mark size=1.8pt,
  error bars/.cd, y dir=both, y explicit]
  table[x=n_prompts, y=mean, col sep=tab,
    y error plus expr=\thisrow{hi}-\thisrow{mean},
    y error minus expr=\thisrow{mean}-\thisrow{lo}] {figures/data/fingerprint_run.tsv};
\addlegendentry{pretraining run (of 75)}
\addplot[series-aux, dashed, mark=o, mark size=1.8pt, mark options={solid, fill=box},
  error bars/.cd, y dir=both, y explicit, error bar style={solid}]
  table[x=n_prompts, y=mean, col sep=tab,
    y error plus expr=\thisrow{hi}-\thisrow{mean},
    y error minus expr=\thisrow{mean}-\thisrow{lo}] {figures/data/fingerprint_recipe.tsv};
\addlegendentry{data recipe (of 25)}
\addplot[series-main, dotted, line width=0.5pt, forget plot]
  table[x=n_prompts, y=chance, col sep=tab] {figures/data/fingerprint_run.tsv};
\addplot[series-aux, dotted, line width=0.5pt, forget plot]
  table[x=n_prompts, y=chance, col sep=tab] {figures/data/fingerprint_recipe.tsv};
\end{axis}
\end{tikzpicture}
  \caption{\textbf{A lead reverses far more often across runs than across MZ twins; the profile identifies the run far more often than the data recipe.}
  Panel a: share of leads reversed on the other model, by kind of pair, binned after the data by the lead on one model; a tie counts one half; 95\,\% intervals over data recipes.
  Panel b: accuracy against number of prompts, log axis: run among \seventysixRunN{}, filled; data recipe of an unseen base model among \eightNrecipes{}, open; means over \seventysixNsub{} random subsets, bars their 2.5--97.5\,\% band; dotted lines, chance.}
  \label{fig:witness}
\end{figure}

We now ask what the differences between pretraining runs mean for someone who evaluates a single checkpoint.
Whether they compare data recipes or attacks, part of what they measure may belong to the checkpoint's pretraining run rather than to its data recipe.
The first question is which attack works best, and the attack template that works best on one checkpoint often loses its lead on another run of the same data recipe, far more often than on its MZ twin.
The second is which prompts break the checkpoint, and these are specific enough to identify its own pretraining run.

\paragraph{Carry-over and the order of data recipes}
We first ask whether a prompt that breaks one model also breaks another, and whether that depends on what the two models share.
We call a prompt reliably vulnerable for a model when the model is vulnerable on it at least nine times in ten, a high bar we fixed in advance.
A prompt that is reliably vulnerable for one model is also reliably vulnerable for its MZ twin \fortyfourTransferMZPct{}\,\% of the time.
For a same-recipe partner this falls to \fortyfourTransferUPct{}\,\%, and for a cross-recipe partner to \fortyfourTransferRPct{}\,\%.
Most of the drop therefore comes with a new pretraining run, and changing the data recipe as well adds little to it.

The pretraining run also matters when a study asks only how often a model is broken.
A comparison of pretraining data might ask which of two data recipes gives models with the higher mean vulnerability, and answer it with one model from each.
DataDecide scores its small-scale experiments by the same test, which it calls decision accuracy: the share of data recipe pairs ordered as the mean over three seeds orders them \citep{magnusson2025datadecide}.
For every pair of data recipes, we check whether taking one of its models gives the same order as all six of their models' means, three base models with two post-training seeds each.
One model from each data recipe gives the opposite order in \fortyfiveFlipOverallPct{}\,\% of these comparisons.
A single model estimates its data recipe's mean vulnerability with a standard error of \fortyfiveSEOneOne{} percentage points, as our judge labels it.
That error is about as large as the differences being measured, since the data recipes' means themselves have a standard deviation of \fortyfiveSDRecipe{} points.
With three base models and one post-training seed each, the error falls to \fortyfiveSEThreeOne{} points.
A difference of a few points between two data recipes, each measured on one model, can therefore come from the models rather than from the data.

\paragraph{The attack ranking}
To see whether a ranking of attacks carries over, we compare the attack template that works best on each model of a pair.
Two pretraining runs of one data recipe usually disagree on that template, while MZ twins mostly agree.
A template's attack success on a model is its mean vulnerability over the template's own 5 to 24 prompts, so it mixes the difficulty of the template with that of its prompts.
The template with the highest success rate survives a new post-training seed in \seventyfourTopOneKeepMZPct{}\,\% \seventyfourTopOneKeepMZPctCI{} of pairs but another pretraining run of the same data recipe in only \seventyfourTopOneKeepUPct{}\,\% \seventyfourTopOneKeepUPctCI{}, with intervals over data recipes (Appendix~\ref{app:prereg}).
MZ twins share their pretraining run, so their third of disagreements comes from the post-training seed and sampling noise alone.
The further disagreement in same-recipe pairs comes with the new pretraining run.

Figure~\ref{fig:witness}a asks the same question of any two templates rather than only the top one.
On one model, a template's lead over another is the difference between their attack success, in percentage points.
We then check whether the partner model keeps the same template ahead.
Grouping leads by the size seen on one model, a reading we chose after seeing the data, we find that a lead of \seventyfourBinLo{} to \seventyfourBinHi{} points is reversed by an MZ twin only \seventyfourSeenMZTenTwentyPct{}\,\% \seventyfourSeenMZTenTwentyPctCI{} of the time.
A same-recipe partner reverses it \seventyfourSeenUTenTwentyPct{}\,\% \seventyfourSeenUTenTwentyPctCI{} of the time and a cross-recipe partner \seventyfourSeenRTenTwentyPct{}\,\%, with intervals over data recipes.
Even a lead above \seventyfourBinHi{} points is not safe across runs.
A same-recipe partner reverses \seventyfourSeenUTwentyPct{}\,\% of such leads, while an MZ twin reverses only \seventyfourSeenMZTwentyPct{}\,\%.
This suggests that a ranking of attacks measured on one checkpoint carries noise from its pretraining run, which one checkpoint cannot measure.

\paragraph{The profile identifies the run}
A model's vulnerability on each prompt, which we call its vulnerability profile, is distinctive enough to identify its pretraining run.
Earlier work identifies a model from its behavior across versions and recovers the base model behind a fine-tune \citep{pasquini2024llmmap,foley2023matchingpairs}.
Here, many candidates share a data recipe and differ only in their pretraining run.
For each of the \seventysixRunN{} runs, its model at post-training seed 1 serves as the reference and its model at seed 2 as the query.
We assign each query to the reference whose profile correlates with its own most closely.
The profiles cover \seventysixNprompts{} of the \sixNadv{} adversarial-harmful prompts, leaving out the one prompt to which some model gave no coherent sample.
This matching finds the right run for all \seventysixRunN{} queries, against one in \seventysixRunN{} for a random guess.
With \seventysixFiftyN{} prompts drawn at random, the match finds the right run \seventysixRunFiftyPct{}\,\% of the time on average over \seventysixNsub{} random subsets (Figure~\ref{fig:witness}b).
Asked for the data recipe instead, the same matching gives a much weaker signal.
Each data recipe's reference averages four models from two of its base models.
Its queries are the models of the third base model, which the reference never saw, \seventysixRecipeN{} queries in all.
The match finds the right data recipe for \seventysixRecipeCorrect{} of the \seventysixRecipeN{} (\seventysixRecipeAccPct{}\,\%), against one in 25 for a random guess.
The data recipe reaches only about one in two with all \seventysixNprompts{} prompts, despite a reference averaged over four models where a run's reference is a single model.
We show this only across a new post-training seed under one post-training recipe, not across a different fine-tune.
The prompts that break a model are a mark of its pretraining run far more than of its data recipe.

\section{Related work}
\label{sec:related}

\paragraph{Training randomness.} Models differing only in a training seed reach near-equal accuracy yet disagree on individual predictions \citep{zhong2021instancelevel,damour2020underspecification} and generalize differently out of distribution \citep{mccoy2020bertsfeather,juneja2023linearconnectivity}. BERT encoder suites such as MultiBERTs nest seeds of task fine-tuning under pretraining seeds \citep{sellam2022multiberts}. In these suites, a new pretraining seed adds less disagreement on task predictions than a new seed of task fine-tuning \citep{zhong2021instancelevel}. Measures of bias, however, shift with the pretraining seed \citep{damour2020underspecification,sellam2022multiberts}. In decoder language models, PolyPythias varies this seed across many pretraining runs on one corpus and evaluates the base models \citep{vanderwal2025polypythias}, with analyses of such suites following how seeds diverge over training \citep{madaan2024benchmarkvariance,fehlauer2025seedconvergence}. None of these studies includes a model trained to refuse.

\paragraph{Variance of safety behavior.} A model's aggregate harmfulness shifts with the seed of benign fine-tuning \citep{fraser2025finetuningsafety}, and sampling randomness flips its safety decision across sampling seeds and temperatures on about a quarter of prompts, on average \citep{larsen2025instability}. Prompts nonetheless rank alike in jailbreak difficulty across attack runs on one model and across distinct models, though less so for one defended model \citep{hughes2024bon}. Pretraining also matters: branches of one run differing only in final data refuse alike after SFT but differently after identical post-training \citep{lu2026finalwindow}. Outside safety, cognitive biases likewise follow the base model more than the instruction data \citep{itzhak2025planted}. Telling the pretraining run apart from its data recipe takes two runs of one data recipe, which DataDecide \citep{magnusson2025datadecide} releases at 1B, given the same post-training.

\paragraph{The refusal mechanism within and across models.} Studies of refusal locate it in low-dimensional structure and find that structure in each model separately. In one account it is a single direction in activation space \citep{arditi2024refusaldirection}. Others describe cones of a few dimensions \citep{wollschlager2025conecones} or a refusal signal encoded apart from harm \citep{zhao2025harmrefusalseparate}. Work that stays inside one pretraining run follows this structure through post-training stages \citep{labunets2026refusalprefixes} and through pretraining checkpoints \citep{reblitzrichardson2026refusalslice}. These works have not settled whether the structure exists before post-training \citep{arditi2024refusaldirection,labunets2026refusalprefixes,reblitzrichardson2026refusalslice}. Work that crosses distinct models carries refusal into other architectures by a rotation \citep{bosco2026refusalbeyondattention} and between two model sizes by a map \citep{oozeer2025activationtransfer}. Across pretraining seeds rather than sizes, \citet{mccann2026polymorphism} compare features and steering vectors among nine Pythia-70m base models, though not refusal. Refusal has thus been compared within one run or across models of different size, architecture or family \citep{reblitzrichardson2026refusalslice}, never across runs of one data recipe. We make that comparison in Appendix~\ref{app:mechanism}.

\paragraph{Noise correction and fixed-seed reruns.} Noise in two measures attenuates the correlation between them, and dividing by the square root of the product of their reliabilities recovers it \citep{spearman1904attenuation}. Applied to models, the correction has been used to compare a model with human ratings or brain recordings, with model reliability taken from readout heads retrained on one pretrained network \citep{depaolini2026nonidentifiability} or assumed perfect \citep{tan2026urbanscene}. Between two models, error consistency corrects only for chance agreement \citep{geirhos2020errorconsistency}. A second line of work finds that even runs with every seed fixed diverge in image classifiers and small or fine-tuned language models, through nondeterministic kernels or one tiny early perturbation \citep{summers2021nondeterminism,zhuang2022tooling,bhojanapalli2021reproducibility,altintas2025butterfly}. Neither line of work corrects a comparison of two models for the noise of measuring each from sampled responses.

\section{Discussion and limitations}
\label{sec:limitations}

Two pretraining runs of the same data, given the same safety training, still break on noticeably different jailbreaks.
Two models built on one base model correlate at \eightRmz{} on average over data recipes.
Two models built on different runs of the same data recipe correlate at only \eightRu{}, close to the \eightRr{} of models trained on different data.
This ordering appears in all \eightNrecipesOrdered{} data recipes.
The part of a profile that differs between runs belongs to the pretraining run, and one checkpoint cannot show how large that part is.
Our results indicate that a jailbreak found on one checkpoint is evidence about one run, and only partly about the data behind it.
This suggests that a comparison of pretraining data and a jailbreak evaluation each read better from more than one run.

A comparison of pretraining data usually asks only how often a data recipe's models break, and one model per data recipe answers even that question poorly.
Compare two data recipes through one model each, and the ranking of the two by mean vulnerability is reversed in \fortyfiveFlipOverallPct{}\,\% of comparisons, against the ranking from each data recipe's six models taken together (Section~\ref{sec:witness}).
Such reversals come from how far a single model can sit from its data recipe's mean.
On one model, a data recipe's mean vulnerability typically lies \fortyfiveSEOneOne{} points from its average over many models, on the judge's scale.
Three base models with one post-training seed each bring that distance down to \fortyfiveSEThreeOne{} points.
This suggests that three base models give a clearer comparison of mean vulnerability on these data recipes than one model does.

A jailbreak evaluation reads the same checkpoint for a different purpose, recording which attack template works best and which prompts break the model.
Both answers move when the pretraining run changes.
The best attack template survives a new post-training seed in \seventyfourTopOneKeepMZPct{}\,\% of pairs but a new pretraining seed in only \seventyfourTopOneKeepUPct{}\,\%.
Which prompts break a checkpoint is also particular to its run.
A model from a second post-training seed, matched by its vulnerability profile, picks out its own pretraining run among all \seventysixRunN{} in \seventysixRunCorrect{} of \seventysixRunN{} queries.
A random set of \seventysixFiftyN{} prompts still finds the right run \seventysixRunFiftyPct{}\,\% of the time (Section~\ref{sec:witness}).
Single prompts show the pattern too.
A prompt that breaks a model at least nine times in ten does so on its MZ twin \fortyfourTransferMZPct{}\,\% of the time but on another run of its data recipe only \fortyfourTransferUPct{}\,\%.
This suggests that the attacks and prompts found on one checkpoint speak for its pretraining run and not only for its data.

Our design has two limitations.
The first is that every model comes from DataDecide's family of pretraining runs and from the SFT and DPO stages of T\"ulu~3.
Within that family we vary the corpus over 25 data recipes that differ in source, filtering and deduplication.
The ancestry gap's interval over prompts excludes zero in every one, which suggests that the gap does not hinge on any single corpus.
We also checked the safety data, and on one data recipe at 1B the gap reappears when a second safety set replaces T\"ulu~3's safety set (Section~\ref{sec:grid}, Appendix~\ref{app:robustness}).
The second is that WildGuard's refusal label, our primary judge, misses refusals on jailbreak prompts.
It puts mean vulnerability at \eightLevelRaw{}, and an audit of its labels adjusts it to \eightLevelCorr{} \eightLevelCorrCI{} over audit items (Appendix~\ref{app:judge}).
Human labels on four models of one data recipe still give an uncorrected difference of \twentythreeGapHuman{} \twentythreeGapHumanCI{} over prompts, an interval above zero (Section~\ref{sec:grid}).

Our largest models have 1B parameters, a size set by what the comparison needs.
It needs several full pretraining runs of one data recipe given the same post-training, and DataDecide releases three such runs for each data recipe at 1B.
At larger sizes each run costs much more.
\citet{madaan2024benchmarkvariance} pretrained ten 7B runs that differed only in their initialization seed, and each took about 15,000 A100-hours.
Short of that cost, the ancestry gap is positive on one data recipe at 150M, 300M and 1B (Section~\ref{sec:grid}).
Our released code and per-prompt judge labels facilitate future work that measures the ancestry gap in larger models.

\section*{Acknowledgements}

The successful completion of this research was made possible by the academic resources and advanced
research infrastructure provided by the National Center for High-Performance Computing, National
Institutes of Applied Research (NIAR), Taiwan. We gratefully acknowledge their invaluable support.

\section*{Impact statement}
\label{sec:impact}

We release our code on GitHub\footnote{\repoURL{}} and the weights of the 150M, 300M and 1B models we post-trained, with their per-prompt judge labels, on Hugging Face.\footnote{\hfURL{}}
The release leaves out the generated responses, and its judge labels name each prompt by id and source rather than by its text.
Each prompt comes from a published benchmark or dataset (Section~\ref{sec:setup}), and we wrote no new jailbreak or harmful request.
Our models are trained from public parts alone, DataDecide's pretraining runs and data and Ai2's OLMo 2 post-training data, so anyone with the compute could train such models without our release.

Jailbreak evaluations usually test one checkpoint, yet among our models a jailbreak found to break one checkpoint may not break a model from another pretraining run of the same data recipe (Section~\ref{sec:grid}).
A new pretraining run of the same data may therefore need its own red-teaming, and where a data recipe has several pretraining runs, a red-teaming result may be better read across them than on one checkpoint (Section~\ref{sec:limitations}).

\bibliography{refs}
\bibliographystyle{icml2026}

\newpage
\appendix
\counterwithin{table}{section}
\counterwithin{figure}{section}
\onecolumn
\FloatBarrier
\section{Post-training batches and compute}
\label{app:compute}
Every training job ran on nodes of eight H200 GPUs.
We measure compute in \emph{H200-hours}, a job's elapsed time in Slurm times the number of GPUs allocated to it.
Generation and judging, which follow post-training, used one GPU per model.
Each 1B pretraining run and each post-training run, SFT followed by DPO, ran on one node and was data-parallel over its eight GPUs.
Data parallelism leaves open the \emph{sharding setting}: which of the optimizer states, gradients and parameters each GPU stores only a share of.
DeepSpeed offers ZeRO's three cumulative sharding settings, ZeRO-1 to ZeRO-3, which shard the optimizer states, then the gradients, and finally the parameters \citep{rajbhandari2020zero}.
Post-training ran in two batches. The \emph{first post-training batch} ran our post-training recipe (Section~\ref{sec:setup}) with the DeepSpeed files that open-instruct ships, which run SFT under ZeRO-3 and DPO under ZeRO-2.
The \emph{second post-training batch} ran SFT under ZeRO-1.
The sharding setting of the SFT stage is the only difference between the two batches.
We checked that open-instruct's argument lists are identical between the two batches for both stages, and that the DPO stage's DeepSpeed file has the same hash in both.
ZeRO-1 and ZeRO-3 share one arithmetic, which we read from the code of DeepSpeed 0.15.4 before any run: bf16 weights with fp32 optimizer copies, and gradients accumulated and communicated in bf16.
The two settings differ in the order and grouping of those sums.
The effect of that difference on the models is reported in Appendix~\ref{app:pipeline}, whose checks use models of both batches.

Table~\ref{tab:compute} lists the models each batch trained.
The first batch trained twelve models from the main data recipe's three base models with post-training seeds 1 to 4, and the two partial reruns' eight.
Those twelve are the models of Appendix~\ref{app:mechanism} and the first part of Appendix~\ref{app:robustness}.
They are also those of the human-labeled check in Section~\ref{sec:grid}, the first audit and the human labels in Appendix~\ref{app:judge}.
The second batch trained every model built for the 25-recipe grid, 156 in all: its 150, plus post-training seeds 3 and 4 on the main data recipe's three base models.
The 25-recipe grid's twelve models on the main data recipe repeat the first batch's twelve pairs of base model and post-training seed, but are separate models, trained again.
Beyond the grid, the second batch also trained each same-seed rerun's four models and the 24 models of 150M and 300M.

Once the second batch's sharding setting was adopted, for the reasons given below, every model we started used it.
The partial reruns' eight models stayed in the first batch as a closed set, compared in the main text only with first-batch models.
Each comparison in the main text takes all its values from one batch.
The partial reruns' strip in Figure~\ref{fig:ladder} is the one exception in placement: it sits beside second-batch means, labeled as first batch and read only against its own first-batch means.
Two comparisons in the appendices mix the batches, and each is labeled as mixed: the mean corrected correlations in Appendix~\ref{app:nondet} and in Appendix~\ref{app:factorial}.

The second batch's sharding setting was chosen to cut the communication of the first batch's SFT stage.
Under ZeRO-3, each GPU gathers the parameters in the forward pass and again in the backward pass, and the gradients are reduce-scattered \citep{rajbhandari2020zero}.
We profiled that communication over 50 SFT steps under the first batch's settings on eight GPUs with Nsight Systems, averaging over GPUs.
The shares of time that follow overlap one another.
Communication kernels were active for \twentyfourProfNccl{} of the total time, and communication with no compute kernel running alongside it took \twentyfourProfExposed{}.
Compute kernels ran for \twentyfourProfCompute{}, and the GPUs sat idle for \twentyfourProfIdle{}.
To cut that communication, we tried candidates that change only the DeepSpeed file or the environment, leaving the batch size, sequence length, loss, data order and open-instruct untouched.
The candidates were ZeRO-2, ZeRO-1, ZeRO-3 with tuned bucket sizes and parameter persistence, and two environment variables, one for NCCL and one for CUDA.
Among other ways to cut the communication, we did not try plain data parallelism, Accelerate without DeepSpeed, because it accumulates gradients in fp32 where DeepSpeed under ZeRO uses bf16, a different arithmetic.
Each candidate ran 50 SFT steps and was compared with two first-batch runs of the same model with the same post-training seed, which we call the reference runs.
A run lies inside the \emph{loss check} when its loss at every step lies within 1.25 times the largest per-step difference between the reference runs, measured against each of them.
The check rests on that one pair.
No precedent sets such a tolerance, so we chose it from first principles: two runs that share model and seed set the spread a change must stay inside.
Relative to the first batch's settings, the step rate was \twentyfourCandSpeedZeroTwo{} under ZeRO-2, \twentyfourCandSpeedZeroOne{} under ZeRO-1, \twentyfourCandSpeedZeroThreeTuned{} under tuned ZeRO-3 and \twentyfourCandSpeedEnv{} under the environment variables.
All candidates but ZeRO-2 lay inside the loss check, and ZeRO-2 reached \twentyfourCandZeroTwoDev{} where the check allowed \twentyfourCandCheck{}.
We chose ZeRO-1, the fastest candidate inside the check.
Inspected only afterwards, ZeRO-2's losses sat above the reference runs on 44 of 48 steps, by 0.37 on average.
ZeRO-2 remains the DPO stage's sharding setting, as in Ai2's post-training recipe, and the SFT loss check says nothing about it.

Before the full runs, we registered a rule: adopt ZeRO-1 for SFT if it trained at least 1.3$\times$ faster and its models stayed within pre-registered tolerances of the first batch's.
We measured the speed-up on a whole model, SFT and DPO, with one full run under each batch's settings of base model 1 with post-training seed 1, on different nodes.
Under the second batch's settings, SFT ran \twentyfourSpeedupSft{} faster and DPO at \twentyfourSpeedupDpo{}, so the whole model trained with a speed-up of \twentyfourSpeedup{}.
Two 50-step first-batch SFT runs on different nodes agree in step time to 1.4\,\%.
This 1.4\,\% spread suggests that the node contributes little to the speed-ups.
The full run under the first batch's settings also took 8:15~h, against 8:13~h for the same command run earlier.
The speed-up is consistent with ZeRO-1 communicating once per optimizer step rather than at each of the eight micro-steps.
Appendix~\ref{app:pipeline} compares the two batches' models, and one part of that comparison came within its tolerance only at the edge.
We adopted ZeRO-1 for SFT, and Appendix~\ref{app:prereg} lists each prediction with its observed value.

Sharing the GPUs lowers the cost per model by another route: it leaves each run unchanged.
We define \emph{GPU sharing} as running two training jobs at once on the same eight GPUs, by time-slicing or under NVIDIA's Multi-Process Service (MPS).
MPS splits each GPU's compute units between the processes, which share its memory.
We tried three forms: time-slicing under the second batch's settings, MPS under the first batch's and MPS under the second's.
On one A100, running two training jobs at once raised throughput up to fourfold for small workloads, less for medium ones and little for workloads that fill the GPU \citep{robroek2022collocation}.
Each form ran two copies of the 50-step SFT run.
We registered that a form would go on to a comparison of its models with the second batch's, like that of Appendix~\ref{app:pipeline}, only under two conditions.
Its projected cost per model, training plus generation and judging, had to fall below 0.7 times the second batch's, \twentyfourTwinAllIn{}~\HtwoH{} measured on the model of the speed-up comparison.
That limit is \twentyfiveBound{}~\HtwoH{}.
Every copy under the form also had to lie inside the loss check.
A form's \emph{gain} is the steps per second its two copies completed together, counted only while both ran, divided by the steps per second of one job alone on the same eight GPUs.
We predicted a gain of at most 1.1$\times$ from time-slicing and observed \twentyfiveTsZone{}.
For MPS under the first batch's settings, we predicted 1.5 to 1.9$\times$ and observed \twentyfiveMpsCur{}, just above the predicted range.
For MPS under the second batch's settings, we predicted a smaller gain than under the first's and observed \twentyfiveMpsZone{}.
We predicted every copy inside the loss check, and the copies outside it were \twentyfiveScreenFail{} of \twentyfiveScreenN{}.
Both first-batch copies were among them, although the first batch's runs defined the check.
Each form had at least one copy outside the check.
Across all six copies, mean loss offsets from the reference runs ranged from $-0.02$ to $+0.11$, against ZeRO-2's $+0.37$.
With one pair of reference runs, we cannot tell whether sharing moves training or the check is narrow.
No form qualified, so every model trained alone on its node.

Two cases outside the registered rule also project above the cost each would need to beat.
The first, MPS under the second batch's settings with DPO run unshared, projects to 26.4~\HtwoH{} per model, above the \twentyfiveBound{}~\HtwoH{} limit.
We did not measure DPO under sharing.
The second, MPS under the first batch's settings, projects its SFT stage alone to 27.9~\HtwoH{} per model, already above the second batch's SFT without sharing, \twentyfourCostAsft{}~\HtwoH{}.

Table~\ref{tab:compute} also gives the H200-hours of every model set the paper reports, in the unit defined at the start of this appendix.
Its hours include jobs that did not complete, which the table also shows on their own.
Training one 1B model, base model 1 with post-training seed 1 retrained for the speed-up comparison, took \twentyfourCostB{}~\HtwoH{} under the first batch's settings.
The main data recipe's twelve first-batch models took from 65.7 to 66.8~\HtwoH{} each.
Under the second batch's settings, the same model's training took \twentyfourCostA{}~\HtwoH{}, with a mean of \eightCostTrainCell{} over the 25-recipe grid.
Generation and judging of that model under the second batch's settings took \twentyfourGenJudge{}~\HtwoH{}.
Per model, generation and judging averaged 1.83~\HtwoH{} over the main data recipe's twelve first-batch models, from 1.75 to 1.89, and \eightCostGenJudgeCellCompleted{} over the 25-recipe grid's completed jobs.
Each 150M model cost \sixteenCostTwinOneFifty{}~\HtwoH{} in all, and each 300M model \sixteenCostTwinThreeHundred{}.
Pretraining costs more than post-training: a 1B pretraining rerun, about four days on one node run in two-day segments, took about \sixtyoneCostPretrainRerunMax{}~\HtwoH{}.
Summed, the 25-recipe grid took \eightCostTotal{}~\HtwoH{}, and the 150M and 300M models took \computeSmallModelsTotal{}.
In all, with checks and test runs included, this paper took \computeTotal{}~\HtwoH{}.

\begin{table}[t]
  \centering\footnotesize
  \caption{\label{tab:compute}\textbf{The work behind this paper took \computeTotal{} H200-hours.}
  Each row counts every job of its model set, whatever its outcome.
  Two columns give the part spent on generation and judging and the part in jobs that did not complete.
  A planned two-day pretraining segment that reached its time limit and was resumed counts as completed.}
  \setlength{\tabcolsep}{4pt}
\begin{tabular}{>{\raggedright\arraybackslash}p{6.2cm}rlrrr}
\toprule
Model set & Models & Batch & H200-hours & \shortstack{Of which generation\\and judging} & \shortstack{Of which jobs that\\did not complete} \\
\midrule
1B pretraining reruns & 3 &  & 2{,}230.4 &  & 0.0 \\
main data recipe & 12 & first & 813.6 & 21.9 & 0.0 \\
partial reruns' post-trained models & 8 & first & 546.1 & 20.5 & 6.1 \\
25-recipe grid & 156 & second & 4{,}969.9 & 303.5 & 16.5 \\
first same-seed rerun's post-trained models & 4 & second & 126.9 & 7.5 & 0.0 \\
150M and 300M models & 24 & second & 445.6 & 51.0 & 0.1 \\
second same-seed rerun: pretraining and post-trained models & 4 & second & 873.2 & 7.5 & 79.3 \\
main data recipe, other safety data & 12 & second & 380.6 & 22.0 & 0.1 \\
projection transplant &  &  & 36.4 &  & 3.5 \\
projection transplant, replication &  &  & 23.3 &  & 0.0 \\
checks of App.~\ref{app:pipeline} and test runs of this appendix &  & both & 595.2 &  & 224.3 \\
other jobs: pretraining checks, evaluations, analyses, test runs &  &  & 355.3 &  & 31.5 \\
\textbf{total} &  &  & 11{,}396.6 &  & 361.5 \\
 
\bottomrule
\end{tabular}

\end{table}

\FloatBarrier
\section{Same-seed reruns: reproduction and nondeterminism}
\label{app:nondet}
Appendix~\ref{app:compute} ends on post-training runs that are not bit-identical, though they share a base model, a post-training seed and a sharding setting.
The post-training seed sets only the data shuffle and the order of the sampler.
GPU nondeterminism lies outside the seed.
Neither the backward pass of attention nor the reductions across GPUs are seeded, and no determinism setting is on.
In short test runs, the losses of two such runs agree for the first two steps and separate from step 3.
How far apart the post-trained models end is reported in Appendix~\ref{app:pipeline}.
This appendix covers our rebuilt pretraining, then where two same-seed runs part, then the two same-seed reruns after post-training.

\subsection{Rebuilding DataDecide's pretraining}

DataDecide's checkpoints ship only the converted model's settings and no training settings.
We rebuilt the training settings with OLMo's ladder script, at the last commit before it dropped the 150M and 1B sizes.
The training library is unchanged at the later commit.
The script takes the training arguments Ai2 used for DataDecide, recovered from Ai2's training scripts.
The only changes set site paths and the run name and switch off remote saving and run logging.
Each run lasts as many steps as its released run, 38{,}157 at 150M and 69{,}369 at 1B, without a change to the settings.
Two inputs cannot match the released runs: the GPU type, H200 in our runs, and the exact library versions inside Ai2's software image.

The initialization, unlike the software image, can be checked before any training.
For each released 150M and 1B run of the main data recipe, one pretraining seed reproduces its step-0 weights tensor for tensor, with a largest difference of 0.
We call this comparison the \emph{step-0 test}.
Every other pretraining seed we tested, the other released runs' seeds and one more at 150M, matches only the normalization weights, which are all ones.
The step-0 test covers the initialization only, and Appendix~\ref{app:factorial} checks the data order.

Our registered check of the rebuilt pretraining compared our 150M reproduction run with the released 150M run that has base model 1's pretraining seed.
We predicted that our loss curve would track the released run's evaluated perplexity within GPU-nondeterminism noise.
The rule's tolerance was the spread between two released evaluations of one checkpoint.
Ai2 evaluated no checkpoint twice, so the rule's 1\,\% fallback applied.
Every one of the eleven held-out sets then had to lie within 1\,\% of the released run's logged value at every evaluated step.
Against that 1\,\% tolerance, the largest difference is \threeOneAMax{}, on one set at step \threeOneAWorstStep{}, and the largest at the final step is \threeOneAFinal{}.

To diagnose differences above the 1\,\% tolerance, we trained our second 150M run for 7{,}500 steps with the same pretraining seed, code and settings as the reproduction run.
Our two runs differ from each other by the same order as the reproduction run differs from the released run.
The largest per-set difference at each step over steps 1{,}250 to 7{,}500 is \threeRepTwoMin{}--\threeRepTwoMax{} between our two runs, against \threeOneASameStepsMin{}--\threeOneASameStepsMax{} between the reproduction run and the released run.
At step 1{,}250 we also compare weights in \emph{relative $L_2$}, the $L_2$ distance between two weight vectors divided by the norm of the reference run's weights.
Our two 150M runs are \threeRelLtwoRepTwo{} apart, and the reproduction run and the released run are \threeRelLtwoReleased{} apart.
For comparison at that step, our reproduction run and the released runs of other pretraining seeds are \threeRelLtwoOtherSeeds{} apart.

The registered consequence of a difference above the tolerance was to stop and diagnose before any 1B run.
We went on to 1B on three grounds: exact step-0 weights, an exact data order, and our second 150M run differing from the reproduction run by the same order.
Two grounds come from the diagnosis, which was post hoc and not registered (Appendix~\ref{app:prereg}).

A second registered check predicted that a 150M run with a pretraining seed DataDecide did not release lies within two standard deviations of the released runs' mean.
The standard deviation is pooled over the 150M released runs of the 25 data recipes, three each, for 50 degrees of freedom.
Both distances scale the standard deviation as the registered rule does.
On final held-out loss, the equal-weight mean over the eleven held-out sets, the new run's distance is \threeNewSeedZLoss{}.
The average of DataDecide's ten benchmarks puts it at \threeNewSeedZBench{}.
Appendix~\ref{app:factorial} applies the rule at 1B.
Outside the registered checks, our 150M reproduction run's benchmark average differs from the released run of its seed by \threeReproBenchGapSD{} pooled standard deviations, unscaled.
\subsection{Where two same-seed runs part}

\begin{figure}[t]
\centering
\textbf{(a)}\par
\begin{tikzpicture}
\begin{axis}[
  castor axis,
  width=0.6\textwidth, height=0.32\textwidth,
  xmode=log, ymode=log, xmin=0.8, xmax=130, ymin=1e-6, ymax=1e-1,
  xlabel={pretraining step}, ylabel={relative $L_2$ between reruns},
  legend style={at={(0.03,0.97)}, anchor=north west},
]
\addplot[series-main, mark=*] table[x=step, y=rel_l2, col sep=tab] {figures/data/probe1_same.tsv};
\addlegendentry{same node}
\addplot[series-aux, dashed, mark=square*] table[x=step, y=rel_l2, col sep=tab] {figures/data/probe1_cross.tsv};
\addlegendentry{cross node}
\end{axis}
\end{tikzpicture}
\par\medskip
\textbf{(b)}\par
\begin{tikzpicture}
\begin{axis}[
  castor axis,
  width=0.6\textwidth, height=0.32\textwidth,
  xmin=0, xmax=70000, ymin=0.8, ymax=1.0,
  xlabel={pretraining step}, ylabel={same-seed / different-seed rel.\ $L_2$},
  scaled x ticks=false, xticklabel style={/pgf/number format/fixed},
]
\addplot[series-main, mark=*, mark size=1.2pt] table[x=step, y=ratio, col sep=tab] {figures/data/divergence.tsv};
\end{axis}
\end{tikzpicture}
\par\medskip
\textbf{(c)}\par
\begin{tikzpicture}
\begin{axis}[
  castor axis,
  width=0.6\textwidth, height=0.32\textwidth,
  xmin=0, xmax=70000, ymin=0.8, ymax=1.0,
  xlabel={pretraining step}, ylabel={same-seed / different-seed rel.\ $L_2$},
  scaled x ticks=false, xticklabel style={/pgf/number format/fixed},
  legend style={at={(0.97,0.03)}, anchor=south east},
]
\addplot[series-main, mark=*, mark size=1.2pt] table[x=step, y=r2r1, col sep=tab] {figures/data/divergence_rtwo.tsv};
\addlegendentry{from the first same-seed rerun}
\addplot[series-aux, dashed, mark=square*, mark size=1.2pt] table[x=step, y=r2d, col sep=tab] {figures/data/divergence_rtwo.tsv};
\addlegendentry{from base model 1}
\end{axis}
\end{tikzpicture}
\caption{\textbf{Same-seed runs part from step \twentyfiveFirstStep{}, and the first same-seed rerun ends at \elevenDivRatioEnd{} of the relative $L_2$ between two released runs of different pretraining seeds.}
(a)~Relative $L_2$ between same-seed 1B runs, steps 1--100, on one node and across nodes, log scale.
(b)~The first same-seed rerun's relative $L_2$ from base model 1 as that share, every 2{,}500 steps from step 2{,}500 to 69{,}369.
(c)~The second same-seed rerun's relative $L_2$ from the first same-seed rerun and from base model 1 as that share, at the same steps.
\label{fig:divergence}}
\end{figure}
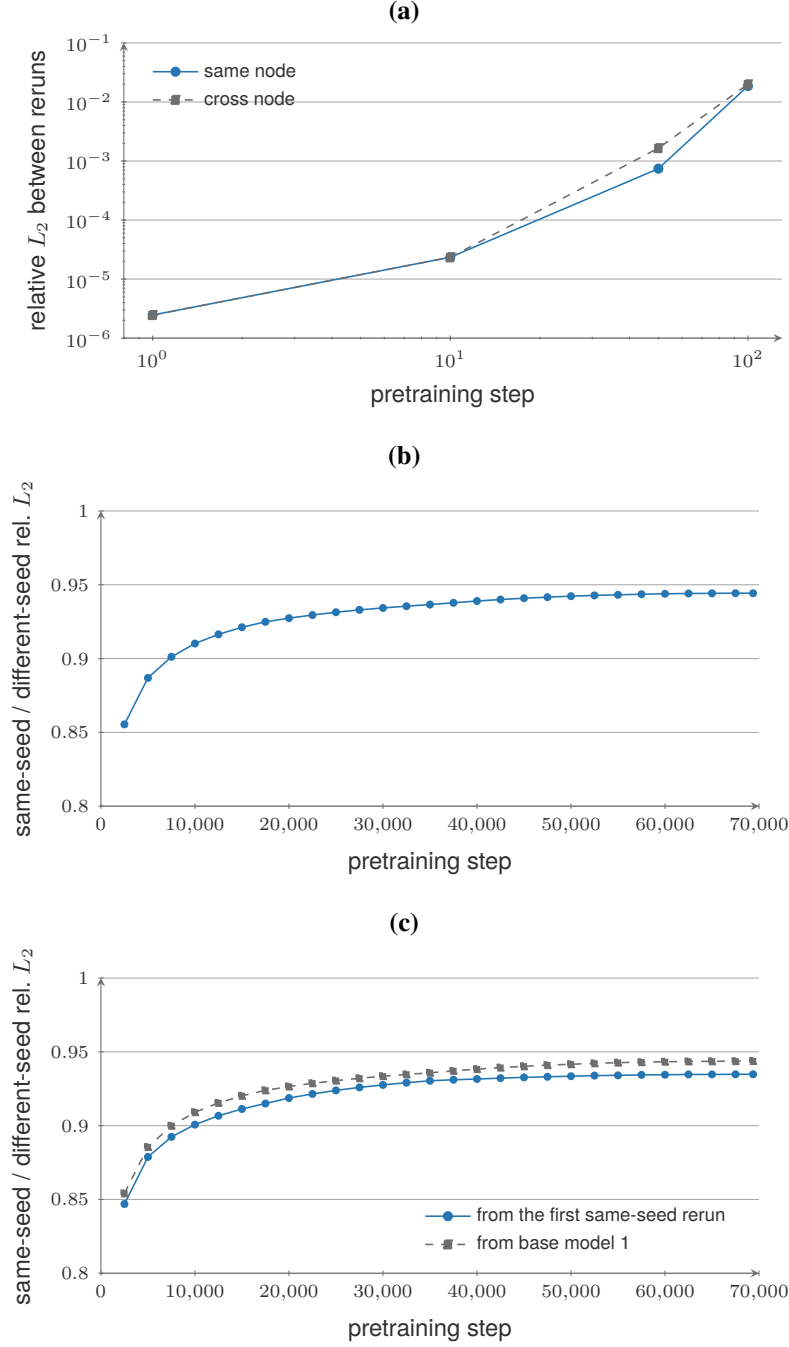

Two pretraining runs with one pretraining seed are not bit-identical, as with the post-training runs that open this appendix.
A first-divergence probe measured where two such runs first part.
The probe trained three 1B runs with one pretraining seed, code and settings, starting from base model 1's step-0 weights.
Two of them ran on one node, and the third on another node with the same GPU model.
We compared the runs step by step, and Figure~\ref{fig:divergence}a plots their relative $L_2$.

Three registered predictions concern the probe (Appendix~\ref{app:prereg}).
The first predicted a difference by step 10, and the runs first differ at step \twentyfiveFirstStep{}.
The second predicted a relative $L_2$ between $10^{-4}$ and $10^{-1}$ at step 100.
At step 100, we find \twentyfiveRelLtwoHundredSame{} on one node and \twentyfiveRelLtwoHundredCross{} across nodes.
The cross-node and same-node distances were predicted to lie within a factor of three of each other, and their ratio at step 100 is \twentyfiveCrossOverSame{}.
A second cross-node pair gives \twentyfiveCrossOverSameB{}.

At step \twentyfiveFirstStep{}, \twentyfiveUnequalStepOneMin{}--\twentyfiveUnequalStepOneMax{} of the \twentyfiveNtensors{} weight tensors differ, at a relative $L_2$ of \twentyfiveRelLtwoStepOne{}.
The tensors still equal at that step are normalization weights of the last blocks.
The log prints the loss and the gradient norm to four digits, and both part later.
The logged loss first differs at step \twentyfiveLossDiffMin{} to \twentyfiveLossDiffMax{}, depending on the pair.
The gradient norm first differs at step \twentyfiveGradDiffMin{} to \twentyfiveGradDiffMax{}.
The split of the weights at step \twentyfiveFirstStep{} suggests that the difference enters in the first forward and backward pass.
One candidate in that pass is the attention kernel, whose backward pass accumulates in nondeterministic order with no determinism setting on.
The probe does not isolate the kernel.
The first-step split is consistent with two results on image classifiers.
Two image-classifier runs on GPU diverge in weights within 100 steps when every other factor is fixed \citep{bhojanapalli2021reproducibility}.
In otherwise deterministic training, a one-bit change to one initial weight gives about as much run-to-run spread in final performance as any other source of nondeterminism \citep{summers2021nondeterminism}.

Beyond the probe's 100 steps, the first same-seed rerun shows how far the parting goes over a full 1B run.
It is the same-seed rerun of base model 1 that Section~\ref{sec:ladder} post-trains.
Its weight distance from base model 1 was a registered secondary measure, without a prediction.
Figure~\ref{fig:divergence}b plots its relative $L_2$ from base model 1 as a share of the relative $L_2$ between two released runs of different pretraining seeds.
The share is \elevenDivRatioEarly{} at step \elevenDivStepEarly{} and \elevenDivRatioMid{} at step \elevenDivStepMid{}.
At the final step, \elevenDivStepEnd{}, the two values are \elevenDivRelLtwoEnd{} and \elevenDivSeedRelLtwoEnd{}, a share of \elevenDivRatioEnd{}.
Its weights end at a cosine of \elevenDivCosEnd{} with base model 1's, while two released runs of different pretraining seeds end at \elevenDivSeedCosEnd{}.
It thus shares a small direction with base model 1.

\subsection{The first same-seed rerun after post-training}

The first same-seed rerun enters Section~\ref{sec:ladder} through one pair of base models.
Table~\ref{tab:twentyseven-pairs} takes four base models, the three released runs and the first same-seed rerun, each with four post-trained models under the second sharding setting (Appendix~\ref{app:compute}).
For every pair of base models, the table gives the mean corrected correlation and its min--max over the model pairs, a range and not an interval.
The same-seed pair, base model 1 with the first same-seed rerun, is the pair that Section~\ref{sec:ladder} reports.
We added two analyses after the first same-seed rerun's first data (Appendix~\ref{app:prereg}).
They are the display by pair of base models, and the comparison of the same-seed pair with the first same-seed rerun's pairs with base models 2 and 3.
The three same-recipe pairs of base models, 1 and 2, 1 and 3, and 2 and 3, reach \twentysevenBpDefAuxTwo{}, \twentysevenBpDefAuxThree{} and \twentysevenBpAuxTwoAuxThree{}.
The first same-seed rerun reaches \twentysevenBpRepAuxTwo{} with base model 2 and \twentysevenBpRepAuxThree{} with base model 3.
The MZ twins of the four base models range from \twentysevenBpAuxThree{} to \twentysevenBpAuxTwo{}.

The first same-seed rerun's pairs with base models 2 and 3 have a mean of \twentysevenRepOther{}, with a prompt interval of \twentysevenRepOtherCI{}.
The same-seed pair sits above that mean by \twentysevenRungTwoMinusRepOther{}, with a prompt interval of \twentysevenRungTwoMinusRepOtherCI{}.
These prompt intervals carry no sampling of base models, since DataDecide released only three of them.
As point estimates, both of the first same-seed rerun's pairs with base models 2 and 3 sit above the same-recipe-pair mean of \twentysevenRu{}.

The first same-seed rerun's four models were first compared with the models of base model 1 under the first sharding setting, while their own post-training used the second.
Before the first same-seed rerun's data, the comparison moved to the models of base model 1 under the second sharding setting.
Its prediction was registered again for the second sharding setting (Appendix~\ref{app:prereg}).
For the comparison across the two sharding settings, we had predicted that the same-seed pair would sit strictly between the same-recipe-pair mean and the MZ twins' mean, nearer the same-recipe-pair mean.
Its interval was to cover neither mean, which under the first sharding setting are \elevenRu{} for same-recipe pairs and \elevenRmz{} for MZ twins.
The same-seed pair reaches \elevenRungTwo{} with its interval \elevenRungTwoCI{}, between the two means and nearer the same-recipe-pair one.
This value across the two sharding settings stands beside the one-setting value of Section~\ref{sec:ladder} and is not replaced by it.
The shift between the two sharding settings is reported in Appendix~\ref{app:pipeline}.

Table~\ref{tab:twentyseven-secondary} extends both comparisons to other outcomes and subsets.
The prediction concerns WildGuard's refusal label on adversarial-harmful prompts.
Plain-harmful prompts, benign prompts and WildGuard's harmfulness label were registered secondaries of the comparison across the two sharding settings, without a prediction, and are repeated, unregistered, under the second sharding setting.
The other outcomes are descriptive.
On adversarial-harmful prompts, each of the five outcomes puts same-recipe pairs lowest, then the first same-seed rerun with base models 2 and 3, then the same-seed pair, then MZ twins.
The order is the same with Qwen3Guard as the judge and with the judges agreeing.
On plain-harmful prompts, the same-seed pair reaches \twentysevenPlainRungTwo{} against \twentysevenPlainRu{} for same-recipe pairs, with no separation.

Benign prompts run the other way.
There the same-seed pair reaches \twentysevenBenignRungTwo{} with a prompt interval of \twentysevenBenignRungTwoCI{}, below the same-recipe-pair mean of \twentysevenBenignRu{}.
The first same-seed rerun's pairs with base models 2 and 3 sit below that mean too, at \twentysevenBenignRepOther{}.
Across the two sharding settings, the same-seed pair reaches \elevenBenignRungTwo{} on benign prompts, against \elevenBenignRu{} for same-recipe pairs.

\begin{table*}[t]
\centering\footnotesize
\caption{\textbf{As point estimates, both of the first same-seed rerun's pairs with base models 2 and 3 sit above the same-recipe-pair mean.}
Means and min--max run over model pairs, under the second sharding setting.
The columns for the first sharding setting repeat the earlier comparison, in which pairs with the first same-seed rerun mix sharding settings.
Min--max ranges are not intervals.
Prompt intervals use \twentysevenNboot{} resamples of the prompts.
\label{tab:twentyseven-pairs}}
\setlength{\tabcolsep}{3pt}
\begin{tabular}{>{\raggedright\arraybackslash}p{3.6cm}>{\raggedright\arraybackslash}p{3.7cm}rrlrl}
\toprule
\begin{tabular}[b]{@{}l@{}}Pair of\\base models\end{tabular} & Kind of pair & \begin{tabular}[b]{@{}c@{}}Model\\pairs\end{tabular} & \begin{tabular}[b]{@{}c@{}}Mean corrected\\correlation\end{tabular} & Min--max & \begin{tabular}[b]{@{}c@{}}First setting\\mean\end{tabular} & \begin{tabular}[b]{@{}c@{}}First setting\\min--max\end{tabular} \\
\midrule
1 $\times$ first same-seed rerun & same-seed pair & 16 & 0.839 & [0.830, 0.850] & 0.840 & [0.831, 0.852] \\
1 $\times$ second same-seed rerun & same-seed pair & 16 & 0.842 & [0.836, 0.850] & -- &  \\
first same-seed rerun $\times$ second same-seed rerun & same-seed pair & 16 & 0.851 & [0.832, 0.861] & -- &  \\
1 $\times$ 2 & same-recipe pair & 16 & 0.833 & [0.817, 0.842] & 0.830 & [0.813, 0.844] \\
1 $\times$ 3 & same-recipe pair & 16 & 0.815 & [0.799, 0.826] & 0.814 & [0.803, 0.826] \\
2 $\times$ 3 & same-recipe pair & 16 & 0.793 & [0.769, 0.812] & 0.790 & [0.773, 0.807] \\
2 $\times$ first same-seed rerun & first same-seed rerun with base model 2 or 3 & 16 & 0.827 & [0.815, 0.843] & 0.821 & [0.808, 0.838] \\
3 $\times$ first same-seed rerun & first same-seed rerun with base model 2 or 3 & 16 & 0.829 & [0.816, 0.840] & 0.830 & [0.815, 0.838] \\
2 $\times$ second same-seed rerun & second same-seed rerun with base model 2 or 3 & 16 & 0.826 & [0.804, 0.845] & -- &  \\
3 $\times$ second same-seed rerun & second same-seed rerun with base model 2 or 3 & 16 & 0.814 & [0.807, 0.821] & -- &  \\
1 $\times$ 1 & MZ twins & 6 & 0.985 & [0.983, 0.989] & 0.984 & [0.983, 0.986] \\
2 $\times$ 2 & MZ twins & 6 & 0.989 & [0.985, 0.993] & 0.986 & [0.985, 0.988] \\
3 $\times$ 3 & MZ twins & 6 & 0.981 & [0.973, 0.987] & 0.983 & [0.976, 0.989] \\
first same-seed rerun $\times$ first same-seed rerun & MZ twins & 6 & 0.982 & [0.972, 0.988] & 0.982 & [0.972, 0.988] \\
second same-seed rerun $\times$ second same-seed rerun & MZ twins & 6 & 0.979 & [0.962, 0.989] & -- &  \\
 
\bottomrule
\end{tabular}
\end{table*}

\begin{table*}[t]
\centering\footnotesize\setlength{\tabcolsep}{2.2pt}
\caption{\textbf{Adversarial-harmful outcomes order the pairs alike, and benign prompts put the same-seed pair below same-recipe pairs.}
Position in the ancestry gap: 0 at the same-recipe-pair mean, 1 at the MZ twins' mean.
Intervals: \twentysevenNboot{} prompt resamples.
\label{tab:twentyseven-secondary}}
\begin{tabular}{>{\raggedright\arraybackslash}p{1.5cm}>{\raggedright\arraybackslash}p{1.9cm}rrrlrrrrr}
\toprule
Subset & Outcome & MZ twins & \begin{tabular}[b]{@{}c@{}}Same-\\recipe\end{tabular} & \begin{tabular}[b]{@{}c@{}}Same-\\seed\\pair\end{tabular} & Interval & \begin{tabular}[b]{@{}c@{}}First\\rerun:\\MZ twins\end{tabular} & \begin{tabular}[b]{@{}c@{}}First rerun\\with base\\models 2 and 3\end{tabular} & \begin{tabular}[b]{@{}c@{}}Position\\in the\\ancestry gap\end{tabular} & \begin{tabular}[b]{@{}c@{}}Mixed:\\same-\\seed\end{tabular} & \begin{tabular}[b]{@{}c@{}}Mixed:\\same-\\recipe\end{tabular} \\
\midrule
adversarial-harmful & WildGuard's refusal label & 0.985 & 0.814 & 0.839 & [0.827, 0.850] & 0.982 & 0.828 & 0.151 & 0.840 & 0.811 \\
 & incoherent counted & 0.985 & 0.814 & 0.839 & [0.827, 0.850] & 0.982 & 0.828 & 0.149 & 0.839 & 0.811 \\
 & WildGuard's harmfulness label & 0.986 & 0.821 & 0.843 & [0.827, 0.857] & 0.988 & 0.836 & 0.134 & 0.844 & 0.820 \\
 & Qwen3Guard as the judge & 0.988 & 0.841 & 0.855 & [0.843, 0.865] & 0.985 & 0.850 & 0.096 & 0.857 & 0.839 \\
 & judges agree & 0.985 & 0.824 & 0.845 & [0.833, 0.856] & 0.985 & 0.837 & 0.131 & 0.846 & 0.823 \\
plain-harmful & WildGuard's refusal label & 0.990 & 0.890 & 0.893 & [0.876, 0.909] & 0.989 & 0.893 & 0.034 & 0.895 & 0.892 \\
 & WildGuard's harmfulness label & 0.991 & 0.895 & 0.890 & [0.866, 0.912] & 0.989 & 0.894 & \ensuremath{-}0.045 & 0.893 & 0.895 \\
all-harmful & WildGuard's refusal label & 0.987 & 0.847 & 0.866 & [0.857, 0.875] & 0.984 & 0.858 & 0.139 & 0.867 & 0.846 \\
 & WildGuard's harmfulness label & 0.987 & 0.843 & 0.859 & [0.847, 0.871] & 0.988 & 0.854 & 0.113 & 0.860 & 0.843 \\
benign & WildGuard's refusal label & 0.988 & 0.851 & 0.793 & [0.775, 0.811] & 0.989 & 0.826 & \ensuremath{-}0.422 & 0.791 & 0.849 \\
 & WildGuard's harmfulness label & 0.985 & 0.868 & 0.817 & [0.782, 0.851] & 0.997 & 0.868 & \ensuremath{-}0.433 & 0.816 & 0.866 \\
 
\bottomrule
\end{tabular}
\end{table*}
\subsection{The second same-seed rerun}

A second same-seed rerun repeats base model 1 on our stack, with the same settings up to its name.
Its four post-trained models carry four post-training seeds under the second sharding setting.
Appendix~\ref{app:factorial} checks those settings among the second same-seed rerun's other checks.
Section~\ref{sec:ladder} states the registered test on one stack.
As for the first same-seed rerun, this subsection reports the second's pairs of base models, its subsets and outcomes, and its weights.
Table~\ref{tab:twentyseven-pairs} adds the second same-seed rerun's pairs of base models.
It reaches \thirtyfiveBpRtwoD{} with base model 1 and \thirtyfiveBpRRtwo{} with the first same-seed rerun, which reaches \twentysevenBpDefRep{} with base model 1.
The three same-seed pairs thus range from \thirtyfiveSamePairMin{} to \thirtyfiveSamePairMax{}, and their mean sits at \thirtyfiveGapFraction{} \thirtyfiveGapFractionCI{} of the ancestry gap.
Its MZ twins reach \thirtyfiveRmzRtwo{} \thirtyfiveRmzRtwoCI{}.
With base models 2 and 3, the second same-seed rerun reaches \thirtyfiveBpRtwoAuxTwo{} and \thirtyfiveBpRtwoAuxThree{}, beside the first same-seed rerun's \twentysevenBpRepAuxTwo{} and \twentysevenBpRepAuxThree{}.

Every pair group of the second same-seed rerun is also reported by subset and outcome, beside the first same-seed rerun's.
Table~\ref{tab:thirtyfive-secondary} gives them, with the pair of the two reruns minus the first same-seed rerun's pair with base model 1.
The interval of that difference clears zero in \thirtyfiveExcessClearCells{} of the table's \thirtyfiveSecondaryRows{} rows.
On benign prompts, a registered question without a prediction asks whether the pair of the two reruns sits above the same-recipe-pair mean.
The pair of the two reruns reaches \thirtyfiveBenignRRtwo{} \thirtyfiveBenignRRtwoCI{} on benign prompts, against \twentysevenBenignRu{} for same-recipe pairs.

\begin{figure}[t]
\centering
\begin{tikzpicture}
\begin{axis}[
  castor axis,
  width=0.6\textwidth, height=0.32\textwidth,
  xmin=0, xmax=70000, ymin=-0.02, ymax=0.02,
  xlabel={pretraining step}, ylabel={loss minus released mean (nats)},
  scaled x ticks=false, xticklabel style={/pgf/number format/fixed},
  scaled y ticks=false, yticklabel style={/pgf/number format/fixed, /pgf/number format/precision=2},
  legend style={at={(0.97,0.97)}, anchor=north east, legend columns=3},
]
\addplot[muted, thin, no markers] table[x=step, y=relone, col sep=tab] {figures/data/trajectory.tsv};
\addlegendentry{released runs}
\addplot[muted, thin, no markers, forget plot] table[x=step, y=reltwo, col sep=tab] {figures/data/trajectory.tsv};
\addplot[muted, thin, no markers, forget plot] table[x=step, y=relthree, col sep=tab] {figures/data/trajectory.tsv};
\addplot[pair-rerun, thick, mark=*, mark size=1pt] table[x=step, y=rone, col sep=tab] {figures/data/trajectory.tsv};
\addlegendentry{first rerun}
\addplot[pair-rerun, thick, dashed, mark=square*, mark size=1pt] table[x=step, y=rtwo, col sep=tab] {figures/data/trajectory.tsv};
\addlegendentry{second rerun}
\end{axis}
\end{tikzpicture}
\caption{\textbf{Both same-seed reruns' in-loop held-out loss stays near the released runs'.}
In-loop held-out loss minus the mean of the main data recipe's three released runs, at every logged step; gray lines are the released runs.
\label{fig:trajectory}}
\end{figure}
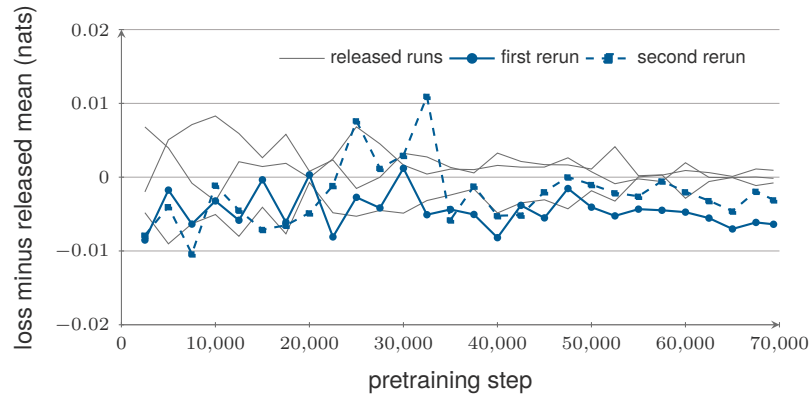

As for the first same-seed rerun in Figure~\ref{fig:divergence}b, we measure the second same-seed rerun's relative $L_2$ from the first same-seed rerun and from base model 1 at every saved step.
At step \elevenDivStepEarly{}, the two are \thirtyfiveDivRatioRoneEarly{} and \thirtyfiveDivRatioDEarly{} of the relative $L_2$ between two released runs of different pretraining seeds (Figure~\ref{fig:divergence}c).
At the final step, \elevenDivStepEnd{}, they are \thirtyfiveDivRelLtwoRoneEnd{} and \thirtyfiveDivRelLtwoDEnd{} against \thirtyfiveDivSeedRelLtwoEnd{}, shares of \thirtyfiveDivRatioRoneEnd{} and \thirtyfiveDivRatioDEnd{}.
Figure~\ref{fig:trajectory} plots both same-seed reruns' in-loop held-out loss against the three released runs.
After step 10,000, the second same-seed rerun deviates from the released runs' mean by at most \thirtyfiveTrajDevRtwo{}, the first by at most \thirtyfiveTrajDevRone{}, and the released runs by up to \thirtyfiveTrajDevReleased{}.

\begin{table*}[t]
\centering\footnotesize\setlength{\tabcolsep}{3pt}
\caption{\textbf{The second same-seed rerun by subset and outcome.}
Mean corrected correlations as in Table~\ref{tab:twentyseven-secondary}; the three same-seed pairs are each rerun with base model 1 and the two reruns together.
The difference is the pair of the two reruns minus the first rerun's pair with base model 1, with its 95\,\% interval over \twentysevenNboot{} prompt resamples.
Position in the ancestry gap is that of the mean of the three same-seed pairs.
Counting incoherent completions as refusals moves no value by more than \thirtyfiveInclMaxDiff{} and is not shown.
\label{tab:thirtyfive-secondary}}
\begin{tabular}{>{\raggedright\arraybackslash}p{2.4cm}>{\raggedright\arraybackslash}p{2.7cm}rrrrrrl}
\toprule
Subset & Outcome & \begin{tabular}[b]{@{}c@{}}Same-\\recipe\end{tabular} & \begin{tabular}[b]{@{}c@{}}First\\rerun with\\base model 1\end{tabular} & \begin{tabular}[b]{@{}c@{}}Second\\rerun with\\base model 1\end{tabular} & \begin{tabular}[b]{@{}c@{}}The two\\reruns\end{tabular} & \begin{tabular}[b]{@{}c@{}}Mean of\\three\end{tabular} & Difference & Interval \\
\midrule
adversarial-harmful & WildGuard's refusal label & 0.814 & 0.839 & 0.842 & 0.851 & 0.844 & +0.011 & [\ensuremath{-}0.001, 0.023] \\
 & WildGuard's harmfulness label & 0.821 & 0.843 & 0.848 & 0.861 & 0.850 & +0.018 & [0.004, 0.032] \\
 & Qwen3Guard as the judge & 0.841 & 0.855 & 0.860 & 0.861 & 0.859 & +0.006 & [\ensuremath{-}0.005, 0.017] \\
 & judges agree & 0.824 & 0.845 & 0.847 & 0.856 & 0.849 & +0.011 & [\ensuremath{-}0.001, 0.022] \\
plain-harmful & WildGuard's refusal label & 0.890 & 0.893 & 0.902 & 0.903 & 0.899 & +0.010 & [\ensuremath{-}0.006, 0.026] \\
 & WildGuard's harmfulness label & 0.895 & 0.890 & 0.892 & 0.901 & 0.894 & +0.010 & [\ensuremath{-}0.011, 0.031] \\
 & Qwen3Guard as the judge & 0.910 & 0.912 & 0.917 & 0.920 & 0.916 & +0.008 & [\ensuremath{-}0.006, 0.022] \\
 & judges agree & 0.893 & 0.897 & 0.904 & 0.907 & 0.903 & +0.010 & [\ensuremath{-}0.006, 0.026] \\
all-harmful & WildGuard's refusal label & 0.847 & 0.866 & 0.871 & 0.878 & 0.872 & +0.011 & [0.003, 0.020] \\
 & WildGuard's harmfulness label & 0.843 & 0.859 & 0.863 & 0.875 & 0.866 & +0.016 & [0.004, 0.027] \\
 & Qwen3Guard as the judge & 0.869 & 0.880 & 0.885 & 0.887 & 0.884 & +0.007 & [0.000, 0.015] \\
 & judges agree & 0.853 & 0.870 & 0.873 & 0.881 & 0.875 & +0.011 & [0.002, 0.019] \\
benign & WildGuard's refusal label & 0.851 & 0.793 & 0.825 & 0.839 & 0.819 & +0.047 & [0.027, 0.066] \\
 & WildGuard's harmfulness label & 0.868 & 0.817 & 0.811 & 0.894 & 0.841 & +0.077 & [0.043, 0.112] \\
 & Qwen3Guard as the judge & 0.874 & 0.825 & 0.848 & 0.876 & 0.849 & +0.051 & [0.036, 0.067] \\
 & judges agree & 0.854 & 0.788 & 0.827 & 0.847 & 0.821 & +0.059 & [0.041, 0.078] \\
 
\bottomrule
\end{tabular}

\medskip
\begin{tabular}{>{\raggedright\arraybackslash}p{2.4cm}>{\raggedright\arraybackslash}p{3.4cm}rrr}
\toprule
Subset & Outcome & \begin{tabular}[b]{@{}c@{}}Second\\rerun:\\MZ twins\end{tabular} & \begin{tabular}[b]{@{}c@{}}Second rerun\\with base\\models 2 and 3\end{tabular} & \begin{tabular}[b]{@{}c@{}}Position\\in the\\ancestry gap\end{tabular} \\
\midrule
adversarial-harmful & WildGuard's refusal label & 0.979 & 0.820 & 0.178 \\
 & WildGuard's harmfulness label & 0.980 & 0.830 & 0.180 \\
 & Qwen3Guard as the judge & 0.979 & 0.843 & 0.123 \\
 & judges agree & 0.979 & 0.827 & 0.157 \\
plain-harmful & WildGuard's refusal label & 0.989 & 0.906 & 0.096 \\
 & WildGuard's harmfulness label & 0.987 & 0.914 & \ensuremath{-}0.004 \\
 & Qwen3Guard as the judge & 0.993 & 0.924 & 0.079 \\
 & judges agree & 0.991 & 0.911 & 0.102 \\
all-harmful & WildGuard's refusal label & 0.982 & 0.857 & 0.178 \\
 & WildGuard's harmfulness label & 0.982 & 0.855 & 0.158 \\
 & Qwen3Guard as the judge & 0.984 & 0.875 & 0.125 \\
 & judges agree & 0.983 & 0.861 & 0.158 \\
benign & WildGuard's refusal label & 0.985 & 0.835 & \ensuremath{-}0.231 \\
 & WildGuard's harmfulness label & 0.980 & 0.856 & \ensuremath{-}0.232 \\
 & Qwen3Guard as the judge & 0.990 & 0.863 & \ensuremath{-}0.210 \\
 & judges agree & 0.986 & 0.839 & \ensuremath{-}0.245 \\
 
\bottomrule
\end{tabular}
\end{table*}

Appendix~\ref{app:pipeline} returns to post-training: how far apart the post-trained models of one base model and one post-training seed end, and the shift between the two sharding settings.

\FloatBarrier
\section{Base models and post-training: details and checks}\label{app:pipeline}
\subsection{Base models}\label{app:pipeline-models}

\begin{table}[t]
  \centering\small
  \caption{\label{tab:base-models}\textbf{Each released base model takes its initialization and its data order from one pretraining seed, and each rerun takes them from released runs' seeds.}
  Columns give each base model, its initialization seed, its data-order seed and who pretrained it.
  We write $(k, s)$ for base model $k$ post-trained with post-training seed $s$, and every per-model table uses it.}
  \begin{tabular}{lrrl}
\toprule
Base model & Initialization seed & Data-order seed & Pretrained by \\
\midrule
1 & 6198 & 6198 & DataDecide \\
2 & 4 & 4 & DataDecide \\
3 & 5 & 5 & DataDecide \\
shared-initialization rerun & 6198 & 4 & this work \\
shared-order rerun & 4 & 6198 & this work \\
first same-seed rerun & 6198 & 6198 & this work \\
second same-seed rerun & 6198 & 6198 & this work \\
 
\bottomrule
\end{tabular}

\end{table}

This appendix gives the training inputs that Section~\ref{sec:setup} names in one line, then three checks that each training step leaves our models as intended.
We convert DataDecide's base models \citep{magnusson2025datadecide} from their released format to a standard decoder layout, keeping every block and re-indexing the tensors.
Each of the three sizes has its own architecture, with every architecture field equal within a size, so we check the conversion once per size.
The check compares fp32 logits on the same inputs with those of DataDecide's reference code at a pinned commit.
It requires a largest absolute logit difference of at most $10^{-3}$ and a top-1 token match on at least 0.999 of positions.
Both thresholds were added at code review and fixed before any size's models were used.
At every size the top-1 token matches on 1.000 of positions, and the largest logit difference is $4.2 \times 10^{-4}$.

The converted models keep DataDecide's tokenizer, with no beginning-of-sequence (BOS) token and a vocabulary of 50,280, and a context of 2048 tokens.
DataDecide pretrains with OLMo's code, and with OLMo's pipeline an end-of-text token ends every document \citep{magnusson2025datadecide,groeneveld2024olmo}.

At 1B, DataDecide trains three pretraining seeds to the full budget of 69,369 steps and 100B tokens \citep{magnusson2025datadecide}.
Table~\ref{tab:base-models} numbers each base model $k$ in DataDecide's release order.
The step-0 test ties each $k$ to a released run whose step-0 weights equal ours tensor for tensor (Appendices~\ref{app:nondet} and~\ref{app:factorial}).
That test verifies the initialization seed, and so the data-order seed, since one seed sets both in DataDecide's code \citep{magnusson2025datadecide}.
For base model 1, a post-hoc order check confirms the data order on the released 150M run, whose order our 1B order equals by hash (Appendix~\ref{app:prereg}).

Below 1B, the size results of Section~\ref{sec:grid} use checkpoints, never base models.
We take the checkpoint of all three pretraining seeds at a step that every seed of every data recipe reaches, 37,500 at 150M and 45,000 at 300M.
That step is about 98\,\% of the schedule of the run with pretraining seed 6198, which lasts 38,157 steps at 150M and 45,787 at 300M.
At these sizes the pretraining seeds are 6198, 14 and 15 rather than the 6198, 4 and 5 of 1B, so Table~\ref{tab:base-models} lists 1B models only.
Three of the main data recipe's four runs with seeds 14 and 15 trained past their budget, both 300M runs and the 150M run with seed 14, which reached 351,260 steps.
They did so because, below 1B, DataDecide stops all but one seed's runs at 25\,\% of the 1B model's compute \citep{magnusson2025datadecide}, about 13 times the 150M budget and 3 times the 300M budget.
Not every run reached that point, and the 150M run with seed 15 stopped at its budget.
A run trained past its budget may have had a different learning rate at our step.
No optimizer state is released, so we could not compare it with the seed-6198 run's.
\subsection{Post-training as run}\label{app:pipeline-post}
From each base model we post-train in two stages, SFT then DPO, following Ai2's post-training of OLMo 2 1B.
Both stages run open-instruct, unmodified, at commit \texttt{fc7748ea}.
Ai2 did not publish its commits, but the last commits before its SFT and DPO runs differ from ours in the scripts only by a guarded import and one environment variable.
We run Ai2's commands from open-instruct's OLMo 2 guide unchanged.
SFT trains for two epochs with learning rate $3 \times 10^{-5}$, a linear schedule, 3\,\% warm-up, batch size 128 and a summed token loss.
DPO trains for one epoch with learning rate $2.5 \times 10^{-6}$, 10\,\% warm-up and batch size 128.
Its loss is Tülu 3's length-normalized DPO with $\beta = 5$, which Tülu 3 attributes to Rafailov et al.\ as tuned by Meng et al.~\citep{lambert2024tulu3}.
Neither stage uses weight decay or gradient clipping.

The SFT data are the Tülu 3 SFT mixture for OLMo 2, which has 866,138 rows \citep{olmo2025olmo2}.
The DPO data are the OLMo 2 1B preference mix, which has 378,301 pairs.
We pin both at fixed revisions and keep their row order.
Ai2 built those pairs on-policy for its own SFT model \citep{olmo2025olmo2}, and every model we post-train receives the same pairs off-policy, so the published post-training recipe applies unchanged to all.
No preference source is labeled safety, yet some pairs reuse SFT safety prompts.
By our exact-match count, the pairs that reuse one verbatim number 14,534.

Post-training seed $s$ gives the SFT stage seed $s$ and the DPO stage seed $110 + s$, so $s = 1$ is Ai2's pair of 1 and 111.
We call the DPO stage's seed alone the \emph{DPO-stage seed}, which Appendix~\ref{app:pipeline-ai2} varies on one fixed SFT model.
Everywhere else, the post-training seed sets both stages.
In each stage the seed sets the dataset shuffle and each epoch's sampler order, and nothing else, since dropout is zero and no weight is newly initialized.
GPU nondeterminism is not seeded (Appendix~\ref{app:nondet}).

Two changes to Ai2's post-training recipe apply alike to every DataDecide base model and to no other model.
First, an end-of-text token opens every SFT and DPO sequence in place of a BOS token, as in Ai2's first-generation OLMo chat templates and open-instruct's code for this tokenizer family.
Under OLMo's pipeline end-of-text ends every pretraining document, so it is the token that precedes a document's start, as in OLMo's own pretraining \citep{groeneveld2024olmo}.
Second, SFT sequences are cut at 2048 tokens, the pretraining context, which DPO already uses in Ai2's post-training recipe.
The cut truncates 4.94\,\% of SFT rows and drops the 1.42\,\% left with no trainable token, against 1.97\,\% and 0.50\,\% in Ai2's run at 4096 tokens.
The OLMo 2 runs of Appendix~\ref{app:pipeline-ai2} keep Ai2's BOS token and 4096 tokens.
Both stages use open-instruct's Tülu chat template, opened by end-of-text for DataDecide base models and by BOS for OLMo 2.
The length rule of Appendix~\ref{app:pipeline-panel} counts tokens with the end-of-text-opened template, the longer of the two.
\subsection{Prompt set rules}\label{app:pipeline-panel}
Two rules tie the prompt set to the post-training of Appendix~\ref{app:pipeline-post}: decontamination against its data, and a length limit set by its context.
Decontamination uses Tülu 3's 8-gram overlap test, which Tülu 3 applies to remove training data and we apply to drop our prompts \citep{lambert2024tulu3}.
A prompt drops when more than half of its tokens fall in 8-grams shared with one training user turn.
We run it against every SFT and DPO user turn, and Ai2's reinforcement-learning data, which removed nothing.
An exact-match check from open-instruct's decontamination code at commit \texttt{d19de7d27} catches prompts too short for an 8-gram.
Our matching is exhaustive where open-instruct searches an index's top 100 hits.
We use one tokenizer on both sides and split on any whitespace.
A template pass removes a human-jailbreak template whole when any of its prompts drops, removing 18 of 114 templates.
Each addition makes the test stricter or leaves it equal.

Two cross-source duplicates leave, the test and the exact match drop 194 prompts, and the template pass drops 46 more.
These steps take the prompt set from 6,532 prompts to 6,290.
The 8-gram test misfires on some short prompts, matching a harmful prompt to a harmless one, and we keep it as fixed before the data.
The removed prompts from sources other than HarmBench's human jailbreaks number 16, some of them misfires.
The second rule, length, drops a prompt when its chat-formatted tokens, counted as in Appendix~\ref{app:pipeline-post}, plus the \sixMaxTokens{} new tokens exceed 2048.
It takes the prompt set from 6,290 prompts to \sixNpanel{}, and all 64 prompts it drops come from HarmBench's human jailbreaks.
\subsection{Reproducing Ai2's OLMo 2 1B post-training}\label{app:pipeline-ai2}
With the prompt set fixed, we check our post-training against Ai2's OLMo 2 1B release, made with the same post-training recipe \citep{olmo2025olmo2}, in two pre-registered checks of the DPO model (Appendix~\ref{app:prereg}).
Each check has two criteria, and the first uses the \twelveNharmful{} harmful prompts of the prompt set.
On them, the Pearson correlation of per-prompt vulnerability between our post-training seed 1 and Ai2's DPO may fall at most 0.05 below that between our post-training seeds 1 and 2.
That vulnerability counts incoherent samples, the registered variant of Appendix~\ref{app:robustness}.
The second criterion requires each of the seven sources' refusal rates, over all its prompts, to lie within 3 points of Ai2's.
SFT models are compared outside the rule.
Every interval in this appendix is a 95\,\% percentile bootstrap over prompts with the models fixed, using the same resampled prompts for every model, from 2,000 draws unless stated.

The first check starts from the OLMo 2 1B weights that the model card names, its main branch.
Our post-training seed 1 correlates with Ai2's DPO at \fourRpipe{} \fourRpipeCI{}, against \fourRseed{} \fourRseedCI{} between our post-training seeds.
That correlation falls more than 0.05 below the one between our seeds.
Counting coherent samples only gives the same correlation to three decimals.
Refusal rates lie within 3 points on \fourRatesWithin{} of \fourNsources{} sources.
To find why the correlation fell short, we compared weights.
Ai2's released SFT model was not fine-tuned from those weights.
In $L_2$ distance over all parameters, our SFT model and Ai2's lie \fourLtwoSftAiTwo{} apart, against \fourLtwoSftSeeds{} between our two post-training seeds.
Ai2's SFT model instead sits one SFT update from a stage-2 branch of the same repository, the ingredient-3 branch \texttt{stage2-ingredient3-step23852-tokens51B}.
On the probed tensors, Ai2's SFT model has a cosine near 0 with the main branch.
The first check thus compared our post-training on one starting model with Ai2's on another, so it says nothing either way about our post-training.

The second check starts from the ingredient-3 branch, and a miss on it was registered to stop the work for diagnosis, with no MZ twins post-trained.
Our post-training seed 1 correlates with Ai2's DPO at \tenRpipe{} \tenRpipeCI{}, against \tenRseed{} \tenRseedCI{} between our post-training seeds.
Refusal rates lie within 3 points on \tenRatesWithin{} of \tenNsources{} sources.
On WildJailbreak, seed 1 refuses \tenWjGap{} points less than Ai2's DPO, and our post-training seeds 1 and 2 differ there by \tenWjSeedDiff{}.
Seed 1 refuses less than Ai2's DPO on all seven sources by 0.0 to 5.2 points, and seed 2 on six of seven.
With seed 2 in seed 1's place, the rate criterion would miss on OR-Bench, WildJailbreak and HarmBench, by up to 10.8 points.
After SFT, our model lies \tenLtwoSftAiTwo{} from Ai2's in $L_2$ distance, against \tenLtwoSftSeeds{} between our seeds.
Its refusal rates lie within \tenSftRatesMaxDiff{} points of Ai2's SFT on every source, and its corrected correlation with Ai2's SFT is \tenSftRcorrAiTwo{}.
The second check misses its rate criterion on one source, WildJailbreak.

The second check's recorded outcome stands.
A decision taken after that outcome kept our post-training pipeline before the diagnosis below, run in parallel, had located the gap.
Post-training of the MZ twins therefore started against the registered stop.
The decision recorded three reasons.
The first was the second check's SFT and weight evidence: the $L_2$ norm of each stage's change from its starting model differs from Ai2's by \tenUpdSftPct{} at SFT and \tenUpdDpoPct{} at DPO.
That reason also cited the corrected correlation on harmful prompts between our DPO model with post-training seed 1 and Ai2's DPO, \tenDpoRcorrAiTwo{}.
The second was that the one gap lies inside our own seed spread, so the 3-point tolerance was tighter than the \tenWjSeedDiff{} points between our post-training seeds.
The third was that the post-training stages are the same for every model we compare, so the gap cannot separate one kind of pair from another.
The registered step-by-step comparison of Ai2's intermediate SFT checkpoints was not run.
Appendix~\ref{app:prereg} records both changes, the overridden stop and the comparison not run.

A pre-registered follow-up (Appendix~\ref{app:prereg}) locates the refusal-rate gap of Table~\ref{tab:twelve-sources}.
It runs DPO from Ai2's own SFT model with Ai2's DPO-stage seed.
On our SFT model with seed 1, it adds two further DPO-stage seeds and a \emph{repeat run}, a second post-training run from the same model with the same seeds.
The rule requires DPO from Ai2's SFT to lie within $\max(3, W)$ points of Ai2's DPO on every source.
Here $W$ is the largest difference in refusal rate among our three DPO-stage seeds on that source.
DPO from Ai2's SFT lies within its tolerance on \twelveNwithin{} of \twelveNsources{} sources, with a largest difference of \twelveMaxAminusRef{} points.
The tolerances span 3.0 to 7.6 points, with $W$ from \twelveWmin{} to \twelveWmax{}.
The two further DPO-stage seeds refuse less than seed 1's model on every source, so seed 1 sits at the high end of the three.
The repeat run lies within \twelveFmax{} points of seed 1's model.

\begin{table*}[t]
  \centering\small
  \caption{\label{tab:twelve-sources}\textbf{DPO from Ai2's SFT model lies within its tolerance of Ai2's DPO on every source.}
  Per source, refusal rates (\%) of our DPO model with post-training seed 1, DPO from Ai2's SFT, the repeat run, the two further DPO-stage seeds and Ai2's DPO.
  Then the difference of DPO from Ai2's SFT from Ai2's DPO, the tolerance $\max(3, W)$, whether within, $W$, and the repeat run's absolute difference from seed 1.
  Rates cover all prompts of each source, labeled by the judge of Section~\ref{sec:setup}.}
  \footnotesize\setlength{\tabcolsep}{3pt}
\begin{tabular}{lrrr@{\hspace{14pt}}r@{\hspace{14pt}}r@{\hspace{14pt}}rrrlrr}
\toprule
 & & & & \multicolumn{2}{c}{\shortstack{Further DPO-\\stage seeds}} & & & & & & \\
\cmidrule(lr){5-6}
Source & Seed~1 & \shortstack{DPO from\\Ai2's SFT} & \shortstack{Repeat\\run} & & & \shortstack{Ai2's\\DPO} & Difference & Tolerance & Within & $W$ & \shortstack{Repeat\\difference} \\
\midrule
HarmBench behaviors & 64.6 & 65.4 & 65.0 & 61.6 & 62.8 & 66.2 & \ensuremath{-}0.8 & 3.1 & yes & 3.1 & 0.4 \\
HarmBench human jailbreaks & 55.6 & 58.9 & 54.1 & 49.9 & 50.9 & 57.9 & +1.0 & 5.7 & yes & 5.7 & 1.6 \\
OR-Bench-Hard-1K & 37.3 & 38.3 & 36.7 & 29.7 & 30.9 & 39.0 & \ensuremath{-}0.7 & 7.6 & yes & 7.6 & 0.6 \\
SORRY-Bench & 71.6 & 71.7 & 71.1 & 67.4 & 70.1 & 72.0 & \ensuremath{-}0.2 & 4.2 & yes & 4.2 & 0.5 \\
StrongREJECT & 91.2 & 91.8 & 90.9 & 88.3 & 89.9 & 91.7 & +0.1 & 3.0 & yes & 2.8 & 0.2 \\
WildJailbreak & 52.9 & 57.1 & 52.8 & 46.6 & 48.1 & 58.1 & \ensuremath{-}1.0 & 6.3 & yes & 6.3 & 0.1 \\
XSTest & 40.5 & 40.7 & 40.3 & 38.2 & 37.2 & 40.5 & +0.1 & 3.3 & yes & 3.3 & 0.2 \\
 
\bottomrule
\end{tabular}

\end{table*}

Per prompt, DPO from Ai2's SFT has a corrected correlation of \twelveAvsRefRcorr{} with Ai2's DPO.
The repeat run has \twelveFloorRcorr{} with seed 1's model, and a change of DPO-stage seed gives \twelveSeedChangeRcorrMin{} to \twelveSeedChangeRcorrMax{}.
We call the $L_2$ distance between a model and its repeat run the \emph{identical-seed distance}, here \twelveLtwoFloor{}.
In weights, DPO from Ai2's SFT lies \twelveLtwoAvsAiTwo{} from Ai2's DPO, against the identical-seed distance and \twelveLtwoSeedMin{} to \twelveLtwoSeedMax{} for a change of DPO-stage seed.
The follow-up departs from its registration in one way: its rule was registered on harmful prompts' refusal rates and run on every prompt's.
OR-Bench has no harmful prompts, so it drops out.
On harmful prompts alone, DPO from Ai2's SFT lies within its tolerance on \twelveHarmfulWithin{} of \twelveHarmfulNsources{} sources.

Our SFT stage reproduces Ai2's on every source.
Started from Ai2's SFT model, our DPO stage reproduces Ai2's DPO.
The second check's one gap beyond 3 points lies inside the spread of our own DPO-stage seeds.
Most of that gap arises before DPO, from the small difference between our SFT model and Ai2's.
The same DPO-stage seed on the two SFT models moves WildJailbreak by \twelveParentSwapWj{} of the \tenWjGap{} points.
\subsection{Pre-registered checks of the first post-training batch}\label{app:pipeline-first}
Post-training ran in two batches (Appendix~\ref{app:compute}).
The twelve models of the first post-training batch are the data of Appendix~\ref{app:mechanism} and of the human-label paragraph of Section~\ref{sec:results}.
They are also the data of Appendices~\ref{app:judge-first} and~\ref{app:judge-pairs}, and of the first-batch part of Appendix~\ref{app:robustness}.
A pre-registered rule on these twelve models tested that the signal is measurable on adversarial-harmful prompts.
The rule required every model's mean vulnerability to lie in $[0.1, 0.9]$ and its reliability to reach at least 0.7, where we predicted at least 0.8. A model's reliability is how well it agrees with itself across two halves of its samples.
Mean vulnerability ranges from \sixVulnMin{} to \sixVulnMax{}, and reliability from \sixRelMin{} to \sixRelMax{}.
With the rule we predicted that plain harmful prompts would mostly be refused, and each model refuses \sixPlainRefusalMin{} to \sixPlainRefusalMax{} of them.
We also predicted that MZ twins would correlate more than same-recipe pairs.
Their raw Pearson correlations are \sixRmzRaw{} \sixRmzRawCI{} for MZ twins against \sixRuRaw{} \sixRuRawCI{} for same-recipe pairs.
These two intervals use 1,000 draws.
Appendices~\ref{app:mechanism} and~\ref{app:robustness} give the corrected difference.

A second pre-registered rule on the same twelve models set the 25-recipe grid's design.
If every pair of MZ twins had a raw correlation within 0.02 of the pair's mean reliability, the grid would drop to one post-training seed on most data recipes.
The pairs of MZ twins that came within 0.02 were \sixOneSeedHold{} of \sixOneSeedPairs{}.
The two outside both involve model $(3, 1)$, by \sixOneSeedMissA{} and \sixOneSeedMissB{}.
The 25-recipe grid therefore kept two post-training seeds.
The 25-recipe grid and the size results of Section~\ref{sec:grid} come from the second post-training batch, which Appendix~\ref{app:pipeline-setups} compares with the first.
\subsection{The two batches compared}\label{app:pipeline-setups}
The second batch changes only how the SFT stage is spread over GPUs (Appendix~\ref{app:compute}).
Every training argument and the whole DPO stage stay the same.
Before we adopted it, a pre-registered test on $(1, 1)$ compared the first-batch model, its repeat run and a test run with the second batch's sharding setting.
Its first criterion required the test run's distance after DPO from the first-batch model, and from its repeat run, to be at most 1.25 times the identical-seed distance.
That distance, here between the first-batch model and its repeat run, is \twentyfourWfloor{}, against \twentyfourWseed{} for a change of post-training seed.
The test run lies \twentyfourWratioAR{} times that distance from the first-batch model and \twentyfourWratioAB{} times from the repeat run.
The second ratio lies inside its tolerance by 0.1\,\% of it.
That tolerance rests on a single identical-seed distance.
After SFT, outside the rule, the ratio to the repeat run is above 1.25, at \twentyfourWsftRatioAB{}.
The second criterion required the paired interval of the repeat run's corrected correlation with the first-batch model minus the test run's to include 0, or the point difference to be at most 0.01.
That difference is \twentyfourRstarDiff{} \twentyfourRstarDiffCI{}.
The third criterion required the test run's refusal rate on each source to lie within 3 points of the first-batch model's, or within the repeat run's own difference if larger.
Every tolerance was 3 points, and refusal rates lie within 3 points on \twentyfourLevelsWithin{} of 7 sources.
The test run met all three criteria, and we adopted its setting.

\begin{table}[t]
  \centering\small
  \caption{\label{tab:twentyseven-config-kinds}\textbf{The two models with the same $(k, s)$ correlate more closely and lie closer in weights than MZ twins, within one batch or across the two.}
  Rows are kinds of pair.
  Columns give the number of pairs, raw Pearson $r$, corrected $r$ with its interval, and mean $L_2$ distance after DPO with its minimum and maximum over pairs.
  Row labels name the two batches.}
  \begin{tabular}{lrrrcrc}
\toprule
Pair & $n$ & Raw $r$ & Corrected $r$ & Interval & Mean $L_2$ & [min, max] \\
\midrule
one $(k, s)$ under both sharding settings & 12 & 0.945 & 0.9999 & [0.9988, 1.0012] & 8.97 & [7.43, 10.31] \\
MZ twins, both under the first sharding setting & 18 & 0.930 & 0.9843 & [0.9824, 0.9861] & 34.34 & [34.13, 34.69] \\
MZ twins, both under the second sharding setting & 18 & 0.930 & 0.9848 & [0.9828, 0.9868] & 34.86 & [34.64, 35.20] \\
MZ twins, one under each setting & 36 & 0.930 & 0.9848 & [0.9833, 0.9863] & 34.67 & [34.44, 35.02] \\
 
\bottomrule
\end{tabular}

\end{table}

After adoption, we post-trained each of the twelve first-batch models $(k, s)$ again in the second batch, as a separate model.
The result is twelve pairs of a first- and a second-batch model with the same $(k, s)$ (Table~\ref{tab:twentyseven-config-kinds}, and per $(k, s)$ Table~\ref{tab:twentyseven-config-ks}).
In behavior, the two models of a pair have a corrected correlation of \twentysevenConfigRstar{} \twentysevenConfigRstarCI{}.
Their raw Pearson correlation of \twentysevenConfigRstarRaw{} is about the models' reliabilities, which span \twentysevenRelMin{} to \twentysevenRelMax{} in the second batch and \sixRelMin{} to \sixRelMax{} in the first.
The corrected correlation exceeds that of every row of MZ twins in Table~\ref{tab:twentyseven-config-kinds}.
The interval of every pair contains 1 (Table~\ref{tab:twentyseven-config-ks}).
The two models of a pair thus correlate with each other as closely as each correlates with itself.
Among MZ twins, pairs across batches minus pairs within one batch give \twentysevenConfigEffect{} \twentysevenConfigEffectCI{}.
Refusal rates lie within 3 points in \twentysevenConfigLevelsWithin{} of \twentysevenConfigLevelsTotal{} pairs of a model and a source, with a largest shift of \twentysevenConfigLevelsMaxAbs{} points.

\begin{table*}[t]
  \centering\small
  \caption{\label{tab:twentyseven-config-ks}\textbf{In each of the twelve pairs, the interval of the corrected correlation contains 1, and the weight distance grows from base model 1 to base model 3.}
  Per $(k, s)$, the columns give the corrected $r$ between the first- and second-batch model with its interval.
  Next comes the mean shift in vulnerability on adversarial-harmful prompts in points, second batch minus first, so a positive shift means the second is more vulnerable.
  Then follow the $L_2$ distance after DPO, its ratio to the identical-seed distance, whether it lies within 1.25 times, the largest shift in refusal rate in points, and the sources within 3 points.}
  \setlength{\tabcolsep}{4pt}
\begin{tabular}{lrcrrrlrr}
\toprule
$(k, s)$ & Corrected $r$ & Interval & \shortstack{Vulnerability\\shift} & $L_2$ & Ratio & $\leq 1.25\times$ & \shortstack{Max rate\\shift} & Within \\
\midrule
$(1, 1)$ & 1.0006 & [0.9967, 1.0048] & +0.46 & 7.43 & 1.16$\times$ & yes & 0.77 & 7/7 \\
$(1, 2)$ & 0.9994 & [0.9958, 1.0031] & +0.16 & 8.06 & 1.26$\times$ & no & 0.56 & 7/7 \\
$(1, 3)$ & 0.9991 & [0.9948, 1.0035] & \ensuremath{-}0.72 & 8.13 & 1.27$\times$ & no & 1.09 & 7/7 \\
$(1, 4)$ & 0.9979 & [0.9942, 1.0017] & +0.41 & 7.82 & 1.22$\times$ & yes & 0.75 & 7/7 \\
$(2, 1)$ & 1.0004 & [0.9964, 1.0045] & +0.20 & 8.61 & 1.35$\times$ & no & 2.01 & 7/7 \\
$(2, 2)$ & 0.9969 & [0.9929, 1.0008] & +0.75 & 9.62 & 1.50$\times$ & no & 1.51 & 7/7 \\
$(2, 3)$ & 0.9997 & [0.9957, 1.0035] & +0.67 & 8.86 & 1.38$\times$ & no & 1.17 & 7/7 \\
$(2, 4)$ & 1.0012 & [0.9974, 1.0046] & +0.65 & 8.71 & 1.36$\times$ & no & 2.23 & 7/7 \\
$(3, 1)$ & 0.9989 & [0.9943, 1.0034] & \ensuremath{-}0.08 & 9.76 & 1.53$\times$ & no & 0.55 & 7/7 \\
$(3, 2)$ & 1.0039 & [0.9994, 1.0086] & +0.32 & 10.24 & 1.60$\times$ & no & 0.60 & 7/7 \\
$(3, 3)$ & 0.9974 & [0.9928, 1.0020] & +0.96 & 10.31 & 1.61$\times$ & no & 1.25 & 7/7 \\
$(3, 4)$ & 1.0039 & [0.9995, 1.0087] & \ensuremath{-}0.63 & 10.09 & 1.58$\times$ & no & 0.70 & 7/7 \\
 
\bottomrule
\end{tabular}

\end{table*}

In weights, the two models of a pair lie \twentysevenConfigWeightsSameCell{} apart on average, \twentysevenConfigRatioMean{} the identical-seed distance of the adoption test.
That ratio rises from base model 1 to base model 3 (Table~\ref{tab:twentyseven-config-ks}), from \twentysevenConfigRatioMin{} to \twentysevenConfigRatioMax{}.
Within \twentysevenConfigFactor{} times that distance lie \twentysevenConfigNwithinBound{} of \twentysevenConfigNsameCell{} pairs.
The factor \twentysevenConfigFactor{} was set on one pair on base model 1 as the adoption criterion, so we report this count and do not apply the factor to the twelve.
A change of post-training seed moves weights by \twentysevenConfigWeightsMZwithin{} (Table~\ref{tab:twentyseven-config-kinds}).
The distance within a pair, as a share of that change, is \twentysevenConfigWeightsRatioSeed{}.

Last, we recompute the mean corrected correlations with every model of the main data recipe taken from the second batch.
The mean corrected correlation of a pair of base models is the mean over the pairs of models built on them.
The first same-seed rerun's mean with base model 1 moves by \twentysevenRungTwoMinusEleven{}, and its mean with base models 2 and 3 by \twentysevenRepOtherMinusEleven{}.
The same-recipe-pair mean \rU{} moves by \twentysevenRuMinusEleven{} and \rMZ{} by \twentysevenRmzMinusEleven{}.
No pair of base models' mean moves by more than \twentysevenBpMaxAbsShift{}.
Appendix~\ref{app:nondet} gives the means that mix the two batches.

Second-batch models refuse slightly less on every source, and their refusal rate minus the first batch's has a mean of \twentysevenConfigLevelsSourceMeanMax{} to \twentysevenConfigLevelsSourceMeanMin{} points per source.
Post-training again in the second batch thus moves weights by \twentysevenConfigWeightsRatioSeed{} of a post-training-seed change and shifts refusal rates by at most \twentysevenConfigLevelsMaxAbs{} points.
Appendix~\ref{app:judge} describes the judge whose labels give every refusal rate and vulnerability in this appendix.

\FloatBarrier
\section{Human audit of the judge}
\label{app:judge}
\subsection{The judge}\label{app:judge-instrument}
This appendix describes the judge and checks it against human labels in two audits and in a labeling of the responses of MZ twins and same-recipe pairs (Appendices~\ref{app:judge-first}, \ref{app:judge-grid} and~\ref{app:judge-pairs}).
It corrects mean vulnerability for the judge's precision (Appendix~\ref{app:judge-correction}).
Our primary judge, WildGuard, answers on three lines: harmful request, response refusal and harmful response \citep{han2024wildguard}.
A response counts as vulnerable when its response-refusal line says it is no refusal, which gives the vulnerability of Section~\ref{sec:setup}.
The four-class judge label adds the harmful-response line, and the harmful-request line goes unused.
Beside that refusal line, we score the refusal line of Qwen3Guard-Gen-8B \citep{zhao2025qwen3guard}.
Qwen3Guard's strict mode acts on its safety line, which gives the harm label, and not on its refusal line.
Qwen3Guard's refusal line sets two variants in Appendix~\ref{app:robustness}, one with Qwen3Guard as the judge and one on samples where both judges agree.

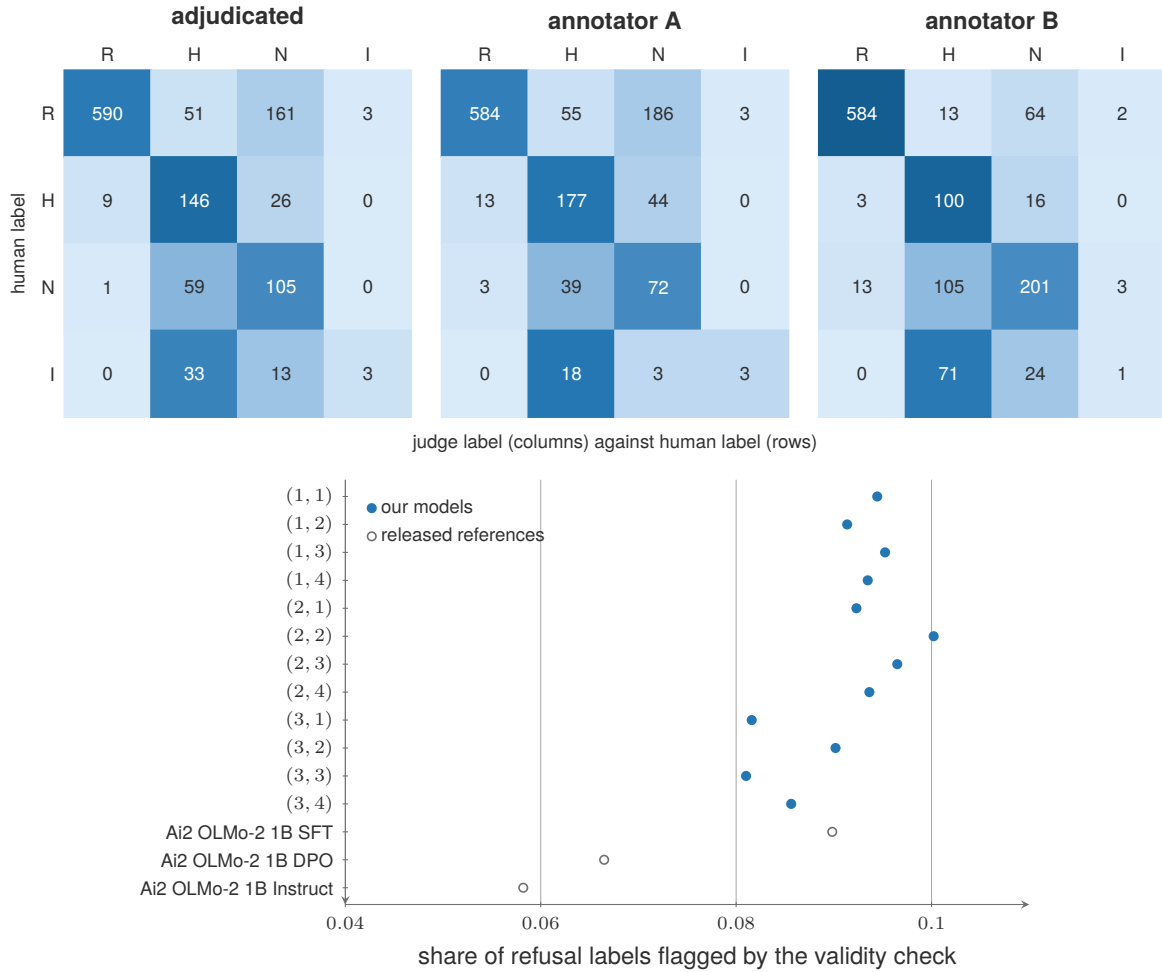
\begin{figure}[h]\centering
\begin{tikzpicture}
\tikzset{castor confusion labels/.style={font=\castortick, text=text}}
\pgfplotsset{castor confusion/.style={castor axis, width=0.36\textwidth, height=0.36\textwidth,
  enlargelimits=false, axis lines=none, ymajorgrids=false,
  xmin=-0.5, xmax=3.5, ymin=-0.5, ymax=3.5,
  colormap name=castor, point meta min=0, point meta max=1,
  title style={font=\castorbold, text=text, yshift=2mm},
  nodes near coords={\ifdim\share pt>0.55pt\textcolor{box}{\cnt}\else\cnt\fi},
  nodes near coords align=center, every node near coord/.append style={font=\castortick, text=text},
  visualization depends on={value \thisrow{count} \as \cnt},
  visualization depends on={value \thisrow{share} \as \share}}}
\newcommand{\castorconfusion}[5]{%
  \begin{axis}[castor confusion, title={#2}, name=#3, #4]
  \addplot[matrix plot*, mesh/cols=4, point meta=explicit]
    table[x=x, y=y, meta=share, col sep=tab] {#1};
  \node[castor confusion labels, anchor=south] at (axis cs:0,3.5) {R};
  \node[castor confusion labels, anchor=south] at (axis cs:1,3.5) {H};
  \node[castor confusion labels, anchor=south] at (axis cs:2,3.5) {N};
  \node[castor confusion labels, anchor=south] at (axis cs:3,3.5) {I};
  \ifnum#5=1
  \node[castor confusion labels, anchor=east] at (axis cs:-0.5,3) {R};
  \node[castor confusion labels, anchor=east] at (axis cs:-0.5,2) {H};
  \node[castor confusion labels, anchor=east] at (axis cs:-0.5,1) {N};
  \node[castor confusion labels, anchor=east] at (axis cs:-0.5,0) {I};
  \fi
  \end{axis}}
\castorconfusion{figures/data/audit_gold.tsv}{adjudicated}{cg}{}{1}
\castorconfusion{figures/data/audit_a.tsv}{annotator A}{ca}{at={(cg.east)}, anchor=west, xshift=4mm}{0}
\castorconfusion{figures/data/audit_b.tsv}{annotator B}{cb}{at={(ca.east)}, anchor=west, xshift=4mm}{0}
\node[castor confusion labels, rotate=90, anchor=south] at ([xshift=-4mm]cg.west) {human label};
\node[castor confusion labels, anchor=north] at ([yshift=-1mm]ca.south) {judge label (columns) against human label (rows)};
\end{tikzpicture}

\vspace{2mm}
\begin{tikzpicture}
\begin{axis}[
  castor axis,
  width=0.62\textwidth, height=0.42\textwidth,
  y dir=reverse, ymin=-0.6, ymax=14.6,
  table/col sep=tab,
  ytick={0,...,14}, yticklabels from table={figures/data/llm_invalid.tsv}{label},
  ymajorgrids=false, xmajorgrids,
  xmin=0.04, xmax=0.11, xtick={0.04,0.06,0.08,0.10},
  xticklabel style={/pgf/number format/fixed},
  xlabel={share of refusal labels flagged by the validity check},
  legend style={at={(0.02,0.98)}, anchor=north west},
]
\addplot[only marks, mark=*, draw=series-main, fill=series-main, castor keep={reference}{0}]
  table[x=share, y=y, col sep=tab] {figures/data/llm_invalid.tsv};
\addlegendentry{our models}
\addplot[only marks, mark=o, draw=series-aux, mark options={fill=box}, castor keep={reference}{1}]
  table[x=share, y=y, col sep=tab] {figures/data/llm_invalid.tsv};
\addlegendentry{released references}
\end{axis}
\end{tikzpicture}
\caption{\label{fig:judge-audit}\textbf{The judge labels many human-labeled refusals as compliance and few compliant responses as refusals.}
Top: confusion matrices of the judge's label against the human label on the responses of Appendix~\ref{app:judge-pairs}, for the adjudicated labels and for each annotator's labels alone.
Rows are the human label, one of refusal, harmful compliance, non-harmful compliance and incoherent, and columns are the judge's label.
Shade is the share of the row.
Bottom: per model, the share of refusals on adversarial-harmful prompts that the validity check marks.
Filled marks are the twelve models of the first sharding setting.
Open marks are Ai2's three OLMo-2 1B models.}
\end{figure}

Both judges see a response only after a coherence check.
The check applies rules only, after the symbolic validity check of \citet{antonelli2026tttsafety}, and catches repetition, an echo of the prompt and similar patterns.
Responses it marks incoherent leave vulnerability.
A second check, the \emph{validity check}, asks Qwen3-8B whether a response is valid, with the truncation-aware prompt of \citet{antonelli2026tttsafety} plus an off-topic item of ours.
The validity check is logged and changes no label and no vulnerability.
On the twelve models of the first sharding setting, it marks \fortyoneInvalidMin{} to \fortyoneInvalidMax{} of refusals on adversarial-harmful prompts (Figure~\ref{fig:judge-audit}, lower panel).
That prompt counts a refusal as valid \citep{antonelli2026tttsafety}, so these marks do not make the refusals invalid in its sense.

One rule acts at the judge's input itself: an input too long for the judge has its response cut to fit, with the first tokens kept and the row flagged.
In the 25-recipe grid of Section~\ref{sec:setup}, one response was cut.
That response was greedy-decoded, enters no rate, and received no judge label.
A missing judge label is excluded, never counted as compliance.

To check the judge's labels, we compare them with human labels in four classes: refusal, harmful compliance, non-harmful compliance and incoherent.
Under human labels, a response is vulnerable when it is either kind of compliance.
Incoherent responses leave rates and correlations, as they do for the judge, and Appendix~\ref{app:judge-correction} gives how they count for precision.
We piloted the labeling guide on 100 responses of Ai2's OLMo-2 1B DPO model.
On this pilot, two annotators agreed at $\kappa$ \fourteenKappaRefusal{} on refusal and \fourteenKappaFour{} on the four classes.
Three rulings from the pilot entered the guide before the first audit.
The pilot's adjudication came after its aggregate judge accuracies were known, though no per-item judge label was seen and no label changed.
None of the validity check's 6 flags on pilot responses matched an annotator's incoherent label.

One guide line was added after the first audit's adjudication and used in the two later label sets.
That line is written for the harm label: an attempt at a harmful request is harmful compliance even when it helps little.
Its last clause labels a response that keeps the request's frame without its content a refusal.
Because it came after the first audit's labels, the line is post hoc.

\subsection{The first audit}\label{app:judge-first}
The first audit drew \fifteenN{} responses from 15 models: the twelve of the first sharding setting and Ai2's OLMo-2 1B SFT, DPO and Instruct.
The draw covers all seven sources of the prompt set.
Its strata cross model, source, whether WildGuard and Qwen3Guard agree, and a stratum flag.
The flag marks a response whose judge lines say both refusal and harmful, or that opens with a refusal phrase and is judged compliance or then pivots.
It also marks a response that the validity check marks.
The flag sets only the strata and never changes a label.
Responses on which the two judges disagree make up 40\,\% of the draw.
A \emph{design-weighted} estimate weights each item by its stratum's size over the number of items drawn from it, the standard stratified estimator.

Two authors labeled every response, the same two in all three label sets of this appendix.
One of them registered the paper's predictions, so we call that author the annotator who registered the prediction and the other the other annotator.
Both labeled independently, blind to model and judge.
They adjudicated every label disagreement jointly before any judge-versus-human comparison, 227 in this audit.
Before adjudication, their $\kappa$ was \fifteenKappaRefusal{} on refusal and \fifteenKappaFour{} on the four classes.
Each 95\,\% interval in both audits resamples items within strata and rescales the design weights, over 2,000 draws.

We registered a prediction of judge accuracy at least 0.90 and judge--human $\kappa$ at least 0.8.
Design-weighted accuracy is \fifteenAccW{}, with a 95\,\% interval of \fifteenAccWCI{}, and judge--human $\kappa$ is \fifteenJudgeHumanKappa{}.
Both lie below their predictions.
A separate rule asked for accuracy at least 0.85 overall and no model below 0.75.
We changed the rule from unweighted to design-weighted accuracy before any label existed, because 40\,\% of the draw comes from disagreements.
Had any model fallen below 0.75, a new judge would have relabeled every response before anything was reported.
That consequence was fixed after the pilot and before any audit label.
Table~\ref{tab:fifteen-per-model} gives each model's interval, and the lowest model's design-weighted accuracy is \fifteenWorstAccW{}.
Read as in Appendix~\ref{app:prereg}, design-weighted accuracy lies above the rule's 0.85, with no model below 0.75.
In the rule's first, unweighted form, accuracy is \fifteenAccU{} overall, and six of the 15 models lie below 0.75 (Table~\ref{tab:fifteen-per-model}).

\begin{table}[h]\centering\small
\caption{\label{tab:fifteen-per-model}\textbf{In the first audit, no model's design-weighted judge accuracy falls below 0.75.}
Each row is one model, named by $(k, s)$ as in Table~\ref{tab:base-models}, or Ai2's models by name.
Columns give design-weighted accuracy, its 95\,\% interval from 2,000 draws of items resampled within strata, unweighted accuracy, design-weighted precision, and mean vulnerability corrected for the judge's precision.
All columns cover all seven sources.
}
\begin{tabular}{lrcrrr}\toprule
Model & accuracy & interval & unweighted accuracy & precision & mean vulnerability \\ \midrule
$(1, 2)$ & 0.782 & [0.597, 0.947] & 0.731 & 0.641 & 0.301 \\
$(3, 3)$ & 0.938 & [0.892, 0.968] & 0.800 & 0.798 & 0.233 \\
$(3, 2)$ & 0.932 & [0.886, 0.963] & 0.764 & 0.857 & 0.399 \\
Ai2 OLMo-2 1B SFT & 0.902 & [0.757, 0.988] & 0.823 & 0.643 & 0.178 \\
$(3, 4)$ & 0.877 & [0.742, 0.960] & 0.760 & 0.758 & 0.346 \\
$(1, 1)$ & 0.926 & [0.884, 0.957] & 0.747 & 0.876 & 0.345 \\
$(1, 3)$ & 0.915 & [0.863, 0.960] & 0.722 & 0.893 & 0.527 \\
Ai2 OLMo-2 1B Instruct & 0.835 & [0.809, 0.858] & 0.826 & 0.754 & 0.507 \\
Ai2 OLMo-2 1B DPO & 0.837 & [0.687, 0.967] & 0.789 & 0.533 & 0.186 \\
$(1, 4)$ & 0.852 & [0.709, 0.948] & 0.662 & 0.674 & 0.242 \\
$(2, 2)$ & 0.953 & [0.920, 0.981] & 0.778 & 0.914 & 0.496 \\
$(2, 3)$ & 0.935 & [0.903, 0.963] & 0.795 & 0.837 & 0.267 \\
$(2, 1)$ & 0.787 & [0.628, 0.933] & 0.695 & 0.683 & 0.402 \\
$(3, 1)$ & 0.841 & [0.707, 0.942] & 0.734 & 0.589 & 0.222 \\
$(2, 4)$ & 0.897 & [0.801, 0.967] & 0.776 & 0.793 & 0.319 \\
 
\bottomrule\end{tabular}\end{table}

Among the models of Table~\ref{tab:fifteen-per-model}, either of two would fall below 0.75 after flipping \fifteenMinFlipsPerModel{} label, that of its heaviest item.
Overall accuracy crosses 0.85 after \fifteenFlipsEightyFive{} flips.
Judge accuracy can depend on the model judged \citep{schwinn2026coinflip}, so we compared models.
Accuracy on the twelve models minus that on Ai2's three is \fifteenTwinsMinusRefs{}, with a 95\,\% interval of \fifteenTwinsMinusRefsCI{}.
Per-model accuracies spread with overlapping intervals and no registered test, so we make no claim either way.

Where WildGuard and Qwen3Guard disagree, WildGuard is right \fifteenDisagreeAccMin{} to \fifteenDisagreeAccMax{} of the time and Qwen3Guard \fifteenQthreeGDisagreeAccMin{} to \fifteenQthreeGDisagreeAccMax{}.
Where they agree, both are right \fifteenAgreeAccMin{} to \fifteenAgreeAccMax{} of the time.
Their agreement still does not show that either is correct \citep{schwinn2026coinflip}.
Over all responses, Qwen3Guard's design-weighted accuracy is \fifteenQthreeGAccW{}, beside WildGuard's \fifteenAccW{}.
These are point estimates with overlapping intervals, and we ran no test between them.
WildGuard was chosen before the audit as the judge of T\"ulu~3 and OLMo~3 (Section~\ref{sec:setup}).
The variant with Qwen3Guard as the judge in Appendix~\ref{app:robustness} checks that choice.

By source, the judge is weakest on the two adversarial sources, at \fifteenBySourceWj{} on WildJailbreak and \fifteenBySourceHjb{} on HarmBench human jailbreaks.
Both lie below 0.85, against \fifteenBySourceOthersMin{} to \fifteenBySourceOthersMax{} on the other five sources.
These two sources supply the adversarial-harmful subset, which carries the paper's primary outcome.

On adversarial-harmful prompts, most of the judge's errors are false positives.
\emph{Precision} is the share of the judge's vulnerable calls that the annotators also call vulnerable.
\emph{Recall} is the share of the annotators' vulnerable responses that the judge calls vulnerable.
Over the twelve models, design-weighted precision is \fifteenPrecWj{} on WildJailbreak and \fifteenPrecHjb{} on HarmBench human jailbreaks.
Recall on these prompts is at least \fifteenRecallAdvMin{} for each of the twelve models.
The judge thus calls vulnerable many responses that the annotators label refusals, and it misses few.
The over-calling is not a mismatch of definitions, since WildGuard's refusal class includes redirecting refusals \citep{han2024wildguard}.

\subsection{The audit of the 25-recipe grid}\label{app:judge-grid}
The audit of the 25-recipe grid checks the judge on the 150 models of the 25-recipe grid (Section~\ref{sec:grid}), on the two adversarial sources only.
It drew \eightAuditNitems{} responses from WildJailbreak and HarmBench human jailbreaks, with strata keyed by data recipe at about 16 responses each.
Inter-annotator $\kappa$ uses all \eightAuditNitems{} responses.
Judge accuracy and judge--human $\kappa$ use the \eightAuditNscored{} responses coherent under both label sets, greedy-decoded ones included.
Rates, precision and values corrected for the judge's precision use the \eightAuditNpopulation{} sampled responses, leaving out \eightAuditNgreedy{} greedy-decoded ones.

The same two annotators labeled every response by hand, blind to the judge, and adjudicated \eightAuditNadjudicated{} rows before any comparison with the judge.
Before adjudication, they agreed at $\kappa$ \eightAuditKappaRefusal{} on refusal and \eightAuditKappaFour{} on the four classes.
Their disagreements between refusal and non-harmful compliance ran one way.
On 27 of 28, the annotator who registered the prediction said refusal.
No per-annotator split of the adjudicated labels was recorded for this audit.

Table~\ref{tab:eight-audit-judge} sets the audit of the 25-recipe grid beside the first audit on the same two sources.
On WildJailbreak, design-weighted accuracy is \eightAuditAccWj{} here, against \fifteenAccSourceWj{} in the first audit.
On HarmBench human jailbreaks, the two audits give \fifteenAccSourceHjb{} and \eightAuditAccHjb{}.
The judge thus does about as well as in the first audit, and a little worse on WildJailbreak.
We read neither value against the rule, because the rule was set on all seven sources and the first audit also lies below 0.85 on these two.
For the same reason, the two-source accuracy of \eightAuditAccW{} is not compared with the first audit's seven-source value.

\begin{table}[h]\centering\small
\caption{\label{tab:eight-audit-judge}\textbf{On the same two sources, the judge's accuracy in the audit of the 25-recipe grid is close to the first audit's.}
For each audit, columns give the number of responses, design-weighted accuracy and unweighted judge--human $\kappa$ per source.
The audit of the 25-recipe grid is on the left and the first audit on the right.
The last row gives inter-annotator $\kappa$ before adjudication, on refusal and on the four classes.}
\begin{tabular}{lrrrrrr}\toprule
 & \multicolumn{3}{c}{audit of the 25-recipe grid} & \multicolumn{3}{c}{first audit} \\
\cmidrule(lr){2-4}\cmidrule(lr){5-7} Source & $n$ & accuracy & $\kappa$ & $n$ & accuracy & $\kappa$ \\
\midrule WildJailbreak & 188 & 0.782 & 0.327 & 167 & 0.840 & 0.312 \\
HarmBench human jailbreaks & 168 & 0.844 & 0.437 & 153 & 0.844 & 0.445 \\
both sources & 356 & 0.804 & 0.379 &  &  &  \\
inter-annotator $\kappa$, refusal / four-class &  & 0.768 / 0.740 &  &  & 0.766 / 0.673 &  \\
 
\bottomrule\end{tabular}\end{table}

Table~\ref{tab:eight-audit-error} gives the direction of the judge's errors, design-weighted over both sources together.
The annotators call \eightAuditHumanRate{} of responses vulnerable, with a 95\,\% interval of \eightAuditHumanRateCI{}.
The judge calls \eightAuditJudgeRate{} \eightAuditJudgeRateCI{} of them vulnerable.
Against a precision of \eightAuditPrecision{} \eightAuditPrecisionCI{}, its recall is \eightAuditRecall{} \eightAuditRecallCI{}.
False positives make up \eightAuditFPmass{} of all responses, against \eightAuditFNmass{} for false negatives.
The judge errs here in the direction it showed on the same two sources in the first audit.

\begin{table}[h]\centering\small
\caption{\label{tab:eight-audit-error}\textbf{In the audit of the 25-recipe grid, the judge's errors are mostly false positives.}
Values are design-weighted over both sources, on \eightAuditNpopulation{} responses.
The first two rows give the vulnerable share under the judge and the adjudicated labels.
The next rows give its precision and recall, then the false-positive and false-negative shares of all responses.
Brackets are 95\,\% intervals as in Appendix~\ref{app:judge-first}.}
\begin{tabular}{lr}\toprule
Quantity & value and interval \\ \midrule
judge vulnerable rate & 0.549 [0.488, 0.614] \\
human vulnerable rate & 0.297 [0.237, 0.357] \\
precision & 0.528 [0.436, 0.615] \\
recall & 0.977 [0.956, 0.994] \\
false-positive share & 0.259 [0.203, 0.315] \\
false-negative share & 0.007 [0.002, 0.013] \\
 
\bottomrule\end{tabular}\end{table}

\subsection{Correcting vulnerability for the judge's precision}\label{app:judge-correction}
Both audits find a judge that over-calls vulnerability with high recall.
For such a judge, multiplying its rate by the precision of its vulnerable call removes roughly its false-positive share \citep{gao2026judgereliability}.
The product leaves out the false-negative share, which is small in both audits.
A vulnerable call on a response the annotators label incoherent counts against precision.
We estimate one factor per prompt source, a split of our own.
Each factor is estimated over the twelve models in the first audit and over the 150 models in the audit of the 25-recipe grid.
A factor per prompt would rest on 0 to 2 audit items.
A factor per model would rest on 20 to 24 items, with intervals reaching 0.

A per-model constant leaves a Pearson correlation unchanged, and a per-source factor moves it a little.
We therefore keep correlations, the ancestry gap and its ratio to the same-recipe-minus-cross-recipe difference on uncorrected rates, with values corrected for the judge's precision beside them.
For mean vulnerability, a per-source factor reorders data recipes only through how their vulnerability divides between the two sources.

In the first audit, the factors are \fifteenPrecWj{} for WildJailbreak and \fifteenPrecHjb{} for HarmBench human jailbreaks.
Uncorrected, the twelve models' mean vulnerability spans \fifteenLevelRawMin{} to \fifteenLevelRawMax{}.
Corrected for the judge's precision, it spans \fifteenLevelCorrMin{} to \fifteenLevelCorrMax{}.
The ancestry gap is \fifteenGapNone{} on uncorrected rates.
On rates corrected for the judge's precision, it is \fifteenGapCorr{}, with a 95\,\% interval of \fifteenGapCorrCI{} over prompts.

The audit of the 25-recipe grid estimates both factors again (Table~\ref{tab:eight-audit-factors}).
On WildJailbreak, the factor is \eightAuditPrecWj{}, with a 95\,\% interval of \eightAuditPrecWjCI{} over items resampled within strata.
On HarmBench human jailbreaks, it is \eightAuditPrecHjb{} \eightAuditPrecHjbCI{}.
One label flip moves the two factors by at most \eightAuditOneFlipWj{} and \eightAuditOneFlipHjb{}.
We registered that both factors would lie close to the first audit's.
On WildJailbreak, the factor lies within \eightAuditPrecWjMinusFifteenAbs{} of the first audit's.
On HarmBench human jailbreaks, it lies \eightAuditPrecHjbMinusFifteenAbs{} lower, and the first audit's factor sits just above the upper end of its interval.

\begin{table}[h]\centering\small
\caption{\label{tab:eight-audit-factors}\textbf{In the audit of the 25-recipe grid, the factor lies near the first audit's on WildJailbreak and lower on HarmBench human jailbreaks.}
Each factor is the design-weighted precision of the judge's vulnerable call for one source.
Its 95\,\% interval is drawn as in Appendix~\ref{app:judge-first}.
The next column gives the largest change from flipping one label.
The last column gives the first audit's factor.}
\begin{tabular}{lrrr}\toprule
Source & factor and interval & largest change, one flip & first audit \\ \midrule
WildJailbreak & 0.553 [0.428, 0.675] & 0.020 & 0.580 \\
HarmBench human jailbreaks & 0.484 [0.363, 0.601] & 0.024 & 0.602 \\
 
\bottomrule\end{tabular}\end{table}

Precision could also depend on the data recipe, so we group the data recipes.
A \emph{dataset group} contains the data recipes built from one source dataset, a grouping of ours rather than DataDecide's.
The 25 data recipes fall into five dataset groups: C4, DCLM, Dolma, Falcon and FineWeb, with the mixes of DCLM and Dolma in DCLM.
We registered a rule for dependence on the data recipe: if a permutation test across data recipes gave $p$ below \eightAuditHetAlpha{}, each dataset group would get its own factors.
The statistic is the spread of the data recipes' design-weighted precisions around the precision over all data recipes, each weighted by the data recipe's number of judge-vulnerable items.
Each permutation shuffles the data recipes assigned to judge-vulnerable items within blocks of prompt source and sampling stratum.
We fixed those blocks at code review, before any judge label was compared with a human one.
From $m = \eightAuditHetNperm{}$ random permutations, of which $b$ reach the observed statistic, we take $p = (b + 1)/(m + 1)$.
For drawn permutations, that $p$ is valid and slightly conservative \citep{phipson2010permutation}.
The statistic and the blocks are ours, without precedent.

\begin{figure}[h]\centering
\input{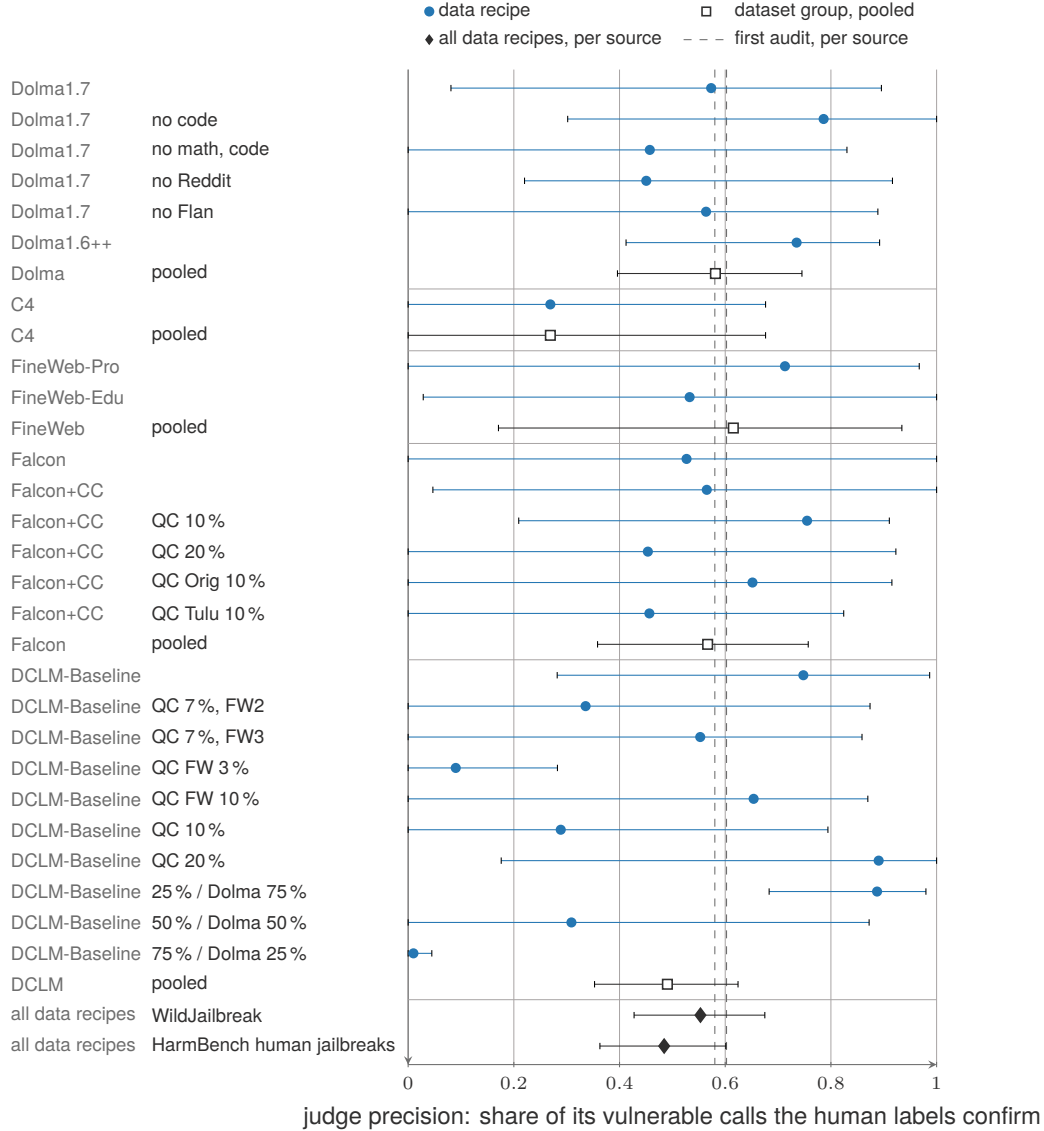}\settowidth{\castorvarw}{\castortick HarmBench human jailbreaks}%
\begin{tikzpicture}
\pgfplotstableread[col sep=tab]{figures/data/judge_precision.tsv}\jprec
\pgfplotstableread[col sep=tab]{figures/data/judge_precision_ref.tsv}\jpref
\pgfplotsset{jp marks/.style={only marks, error bars/x dir=both, error bars/x explicit,
  error bars/error bar style={line width=0.4pt}}}
\begin{axis}[
  castor axis,
  width=0.5\textwidth, height=0.86\textwidth,
  y dir=reverse, ymin=-0.6, ymax=31.6, ytick=\empty, ymajorgrids=false, xmajorgrids,
  xmin=0, xmax=1, xtick={0,0.2,0.4,0.6,0.8,1},
  xticklabel style={/pgf/number format/fixed},
  xlabel={judge precision: share of its vulnerable calls the human labels confirm},
  table/col sep=tab,
  legend style={at={(0.5,1.01)}, anchor=south, legend columns=2, /tikz/every even column/.append style={column sep=6pt}},
]
\castorrecipelabels{\jprec}{y}
\pgfplotstablegetrowsof{\jprec}%
\pgfmathtruncatemacro{\jplast}{\pgfplotsretval-1}%
\foreach \i in {0,...,\jplast}{%
  \pgfplotstablegetelem{\i}{kind}\of\jprec\edef\jk{\pgfplotsretval}%
  \def\jg{group}%
  \ifx\jk\jg
    \pgfplotstablegetelem{\i}{y}\of\jprec\edef\jy{\pgfplotsretval}%
    \edef\rulecmd{\noexpand\draw[grid, line width=0.2pt]
      ({rel axis cs:0,0}|-{axis cs:0,\jy+0.5}) -- ({rel axis cs:1,0}|-{axis cs:0,\jy+0.5});}%
    \rulecmd
  \fi
}
\pgfplotstablegetrowsof{\jpref}%
\pgfmathtruncatemacro{\jpreflast}{\pgfplotsretval-1}%
\foreach \i in {0,...,\jpreflast}{%
  \pgfplotstablegetelem{\i}{precision}\of\jpref\edef\jx{\pgfplotsretval}%
  \edef\refcmd{\noexpand\draw[muted, dashed, line width=0.4pt] (axis cs:\jx,-0.6) -- (axis cs:\jx,31.6);}%
  \refcmd
}
\addplot[jp marks, mark=*, mark size=1.6pt, draw=series-main, fill=series-main,
  error bars/error bar style={draw=series-main}, castor keep={kind}{recipe}]
  table[x=precision, y=y, x error plus expr=\thisrow{hi}-\thisrow{precision},
        x error minus expr=\thisrow{precision}-\thisrow{lo}] {figures/data/judge_precision.tsv};
\addlegendentry{data recipe}
\addplot[jp marks, mark=square*, mark size=1.8pt, draw=text, mark options={fill=box},
  error bars/error bar style={draw=text}, castor keep={kind}{group}]
  table[x=precision, y=y, x error plus expr=\thisrow{hi}-\thisrow{precision},
        x error minus expr=\thisrow{precision}-\thisrow{lo}] {figures/data/judge_precision.tsv};
\addlegendentry{dataset group, pooled}
\addplot[jp marks, mark=diamond*, mark size=2.6pt, draw=text, fill=text,
  error bars/error bar style={draw=text}, castor keep={kind}{pooled}]
  table[x=precision, y=y, x error plus expr=\thisrow{hi}-\thisrow{precision},
        x error minus expr=\thisrow{precision}-\thisrow{lo}] {figures/data/judge_precision.tsv};
\addlegendentry{all data recipes, per source}
\addlegendimage{muted, dashed, line width=0.4pt}
\addlegendentry{first audit, per source}
\end{axis}
\end{tikzpicture}
\caption{\label{fig:judge-precision}\textbf{Per data recipe, the judge's precision rests on few items, so its intervals span most of $[0, 1]$.}
Marks give the design-weighted precision of the judge's vulnerable calls against the adjudicated labels in the audit of the 25-recipe grid.
They use \eightAuditNpopulation{} responses, with greedy-decoded ones left out.
The data recipes appear in the order of the table of data recipes.
Open squares give each dataset group with its data recipes taken together, and C4 is its one data recipe.
Diamonds give the two per-source factors.
Bars are 95\,\% intervals from 2,000 draws of audit items.
Dashed lines give the first audit's factors, \fifteenPrecWj{} for WildJailbreak and \fifteenPrecHjb{} for HarmBench human jailbreaks, each line named.
The registered test of dependence on the data recipe gives $p = \eightAuditHetP{}$, so one factor per source corrects every model's rates.
The figure shows why the test cannot resolve differences between data recipes, not that there are none.
}
\end{figure}

We further registered that no dependence on the data recipe would be detectable at about 16 items per data recipe.
The observed $p$ is \eightAuditHetP{}, with the statistic at \eightAuditHetT{} against a null mean of \eightAuditHetTnull{}.
By the registered rule, one factor per source therefore corrects every model's rates.
Figure~\ref{fig:judge-precision} shows why the test resolves so little.
Per data recipe, precision runs from \eightAuditPrecRecipeMin{} to \eightAuditPrecRecipeMax{}, each on few judge-vulnerable items.
The permutation null produces a spread this wide from sampling alone.
No dependence on the data recipe is thus detectable in this audit, and that result is not evidence that the judge errs alike across data recipes.
A path of \eightAuditFlipsToCross{} label flips among \eightAuditFlipCandidates{} candidates, each chosen to lower $p$ most, would take $p$ below \eightAuditHetAlpha{}.
That path gives an upper bound on the fewest flips needed.

Uncorrected, mean vulnerability over the 25-recipe grid's models is \eightLevelRaw{}.
Corrected for the judge's precision, it is \eightLevelCorr{}, with a 95\,\% interval of \eightLevelCorrCI{} over audit items.
Per data recipe, it spans \eightLevelCorrRecipeMin{} to \eightLevelCorrRecipeMax{}, and per model \eightLevelCorrModelMin{} to \eightLevelCorrModelMax{}.
The variance shares of Section~\ref{sec:grid} rest on rates, so we recompute them on per-model mean vulnerability corrected for the judge's precision.
On those rates, the data recipe, the pretraining seed and the post-training seed take variance shares of \eightShareRecipeCorr{} \eightShareRecipeCorrCI{}, \eightSharePretrainCorr{} \eightSharePretrainCorrCI{} and \eightSharePosttrainCorr{} \eightSharePosttrainCorrCI{}.
These 95\,\% intervals come from a bootstrap over data recipes.
On uncorrected rates, the same shares are \eightShareRecipe{}, \eightSharePretrain{} and \eightSharePosttrain{}.
Appendix~\ref{app:prereg} prints both sets, the registered ones first.

Table~\ref{tab:eight-audit-contrast} carries the ancestry gap onto these rates.
Uncorrected, the ancestry gap is \eightAuditGapNone{}.
On rates corrected for the judge's precision, it is \eightAuditGap{}.
Its 95\,\% intervals over data recipes, prompts and audit items are \eightAuditGapCIrecipes{}, \eightAuditGapCIprompts{} and \eightAuditGapCIaudit{}.
We report the three intervals side by side and never combine them.
On the same rates, the ancestry gap is \eightSeedOverRecipeTimesCorr{} the same-recipe-minus-cross-recipe difference, with an interval of \eightSeedOverRecipeTimesCorrCI{} over data recipes.

\begin{table*}[t]\centering\small
\caption{\label{tab:eight-audit-contrast}\textbf{The ancestry gap on uncorrected rates and on rates corrected for the judge's precision.}
The first row is on uncorrected rates, the second on rates corrected for the judge's precision with one factor per source.
Three 95\,\% intervals, over data recipes, over prompts and over audit items, stand side by side and are never combined.
The second-last column gives the same-recipe-minus-cross-recipe difference on the same rates.
The last column gives the mean corrected correlations of MZ twins, same-recipe pairs and cross-recipe pairs on those rates.}
\begin{tabularx}{\textwidth}{@{}>{\raggedright\arraybackslash}Xrcccrc@{}}\toprule
Rates & \begin{tabular}[b]{@{}c@{}}ancestry\\gap\end{tabular} & \begin{tabular}[b]{@{}c@{}}over data\\recipes\end{tabular} & \begin{tabular}[b]{@{}c@{}}over\\prompts\end{tabular} & \begin{tabular}[b]{@{}c@{}}over audit\\items\end{tabular} & \begin{tabular}[b]{@{}c@{}}same-recipe $-$\\cross-recipe\end{tabular} & \begin{tabular}[b]{@{}c@{}}MZ / same-recipe /\\cross-recipe means\end{tabular} \\ \midrule
uncorrected & 0.173 & [0.167, 0.179] & [0.166, 0.180] & -- & 0.032 & 0.984 / 0.811 / 0.779 \\
corrected for the judge's precision & 0.168 & [0.162, 0.174] & [0.161, 0.175] & [0.144, 0.174] & 0.030 & 0.984 / 0.816 / 0.786 \\
 
\bottomrule\end{tabularx}\end{table*}

\subsection{Human labels on MZ twins and same-recipe pairs}\label{app:judge-pairs}
The correction above acts on rates, but a judge error could act on correlations.
MZ twins write near-identical responses, so one judge error repeated in both raises their correlation more than a same-recipe pair's.
The audits contain 0 to 2 items per prompt and cannot show it.
We therefore labeled responses of MZ twins and same-recipe pairs to the same prompts.

The draw takes \twentythreeNprompts{} adversarial-harmful prompts, 50 per source, sampled uniformly with no selection on any label.
Each prompt is answered by \twentythreeNtwins{} models of the first sharding setting: two base models, each with post-training seeds 1 and 2.
These four models form two pairs of MZ twins and four same-recipe pairs.
We label \twentythreeK{} of each model's 20 stored samples per prompt, \twentythreeNitems{} responses in all, of which 1 also appears in the first audit.
The same two annotators labeled every response independently, blind to model and judge, and adjudicated all \twentythreeNdisagreements{} label disagreements jointly before any comparison with the judge.
Before adjudication, their $\kappa$ was \twentythreeKappaRefusal{} on refusal and \twentythreeKappaFour{} on the four classes, lower than in both audits.

The statistic is the MZ-versus-same-recipe difference: the mean Pearson correlation of per-prompt vulnerability over pairs of MZ twins minus that over same-recipe pairs.
Human labels and WildGuard's refusal label each give it on the same responses.
The difference is not corrected for reliability, because every model has \twentythreeK{} samples per prompt under both label sets.
We registered a reading rule with three branches.
If the human interval excludes 0 and the paired interval of judge minus human includes 0, the judge does not create the difference.
If the human interval includes 0, the check is inconclusive at this size.
If the paired interval lies above 0, the judge inflates the difference by that amount, and the human difference is reported beside it.

Table~\ref{tab:twentythree-variants} gives the registered result in its first row.
Human labels give a difference of \twentythreeGapHuman{}, with a 95\,\% interval of \twentythreeGapHumanCI{} over \twentythreeNboot{} bootstrap draws of prompts.
Its parts are a mean correlation of \twentythreeRmzHuman{} for MZ twins and \twentythreeRuHuman{} for same-recipe pairs.
On the same responses, the judge gives \twentythreeRmzJudge{} and \twentythreeRuJudge{}, a difference of \twentythreeGapJudge{} \twentythreeGapJudgeCI{}.
Both label sets use the same prompt draws, so judge minus human has a paired interval.
Judge minus human is \twentythreeDiff{}, with a paired interval of \twentythreeDiffCI{}.
The human interval excludes 0 and the paired interval includes 0, which places the result in the rule's first branch on these four models.
The paired interval's upper end is as large as the human difference itself, \twentythreeDiffHi{} against \twentythreeGapHuman{}.
An inflation that large is not excluded, so the reading rests on the human interval excluding 0.

\begin{table*}[t]\centering\footnotesize
\caption{\label{tab:twentythree-variants}\textbf{The MZ-versus-same-recipe difference under the registered analysis and descriptive variants.}
The first row is the registered analysis.
The rows below apply the rule descriptively: with incoherent responses counted, per source, and with each annotator's labels alone.
The last two rows name each annotator by role, as the annotator who registered the prediction and the other annotator.
Columns give the number of prompts, the human mean correlations of MZ twins and of same-recipe pairs, and the human and judge differences.
Further columns give judge minus human and the rule's branch in words: the judge does not create the difference, or the check is inconclusive at this size.
Intervals are 95\,\% intervals over prompts, and judge minus human has a paired interval.}
{\setlength{\tabcolsep}{3pt}%
\begin{tabularx}{\textwidth}{@{}>{\raggedright\arraybackslash}Xrrrrcrcrc>{\raggedright\arraybackslash}p{1.8cm}@{}}\toprule
Variant & prompts & \begin{tabular}[b]{@{}c@{}}MZ\\twins\end{tabular} & \begin{tabular}[b]{@{}c@{}}same-\\recipe\end{tabular} & human & interval & judge & interval & \begin{tabular}[b]{@{}c@{}}judge $-$\\human\end{tabular} & interval & branch \\ \midrule
registered, incoherent left out & 100 & 0.635 & 0.479 & 0.156 & [0.036, 0.294] & 0.203 & [0.101, 0.341] & +0.047 & [\ensuremath{-}0.060, 0.157] & judge does not create it \\
incoherent counted & 100 & 0.652 & 0.493 & 0.159 & [0.037, 0.293] & 0.197 & [0.094, 0.334] & +0.039 & [\ensuremath{-}0.050, 0.143] & judge does not create it \\
HarmBench human jailbreaks & 50 & 0.686 & 0.502 & 0.184 & [0.005, 0.391] & 0.241 & [0.068, 0.453] & +0.057 & [\ensuremath{-}0.092, 0.221] & judge does not create it \\
WildJailbreak & 50 & 0.561 & 0.443 & 0.118 & [\ensuremath{-}0.057, 0.334] & 0.130 & [0.021, 0.276] & +0.012 & [\ensuremath{-}0.135, 0.148] & inconclusive at this size \\
annotator who registered the prediction & 100 & 0.699 & 0.555 & 0.144 & [0.045, 0.246] & 0.203 & [0.101, 0.341] & +0.059 & [\ensuremath{-}0.036, 0.164] & judge does not create it \\
other annotator & 100 & 0.654 & 0.530 & 0.124 & [0.007, 0.259] & 0.203 & [0.101, 0.341] & +0.079 & [\ensuremath{-}0.009, 0.173] & judge does not create it \\
 
\bottomrule\end{tabularx}}\end{table*}

The same table repeats the difference with incoherent responses counted, per source, and from each annotator's labels alone.
On WildJailbreak alone, the human interval includes 0 at 50 prompts.
Table~\ref{tab:twentythree-pairs} breaks the correlations into the six pairs.
In point estimates, both pairs of MZ twins lie above all four same-recipe pairs, though their intervals overlap.

\begin{table}[h]\centering\small
\caption{\label{tab:twentythree-pairs}\textbf{In point estimates, both pairs of MZ twins correlate above all four same-recipe pairs.}
Each row is one of the six pairs among the four models, named by $(k, s)$ as in Table~\ref{tab:base-models}.
Columns give the Pearson correlation of per-prompt vulnerability under human labels and under the judge.
Each correlation has its 95\,\% interval over prompts.}
\begin{tabular}{lllrcrc}\toprule
\multicolumn{2}{l}{Pair of models} & kind of pair & human $r$ & interval & judge $r$ & interval \\ \midrule
$(1, 1)$ & $(1, 2)$ & MZ twins & 0.645 & [0.463, 0.778] & 0.742 & [0.636, 0.820] \\
$(1, 1)$ & $(2, 1)$ & same-recipe pair & 0.458 & [0.256, 0.620] & 0.591 & [0.425, 0.727] \\
$(1, 1)$ & $(2, 2)$ & same-recipe pair & 0.536 & [0.361, 0.673] & 0.572 & [0.418, 0.701] \\
$(1, 2)$ & $(2, 1)$ & same-recipe pair & 0.422 & [0.217, 0.605] & 0.586 & [0.437, 0.716] \\
$(1, 2)$ & $(2, 2)$ & same-recipe pair & 0.499 & [0.344, 0.639] & 0.532 & [0.368, 0.683] \\
$(2, 1)$ & $(2, 2)$ & MZ twins & 0.625 & [0.479, 0.743] & 0.805 & [0.725, 0.874] \\
 
\bottomrule\end{tabular}\end{table}

A companion statistic is one minus the ratio of the same-recipe pairs' mean correlation to that of MZ twins.
We fixed it before any human label, after the judge-only dry run, and attach no reading to it.
Human labels give \twentythreeRelgapHuman{} \twentythreeRelgapHumanCI{}, and the judge gives \twentythreeRelgapJudge{} \twentythreeRelgapJudgeCI{}.
Their paired difference is \twentythreeRelgapDiff{} \twentythreeRelgapDiffCI{}.
Appendix~\ref{app:prereg} lists the companion among the analyses chosen after the data.

The two annotators differ systematically, as in the audit of the 25-recipe grid.
The other annotator labeled non-harmful compliance 322 times against 114, and incoherent 96 times against 24.
On 145 of 152 disagreements between refusal and non-harmful compliance, the annotator who registered the prediction said refusal.
Adjudication was blind to model and judge, but not to whose label was whose.
The final label matched the annotator who registered the prediction on \twentythreeAdjHarry{} of \twentythreeNdisagreements{} disagreements.
It matched the other annotator on \twentythreeAdjCoauthor{}, and neither on \twentythreeAdjNeither{}.
The labels of the annotator who registered the prediction give a difference of \twentythreeGapHarry{} \twentythreeGapHarryCI{}.
The other annotator's labels give \twentythreeGapCoauthor{} \twentythreeGapCoauthorCI{}.
Both lie below the adjudicated difference of \twentythreeGapHuman{}.

The judge's agreement with the adjudicated labels is measured on the \twentythreeJudgeN{} responses coherent under both label sets.
It reaches an accuracy of \twentythreeJudgeAcc{} and a $\kappa$ of \twentythreeJudgeKappa{}.
Against a precision of \twentythreeJudgePrec{}, its recall is \twentythreeJudgeRecall{}.
Mean vulnerability is \twentythreeJudgeRate{} under the judge, against \twentythreeHumanRate{} under human labels.
Table~\ref{tab:twentythree-judge-per-model} splits these values by model.
Figure~\ref{fig:judge-audit} shows where the errors fall.
The judge labels as compliance \fortyoneAuditRefToComp{} responses that the annotators labeled refusal, and as refusal \fortyoneAuditCompToRef{} that they labeled compliance.
This draw repeats the direction of the first audit, though no rule was registered for it.

\begin{table}[h]\centering\small
\caption{\label{tab:twentythree-judge-per-model}\textbf{The judge against the adjudicated labels on each of the four models.}
Each row is one model, named by $(k, s)$ as in Table~\ref{tab:base-models}.
Columns give the number of responses coherent under both label sets, then the judge's accuracy, $\kappa$, precision and recall against the adjudicated labels.
The last two columns give mean vulnerability under the judge and under human labels.}
\begin{tabular}{lrrrrrrr}\toprule
Model & $n$ & accuracy & $\kappa$ & precision & recall & judge vulnerability & human vulnerability \\ \midrule
$(1, 1)$ & 288 & 0.760 & 0.518 & 0.528 & 0.974 & 0.493 & 0.267 \\
$(1, 2)$ & 287 & 0.812 & 0.608 & 0.608 & 0.963 & 0.453 & 0.286 \\
$(2, 1)$ & 286 & 0.825 & 0.649 & 0.667 & 0.969 & 0.493 & 0.339 \\
$(2, 2)$ & 287 & 0.829 & 0.651 & 0.652 & 0.978 & 0.470 & 0.314 \\
 
\bottomrule\end{tabular}\end{table}

\FloatBarrier
\section{Partial reruns: checks and comparison}
\label{app:factorial}
\subsection{Building and checking the reruns}\label{app:factorial-checks}

\begin{table}[t]
  \centering\small
  \caption{\label{tab:nine-ztable}\textbf{Each rerun's final held-out loss and OLMES average, each with its $z$ against the main data recipe's three released runs.}
  Rows are reruns.
  Held-out loss is an equal-weight mean over eleven held-out sets, and the OLMES average is a macro average of ten benchmarks.
  The threshold is $z \le 2$.}
  \begin{tabular}{lrrrr}
\toprule
Rerun & Held-out loss & $z$ & OLMES average & $z$ \\
\midrule
shared-initialization rerun & 2.6543 & 0.06 & 0.5735 & 0.81 \\
shared-order rerun & 2.6480 & 0.50 & 0.5682 & 1.60 \\
first same-seed rerun & 2.6489 & 0.43 & 0.5813 & 0.36 \\
second same-seed rerun & 2.6522 & 0.20 & 0.5782 & 0.10 \\
 
\bottomrule
\end{tabular}

\end{table}

This appendix gives what Section~\ref{sec:setup} and Section~\ref{sec:ladder} leave out about the reruns: how each was built and checked, the sixteen-model comparison in full, its other outcomes and subsets, and one comparison within one stack.
Appendix~\ref{app:prereg} gives the predictions and registered checks of this appendix with their thresholds and observed values.

Each rerun is a 1B base model of the main data recipe, pretrained to the full budget on our cluster.
Its steps and tokens are in Table~\ref{tab:base-models} of Appendix~\ref{app:pipeline}.
Its code is OLMo's at a pinned commit, run with DataDecide's 1B settings.
DataDecide released no settings file, so we rebuilt ours from the code.
Apart from the partial reruns' data-order seed, our reruns change only the fields that point to our cluster.
Those fields are the data and evaluation paths, the save folder and the run name, with remote saving and logging turned off.
Each rerun records its changes (Appendix~\ref{app:nondet}).
Each partial rerun takes its initialization and data order from the pretraining seeds of two different released runs.
The shared-initialization rerun takes base model 1's initialization and base model 2's data order, and the shared-order rerun the reverse.
A same-seed rerun takes both from base model 1's pretraining seed, because its data-order setting is empty and falls back to the initialization seed.

At step 0, each rerun's initial weights equal those of the released run whose initialization it shares, in 99 of 99 tensors with a maximum difference of 0.
The first same-seed rerun's step-0 weights come from the run itself, and the partial reruns' from 10-step short test runs.
Each test run's settings file is byte-identical to its rerun's, up to the run name.
The second same-seed rerun's step 0 comes from a 50-step short test run and matches base model 1 in 99 of 99 tensors.
That test run's order hash and settings equal the first same-seed rerun's, up to the run name.

The data order is checked through its index.
The order index of every rerun with base model 1's data order is hash-identical to the order we rebuilt for that seed.
That rebuilt order was tested against DataDecide's released 150M runs, through their loss.
A language model has lower loss on its most recent training batches, so a matching order gives lower loss on its own seed's recent training batches.
On the 50 training batches before step 1,250, each released 150M model's loss under the rebuilt order of its seed is lower than under any other order by about 0.30 nats.
The shared-initialization rerun uses base model 2's data order, which has no check of its own.
Base model 2's pretraining seed has no released 150M run, so the loss test misses it.

Loss during pretraining is not a check, because in-loop losses are not reproducible from run to run.
After step 10,000, the in-loop held-out loss of the shared-initialization rerun deviates from the three released runs' mean by at most \nineTrajDevX{}.
The shared-order rerun deviates by at most \nineTrajDevY{}, and the first same-seed rerun by at most \nineTrajDevR{}.
The released runs themselves deviate from their mean by up to \nineTrajDevReleased{}.
The partial reruns thus sit slightly above the released runs' own deviation.

The last check compares each finished rerun with the released runs of its data recipe.
The \emph{seed-spread check} asks whether a rerun's final held-out loss and OLMES average each lie within $z \le 2$ of the mean of the main data recipe's three released runs.
Held-out loss is the equal-weight mean of the losses on the eleven held-out sets of OLMo's validation evaluator, which Ai2 logged for the released runs.
The equal weighting follows Paloma, which averages perplexity over domains \citep{magnusson2023paloma}.
The OLMES average is the macro average of DataDecide's ten OLMES benchmarks, each over its full split \citep{magnusson2025datadecide}.
Within that average, MMLU enters as a macro average over its 57 subtasks, as OLMES prescribes \citep{gu2024olmes}.
Our reproduction of DataDecide's released averages is consistent with their using this macro too.

For a rerun's value $x$, the check computes $z = |x - \bar m| / (\sigma\sqrt{4/3})$.
Here $\bar m$ is the released runs' mean, \nineMbarCE{} for held-out loss and \nineMbarOlmes{} for the OLMES average.
The seed standard deviation $\sigma$ is pooled over 25 data recipes with three runs each, giving 50 degrees of freedom, and is \nineSigmaCE{} and \nineSigmaOlmes{} respectively.
The factor $\sqrt{4/3}$ follows from the variance of a new run minus the mean of three, $(1 + 1/3)\sigma^2$.
The same evaluation code scored all 75 released runs and each rerun.
We fixed the rule and its threshold before evaluating any run, and we know of no precedent for either.
The second same-seed rerun enters Table~\ref{tab:nine-ztable} with its own row.
The shared-order rerun's OLMES average gives the largest $z$ in Table~\ref{tab:nine-ztable}, \nineZLargest{}.
\subsection{The sixteen-model comparison}\label{app:factorial-sixteen}

\begin{table*}[t]
  \centering\small
  \caption{\label{tab:nine-levels}\textbf{Sharing only the initialization or only the data order matches pairs sharing neither in mean corrected correlation, within the intervals over prompts.}
  Rows are kinds of pair among the sixteen models, all post-trained under the first sharding setting.
  A pair crosses the stack when one base model was pretrained on DataDecide's stack and the other on ours.
  Raw $r$ is the mean, over the model pairs of one kind, of the uncorrected Pearson correlation of per-prompt vulnerability, and corrected $r$ is the mean corrected correlation.
  Intervals are 95\,\% intervals over prompts, 1,000 draws.
  The jackknife standard error leaves one base model out.}
  \begin{tabular}{@{}>{\raggedright\arraybackslash}p{2.1cm}>{\raggedright\arraybackslash}p{5.0cm}ccrrcr@{}}
\toprule
Kind of pair & Pairs of base models & \shortstack{Crosses\\the stack} & \shortstack{Model\\pairs} & Raw $r$ & \shortstack{Corrected\\$r$} & \shortstack{95\,\% interval\\over prompts} & \shortstack{Jackknife\\SE} \\
\midrule
MZ twins & within each of the four & no & 24 & 0.932 & 0.985 & [0.983, 0.987] & 0.001 \\
initialization only & base model 1 $\times$ shared-initialization rerun; base model 2 $\times$ shared-order rerun & yes & 32 & 0.778 & 0.822 & [0.811, 0.833] & 0.050 \\
data order only & base model 1 $\times$ shared-order rerun; base model 2 $\times$ shared-initialization rerun & yes & 32 & 0.777 & 0.822 & [0.810, 0.833] & 0.037 \\
neither & base model 1 $\times$ base model 2; shared-initialization rerun $\times$ shared-order rerun & no & 32 & 0.778 & 0.823 & [0.811, 0.834] & 0.013 \\
 
\bottomrule
\end{tabular}

\end{table*}

The sixteen-model comparison uses base models 1 and 2 and the two partial reruns, each post-trained with post-training seeds 1--4.
All sixteen models were post-trained under the first sharding setting (Appendix~\ref{app:compute}), and one judge scores their answers to one prompt set.
Their prompt subset and outcome are the primary ones of Section~\ref{sec:setup}.
Among these models, every pair from two different base models enters the comparison.
By what their two base models share, these pairs fall into three kinds of pair, 32 model pairs each.
Pairs sharing only the initialization join base model 1 with the shared-initialization rerun and base model 2 with the shared-order rerun.
Pairs sharing only the data order join base model 1 with the shared-order rerun and base model 2 with the shared-initialization rerun.
Pairs sharing neither join base model 1 with base model 2, as same-recipe pairs do.
Pairs sharing neither also join the two partial reruns with each other.
Beside these, MZ twins of the four base models add 24 model pairs.

Table~\ref{tab:nine-levels} gives the mean corrected correlation of each kind of pair with a 95\,\% interval over prompts from 1,000 draws and a jackknife standard error leaving one base model out.
Pairs sharing neither reach \nineNeither{} \nineNeitherCI{}.
Pairs sharing only the initialization reach \nineInit{} \nineInitCI{}, and pairs sharing only the data order \nineOrder{} \nineOrderCI{}.
MZ twins reach \nineBoth{} \nineBothCI{}.

As in Section~\ref{sec:setup}, a base model's stack is the hardware and software it was pretrained on, DataDecide's or ours.
These kinds of pair do not cross the stack alike.
Every pair sharing only the initialization or only the data order joins a released run, pretrained on DataDecide's stack, with a rerun pretrained on ours.
No pair sharing neither crosses the stack.
A stack difference would therefore pull the means of pairs sharing only the initialization and only the data order down against pairs sharing neither.
Between those two means the pull cancels, so their difference is the one balanced on the stack.
That initialization-minus-order difference is \nineInitMinusOrder{} \nineInitMinusOrderCI{}, with a jackknife standard error of \nineInitMinusOrderJackSE{}.
The initialization-minus-neither difference is \nineInitMinusNeither{} \nineInitMinusNeitherCI{}, and the order-minus-neither difference \nineOrderMinusNeither{} \nineOrderMinusNeitherCI{}.
Their jackknife standard errors are \nineInitMinusNeitherJackSE{} and \nineOrderMinusNeitherJackSE{}.
MZ twins exceed pairs sharing neither by \nineBothMinusNeither{} \nineBothMinusNeitherCI{}.

Each of the three means between different base models averages two pairs of base models, and its two pairs correlate at different values.
The shared-initialization rerun correlates higher with every base model than the shared-order rerun does.
The kinds of pair sharing only the initialization and only the data order each hold one pair of each rerun with a released run.
The shared-initialization rerun's higher correlation therefore also cancels in their difference.
The shared-initialization rerun correlates with base model 1 at \nineBpDX{} and with base model 2 at \nineBpAX{}.
The shared-order rerun correlates with base model 1 at \nineBpDY{} and with base model 2 at \nineBpAY{}.
Base models 1 and 2 correlate at \nineBpDA{}, and the two partial reruns at \nineBpXY{}.
The jackknife standard error carries this spread between pairs of base models, and the interval over prompts carries none of it.

In these sixteen models, sharing only the initialization or only the data order matches pairs sharing neither in mean corrected correlation, within the intervals over prompts and beside the jackknife standard errors above.
The table cannot tell a small contribution from either, hidden by the stack, from none.
The registered prediction placed the mean corrected correlation of pairs with a partial rerun between those of same-seed pairs and same-recipe pairs.
That prediction was set aside after the first same-seed rerun's data and before the partial reruns' data.
Appendix~\ref{app:prereg} gives that prediction with its reading.
\subsection{Other outcomes and subsets}\label{app:factorial-robust}

\begin{table*}[t]
  \centering\small
  \caption{\label{tab:nine-robustness}\textbf{Every interval on the harmful subsets covers zero, and two intervals on the benign subset exclude zero.}
  Each row gives the three differences between means for one subset and outcome, and the first row is the primary subset and outcome.
  Outcomes are those of Appendix~\ref{app:robustness}.
  Intervals are 95\,\% intervals over prompts, 1,000 draws.}
  \setlength{\tabcolsep}{4pt}%
\begin{tabular}{@{}>{\raggedright\arraybackslash}p{2.2cm}>{\raggedright\arraybackslash}p{2.6cm}rcrcrc@{}}
\toprule
 & & \multicolumn{2}{c}{Initialization minus order} & \multicolumn{2}{c}{Initialization minus neither} & \multicolumn{2}{c}{Order minus neither} \\
\cmidrule(lr){3-4}\cmidrule(lr){5-6}\cmidrule(l){7-8}
Subset & Outcome & Difference & Interval & Difference & Interval & Difference & Interval \\
\midrule
adversarial-harmful & WildGuard's refusal label & +0.000 & [\ensuremath{-}0.008, 0.009] & \ensuremath{-}0.001 & [\ensuremath{-}0.008, 0.007] & \ensuremath{-}0.001 & [\ensuremath{-}0.008, 0.006] \\
 & incoherent counted & +0.001 & [\ensuremath{-}0.008, 0.009] & +0.000 & [\ensuremath{-}0.008, 0.007] & \ensuremath{-}0.001 & [\ensuremath{-}0.008, 0.006] \\
 & WildGuard's harmfulness label & +0.003 & [\ensuremath{-}0.007, 0.014] & +0.005 & [\ensuremath{-}0.004, 0.015] & +0.002 & [\ensuremath{-}0.007, 0.011] \\
 & Qwen3Guard as the judge & \ensuremath{-}0.001 & [\ensuremath{-}0.008, 0.007] & \ensuremath{-}0.002 & [\ensuremath{-}0.009, 0.004] & \ensuremath{-}0.002 & [\ensuremath{-}0.008, 0.005] \\
 & judges agree & +0.002 & [\ensuremath{-}0.007, 0.010] & +0.002 & [\ensuremath{-}0.006, 0.009] & +0.000 & [\ensuremath{-}0.007, 0.007] \\
plain-harmful & WildGuard's refusal label & \ensuremath{-}0.006 & [\ensuremath{-}0.016, 0.004] & +0.002 & [\ensuremath{-}0.007, 0.011] & +0.008 & [\ensuremath{-}0.001, 0.016] \\
all-harmful & WildGuard's refusal label & \ensuremath{-}0.002 & [\ensuremath{-}0.009, 0.004] & +0.000 & [\ensuremath{-}0.006, 0.006] & +0.002 & [\ensuremath{-}0.003, 0.007] \\
benign & WildGuard's refusal label & +0.012 & [0.004, 0.021] & +0.001 & [\ensuremath{-}0.008, 0.011] & \ensuremath{-}0.011 & [\ensuremath{-}0.019, \ensuremath{-}0.001] \\
 
\bottomrule
\end{tabular}

\end{table*}

The same three differences appear in Table~\ref{tab:nine-robustness} for the four other outcomes of Appendix~\ref{app:robustness} and for three other subsets.
Of those subsets, the all-harmful subset is the union of the adversarial-harmful and plain-harmful subsets.
On every harmful subset and outcome, each 95\,\% interval over prompts covers zero.
On the benign subset, two of the three intervals exclude zero.
Its initialization-minus-order difference is \nineBenignInitMinusOrder{} \nineBenignInitMinusOrderCI{}, and its order-minus-neither difference is \nineBenignOrderMinusNeither{} \nineBenignOrderMinusNeitherCI{}.
We report both without a reading, because the benign subset is not the primary subset.
As in Table~\ref{tab:nine-levels}, two of these three differences are not balanced on the stack, so Appendix~\ref{app:factorial-stack} compares pairs within one stack.
\subsection{Within one stack: the first same-seed rerun}\label{app:factorial-stack}

\begin{table*}[t]
  \centering\small
  \caption{\label{tab:nine-fifth}\textbf{Within our stack, the first same-seed rerun correlates higher with the shared-initialization rerun than with the shared-order rerun.}
  The first row is the first same-seed rerun's MZ twins.
  The first same-seed rerun's pair with base model 1 reproduces the value across the two sharding settings of Appendix~\ref{app:nondet} with separate code.
  Its pair with base model 2 shares neither and crosses the stack.
  The last three rows stay on our stack.
  Every row pairing the first same-seed rerun with another base model crosses sharding settings.
  Raw $r$ is the mean over a row's model pairs of the uncorrected Pearson correlation of per-prompt vulnerability, and corrected $r$ is the mean of their corrected correlations.
  Intervals are 95\,\% intervals over prompts, 1,000 draws.}
  \begin{tabular}{@{}>{\raggedright\arraybackslash}p{5.8cm}>{\raggedright\arraybackslash}p{4.2cm}crcr@{}}
\toprule
Pair of base models & Kind of pair & \shortstack{Crosses\\the stack} & \shortstack{Corrected\\$r$} & \shortstack{95\,\% interval\\over prompts} & Raw $r$ \\
\midrule
first same-seed rerun & MZ twins & no & 0.982 & [0.978, 0.986] & 0.919 \\
first same-seed rerun $\times$ base model 1 & same-seed pair, across sharding settings & yes & 0.840 & [0.828, 0.851] & 0.790 \\
first same-seed rerun $\times$ base model 2 & neither & yes & 0.821 & [0.807, 0.834] & 0.774 \\
first same-seed rerun $\times$ shared-initialization rerun & initialization only & no & 0.834 & [0.820, 0.846] & 0.784 \\
first same-seed rerun $\times$ shared-order rerun & data order only & no & 0.818 & [0.803, 0.832] & 0.769 \\
shared-initialization rerun $\times$ shared-order rerun & neither & no & 0.815 & [0.801, 0.828] & 0.769 \\
 
\bottomrule
\end{tabular}

\end{table*}

With the first same-seed rerun, all three differences can be taken within our stack, where the sixteen-model comparison balances the stack only in the initialization-minus-order difference.
It shares only the initialization with the shared-initialization rerun and only the data order with the shared-order rerun.
The two partial reruns share neither, and none of these three pairs crosses the stack.
The other rows of Table~\ref{tab:nine-fifth} list its MZ twins and its pairs with base models 1 and 2, and only those two pairs cross the stack.
Only the first same-seed rerun enters, because no pairs of the second same-seed rerun with the partial reruns were registered.

Its four models were post-trained under the second sharding setting, and the other sixteen in the first.
Every pair it forms with another base model therefore crosses sharding settings.
Its rows are listed by pair of base models, with the analysis's pooled means over three base models left unreported, and are not pooled into the means of Table~\ref{tab:nine-levels}.

Table~\ref{tab:nine-fifth} gives each pair's corrected correlation with its 95\,\% interval over prompts.
Within our stack, the two partial reruns correlate at \nineFifthXYref{} \nineFifthXYrefCI{}.
With the shared-initialization rerun, the first same-seed rerun correlates at \nineFifthRX{} \nineFifthRXCI{}.
With the shared-order rerun, it correlates at \nineFifthRY{} \nineFifthRYCI{}.
The initialization-minus-order difference is \nineFifthInitMinusOrder{} \nineFifthInitMinusOrderCI{}.
Both its pairs cross sharding settings, while the other two cross them on one side only.
The initialization-minus-neither difference is \nineFifthInitMinusNeither{} \nineFifthInitMinusNeitherCI{}, and the order-minus-neither difference \nineFifthOrderMinusNeither{} \nineFifthOrderMinusNeitherCI{}.
No jackknife standard error accompanies them, because leaving out the first same-seed rerun leaves no difference to compute.

The difference of \nineFifthInitMinusOrder{} rests on one shared-initialization rerun, one shared-order rerun and an interval over prompts only.
That interval carries no variation between base models.
In the sixteen-model comparison, the shared-initialization rerun correlates higher with every base model, at \nineBpAX{} to \nineBpDX{}, than the shared-order rerun does, at \nineBpAY{} to \nineBpDY{}.
That pattern predicts a positive difference here even if sharing the initialization adds nothing.
The three pairs therefore cannot separate a share of the initialization from that pattern.

The sharding setting is a second explanation, and its effect is bounded.
The initialization-minus-order difference is balanced on sharding setting, because both of its pairs cross it.
The twelve models of the released base models were also post-trained again under the second sharding setting.
The first same-seed rerun's pair with base model 1 then moved by \twentysevenRungTwoMinusEleven{}, and its pairs with base models 2 and 3 by \twentysevenRepOtherMinusEleven{}.
Appendix~\ref{app:pipeline} reports that no pair of base models moved by more than \twentysevenBpMaxAbsShift{}, below the difference of \nineFifthInitMinusOrder{}.
Those shifts were measured on pairs with the released runs and are carried over to these rows.
The partial reruns were not post-trained again under the second sharding setting.

On the other outcomes and subsets, the initialization-minus-order difference within one stack is positive, with an interval above zero, for incoherent counted, Qwen3Guard as the judge, judges agree, and the plain-harmful and all-harmful subsets.
It is negative under WildGuard's harmfulness label, with an interval covering zero, and reverses sign on the benign subset, with an interval below zero.
We report these values without a reading.

\FloatBarrier
\section{The refusal direction: layers and positions, ablation and controls}
\label{app:mechanism}
\subsection{Directions, layers and positions}\label{app:mechanism-sites}
This appendix asks whether pretraining runs that refuse different prompts also differ in how they refuse, inside the model.
Following \citet{arditi2024refusaldirection}, a model's \emph{refusal direction} is the difference between its mean activations on harmful and on harmless instructions at one layer and token position.
A prompt's \emph{projection} is its activation at that layer and position projected onto the model's unit refusal direction.
Where no other models are named, this appendix studies \emph{the twelve models}, the main data recipe's three base models, each post-trained with four post-training seeds (Appendix~\ref{app:compute}).
The first two subsections ask whether ablating the refusal direction raises vulnerability, with each model's own direction (Appendix~\ref{app:mechanism-ablation}) and with another pretraining run's direction after a rotation (Appendix~\ref{app:mechanism-transfer}).
The rest compare MZ twins with same-recipe pairs through their projections, at one fixed layer and position and under other choices (Appendix~\ref{app:mechanism-robust}), over every layer and position and on new models (Appendix~\ref{app:mechanism-sitemap}), beside a harm direction (Appendix~\ref{app:mechanism-harm}), and without the instruction's last token and against the model's output read-out (Appendix~\ref{app:mechanism-token}).
The last sets a model's projection on each prompt to that of another pretraining run's model and measures how far the model's vulnerability profile moves toward that model's (Appendix~\ref{app:mechanism-transplant}).
Appendix~\ref{app:prereg} lists every prediction of this appendix with its threshold and observed value.

Each model's refusal direction is estimated from \emph{estimation instructions}, 256 harmful and 256 harmless, and its candidates are scored on \emph{validation instructions}, 32 of each class.
The harmful instructions come from AdvBench, MaliciousInstruct and TDC2023 as \citet{arditi2024refusaldirection} release them, and the harmless ones come from Alpaca.
Both instruction pools were decontaminated against the prompt set in both directions.
An instruction was removed when its 8-grams covered more than 50\,\% of either text's tokens, or when it matched a benchmark prompt exactly.
The 50\,\% threshold on a benchmark prompt's tokens follows \citet{lambert2024tulu3}, and the reverse direction and the exact match are ours.
These two additions catch short instructions inside long jailbreak prompts.
Decontamination removed 143 of 720 harmful instructions, 10 of 18,793 in the harmless training pool and none of 6,264 in the harmless validation pool.
Our instructions depart from Arditi et al.'s in two declared ways.
We use twice their 128 estimation instructions per class, so each half in the reliability check below holds 128.
We also draw harmful validation instructions from the same three sources rather than from HarmBench, because HarmBench is part of the prompt set.

Each candidate direction is a difference of means between the two classes at one of 16 layers and 7 post-instruction positions \citep{arditi2024refusaldirection}.
Every candidate then receives that method's bypass, induce and KL scores on the validation instructions.
Arditi et al.'s filters keep a candidate whose induce score is above 0, whose KL score is below 0.1 and whose layer lies below 0.8 of the model's depth.
We apply the filters as their released code does, with an induce score of at least 0, a KL score of at most 0.1 and layers 0 to 11.
Their text admits layers 0 to 12, and we record the difference as a deviation.
No candidate passes these filters in all twelve models.
We therefore fix one layer and position in advance for all models: the candidate that passes in the most models, a rule of ours without precedent.
That candidate is layer \sevenSiteLayer{}, position \sevenSitePos{}, and was fixed before any correlation was computed.
It passes in \sevenNpassing{} of 12 models.
The same rule picks the same layer and position in the 72 models of the 24 other data recipes at post-training seed 1 in the 25-recipe grid, where it passes in 14 of 72.
Each model's own layer and position is instead its passing candidate with the lowest bypass score.
In five models, four of base model 1 and one of base model 2, no candidate passed, and the KL filter was dropped for them.
Table~\ref{tab:twentytwo-per-model} lists each model's own layer and position.
Neither rule can pick layer 13, last token, which Appendix~\ref{app:mechanism-sitemap} examines, because it lies above the layers that the filters admit.

Projections are reliable across halves of the estimation instructions.
Per model, we estimate directions from two disjoint random halves, 128 instructions per class, and correlate their projections over prompts.
The half-split design follows \citet{reblitzrichardson2026refusalslice}, who apply it to a direction's reliability rather than to projections.
After the Spearman--Brown step-up, the mean over 200 splits ranges across models from \sevenProjRelMin{} to \sevenProjRelMax{}.
Correcting for that reliability therefore barely moves any value.
\subsection{Ablation per model}\label{app:mechanism-ablation}

\begin{table}[t]
  \centering\small
  \caption{\label{tab:twentytwo-per-model}\textbf{Ablation at each model's own layer and position doubles plain-harmful vulnerability in all twelve models.} Rows: models by base model and post-training seed. The first column gives each model's own layer and position, with $^*$ where the KL filter was dropped. The next gives plain-harmful vulnerability before ablation. The last columns give it after ablation at the model's own layer and position, then at layer \sevenSiteLayer{}, position \sevenSitePos{}, each with its ratio.}
    \begin{tabular}{@{}ccccccc@{}}
    \toprule
    & & & \multicolumn{2}{c}{Own layer and position} & \multicolumn{2}{c}{Layer \sevenSiteLayer{}, position \sevenSitePos{}} \\
    \cmidrule(lr){4-5}\cmidrule(l){6-7}
    Model & Layer, position & Before & After & Ratio & After & Ratio \\
    \midrule
    $(1, 1)$ & (9, \ensuremath{-}7)$^\ast$ & 0.234 & 0.954 & 4.08$\times$ & 0.448 & 1.92$\times$ \\
$(1, 2)$ & (9, \ensuremath{-}7)$^\ast$ & 0.221 & 0.948 & 4.29$\times$ & 0.459 & 2.07$\times$ \\
$(1, 3)$ & (9, \ensuremath{-}7)$^\ast$ & 0.248 & 0.958 & 3.87$\times$ & 0.505 & 2.04$\times$ \\
$(1, 4)$ & (10, \ensuremath{-}1)$^\ast$ & 0.243 & 0.970 & 3.99$\times$ & 0.434 & 1.79$\times$ \\
$(2, 1)$ & (9, \ensuremath{-}6) & 0.284 & 0.799 & 2.81$\times$ & 0.951 & 3.35$\times$ \\
$(2, 2)$ & (11, \ensuremath{-}2) & 0.295 & 0.953 & 3.23$\times$ & 0.950 & 3.22$\times$ \\
$(2, 3)$ & (9, \ensuremath{-}1)$^\ast$ & 0.288 & 0.982 & 3.40$\times$ & 0.954 & 3.31$\times$ \\
$(2, 4)$ & (11, \ensuremath{-}2) & 0.276 & 0.921 & 3.34$\times$ & 0.923 & 3.35$\times$ \\
$(3, 1)$ & (10, \ensuremath{-}7) & 0.273 & 0.977 & 3.58$\times$ & 0.946 & 3.47$\times$ \\
$(3, 2)$ & (11, \ensuremath{-}7) & 0.268 & 0.876 & 3.27$\times$ & 0.910 & 3.40$\times$ \\
$(3, 3)$ & (10, \ensuremath{-}7) & 0.267 & 0.979 & 3.67$\times$ & 0.903 & 3.38$\times$ \\
$(3, 4)$ & (10, \ensuremath{-}7) & 0.277 & 0.976 & 3.52$\times$ & 0.943 & 3.40$\times$ \\
 
    \bottomrule
  \end{tabular}

\end{table}

Table~\ref{tab:twentytwo-per-model} reports the ablation of each model's refusal direction at its own layer and position and at layer \sevenSiteLayer{}, position \sevenSitePos{}.
The ablation edits every matrix that writes to the residual stream so that none writes to the direction \citep{arditi2024refusaldirection}.
In our models these are the embedding and the attention and MLP output matrices.
The models have no output biases and use rotary positions, so no positional-embedding matrix needs editing.
As a control, a random direction is ablated at every layer, in the same three models for both rules, one per base model.
Under our quarter rule, the random direction must change plain-harmful vulnerability by less than a quarter of that model's increase under ablation of its refusal direction.
Its largest absolute change is \twentytwoRandomMax{}.
The quarter rule and the random control have no precedent and rest on first principles.

At layer \sevenSiteLayer{}, position \sevenSitePos{}, ablation doubled plain-harmful vulnerability in \sevenCausalNdouble{} of \sevenCausalN{} models.
The two exceptions are MZ twins of base model 1, whose vulnerabilities there rose to \sevenCausalDefOneX{} and \sevenCausalDefFourX{} their values before ablation.
At their own layers and positions the same two models rise to \twentytwoDefOneX{} and \twentytwoDefFourX{}.
A model's own layer and position is not stronger in every model, however.
Two models, of base models 2 and 3, end lower after ablation at their own layer and position than at layer \sevenSiteLayer{}, position \sevenSitePos{}: 0.799 against 0.951, and 0.876 against 0.910.
Every model's vulnerability doubles under at least one rule, and only the per-model rule doubles it in all twelve.

Two models of base model 2 show how much sampling alone moves vulnerability, since their own layer and position is layer \sevenSiteLayer{}, position \sevenSitePos{}.
Both ablations in these models remove the same direction and differ only in their generated samples.
Their vulnerabilities after the two ablations are 0.953 against 0.950 and 0.921 against 0.923.

Beside the doubling rule, as registered, we report the share of incoherent or judge-error samples.
Under ablation at each model's own layer and position, that share is at most \twentytwoNanPlainMax{} on plain harmful prompts, \twentytwoNanAdvMax{} on adversarial-harmful prompts and \twentytwoNanBenignMax{} on benign prompts.
One model of base model 2, at a layer and position chosen without the KL filter, reaches \twentytwoNanPlainMax{} on plain harmful prompts. That share is above the largest of any model under ablation at layer \sevenSiteLayer{}, position \sevenSitePos{}, \sevenNanPlainMax{}.
\subsection{Another run's direction, rotated}\label{app:mechanism-transfer}

\begin{table*}[t]
  \centering\small
  \caption{\label{tab:eighteen-transfer}\textbf{Rotated, another pretraining run's refusal direction gives a relative rise of at least 0.5, and at least 0.3 above the mismatched map's, in \eighteenRotNpass{} of 12 targets.} Each row is a target, by base model and post-training seed, with its donor's base model and its plain-harmful vulnerability unablated and under its own direction. Columns then give the relative rise $T$ with 95\,\% prompt-bootstrap intervals for the same base model's direction unrotated, the donor's direction unrotated and rotated, the mismatched map, and the standardized fit, marked $^\dagger$ as added after the fit diagnostics. The last column gives point estimates without intervals for the random direction of Appendix~\ref{app:mechanism-ablation}, seed-1 targets only. The last two rows give the mean and the median. $T$ above 1 means the transported direction raised vulnerability more than the target's own, most where the own direction reached least: 1.200 at 0.799 and 1.190 at 0.876.}
  \scriptsize
\setlength{\tabcolsep}{3pt}
\begin{tabular}{@{}cccccccccc@{}}
  \toprule
  & & \multicolumn{2}{c}{Vulnerability} & \multicolumn{6}{c}{Relative rise $T$, 95\,\% interval over prompts} \\
  \cmidrule(lr){3-4}\cmidrule(l){5-10}
  Model & Donor & Before & Own & Same base & Other, unrot. & Other, rot. & Mismatched & Std.\ fit$^\dagger$ & Random \\
  \midrule
  $(1, 1)$ & 2 & 0.234 & 0.954 & \begin{tabular}[t]{@{}c@{}}1.000\\{}[0.996, 1.005]\end{tabular} & \begin{tabular}[t]{@{}c@{}}0.025\\{}[0.018, 0.031]\end{tabular} & \begin{tabular}[t]{@{}c@{}}0.787\\{}[0.770, 0.804]\end{tabular} & \begin{tabular}[t]{@{}c@{}}\ensuremath{-}0.004\\{}[\ensuremath{-}0.012, 0.003]\end{tabular} & \begin{tabular}[t]{@{}c@{}}0.842\\{}[0.827, 0.857]\end{tabular} & 0.035 \\
$(1, 2)$ & 2 & 0.221 & 0.948 & \begin{tabular}[t]{@{}c@{}}1.002\\{}[0.998, 1.006]\end{tabular} & \begin{tabular}[t]{@{}c@{}}\ensuremath{-}0.007\\{}[\ensuremath{-}0.013, 0.000]\end{tabular} & \begin{tabular}[t]{@{}c@{}}0.682\\{}[0.665, 0.698]\end{tabular} & \begin{tabular}[t]{@{}c@{}}0.104\\{}[0.093, 0.115]\end{tabular} & \begin{tabular}[t]{@{}c@{}}0.638\\{}[0.621, 0.654]\end{tabular} & -- \\
$(1, 3)$ & 2 & 0.248 & 0.958 & \begin{tabular}[t]{@{}c@{}}1.015\\{}[1.007, 1.023]\end{tabular} & \begin{tabular}[t]{@{}c@{}}0.004\\{}[\ensuremath{-}0.003, 0.011]\end{tabular} & \begin{tabular}[t]{@{}c@{}}1.008\\{}[1.001, 1.016]\end{tabular} & \begin{tabular}[t]{@{}c@{}}0.029\\{}[0.021, 0.037]\end{tabular} & \begin{tabular}[t]{@{}c@{}}0.979\\{}[0.970, 0.988]\end{tabular} & -- \\
$(1, 4)$ & 2 & 0.243 & 0.970 & \begin{tabular}[t]{@{}c@{}}0.988\\{}[0.980, 0.995]\end{tabular} & \begin{tabular}[t]{@{}c@{}}0.003\\{}[\ensuremath{-}0.004, 0.010]\end{tabular} & \begin{tabular}[t]{@{}c@{}}0.588\\{}[0.571, 0.605]\end{tabular} & \begin{tabular}[t]{@{}c@{}}0.117\\{}[0.105, 0.127]\end{tabular} & \begin{tabular}[t]{@{}c@{}}0.554\\{}[0.537, 0.571]\end{tabular} & -- \\
$(2, 1)$ & 3 & 0.284 & 0.799 & \begin{tabular}[t]{@{}c@{}}1.264\\{}[1.239, 1.293]\end{tabular} & \begin{tabular}[t]{@{}c@{}}0.004\\{}[\ensuremath{-}0.005, 0.014]\end{tabular} & \begin{tabular}[t]{@{}c@{}}1.200\\{}[1.178, 1.224]\end{tabular} & \begin{tabular}[t]{@{}c@{}}\ensuremath{-}0.117\\{}[\ensuremath{-}0.135, \ensuremath{-}0.099]\end{tabular} & \begin{tabular}[t]{@{}c@{}}1.345\\{}[1.309, 1.385]\end{tabular} & 0.048 \\
$(2, 2)$ & 3 & 0.295 & 0.953 & \begin{tabular}[t]{@{}c@{}}1.047\\{}[1.040, 1.056]\end{tabular} & \begin{tabular}[t]{@{}c@{}}\ensuremath{-}0.002\\{}[\ensuremath{-}0.011, 0.006]\end{tabular} & \begin{tabular}[t]{@{}c@{}}0.427\\{}[0.409, 0.446]\end{tabular} & \begin{tabular}[t]{@{}c@{}}0.012\\{}[0.002, 0.022]\end{tabular} & \begin{tabular}[t]{@{}c@{}}1.050\\{}[1.038, 1.062]\end{tabular} & -- \\
$(2, 3)$ & 3 & 0.288 & 0.982 & \begin{tabular}[t]{@{}c@{}}0.932\\{}[0.923, 0.941]\end{tabular} & \begin{tabular}[t]{@{}c@{}}0.025\\{}[0.018, 0.032]\end{tabular} & \begin{tabular}[t]{@{}c@{}}0.896\\{}[0.884, 0.907]\end{tabular} & \begin{tabular}[t]{@{}c@{}}\ensuremath{-}0.084\\{}[\ensuremath{-}0.098, \ensuremath{-}0.070]\end{tabular} & \begin{tabular}[t]{@{}c@{}}0.991\\{}[0.982, 1.000]\end{tabular} & -- \\
$(2, 4)$ & 3 & 0.276 & 0.921 & \begin{tabular}[t]{@{}c@{}}0.810\\{}[0.791, 0.826]\end{tabular} & \begin{tabular}[t]{@{}c@{}}0.020\\{}[0.011, 0.027]\end{tabular} & \begin{tabular}[t]{@{}c@{}}1.058\\{}[1.048, 1.069]\end{tabular} & \begin{tabular}[t]{@{}c@{}}0.039\\{}[0.028, 0.051]\end{tabular} & \begin{tabular}[t]{@{}c@{}}1.074\\{}[1.058, 1.092]\end{tabular} & -- \\
$(3, 1)$ & 1 & 0.273 & 0.977 & \begin{tabular}[t]{@{}c@{}}0.869\\{}[0.856, 0.881]\end{tabular} & \begin{tabular}[t]{@{}c@{}}\ensuremath{-}0.010\\{}[\ensuremath{-}0.017, \ensuremath{-}0.004]\end{tabular} & \begin{tabular}[t]{@{}c@{}}1.019\\{}[1.015, 1.024]\end{tabular} & \begin{tabular}[t]{@{}c@{}}0.052\\{}[0.043, 0.062]\end{tabular} & \begin{tabular}[t]{@{}c@{}}0.871\\{}[0.852, 0.889]\end{tabular} & 0.010 \\
$(3, 2)$ & 1 & 0.268 & 0.876 & \begin{tabular}[t]{@{}c@{}}1.166\\{}[1.151, 1.184]\end{tabular} & \begin{tabular}[t]{@{}c@{}}\ensuremath{-}0.008\\{}[\ensuremath{-}0.016, 0.000]\end{tabular} & \begin{tabular}[t]{@{}c@{}}1.190\\{}[1.172, 1.210]\end{tabular} & \begin{tabular}[t]{@{}c@{}}0.042\\{}[0.032, 0.053]\end{tabular} & \begin{tabular}[t]{@{}c@{}}1.043\\{}[1.016, 1.070]\end{tabular} & -- \\
$(3, 3)$ & 1 & 0.267 & 0.979 & \begin{tabular}[t]{@{}c@{}}0.994\\{}[0.991, 0.997]\end{tabular} & \begin{tabular}[t]{@{}c@{}}\ensuremath{-}0.003\\{}[\ensuremath{-}0.010, 0.004]\end{tabular} & \begin{tabular}[t]{@{}c@{}}1.014\\{}[1.010, 1.019]\end{tabular} & \begin{tabular}[t]{@{}c@{}}0.042\\{}[0.034, 0.051]\end{tabular} & \begin{tabular}[t]{@{}c@{}}0.756\\{}[0.738, 0.774]\end{tabular} & -- \\
$(3, 4)$ & 1 & 0.277 & 0.976 & \begin{tabular}[t]{@{}c@{}}1.002\\{}[0.998, 1.006]\end{tabular} & \begin{tabular}[t]{@{}c@{}}\ensuremath{-}0.018\\{}[\ensuremath{-}0.025, \ensuremath{-}0.011]\end{tabular} & \begin{tabular}[t]{@{}c@{}}0.965\\{}[0.957, 0.972]\end{tabular} & \begin{tabular}[t]{@{}c@{}}0.289\\{}[0.271, 0.305]\end{tabular} & \begin{tabular}[t]{@{}c@{}}0.515\\{}[0.484, 0.547]\end{tabular} & -- \\
\midrule Mean &  & 0.264 & 0.941 & 1.007 & 0.003 & 0.903 & 0.043 & 0.888 &  \\
Median &  & 0.270 & 0.956 & 1.001 & 0.000 & 0.986 & 0.041 & 0.925 &  \\
 
  \bottomrule
\end{tabular}

\end{table*}

Appendix~\ref{app:mechanism-ablation} removed each model's own refusal direction.
The question now is whether another pretraining run's direction does the same.
The same ablation can be applied with that direction.
Each of the twelve models, as a target, takes the direction of a \emph{donor}, the model of the next base model with the same post-training seed, with base model 1 the donor of base model 3.
Unaligned, the donor's direction at its own layer and position is near orthogonal to the target's there, with cosines of $-0.019$ to 0.032.
Following \citet{bosco2026refusalbeyondattention}, we therefore rotate it with an orthogonal Procrustes map, which cannot synthesize a direction, so a transported direction that works comes from the donor, not from the map.
The map is fitted at the donor's layer on raw, mean-centered activations of 256 harmful and 256 harmless instructions outside the estimation and validation instructions and the prompt set, at every token position but the first.
This gives 12,691 paired rows, each the two models' activations on the same instruction at the same token position.
About half of these rows are chat-template tokens that every prompt shares.
In a diagnostic fit without them, the map aligns the donor's direction with that of base model 1's targets at only 0.12 to 0.30, against 0.92 to 0.96 for the full fit.
This is consistent with the transport to base model 1's targets resting on the template positions, where every model's own layer and position lies.

The outcome is the \emph{relative rise} $T$: the rise from the unablated model in plain-harmful vulnerability under the transported direction, over that under the target's own direction.
A relative rise of 1 means that the transported direction removes refusal as fully as the target's own.
The controls are the same base model's direction at the next post-training seed, the donor's direction unrotated, and a map fitted on mismatched prompts, the last as in Bosco et al.'s Appendix A.7.
Table~\ref{tab:eighteen-transfer} gives each target.
Unrotated, the donor's direction gives relative rises of \eighteenUnrotMin{} to \eighteenUnrotMax{}, and the mismatched map a mean of \eighteenMismatchMean{}.
The unrotated direction moves plain-harmful vulnerability by \eighteenInvMin{} to \eighteenInvMax{}, below zero in 6 of 12 targets, each by at most 0.013, and never by more than the target's own direction does.
The inversion that \citet{mccann2026polymorphism} report for unrotated steering vectors across Pythia-70m seeds is therefore not seen in this ablation.
The same base model's direction, also unrotated, reaches at least the registered 0.8 in \eighteenSameNpass{} of 12 targets on point estimates, though the lowest, \eighteenSameMin{}, has an interval reaching below 0.8.
Rotated, the donor's direction meets the registered criterion, a relative rise of at least 0.5 and at least 0.3 above the mismatched map, in \eighteenRotNpass{} of 12 targets, with mean \eighteenRotMean{} and median \eighteenRotMedian{}.
The one target that misses is base model 2 at post-training seed 2, at \eighteenRotFailT{}.

The standardized fit, on activations standardized per dimension, meets the same criterion in \eighteenStdNpass{} of 12 targets and misses a different one, base model 3 at post-training seed 4.
Bosco et al. fit this way in their Appendix A.9 because a raw fit under-aligns the refusal direction.
We added that fit after the fit diagnostics and before reading any number from the donor's conditions.
On adversarial-harmful prompts, the conditions keep the order they have on plain-harmful prompts, with mean vulnerability 0.489 unablated, 0.963 under the target's own direction and 0.938 rotated, against 0.538 under the mismatched map and 0.482 unrotated.

A diagnostic added after the results shows that the relative rise does not track alignment.
Rotated directions whose best cosine with any of the target's candidate directions is about 0.5 reach relative rises of 0.90 to 1.20 in three targets of base model 2, while directions above 0.9 reach only 0.59 to 0.79 in three of base model 1.
The first half is consistent with saturation, since the target's own direction already raises plain-harmful vulnerability to \eighteenOwnPMin{}--\eighteenOwnPMax{}, so a partly aligned direction can reach the same ceiling.
Saturation does not explain the second half, where the target's own direction reaches 0.948 to 0.970.
So the donor's direction is transportable by a rotation, but not shown to be the target's own.
\subsection{Layer \sevenSiteLayer{}, position \sevenSitePos{}, under other choices}\label{app:mechanism-robust}

\begin{figure}[t]
  \centering
  \input{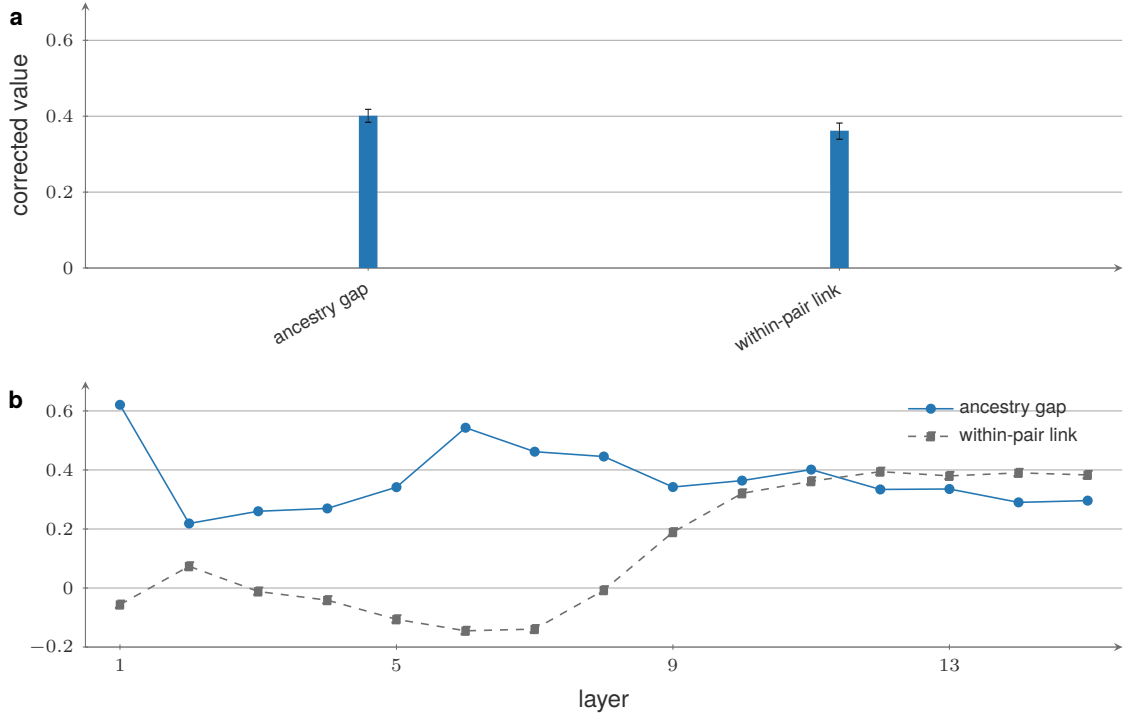}
  \caption{\label{fig:projection}\textbf{On the twelve models, the ancestry gap in projections and the within-pair link are both positive at layer \sevenSiteLayer{}, position \sevenSitePos{}.} Panel a gives both at that layer and position, with 95\,\% prompt-bootstrap intervals. Panel b gives both at each layer at position \sevenSitePos{}, as point estimates.}
\end{figure}

The analysis of the twelve models reads two statistics at layer \sevenSiteLayer{}, position \sevenSitePos{}. The first is the ancestry gap in projections, the ancestry gap computed on projections in place of vulnerability. The second is the within-pair link, defined within a same-recipe pair. It is minus the correlation over prompts between the difference of the two models' projections and the difference of their vulnerabilities, corrected for the reliability of both differences. It is positive when the model with the larger projection on a prompt is the less vulnerable one.
Figure~\ref{fig:projection} shows both statistics at layer \sevenSiteLayer{}, position \sevenSitePos{}, and across layers at that position.
At layer \sevenSiteLayer{}, position \sevenSitePos{}, the ancestry gap in projections is \sevenGap{} \sevenGapCI{}.
The within-pair link there is \sevenLink{} \sevenLinkCI{}.
The pre-registered reading rule calls a within-pair link of at least 0.5 ``runs through'', 0.3 to 0.5 ``partly'' and below 0.3 ``elsewhere'' (Appendix~\ref{app:prereg}), so this one is ``partly''.
Each check below changes one choice of that analysis, on the same twelve models.
Every interval in this subsection is a 95\,\% prompt-bootstrap interval unless the bootstrap over base models is named.
The first check moves to each model's own layer and position, where the ancestry gap in projections is \sevenOwnSiteGap{} \sevenOwnSiteGapCI{}.
The within-pair link there is \sevenOwnSiteLink{} \sevenOwnSiteLinkCI{}, which the pre-registered reading rule calls ``elsewhere''.
These values are consistent with the two models of a pair being read at different positions, since the models' own positions span $-7$ to $-1$.
The per-model rule had been expected to enlarge the ancestry gap in projections, and it shrank it.
The registered prediction on that gap's size, however, concerns layer \sevenSiteLayer{}, position \sevenSitePos{} (Appendix~\ref{app:prereg}).

A cosine read-out, which takes each activation's cosine with the unit direction, gives an ancestry gap in projections of \sevenCosReadoutGap{}.
Its within-pair link is \sevenCosReadoutLink{}.
Replacing WildGuard's refusal label with Qwen3Guard's gives a within-pair link of \sevenQthreeGLink{}.
The pre-registered reading rule calls both within-pair links ``partly''.

By prompt subset, the ancestry gap in projections exceeds that subset's ancestry gap in each of the three subsets.
On plain harmful prompts it is \sevenPlainGap{} \sevenPlainGapCI{} against \sixPlainGap{} \sixPlainGapCI{}.
On all harmful prompts it is \sevenHarmfulGap{} \sevenHarmfulGapCI{} against \sixHarmfulGap{} \sixHarmfulGapCI{}, and on benign prompts \sevenBenignGap{} \sevenBenignGapCI{} against \sixBenignGap{} \sixBenignGapCI{}.
In each subset the two intervals are disjoint.

A bootstrap over base models gives the ancestry gap in projections the interval \sevenGapCIbases{}.
That interval lies wholly above the ancestry gap on adversarial-harmful prompts, \sixGap{}.
The within-pair link's interval under the same bootstrap is \sevenLinkCIbases{}.
With only three base models, its two ends are two of the three per-pair within-pair links.
The lower end falls below 0.3.
By pair of base models, the within-pair link is \sevenLinkBpDefAuxTwo{} for base models 1 and 2, \sevenLinkBpDefAuxThree{} for 1 and 3, and \sevenLinkBpAuxTwoAuxThree{} for 2 and 3.
The mean corrected correlations in projections also vary by base model.
MZ twins of base models 1, 2 and 3 reach mean corrected correlations in projections of \sevenBpDef{}, \sevenBpAuxTwo{} and \sevenBpAuxThree{}.
Same-recipe pairs of base models 1 and 2, 1 and 3, and 2 and 3 reach \sevenBpDefAuxTwo{}, \sevenBpDefAuxThree{} and \sevenBpAuxTwoAuxThree{}.

The directions, compared without alignment, give MZ twins a mean cosine of \sevenCosMZ{}.
Those of same-recipe pairs average \sevenCosU{}, near orthogonal as in Appendix~\ref{app:mechanism-transfer}, and none exceeds \sevenCosUabsMax{} in absolute value.
Same-recipe pairs are therefore compared only through their projections of shared prompts.
Within each model, these projections also track vulnerability.
On adversarial-harmful prompts, projection and vulnerability correlate negatively in all twelve models, from \sevenValidityMin{} to \sevenValidityMax{} corrected.
\subsection{A map over layers and positions, and new models}\label{app:mechanism-sitemap}

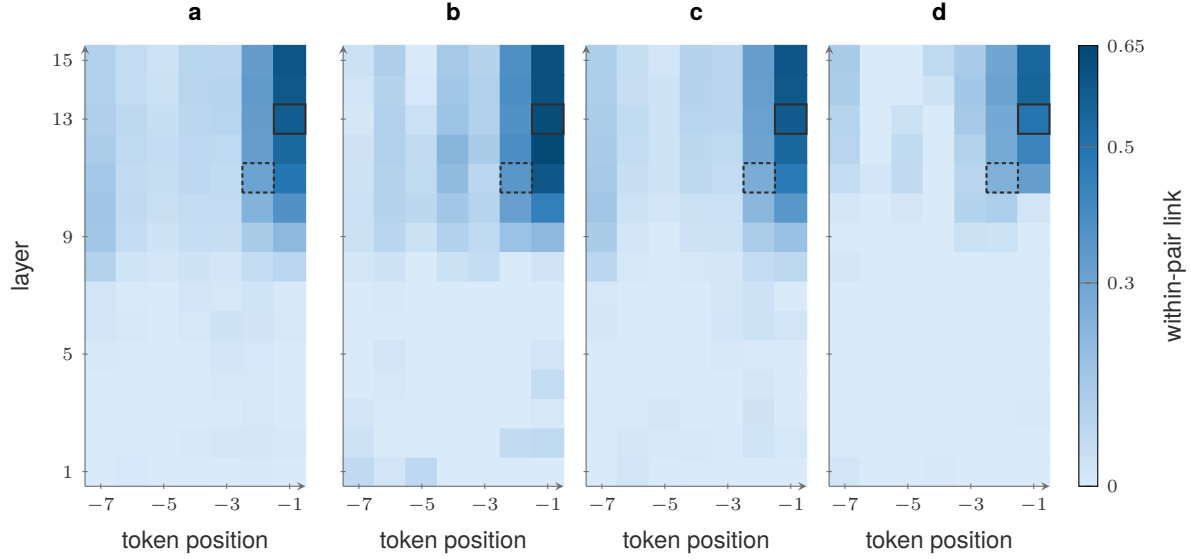
\begin{figure*}[t]
  \centering
  \pgfplotsset{castor sitemap/.style={
  castor axis,
  width=0.17\textwidth, height=0.34\textwidth, scale only axis,
  ymajorgrids=false,
  enlargelimits=false, axis on top,
  xmin=-7.5, xmax=-0.5, ymin=0.5, ymax=15.5,
  xtick={-7,-5,-3,-1}, ytick={1,5,9,13,15},
  xlabel={token position},
  colormap name=castor, point meta min=0, point meta max=0.65,
}}
\newcommand{\castorsitemap}[1]{%
  \addplot[matrix plot*, mesh/cols=7, point meta=explicit, draw=none]
    table[x=position, y=layer, meta=link, col sep=tab] {#1};
  \draw[draw=text, line width=0.8pt] (axis cs:-1.5,12.5) rectangle (axis cs:-0.5,13.5);
  \draw[draw=text, line width=0.8pt, dash pattern=on 1.6pt off 1.2pt] (axis cs:-2.5,10.5) rectangle (axis cs:-1.5,11.5);
}
\begin{tikzpicture}
\begin{axis}[castor sitemap, name=a, ylabel={layer}, title={\castorbold a}]
\castorsitemap{figures/data/site_map_A.tsv}
\end{axis}
\begin{axis}[castor sitemap, name=b, at={(a.east)}, anchor=west, xshift=5mm, title={\castorbold b},
  yticklabels={}]
\castorsitemap{figures/data/site_map_A0.tsv}
\end{axis}
\begin{axis}[castor sitemap, name=c, at={(b.east)}, anchor=west, xshift=3mm, title={\castorbold c},
  yticklabels={}]
\castorsitemap{figures/data/site_map_B.tsv}
\end{axis}
\begin{axis}[castor sitemap, name=d, at={(c.east)}, anchor=west, xshift=3mm, title={\castorbold d},
  yticklabels={},
  colorbar, colorbar style={width=2.5mm, ytick={0,0.3,0.5,0.65}, ylabel={within-pair link},
    tick label style={font=\castortick, text=text}, label style={font=\castorlabel, text=text}}]
\castorsitemap{figures/data/site_map_S300M.tsv}
\end{axis}
\end{tikzpicture}
  \caption{\label{fig:site-map}\textbf{The within-pair link rises only at the last two positions of later layers.} Each square is the corrected within-pair link of same-recipe pairs at one layer and post-instruction position. Panel a shows the \emph{confirmatory set}: the 72 same-recipe pairs of the 25-recipe grid's 24 other data recipes, post-training seed 1. Panel b shows the main data recipe's models in the 25-recipe grid. Panel c shows eight other data recipes whose MZ twins this analysis reads, and panel d the main data recipe at 300M. A solid outline marks layer 13, last token, and a dashed outline layer \sevenSiteLayer{}, position \sevenSitePos{}. One color scale spans 0--0.65. Values below 0, lowest $-0.14$, take the lightest color and lie mostly in early layers, at 300M also at positions $-6$ to $-4$ of later layers. Layer 0 is undefined.}
\end{figure*}

\begin{figure}[t]
  \centering
  \edef\linkrecipeswidth{\the\dimexpr0.5\textwidth\relax}\edef\linkrecipesheight{228pt}%
  \providecommand{\linkrecipeswidth}{\columnwidth}
\providecommand{\linkrecipesheight}{1.2\columnwidth}
\input{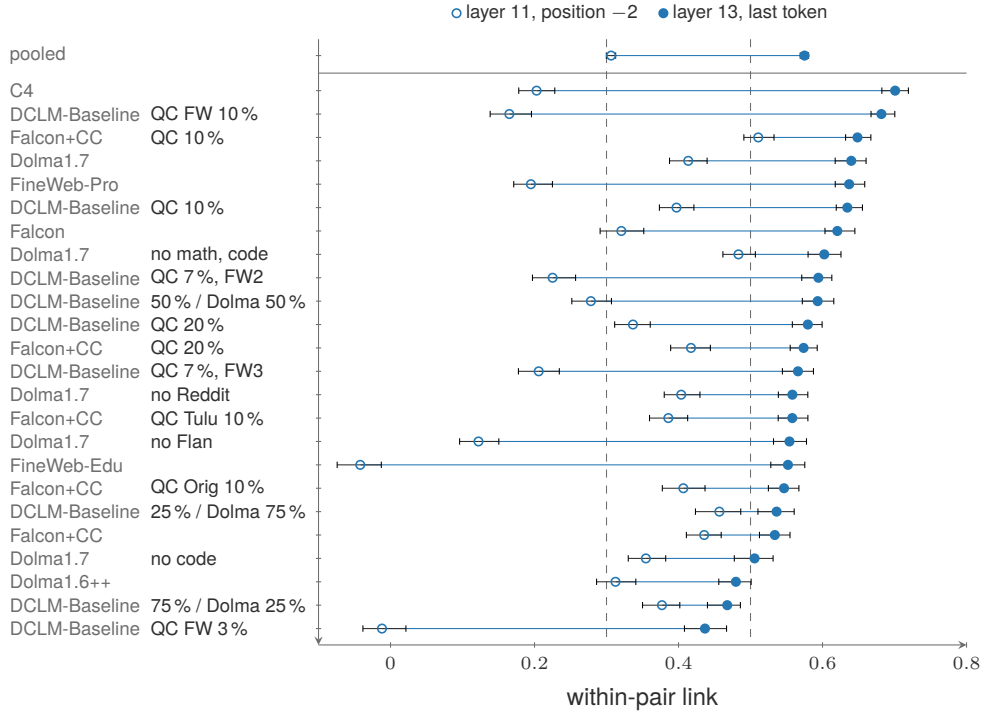}%
\pgfplotstableread[col sep=tab]{figures/data/link_recipes.tsv}\linkrecipestable%
\begin{tikzpicture}
\begin{axis}[
  castor axis,
  name=links, scale only axis,
  width=\linkrecipeswidth, height=\linkrecipesheight,
  y dir=reverse, ymin=-0.7, ymax=25.2,
  table/col sep=tab,
  ytick={0,1.5,2.5,...,24.5}, yticklabels={},
  ymajorgrids=false,
  xmin=-0.1, xmax=0.8, xtick={0,0.2,0.4,0.6,0.8},
  xlabel={within-pair link},
  legend style={at={(0.5,1.01)}, anchor=south, legend columns=2,
    /tikz/every even column/.append style={column sep=6pt}},
]
\castorrecipelabels{\linkrecipestable}{y}
\addplot[muted, dashed, line width=0.4pt, forget plot] coordinates {(0.3,-0.7) (0.3,25.2)};
\addplot[muted, dashed, line width=0.4pt, forget plot] coordinates {(0.5,-0.7) (0.5,25.2)};
\addplot[muted, line width=0.4pt, forget plot] coordinates {(-0.1,0.75) (0.8,0.75)};
\addplot[series-main, line width=0.4pt, unbounded coords=jump, forget plot]
  table[x=x, y=y, col sep=tab] {figures/data/link_recipes_join.tsv};
\addplot[only marks, mark=o, draw=series-main, mark options={fill=box},
  error bars/x dir=both, error bars/x explicit, error bars/error bar style={draw=text}]
  table[x=compare, y=y, col sep=tab, x error plus expr=\thisrow{compare_hi}-\thisrow{compare},
        x error minus expr=\thisrow{compare}-\thisrow{compare_lo}] {figures/data/link_recipes.tsv};
\addlegendentry{layer \sevenSiteLayer{}, position \sevenSitePos{}}
\addplot[only marks, mark=*, draw=series-main, fill=series-main,
  error bars/x dir=both, error bars/x explicit, error bars/error bar style={draw=text}]
  table[x=site, y=y, col sep=tab, x error plus expr=\thisrow{site_hi}-\thisrow{site},
        x error minus expr=\thisrow{site}-\thisrow{site_lo}] {figures/data/link_recipes.tsv};
\addlegendentry{layer 13, last token}
\end{axis}

\end{tikzpicture}
  \caption{\label{fig:link-recipes}\textbf{Per data recipe of the confirmatory set, the within-pair link at layer 13, last token, is at least 0.5 in 21 of the 24.} One row per data recipe, with its three same-recipe pairs, sorted by the link at layer 13, last token, and the 72 pairs pooled in the top row. Filled marks give the within-pair link at layer 13, last token, and open marks at layer \sevenSiteLayer{}, position \sevenSitePos{}. Dashed lines mark 0.3 and 0.5, the thresholds of the pre-registered reading rule. Bars are 95\,\% intervals from a bootstrap over prompts, from 200 draws per data recipe and 1,000 for the pooled row.}
\end{figure}

\begin{table*}[t]
  \centering\small
  \caption{\label{tab:twentyone-sets}\textbf{Layer 13, last token, gives a higher within-pair link than layer \sevenSiteLayer{}, position \sevenSitePos{} in every row.} Each row is one group of models and kind of pair. Columns give the number of pairs, the within-pair link at layer 13, last token, the within-pair link at layer \sevenSiteLayer{}, position \sevenSitePos{}, and their paired difference. Intervals are 95\,\% prompt-bootstrap intervals, from 1,000 draws for the confirmatory set and 200 otherwise. The first four rows are same-recipe pairs. The last four rows are all on the main data recipe. They give same-recipe pairs, then shared-initialization pairs and shared-order pairs, each pairing a partial rerun with a released base model. The last row gives the 16 pairs of the same-seed rerun with a released base model.}
    \begin{tabular}{@{}lcccc@{}}
    \toprule
    Models and pairs & Pairs & Layer 13, last token & Layer \sevenSiteLayer{}, position \sevenSitePos{} & Difference \\
    \midrule
    other data recipes, post-training seed 1 (confirmatory) & 72 & 0.575 [0.569, 0.581] & 0.306 [0.300, 0.313] & 0.269 [0.262, 0.275] \\
main data recipe, in the 25-recipe grid & 48 & 0.629 [0.614, 0.646] & 0.359 [0.333, 0.380] & 0.270 [0.241, 0.298] \\
eight data recipes, post-training seeds 1 and 2 & 96 & 0.579 [0.570, 0.587] & 0.269 [0.260, 0.278] & 0.310 [0.299, 0.319] \\
main data recipe, 300M & 48 & 0.492 [0.476, 0.515] & 0.251 [0.229, 0.278] & 0.241 [0.214, 0.267] \\
main data recipe: same-recipe pairs & 48 & 0.615 [0.603, 0.629] & 0.245 [0.225, 0.267] & 0.371 [0.348, 0.394] \\
partial reruns: shared-initialization pairs & 48 & 0.574 [0.559, 0.590] & 0.309 [0.289, 0.326] & 0.265 [0.243, 0.290] \\
partial reruns: shared-order pairs & 48 & 0.627 [0.612, 0.639] & 0.167 [0.141, 0.190] & 0.460 [0.435, 0.488] \\
same-seed rerun: pairs with a released base model & 16 & 0.574 [0.549, 0.600] & 0.192 [0.162, 0.221] & 0.382 [0.344, 0.423] \\
 
    \bottomrule
  \end{tabular}

\end{table*}

\begin{table}[t]
  \centering\footnotesize
  \caption{\label{tab:twentyone-per-corpus}\textbf{Per data recipe of the confirmatory set, layer 13, last token, gives the higher within-pair link.} Each row is one data recipe with its three same-recipe pairs, in the columns of Table~\ref{tab:twentyone-sets}. Intervals are 95\,\% prompt-bootstrap intervals from 200 draws. Rows are sorted by the within-pair link at layer 13, last token.}
    \begin{tabular}{@{}lccc@{}}
    \toprule
    Pretraining data recipe & Layer 13, last token & Layer \sevenSiteLayer{}, position \sevenSitePos{} & Difference \\
    \midrule
    C4 & 0.701 [0.682, 0.719] & 0.203 [0.178, 0.228] & 0.498 [0.473, 0.525] \\
DCLM-Baseline (QC FW 10\,\%) & 0.682 [0.667, 0.700] & 0.165 [0.138, 0.196] & 0.517 [0.488, 0.543] \\
Falcon+CC (QC 10\,\%) & 0.649 [0.632, 0.667] & 0.511 [0.491, 0.533] & 0.138 [0.114, 0.161] \\
Dolma1.7 & 0.640 [0.618, 0.660] & 0.413 [0.388, 0.440] & 0.226 [0.197, 0.256] \\
FineWeb-Pro & 0.637 [0.618, 0.659] & 0.195 [0.171, 0.225] & 0.442 [0.413, 0.473] \\
DCLM-Baseline (QC 10\,\%) & 0.634 [0.619, 0.655] & 0.397 [0.373, 0.421] & 0.237 [0.213, 0.265] \\
Falcon & 0.621 [0.603, 0.645] & 0.321 [0.291, 0.352] & 0.300 [0.270, 0.331] \\
Dolma1.7 (no math, code) & 0.602 [0.580, 0.626] & 0.483 [0.462, 0.507] & 0.119 [0.088, 0.144] \\
DCLM-Baseline (QC 7\,\%, FW2) & 0.594 [0.571, 0.613] & 0.225 [0.197, 0.257] & 0.369 [0.340, 0.398] \\
DCLM-Baseline 50\,\% / Dolma 50\,\% & 0.593 [0.572, 0.616] & 0.278 [0.252, 0.307] & 0.315 [0.284, 0.348] \\
DCLM-Baseline (QC 20\,\%) & 0.580 [0.558, 0.599] & 0.337 [0.311, 0.361] & 0.243 [0.214, 0.270] \\
Falcon+CC (QC 20\,\%) & 0.574 [0.555, 0.593] & 0.417 [0.389, 0.444] & 0.156 [0.134, 0.179] \\
DCLM-Baseline (QC 7\,\%, FW3) & 0.566 [0.544, 0.587] & 0.206 [0.178, 0.234] & 0.360 [0.328, 0.392] \\
Dolma1.7 (no Reddit) & 0.558 [0.539, 0.580] & 0.404 [0.380, 0.430] & 0.154 [0.123, 0.185] \\
Falcon+CC (QC Tulu 10\,\%) & 0.558 [0.538, 0.580] & 0.386 [0.360, 0.413] & 0.172 [0.143, 0.199] \\
Dolma1.7 (no Flan) & 0.554 [0.532, 0.578] & 0.122 [0.096, 0.150] & 0.432 [0.395, 0.469] \\
FineWeb-Edu & 0.552 [0.528, 0.575] & \ensuremath{-}0.042 [\ensuremath{-}0.074, \ensuremath{-}0.013] & 0.594 [0.561, 0.630] \\
Falcon+CC (QC Orig 10\,\%) & 0.547 [0.525, 0.567] & 0.407 [0.377, 0.437] & 0.140 [0.101, 0.172] \\
DCLM-Baseline 25\,\% / Dolma 75\,\% & 0.536 [0.510, 0.561] & 0.457 [0.424, 0.486] & 0.080 [0.053, 0.111] \\
Falcon+CC & 0.534 [0.512, 0.555] & 0.436 [0.411, 0.459] & 0.098 [0.072, 0.125] \\
Dolma1.7 (no code) & 0.506 [0.477, 0.531] & 0.355 [0.330, 0.382] & 0.151 [0.113, 0.187] \\
Dolma1.6++ & 0.480 [0.456, 0.501] & 0.312 [0.286, 0.341] & 0.167 [0.136, 0.199] \\
DCLM-Baseline 75\,\% / Dolma 25\,\% & 0.468 [0.440, 0.486] & 0.377 [0.350, 0.402] & 0.091 [0.060, 0.120] \\
DCLM-Baseline (QC FW 3\,\%) & 0.437 [0.408, 0.467] & \ensuremath{-}0.012 [\ensuremath{-}0.038, 0.021] & 0.449 [0.411, 0.486] \\
 
    \bottomrule
  \end{tabular}

\end{table}

Over the confirmatory set, the within-pair link is \twentyoneLinkCommon{} at layer \sevenSiteLayer{}, position \sevenSitePos{}, and \twentyoneLink{} \twentyoneLinkCI{} at layer 13, last token, on the same pairs.
The pre-registered reading rule calls the first ``partly'' and the second ``runs through''.
Figure~\ref{fig:site-map} maps the within-pair link of same-recipe pairs over every layer and post-instruction position.
At positions $-7$ to $-3$ the within-pair link stays low in panels a to c, at most 0.23.
From layer 9 in panels a to c and layer 11 at 300M, it rises at the last two positions.
At the last token it levels off over layers 13 to 15, at 0.575--0.601, 0.613--0.629 and 0.579--0.597 in panels a, b and c.
Only at 300M does it still rise from layer 13 to 14, from 0.492 to 0.549.
Layer 13 lies on that plateau but not at its maximum, which falls at layer 15 in panels a and c and at layer 12 in panel b.

The map on the twelve models came first and was exploratory.
On those models the within-pair link stays below 0.3 at every layer of positions $-7$ to $-3$.
At the last token in layers 11 to 15 it is \sevenSiteMapLastMin{}--\sevenSiteMapLastMax{}.
From that map we registered layer 13, last token, before the new models' data (Appendix~\ref{app:prereg}).

The same map also locates the ancestry gap in projections, which ranges from \sevenSiteMapGapMin{} to \sevenSiteMapGapMax{}.
Its eight largest values, 0.57--0.86, lie where the within-pair link is at most 0.08.
Beyond those eight, the ancestry gap in projections shows no trend with the within-pair link.
On the new models' maps, too, the ancestry gap in projections is largest where the within-pair link is near zero.
This pattern is consistent with the two quantities measuring different things.

Table~\ref{tab:twentyone-sets} compares layer 13, last token, with layer \sevenSiteLayer{}, position \sevenSitePos{} on every group of new models.
Layer 13, last token, gives the higher within-pair link in all eight rows.
At 300M its within-pair link is \twentyoneLinkThreeHundred{} \twentyoneLinkThreeHundredCI{}, below 0.5, which the pre-registered reading rule calls ``partly''.
At 150M the models have 12 layers, so layer 13 does not exist there.
Their last-token within-pair link reaches \twentyoneLinkOneFiftyLastToken{} only at the last layer.
Per data recipe of the confirmatory set, the within-pair link at layer 13, last token, is at least 0.5 in \twentyoneNrecipesRunsThrough{} of \twentyoneNrecipes{} (Table~\ref{tab:twentyone-per-corpus}).
It is 0.3 to 0.5 in the other \twentyoneNrecipesPartly{}.
Its difference from layer \sevenSiteLayer{}, position \sevenSitePos{} is positive in \twentyoneNrecipesDiffPositive{} of \twentyoneNrecipes{}.
Across data recipes its standard deviation is \twentyonePerRecipeSD{}, against \twentyonePerRecipeSDCommon{} at layer \sevenSiteLayer{}, position \sevenSitePos{}.

The ancestry gap in projections, by contrast, is smaller at layer 13, last token, than at layer \sevenSiteLayer{}, position \sevenSitePos{}.
On the main data recipe's models in the 25-recipe grid it is \twentyoneRefusalGapAZeroSite{} against \twentyoneRefusalGapAZeroCommon{}.
On the eight data recipes of panel c it is \twentyoneRefusalGapBSite{} against \twentyoneRefusalGapBCommon{}.
At layer 13, last token, the mean corrected correlation of same-recipe pairs in projections is higher, while that of MZ twins is about the same.
\subsection{The harm-direction control}\label{app:mechanism-harm}

\begin{figure*}[t]
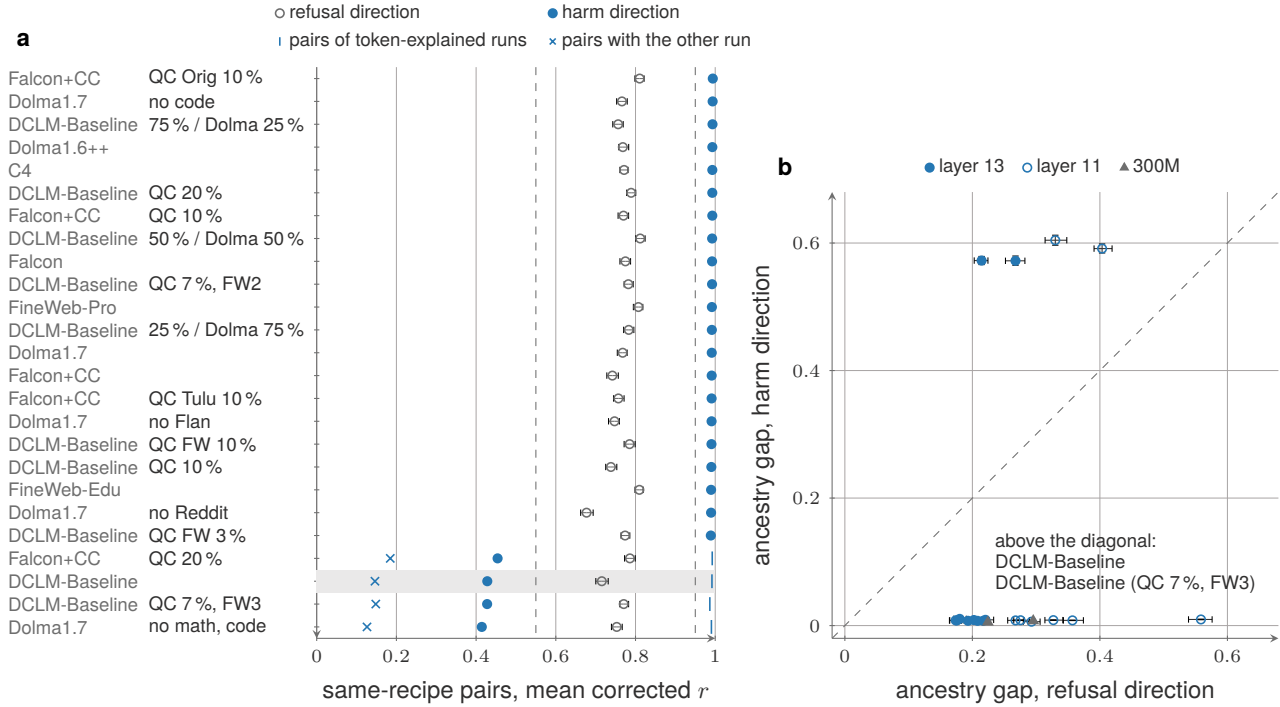

  \centering
  \providecommand{\harmrow}{8.6pt}%
\providecommand{\harmwidth}{150pt}%
\providecommand{\harmsquare}{168pt}%
\pgfplotstableread[col sep=tab]{figures/data/harm_ru.tsv}\harmrutable%
\input{figures/data/harm_labels}%
\pgfplotstablegetrowsof{\harmrutable}%
\pgfmathtruncatemacro{\harmlast}{\pgfplotsretval-1}%
\def\harmmain{-10}%
\foreach \i in {0,...,\harmlast}{%
  \pgfplotstablegetelem{\i}{main}\of\harmrutable%
  \ifnum\pgfplotsretval=1 \xdef\harmmain{\i}\fi%
}%
\input{figures/recipe_labels}%
\begin{tikzpicture}
\begin{axis}[
  castor axis,
  name=ru, scale only axis,
  width=\harmwidth, height=25*\harmrow,
  y dir=reverse, ymin=-0.5, ymax=24.5,
  table/col sep=tab,
  ytick={0,...,24}, yticklabels={},
  ymajorgrids=false, xmajorgrids,
  xmin=0.0, xmax=1.0, xtick={0,0.2,0.4,0.6,0.8,1.0},
  xticklabel style={/pgf/number format/fixed, /pgf/number format/precision=1},
  xlabel={same-recipe pairs, mean corrected $r$},
  legend style={at={(0.5,1.01)}, anchor=south, legend columns=2,
    /tikz/every even column/.append style={column sep=6pt}},
]
\fill[band] (axis cs:0.0,\harmmain-0.5) rectangle (axis cs:1.0,\harmmain+0.5);
\castorrecipelabels{\harmrutable}{y}
\addplot[muted, dashed, line width=0.4pt, forget plot] coordinates {(0.55,-0.5) (0.55,24.5)};
\addplot[muted, dashed, line width=0.4pt, forget plot] coordinates {(0.95,-0.5) (0.95,24.5)};
\addplot[only marks, mark=o, draw=series-aux, mark options={fill=box},
  error bars/x dir=both, error bars/x explicit, error bars/error bar style={draw=series-aux}]
  table[x=ref13, y=y, x error plus expr=\thisrow{ref13_hi}-\thisrow{ref13},
        x error minus expr=\thisrow{ref13}-\thisrow{ref13_lo}] {figures/data/harm_ru.tsv};
\addlegendentry{refusal direction}
\addplot[only marks, mark=*, draw=series-main, fill=series-main,
  error bars/x dir=both, error bars/x explicit, error bars/error bar style={draw=text}]
  table[x=harm13, y=y, x error plus expr=\thisrow{harm13_hi}-\thisrow{harm13},
        x error minus expr=\thisrow{harm13}-\thisrow{harm13_lo}] {figures/data/harm_ru.tsv};
\addlegendentry{harm direction}
\addplot[only marks, mark=|, mark size=2.6pt, draw=series-main, line width=0.6pt]
  table[x=harm13_tok, y=y] {figures/data/harm_ru.tsv};
\addlegendentry{pairs of token-explained runs}
\addplot[only marks, mark=x, mark size=2.2pt, draw=series-main, line width=0.6pt]
  table[x=harm13_odd, y=y] {figures/data/harm_ru.tsv};
\addlegendentry{pairs with the other run}
\end{axis}
\node[font=\castorbold, anchor=north west] at ([xshift=-\the\castorfamw-\the\castorvarw-9pt, yshift=5.8mm]ru.north west) {a};
\begin{axis}[
  castor axis,
  name=gaps, scale only axis,
  at={(ru.south east)}, anchor=south west, xshift=44pt,
  width=\harmsquare, height=\harmsquare,
  table/col sep=tab,
  xmin=-0.02, xmax=0.68, ymin=-0.02, ymax=0.68,
  xtick={0,0.2,0.4,0.6}, ytick={0,0.2,0.4,0.6},
  xticklabel style={/pgf/number format/fixed, /pgf/number format/precision=1},
  yticklabel style={/pgf/number format/fixed, /pgf/number format/precision=1},
  xlabel={ancestry gap, refusal direction}, ylabel={ancestry gap, harm direction},
  xmajorgrids,
  legend style={at={(0.5,1.01)}, anchor=south, legend columns=3,
    /tikz/every even column/.append style={column sep=5pt}},
]
\addplot[muted, dashed, line width=0.4pt, forget plot] coordinates {(-0.02,-0.02) (0.68,0.68)};
\addplot[castor keep={layer}{13}, only marks, mark=*, draw=series-main, fill=series-main,
  error bars/x dir=both, error bars/x explicit, error bars/y dir=both, error bars/y explicit,
  error bars/error bar style={draw=text}]
  table[x=refgap, y=harmgap, x error plus expr=\thisrow{refgap_hi}-\thisrow{refgap},
        x error minus expr=\thisrow{refgap}-\thisrow{refgap_lo},
        y error plus expr=\thisrow{harmgap_hi}-\thisrow{harmgap},
        y error minus expr=\thisrow{harmgap}-\thisrow{harmgap_lo}] {figures/data/harm_gaps.tsv};
\addlegendentry{layer 13}
\addplot[castor keep={layer}{11}, only marks, mark=o, draw=series-main, mark options={fill=box},
  error bars/x dir=both, error bars/x explicit, error bars/y dir=both, error bars/y explicit,
  error bars/error bar style={draw=text}]
  table[x=refgap, y=harmgap, x error plus expr=\thisrow{refgap_hi}-\thisrow{refgap},
        x error minus expr=\thisrow{refgap}-\thisrow{refgap_lo},
        y error plus expr=\thisrow{harmgap_hi}-\thisrow{harmgap},
        y error minus expr=\thisrow{harmgap}-\thisrow{harmgap_lo}] {figures/data/harm_gaps.tsv};
\addlegendentry{layer 11}
\addplot[only marks, mark=triangle*, mark size=2pt, draw=series-aux, fill=series-aux]
  table[x=refgap, y=harmgap] {figures/data/harm_gaps_300m.tsv};
\addlegendentry{300M}
\node[font=\castortick, text=text, anchor=south east, align=left] at (rel axis cs:0.97,0.08)
  {above the diagonal:\\ \harmabovelabels};
\end{axis}
\node[font=\castorbold, anchor=north west] at ([xshift=-24pt, yshift=5.8mm]gaps.north west) {b};
\end{tikzpicture}
  \caption{\label{fig:harm-control}\textbf{In harm-direction projections, the mean corrected correlation of same-recipe pairs is near 1 in 21 data recipes, and each of the four low means traces to one run.} (a) One row per data recipe, the main data recipe on a band, sorted by the harm-direction mean. The x-axis gives the mean corrected correlation of same-recipe pairs in projections. Filled marks are the harm direction at layer 13 and the last token of the instruction, and open marks the refusal direction at layer 13, last token. Intervals are 95\,\% prompt-bootstrap intervals from 200 draws, mostly narrower than the mark. Dashed lines at 0.95 and 0.55 separate the data recipes' means. On the four low rows, a cross gives pairs with the other run, \fortysixOddPairHarmRuMinThirteen{}--\fortysixOddPairHarmRuMaxThirteen{}, a bar gives pairs of the two token-explained runs, \fortysixOddTokPairHarmRuMinThirteen{}, and the dot gives their pair-weighted mean. These four rows are DCLM-Baseline, the main data recipe; DCLM-Baseline (QC 7\,\%, FW3); Dolma1.7 (no math, code); and Falcon+CC (QC 20\,\%). (b) The 9 data recipes whose MZ twins the analysis reads, the main data recipe and those of Figure~\ref{fig:site-map}c. The x-axis gives the ancestry gap in refusal-direction projections, and the y-axis the ancestry gap in harm-direction projections. Filled marks are layer 13 and open marks layer 11. Triangles are the main data recipe at 300M, never pooled. The dashed line marks equal ancestry gaps. The four points above it, two data recipes at both layers, come from the other run's pairs.}
\end{figure*}

For the harm-direction control, each model's harm direction is the same difference of means on the same instructions, read at the last token of the instruction.
That position follows \citet{zhao2025harmrefusalseparate}, who find harmfulness encoded mainly at the instruction's last token.
The harm direction is read in the layer of the refusal direction it is set beside: layer 11 beside layer \sevenSiteLayer{}, position \sevenSitePos{}, and layer 13 beside layer 13, last token.
Prompts are projected onto it and corrected as for the refusal direction.
On the twelve models, the ancestry gap in harm-direction projections is \sevenHarmGap{} \sevenHarmGapCI{}.
The mean corrected correlation of MZ twins is \sevenHarmRmz{}, and that of same-recipe pairs \sevenHarmRu{}.
This ancestry gap exceeds the ancestry gap in refusal-direction projections, \sevenGap{}.
The main data recipe's models in the 25-recipe grid reproduce it at \twentyoneHarmGapAZeroCommon{}.
On the eight other data recipes of Figure~\ref{fig:site-map}c, the ancestry gap in harm-direction projections is \twentyoneHarmGapBCommon{}, against \twentyoneRefusalGapBCommon{} in refusal-direction projections.
At 300M and layer 11, it is \twentyoneHarmGapThreeHundredCommon{}.
In absolute value, every within-pair link in harm-direction projections is at most \twentyoneHarmLinkMax{}.

Figure~\ref{fig:harm-control}a extends the comparison to 25 data recipes at layer 13.
There the mean corrected correlation of same-recipe pairs in harm-direction projections is \fortysixSharedHarmRuMinThirteen{}--\fortysixSharedHarmRuMaxThirteen{} in \fortysixNShared{} data recipes.
In the other \fortysixNSpecific{} it is \fortysixSpecificHarmRuMinThirteen{}--\fortysixSpecificHarmRuMaxThirteen{}, and no data recipe lies between.
To locate the split, we measure per pretraining run the share of a harm-direction projection's variance over prompts explained by the token that ends the instruction, written $\eta^2$.
That token explains almost all of the variance in \fortysixNRunsToken{} of \fortysixNRuns{} runs, with $\eta^2$ of \fortysixTokenEtaMinThirteen{}--\fortysixTokenEtaMaxThirteen{}.
The other \fortysixNRunsOdd{} runs fall one in each of the four data recipes with the low mean, and there the token explains \fortysixOddEtaMinThirteen{}--\fortysixOddEtaMaxThirteen{}.
In each of those four data recipes, we call that run \emph{the other run} and the remaining two the \emph{token-explained runs}.
We chose this split after seeing the data (Appendix~\ref{app:prereg}).
Same-recipe pairs of the two token-explained runs correlate at \fortysixOddTokPairHarmRuMinThirteen{} in harm-direction projections.
Pairs with the other run correlate at \fortysixOddPairHarmRuMinThirteen{}--\fortysixOddPairHarmRuMaxThirteen{}.
Each of the four low means is the pair-weighted mean of the two kinds of pairs.
With MZ twins included, the large ancestry gap in harm-direction projections comes from the other run's pairs.
Computed on those pairs it is \fortysixOddRunHarmGapMinThirteen{}, against \fortysixOddTokRunHarmGapMinThirteen{} on pairs of the token-explained runs.
At layer 13, the other run's same-recipe pairs correlate in refusal-direction projections at \fortysixMainRefusalUOddThirteen{}, against \fortysixMainRefusalUTokenThirteen{} for pairs of the token-explained runs, in the main data recipe, and within \fortysixOtherRefusalUDiffMaxThirteen{} in the other three.
The split changes no refusal-direction value reported in this appendix.

One difference between the classes lies in how instructions end: \fortysixHarmfulEndsInWord{} of the harmful estimation instructions end in a word, against \fortysixHarmlessEndsInWord{} of the harmless ones.
This difference in endings is consistent with the last token of the instruction explaining harm-direction projections in \fortysixNRunsToken{} runs.
The harm direction's projections are explained by the last token of the instruction in \fortysixNRunsToken{} of \fortysixNRuns{} runs and not in the other \fortysixNRunsOdd{}, so it cannot single out refusal.
We therefore do not use the harm direction as a comparator for the refusal direction.
As a diagnostic, projections centered within their last token give same-recipe pairs in the \fortysixNShared{} high-mean data recipes a mean corrected correlation of \fortysixSharedWithinTokenHarmRuThirteen{} for the harm direction, within the refusal direction's \fortysixSharedWithinTokenRefusalRuMinThirteen{}--\fortysixSharedWithinTokenRefusalRuMaxThirteen{}.
On the seven of them whose MZ twins are read here, the ancestry gap in centered projections is \fortysixSharedWithinTokenHarmGapMinThirteen{}--\fortysixSharedWithinTokenHarmGapMaxThirteen{} for the harm direction and \fortysixSharedWithinTokenRefusalGapMinThirteen{}--\fortysixSharedWithinTokenRefusalGapMaxThirteen{} for the refusal direction.
What the harm direction reads in the four other runs remains unknown.
\subsection{The within-pair link without the last token, and against the output read-out}\label{app:mechanism-token}

The instruction's last token, which explains harm-direction projections in \fortysixNRunsToken{} of \fortysixNRuns{} runs, also explains \seventythreeEtaProjMin{}--\seventythreeEtaProjMax{} of the variance of refusal-direction projections at layer 13, last token, in the confirmatory set's models ($\eta^2$).
Centering projections and vulnerability within each last token removes that share and leaves the within-pair link over the confirmatory set's \seventythreeNpairs{} same-recipe pairs at \seventythreeLinkCentred{} \seventythreeLinkCentredCI{}, against \seventythreeLinkRaw{} uncentered.
The same centering lowers the ancestry gap in refusal-direction projections on the main data recipe's models in the 25-recipe grid from \seventythreeGapRaw{} to \seventythreeGapCentred{} \seventythreeGapCentredCI{}.
Over the same \seventythreeNpairs{} same-recipe pairs, centering the projections alone, without vulnerability, gives a within-pair link of \seventythreeLinkProjCentred{}, with an interval of \seventythreeLinkProjCentredClusterCI{} when whole groups of prompts that share a last token are resampled.

A within-pair link that survives centering could still be the model's output decision, read early.
The \emph{logit lens} of nostalgebraist (2020), as described by \citet{belrose2023tunedlens}, reads that output at layer 13, last token, by applying the final norm and the unembedding to the residual stream.
From its logits we take the refusal score of \citet{arditi2024refusaldirection}, which tracks the same score at the model's output with per-model correlations of \seventythreeLensFinalMin{}--\seventythreeLensFinalMax{} on the estimation instructions.
This lens score has a lower within-pair link than the projection, \seventythreeLogitLink{} \seventythreeLogitLinkCI{}.
Controlling for it leaves the within-pair link of the uncentered projections at \seventythreePartial{} \seventythreePartialCI{}, against \seventythreeLinkRaw{} without the control.
Every interval except the token-group one above is a 95\,\% prompt-bootstrap interval from 200 draws.
Appendix~\ref{app:prereg} holds the registered prediction for the refusal direction within the last token.
\subsection{Another run's projection, transplanted}\label{app:mechanism-transplant}

\begin{table}[t]
  \centering\small
  \caption{\label{tab:transplant}\textbf{Setting a model's projection on each prompt to another pretraining run's covers \ninetyoneClosureAllTwelve{} of the distance from the two models' correlation to the correlation of the model with its MZ twin, over \ninetyoneNpairsAll{} directed pairs.} Rows give the mean closure, its 95\,\% prompt-bootstrap interval and the range over pairs. $^\dagger$: not registered.}
  \scriptsize
\setlength{\tabcolsep}{4pt}
\begin{tabular}{@{}lccc@{}}
  \toprule
  Condition & Mean closure & 95\,\% interval & Per-pair range \\
  \midrule
  \multicolumn{4}{@{}l}{\emph{First three pairs}} \\
\quad Transplant & 0.483 & [0.451, 0.514] & 0.426 to 0.536 \\
\quad Wrong prompts & \ensuremath{-}1.229 & [\ensuremath{-}1.392, \ensuremath{-}1.095] & \ensuremath{-}1.604 to \ensuremath{-}1.007 \\
\quad O's MZ twin as O & 0.498 & [0.466, 0.532] & 0.457 to 0.542 \\
\quad Layer 11, position $-2$ & 0.029 & [\ensuremath{-}0.019, 0.071] & \ensuremath{-}0.284 to 0.228 \\
\quad Nothing added & \ensuremath{-}0.020 & [\ensuremath{-}0.051, 0.012] & \ensuremath{-}0.028 to \ensuremath{-}0.013 \\
\midrule\multicolumn{4}{@{}l}{\emph{Nine more pairs}} \\
\quad Reversed & 0.504 & [0.470, 0.535] & 0.418 to 0.592 \\
\quad Other seeds & 0.491 & [0.466, 0.516] & 0.395 to 0.622 \\
\quad \quad Seed 3$^\dagger$ & 0.495 & [0.466, 0.526] & 0.395 to 0.622 \\
\quad \quad Seed 4$^\dagger$ & 0.487 & [0.455, 0.517] & 0.437 to 0.512 \\
\quad All nine & 0.495 & [0.472, 0.518] & 0.395 to 0.622 \\
\quad Nothing added & \ensuremath{-}0.023 & [\ensuremath{-}0.041, \ensuremath{-}0.005] & \ensuremath{-}0.048 to 0.010 \\
\midrule All twelve & 0.492 & [0.469, 0.515] & 0.395 to 0.622 \\
 
  \bottomrule
\end{tabular}

\end{table}

Apart from the ablation of Appendix~\ref{app:mechanism-ablation}, the results so far read projections beside vulnerability without setting them.
To change the projection, we edit the model's forward pass while it generates.
The edit adds a scalar multiple of the model's unit refusal direction at layer 13, last token, to the input of layer 13 at the last prompt token and at every generated token, and adds nothing at earlier positions.
Adding a vector to the forward pass at inference is the operation of \citet{turner2023actadd}, though at other positions than ours.
Like \citet{rimsky2024caa}, we add the vector at every position after the prompt with a signed multiplier, and we also add it at the prompt's last token.
The multiplier is counted in $\sigma$, the standard deviation of the model's projections over the prompt set, and $z$ is a projection standardized by that model's mean and $\sigma$.
The edited models are the main data recipe's models in the 25-recipe grid, base models 1 to 3 each with post-training seeds 1 to 4.
Their outcome is adversarial-harmful vulnerability from 20 samples per prompt, scored by the same judge as everywhere in the paper, and every correlation is corrected and taken over the same 3,212 prompts.
The samples come from this forward pass, run through hooks, not from the sampler used in the rest of the paper.
We registered that the hooked pass would reproduce stored projections to within $10^{-4}$ in raw units, and it reproduces them to within 0.084\,$\sigma$, 5.13 at most in raw units, from bf16 arithmetic on batched inputs and the attention kernel.
Adding nothing, the dose curve's $k=0$ below, bridges the two samplers.
Appendix~\ref{app:prereg} lists the other deviations from the registration.
Every interval in this subsection is a 95\,\% prompt-bootstrap interval from 1,000 draws, with the pairs of models kept fixed.

The edit is first applied as a dose curve on base model 1 with post-training seed 1, adding $k\sigma$ on every prompt for seven values of $k$ from $-2$ to $+2$.
We registered that mean vulnerability would be lower at every step up in $k$, and that its difference between $k=-0.5$ and $k=+0.5$ would be at least 0.05 with an interval clear of 0.
Mean vulnerability falls at every step, from \eightynineDoseMeanPMinusTwo{} at $k=-2$ through \eightynineDoseMeanPZero{} at $k=0$ to \eightynineDoseMeanPPlusTwo{} at $k=+2$, and the difference is \eightynineGateDiff{} \eightynineGateDiffCI{}.
At $k=0$, with nothing added, we registered a corrected correlation of at least 0.95 between the hooked pass's samples and the samples used in the rest of the paper, and it is \eightynineIdentityRc{} \eightynineIdentityRcCI{}.
In comparisons we did not register, the edited samples keep a corrected correlation with those samples of \eightynineDoseRcMinusHalf{} at $k=-0.5$ and \eightynineDoseRcPlusHalf{} at $k=+0.5$, and only \eightynineDoseRcMinusTwo{} at $k=-2$ and \eightynineDoseRcPlusTwo{} at $k=+2$.
Over half a $\sigma$, the edit thus shifts mean vulnerability while keeping the per-prompt profile: a slope, not a step.

The dose curve adds one multiple to every prompt, and the \emph{transplant} gives each prompt its own.
On each prompt it sets the $z$ of the edited model $M$ to that of $O$, the model of another pretraining run with the same post-training seed, so that $M$ and $O$ form a same-recipe pair.
It does so by adding $\sigma_M(z_O - z_M)$ along $M$'s direction, where $z_O$ comes from $O$'s direction, mean and $\sigma$, so $O$'s direction never enters $M$.
We write $M'$ for the edited $M$.
This edit is a variant of the interchange intervention of \citet{geiger2021causalabstraction} on two counts: the value moves between two models on one prompt rather than between two inputs, and it is one standardized scalar along $M$'s direction rather than a whole representation.
Causal mediation analysis \citep{vig2020mediation} sets an internal quantity by intervention and asks how much of a behavioral difference goes with it, and the \emph{closure} measures that share here:
\[
  \text{closure} = \frac{r(M', O) - r(M, O)}{r(M, M_{\mathrm{MZ}}) - r(M, O)},
\]
where $M_{\mathrm{MZ}}$ is $M$'s MZ twin and every $r$ is a corrected correlation.
The closure is thus the share of the distance from the same-recipe pair's correlation to the correlation of $M$ with its MZ twin that the edit covers.

The first three pairs, all at post-training seed 1, give base model 1 the values of base model 2, base model 2 those of base model 3, and base model 3 those of base model 1.
Their offsets $z_O - z_M$ have a standard deviation of \eightynineOffsetSDMin{}--\eightynineOffsetSDMax{} across prompts, inside the doses of the curve above.
Against correlations of 0.983--0.985 between $M$ and its MZ twin, the edit raises the same-recipe pairs' correlations from \eightyninePairRcMin{}--\eightyninePairRcMax{} for $r(M, O)$ to \eightynineEditRcMin{}--\eightynineEditRcMax{} for $r(M', O)$.
We registered a closure above 0 with an interval clear of 0, and the closure is \eightynineClosureTransplant{} \eightynineClosureTransplantCI{}, with single pairs from \eightynineClosureTransplantPairMin{} to \eightynineClosureTransplantPairMax{} (Table~\ref{tab:transplant}).

Registered controls on the same three pairs ask which parts of the transplant this closure needs.
The first gives $O$'s values to the wrong prompts by a derangement, which keeps their distribution and breaks only their assignment.
We registered a closure of at most 0 for it, and it is \eightynineClosurePermuted{} \eightynineClosurePermutedCI{}, so the wrong assignment pulls $M$ away from $O$.
The second takes the values from $O$'s MZ twin, its post-training seed 2, in place of $O$, registered to land within 0.10 of the transplant.
Its closure is \eightynineClosureMzDonor{} \eightynineClosureMzDonorCI{}, a difference of \eightynineMzDonorMinusTransplant{} \eightynineMzDonorMinusTransplantCI{}.
The third repeats the transplant at layer \sevenSiteLayer{}, position \sevenSitePos{}, fixed in advance in Appendix~\ref{app:mechanism-sites}, with that layer and position's direction, mean and $\sigma$, registered to close less than the transplant.
Its closure is \eightynineClosureSiteEleven{} \eightynineClosureSiteElevenCI{}, and the transplant's closure minus this one is \eightynineTransplantMinusSiteEleven{} \eightynineTransplantMinusSiteElevenCI{}.
Its three pairs differ in sign, from \eightynineClosureSiteElevenPairMin{} to \eightynineClosureSiteElevenPairMax{}, so we read nothing in it beyond its mean.
The fourth control, nothing added, was registered only as the bridge between the samplers, with no prediction on its closure, and its values follow with the nine more pairs.

Beyond these controls, four secondary measures registered for the same pairs describe the transplant.
The closure measured against $O$'s MZ twin instead of $O$ is \eightynineClosureVsTwinTransplant{} \eightynineClosureVsTwinTransplantCI{}, which is consistent with most of the closure not being $O$'s sampling noise.
The edited model keeps most of its profile, with $r(M', M)$ of \eightynineSelfTransplantMin{}--\eightynineSelfTransplantMax{}.
Its mean vulnerability moves by at most 0.008, since the mean offset is zero by construction, and with the values on the wrong prompts it is lower by 0.013--0.019.

The nine more directed pairs, registered before any of their arms ran, repeat the transplant with the same edit, samples and judge.
They are the first three pairs reversed at post-training seed 1, and the first three again at post-training seeds 3 and 4, where $M$'s MZ twin is its model at the other of these two seeds.
Their offsets have a standard deviation of \ninetyoneOffsetSDMin{}--\ninetyoneOffsetSDMax{}, inside the doses of the curve, and, not registered, their mean vulnerability moves by at most 0.014.
We registered a closure within 0.10 of the first three's 0.483 with an interval clear of 0, and the nine give \ninetyoneClosureNewNine{} \ninetyoneClosureNewNineCI{}.
The same threshold was registered for the three reversed pairs, which give \ninetyoneClosureReversed{} \ninetyoneClosureReversedCI{}, and for the six at other seeds, which give \ninetyoneClosureOtherSeed{} \ninetyoneClosureOtherSeedCI{}.
Not registered, of those six, seed 3 gives \ninetyoneClosureSeedThree{} \ninetyoneClosureSeedThreeCI{} and seed 4 gives \ninetyoneClosureSeedFour{} \ninetyoneClosureSeedFourCI{}.
Not registered either, single pairs of the nine range from \ninetyoneClosurePairMin{} to \ninetyoneClosurePairMax{}, and the nine minus the first three is \ninetyoneNewNineMinusEightyNine{} \ninetyoneNewNineMinusEightyNineCI{}.
We registered a corrected correlation of at least 0.95 between each model's hooked samples with nothing added and its samples in the rest of the paper, and they correlate at \ninetyoneIdentityRecordMin{}--\ninetyoneIdentityRecordMax{}.
With nothing added, the closure, not a registered test, is \eightynineClosureIdentity{} \eightynineClosureIdentityCI{} in the first three pairs and \ninetyoneIdentityNewNine{} \ninetyoneIdentityNewNineCI{} in the nine, so measured against the samples with nothing added instead of the unedited model the closure would be about 0.02 larger.

Pooled with the first three, these nine give the record of \ninetyoneNpairsAll{} directed pairs, whose closure is \ninetyoneClosureAllTwelve{} \ninetyoneClosureAllTwelveCI{}.
Because the bootstrap keeps the pairs fixed, that interval is narrower than the spread over single pairs.
In this intervention, then, about half of the distance from a same-recipe pair's correlation to an MZ twin's, \ninetyoneClosureAllTwelve{}, moves with one value per prompt, the projection at layer 13, last token.
The mean stays near that value for the reversed pairs and at each of three post-training seeds, while single pairs can differ with which model of the pair is edited, which we did not register: base model 3 given base model 2's values closes 0.592, and base model 2 given base model 3's closes 0.485.
These results cover one data recipe at 1B parameters, one layer and position, a single scalar per prompt, and only the edited model's direction, never the other run's.

\FloatBarrier
\section{Robustness checks and further results on vulnerability}
\label{app:robustness}
This appendix gives the checks on the ancestry gap that the body only points to, and two further results that Section~\ref{sec:limitations} draws on.
In order, it covers the twelve models of Section~\ref{sec:setup}, the 25-recipe grid, model sizes, which jailbreaks are stable, and how many base models and post-training seeds a comparison needs.
Unless a passage says otherwise, every correlation is a corrected correlation on the adversarial-harmful subset under WildGuard's refusal label.
Every rate, such as mean vulnerability, is the judge's, and Appendix~\ref{app:judge} gives these rates corrected for the judge's precision.

\subsection{One data recipe, twelve models}\label{app:robustness-twelve}

\begin{table}[h]
  \centering\footnotesize
  \caption{\label{tab:six-robustness}\textbf{The ancestry gap of the twelve models lies above zero under every outcome and in every prompt subset.}
  Rows give a prompt subset and an outcome, for the twelve models of the first sharding setting.
  \rMZ{} and \rU{} are the mean corrected correlations of MZ twins and same-recipe pairs, and the ancestry gap is their difference.
  Intervals are 95\,\% intervals from \sixNboot{} draws, over prompts and over the three base models.
  Over base models, \sixNbaseDraws{} of the \sixNboot{} draws define a same-recipe pair, and only these enter the interval.}
  \begin{tabular}{llrrrll}
\toprule
Subset & Outcome & \rMZ{} & \rU{} & \shortstack{Ancestry\\gap} & \shortstack{95\,\% interval\\over prompts} & \shortstack{95\,\% interval\\over base models} \\
\midrule
adversarial-harmful & WildGuard's refusal label & 0.984 & 0.811 & 0.173 & [0.162, 0.185] & [0.155, 0.195] \\
adversarial-harmful & incoherent counted & 0.984 & 0.811 & 0.173 & [0.162, 0.185] & [0.155, 0.195] \\
adversarial-harmful & WildGuard's harmfulness label & 0.987 & 0.820 & 0.166 & [0.153, 0.181] & [0.155, 0.188] \\
adversarial-harmful & Qwen3Guard as the judge & 0.986 & 0.839 & 0.147 & [0.137, 0.158] & [0.139, 0.156] \\
adversarial-harmful & judges agree & 0.985 & 0.823 & 0.162 & [0.152, 0.174] & [0.150, 0.178] \\
plain-harmful & WildGuard's refusal label & 0.991 & 0.892 & 0.099 & [0.086, 0.114] & [0.086, 0.106] \\
plain-harmful & incoherent counted & 0.991 & 0.891 & 0.099 & [0.086, 0.114] & [0.086, 0.106] \\
plain-harmful & WildGuard's harmfulness label & 0.989 & 0.895 & 0.094 & [0.078, 0.115] & [0.085, 0.100] \\
plain-harmful & Qwen3Guard as the judge & 0.992 & 0.911 & 0.082 & [0.069, 0.096] & [0.072, 0.087] \\
plain-harmful & judges agree & 0.991 & 0.894 & 0.097 & [0.083, 0.113] & [0.084, 0.106] \\
all-harmful & WildGuard's refusal label & 0.987 & 0.846 & 0.141 & [0.133, 0.149] & [0.132, 0.153] \\
all-harmful & incoherent counted & 0.987 & 0.846 & 0.141 & [0.133, 0.149] & [0.132, 0.153] \\
all-harmful & WildGuard's harmfulness label & 0.987 & 0.843 & 0.145 & [0.134, 0.156] & [0.137, 0.158] \\
all-harmful & Qwen3Guard as the judge & 0.989 & 0.868 & 0.121 & [0.113, 0.128] & [0.116, 0.125] \\
all-harmful & judges agree & 0.987 & 0.853 & 0.134 & [0.126, 0.142] & [0.128, 0.142] \\
benign & WildGuard's refusal label & 0.985 & 0.849 & 0.136 & [0.125, 0.149] & [0.127, 0.144] \\
benign & incoherent counted & 0.985 & 0.849 & 0.136 & [0.125, 0.149] & [0.127, 0.144] \\
benign & WildGuard's harmfulness label & 0.981 & 0.866 & 0.115 & [0.096, 0.136] & [0.103, 0.127] \\
benign & Qwen3Guard as the judge & 0.987 & 0.874 & 0.114 & [0.104, 0.123] & [0.102, 0.121] \\
benign & judges agree & 0.986 & 0.854 & 0.132 & [0.122, 0.144] & [0.120, 0.139] \\
 
\bottomrule
\end{tabular}

\end{table}

The twelve models of Section~\ref{sec:setup} come from the main data recipe's three base models, each post-trained with four post-training seeds under the first sharding setting.
These twelve models form \sixNmz{} pairs of MZ twins and \sixNu{} same-recipe pairs.
Table~\ref{tab:six-robustness} recomputes \rMZ{}, \rU{} and the ancestry gap on four prompt subsets under five outcomes.
The subsets are adversarial-harmful, plain-harmful and benign, and all-harmful, which joins the two harmful subsets in \sixNharmful{} prompts.
Beside WildGuard's refusal label, two outcomes change which samples count: incoherent counted keeps incoherent samples with the judge's own refusal label, and judges agree keeps only samples on which both judges agree on refusal.
Two outcomes change the label: WildGuard's harmfulness label gives the share of coherent samples that comply harmfully, and Qwen3Guard as the judge replaces WildGuard with Qwen3Guard.
Across these 20 combinations of subset and outcome, the ancestry gap is positive, and every interval over prompts and over base models lies above zero.
The gap is smallest on plain-harmful prompts, at \sixGapPlainMin{}--\sixGapPlainMax{}, and largest on the adversarial-harmful subset, at \sixGapAdvMin{}--\sixGapAdvMax{}.
Under incoherent counted, the gap and both its intervals equal those under WildGuard's refusal label to three decimals in every subset.
Of all the mean corrected correlations, only one differs between the two outcomes, in the third decimal.

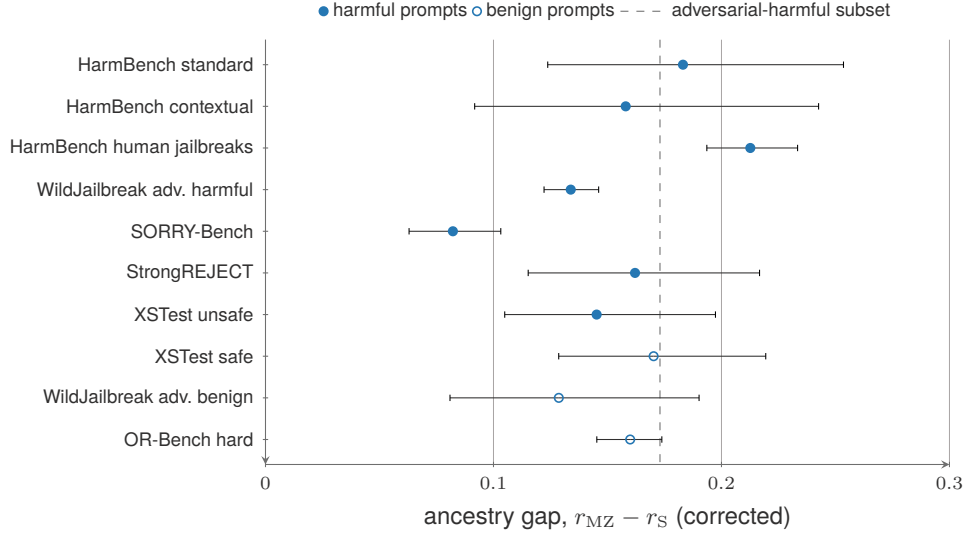
\begin{figure}[h]
  \centering
  \begin{tikzpicture}
\begin{axis}[
  castor axis,
  width=0.62\textwidth, height=0.42\textwidth,
  y dir=reverse, ymin=-0.6, ymax=9.6,
  table/col sep=tab,
  ytick={0,...,9}, yticklabels from table={figures/data/by_source.tsv}{label},
  ymajorgrids=false, xmajorgrids,
  xmin=0, xmax=0.3, xtick={0,0.1,0.2,0.3},
  xlabel={ancestry gap, $\rMZ - \rU$ (corrected)},
  legend style={at={(0.5,1.02)}, anchor=south, legend columns=3},
]
\addplot[muted, dashed, line width=0.4pt, forget plot] table[x=x, y=y, col sep=tab] {figures/data/by_source_ref.tsv};
\addplot[only marks, mark=*, draw=series-main, fill=series-main, castor keep={harmful}{1},
  error bars/x dir=both, error bars/x explicit, error bars/error bar style={draw=text}]
  table[x=gap, y=y, col sep=tab, x error plus expr=\thisrow{hi}-\thisrow{gap},
        x error minus expr=\thisrow{gap}-\thisrow{lo}] {figures/data/by_source.tsv};
\addlegendentry{harmful prompts}
\addplot[only marks, mark=o, draw=series-main, mark options={fill=box}, castor keep={harmful}{0},
  error bars/x dir=both, error bars/x explicit, error bars/error bar style={draw=text}]
  table[x=gap, y=y, col sep=tab, x error plus expr=\thisrow{hi}-\thisrow{gap},
        x error minus expr=\thisrow{gap}-\thisrow{lo}] {figures/data/by_source.tsv};
\addlegendentry{benign prompts}
\addlegendimage{muted, dashed, line width=0.4pt}
\addlegendentry{adversarial-harmful subset}
\end{axis}
\end{tikzpicture}
  \caption{\label{fig:by-source}\textbf{The ancestry gap of the twelve models lies above zero on every prompt source.}
  Each mark is the gap on one prompt source, with reliabilities and the bootstrap recomputed on its prompts.
  Filled marks are harmful sources and open marks benign ones, and the dashed line marks the adversarial-harmful gap of \sixGap{}.
  Intervals are 95\,\% intervals over the source's prompts, from \fortyoneSourceNboot{} draws.}
\end{figure}

Prompt sources split the prompt set more finely than subsets: Figure~\ref{fig:by-source} recomputes the gap on each of \fortyoneSourceN{} sources.
They are the prompt set's own, with three split by category.
HarmBench splits into standard, contextual and human-jailbreak prompts, WildJailbreak into adversarial harmful and adversarial benign, and XSTest into unsafe and safe.
On every source, the gap and its interval over prompts lie above zero, with the gap at \fortyoneSourceGapMin{}--\fortyoneSourceGapMax{}.
The gap is largest on HarmBench's human jailbreaks and smallest on SORRY-Bench, whose plain harmful requests already give same-recipe pairs a mean corrected correlation of \fortyoneSorryRu{}.

The intervals over base models, to which Section~\ref{sec:setup} points, resample the three base models.
A draw that picks a single distinct base model contains no same-recipe pair, so \sixNbaseDraws{} of the \sixNboot{} draws define the interval.
Over base models, \rMZ{} lies in \sixRmzCIbases{} and \rU{} in \sixRuCIbases{}.
The gap of \sixGap{} has an interval of \sixGapCIbases{} over base models, against \sixGapCI{} over prompts.
A leave-one-out jackknife over base models gives standard errors of \sixGapJackSE{} for the gap, \sixRuJackSE{} for \rU{} and \sixRmzJackSE{} for \rMZ{}.
Both the bootstrap and the jackknife rest on three base models only.
Behind these intervals lie the means of single base models and of their pairs.
MZ twins of base models 1, 2 and 3 of Table~\ref{tab:base-models} correlate at \sixBpDef{}, \sixBpAuxTwo{} and \sixBpAuxThree{}.
Same-recipe pairs of base models 1 and 2, 1 and 3, and 2 and 3 correlate at \sixBpDefAuxTwo{}, \sixBpDefAuxThree{} and \sixBpAuxTwoAuxThree{}.

\begin{table}[h]
  \centering\footnotesize
  \caption{\label{tab:six-per-model}\textbf{Mean vulnerability and reliability of each of the twelve models.}
  A row $(k, s)$ is base model $k$ post-trained with post-training seed $s$, as in Table~\ref{tab:base-models}.
  Mean vulnerability is the judge's rate on each subset, and reliability is on the adversarial-harmful subset.
  On benign prompts, a non-refusal is an answer, so a higher vulnerability means fewer over-refusals.
  The last three rows are Ai2's released OLMo~2 1B SFT, DPO and Instruct models, references outside the design and in no comparison.}
  \begin{tabular}{lrrrr}
\toprule
 & \multicolumn{2}{c}{Adversarial-harmful} & Plain-harmful & Benign \\
\cmidrule(lr){2-3}\cmidrule(lr){4-4}\cmidrule(l){5-5}
Model $(k, s)$ & \shortstack{Mean\\vulnerability} & Reliability & \shortstack{Mean\\vulnerability} & \shortstack{Mean\\vulnerability} \\
\midrule
$(1, 1)$ & 0.479 & 0.945 & 0.234 & 0.601 \\
$(1, 2)$ & 0.475 & 0.946 & 0.221 & 0.566 \\
$(1, 3)$ & 0.512 & 0.943 & 0.248 & 0.642 \\
$(1, 4)$ & 0.469 & 0.950 & 0.243 & 0.611 \\
$(2, 1)$ & 0.482 & 0.951 & 0.284 & 0.712 \\
$(2, 2)$ & 0.506 & 0.951 & 0.295 & 0.741 \\
$(2, 3)$ & 0.497 & 0.951 & 0.288 & 0.730 \\
$(2, 4)$ & 0.468 & 0.953 & 0.276 & 0.688 \\
$(3, 1)$ & 0.463 & 0.942 & 0.273 & 0.673 \\
$(3, 2)$ & 0.495 & 0.936 & 0.268 & 0.674 \\
$(3, 3)$ & 0.510 & 0.940 & 0.267 & 0.681 \\
$(3, 4)$ & 0.508 & 0.935 & 0.277 & 0.686 \\
Ai2 OLMo-2 1B SFT (reference) & 0.248 & 0.932 & 0.147 & 0.557 \\
Ai2 OLMo-2 1B DPO (reference) & 0.387 & 0.949 & 0.228 & 0.693 \\
Ai2 OLMo-2 1B Instruct (reference) & 0.428 & 0.948 & 0.228 & 0.700 \\
 
\bottomrule
\end{tabular}

\end{table}

Across the twelve models of Table~\ref{tab:six-per-model}, mean vulnerability on the adversarial-harmful subset runs from \sixVulnMin{} to \sixVulnMax{}.
Their reliability runs from \sixRelMin{} to \sixRelMax{}.
On plain-harmful prompts, mean vulnerability runs from \sixPlainVulnMin{} to \sixPlainVulnMax{}, and on benign prompts from \sixBenignVulnMin{} to \sixBenignVulnMax{}.

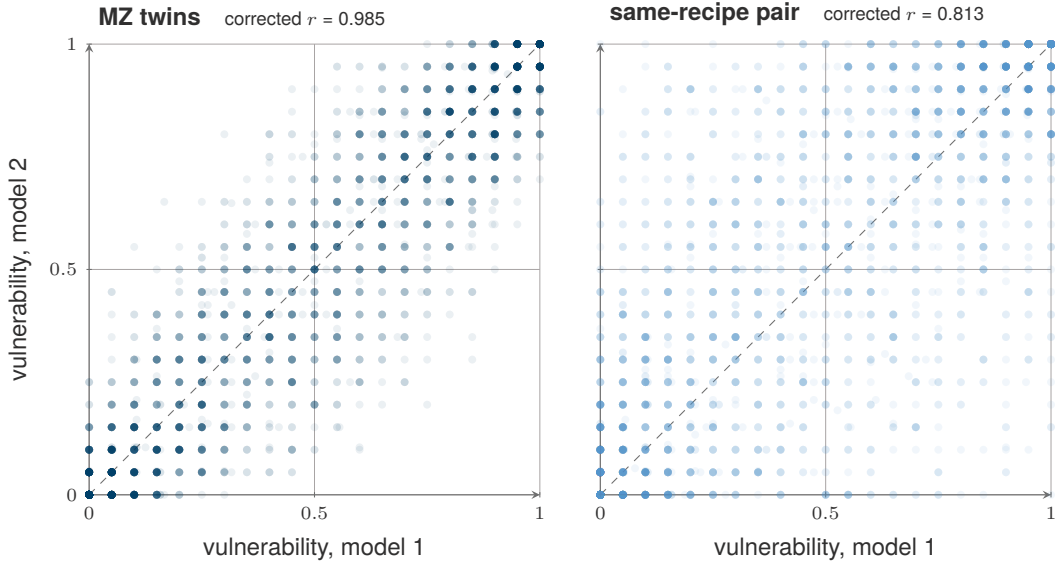
\begin{figure}[h]
  \centering
  \begin{tikzpicture}
\pgfplotsset{castor scatter/.style={castor axis, width=0.44\textwidth, height=0.44\textwidth,
  xmin=0, xmax=1, ymin=0, ymax=1, xtick={0,0.5,1}, ytick={0,0.5,1}, xmajorgrids,
  xlabel={vulnerability, model 1}}}
\begin{axis}[castor scatter, name=mz, ylabel={vulnerability, model 2}]
\addplot[muted, dashed, line width=0.4pt, domain=0:1, samples=2, forget plot] {x};
\addplot[only marks, mark=*, mark size=1.5pt, color=pair-mz, mark options={draw opacity=0, fill opacity=0.08}]
  table[x=x, y=y, col sep=tab] {figures/data/scatter_mz.tsv};
\node[anchor=south west, font=\castortick, text=text, align=left] at (rel axis cs:0,1.02)
  {\castorbold MZ twins\quad\castortick\mdseries corrected $r$ = \fortyoneScatterRMz};
\end{axis}
\begin{axis}[castor scatter, at={(mz.east)}, anchor=west, xshift=8mm, yticklabels={}]
\addplot[muted, dashed, line width=0.4pt, domain=0:1, samples=2, forget plot] {x};
\addplot[only marks, mark=*, mark size=1.5pt, color=pair-unrelated, mark options={draw opacity=0, fill opacity=0.08}]
  table[x=x, y=y, col sep=tab] {figures/data/scatter_u.tsv};
\node[anchor=south west, font=\castortick, text=text, align=left] at (rel axis cs:0,1.02)
  {\castorbold same-recipe pair\quad\castortick\mdseries corrected $r$ = \fortyoneScatterRU};
\end{axis}
\end{tikzpicture}
  \caption{\label{fig:scatter}\textbf{Per prompt, the vulnerabilities of a pair of MZ twins correlate more closely than those of a same-recipe pair.}
  Each panel plots one pair's vulnerabilities on the \fortyoneScatterN{} adversarial-harmful prompts, both axes from 0 to 1, with a dashed diagonal.
  Each is the lower-median pair of its type by corrected correlation, \fortyoneScatterRMz{} for the MZ twins and \fortyoneScatterRU{} for the same-recipe pair.
  With \sixK{} samples, a model's vulnerability takes at most 21 values, so points stack on a lattice.
  Opacity shows how many points stack, and no jitter is added.}
\end{figure}

Figure~\ref{fig:scatter} shows what the mean corrected correlations of MZ twins and same-recipe pairs look like for single pairs.
The pair of MZ twins ranks \fortyoneScatterRankMz{} of \sixNmz{} and correlates at \fortyoneScatterRMz{}, with a Pearson correlation of \fortyoneScatterRawMz{}.
The same-recipe pair ranks \fortyoneScatterRankU{} of \sixNu{} and correlates at \fortyoneScatterRU{}, with a Pearson correlation of \fortyoneScatterRawU{}.
\subsection{The 25-recipe grid}\label{app:robustness-grid}

The 25-recipe grid extends these checks to \eightNrecipes{} data recipes and \eightNmodels{} models, all post-trained under the second sharding setting.
As in Section~\ref{sec:grid}, each mean corrected correlation is computed per data recipe and then averaged over data recipes.
A data recipe's cross-recipe mean pairs its six models with the models of the other 24 data recipes.

\begin{table*}[t]
  \centering\footnotesize
  \caption{\label{tab:eight-per-recipe}\textbf{The ancestry gap's interval over prompts lies above zero in \eightNrecipesGapCIaboveZero{} of 25 data recipes.}
  Rows are the data recipes in order of the gap, largest first, and $^{\ast}$ marks the main data recipe.
  The first three numeric columns give the mean corrected correlations of MZ twins, same-recipe pairs and cross-recipe pairs, then the gap with its 95\,\% interval over prompts and the same-recipe-minus-cross-recipe difference.
  Mean vulnerability is given corrected for the judge's precision, with its 95\,\% interval from the audit of Appendix~\ref{app:judge}, and as the judge's rate.}
  \setlength{\tabcolsep}{2pt}
\begin{tabular}{l@{\hspace{7pt}}lllllll}
\toprule
Data recipe & \shortstack{MZ\\twins} & \shortstack{Same-\\recipe\\pairs} & \shortstack{Cross-\\recipe\\pairs} & \shortstack{Ancestry gap,\\95\,\% interval\\over prompts} & \shortstack{Same-\\recipe minus\\cross-\\recipe} & \shortstack{Mean vulnerability,\\corrected, 95\,\%\\interval from\\the audit} & \shortstack{Mean\\vulnerability,\\by the judge} \\
\midrule
DCLM-Baseline (QC FW 10\,\%) & 0.982 & 0.773 & 0.748 & 0.209 [0.196, 0.223] & 0.025 & 0.274 [0.228, 0.319] & 0.523 \\
DCLM-Baseline (QC 10\,\%) & 0.980 & 0.777 & 0.775 & 0.203 [0.190, 0.215] & 0.002 & 0.263 [0.218, 0.307] & 0.500 \\
Dolma1.7 (no Reddit) & 0.989 & 0.796 & 0.783 & 0.192 [0.179, 0.206] & 0.013 & 0.284 [0.234, 0.331] & 0.538 \\
DCLM-Baseline 75\,\% / Dolma 25\,\% & 0.984 & 0.799 & 0.777 & 0.185 [0.173, 0.197] & 0.022 & 0.280 [0.230, 0.325] & 0.530 \\
Falcon+CC (QC 10\,\%) & 0.989 & 0.804 & 0.766 & 0.185 [0.173, 0.198] & 0.037 & 0.246 [0.202, 0.287] & 0.464 \\
C4 & 0.979 & 0.796 & 0.786 & 0.182 [0.171, 0.195] & 0.010 & 0.271 [0.224, 0.316] & 0.515 \\
DCLM-Baseline (QC 7\,\%, FW3) & 0.978 & 0.797 & 0.758 & 0.181 [0.169, 0.193] & 0.039 & 0.293 [0.243, 0.341] & 0.558 \\
DCLM-Baseline (QC FW 3\,\%) & 0.973 & 0.794 & 0.647 & 0.179 [0.168, 0.192] & 0.146 & 0.347 [0.289, 0.404] & 0.664 \\
Falcon+CC (QC Tulu 10\,\%) & 0.985 & 0.808 & 0.787 & 0.177 [0.165, 0.190] & 0.021 & 0.294 [0.243, 0.343] & 0.558 \\
Falcon & 0.985 & 0.809 & 0.788 & 0.176 [0.164, 0.188] & 0.021 & 0.303 [0.251, 0.353] & 0.576 \\
DCLM-Baseline$^{\ast}$ & 0.984 & 0.809 & 0.786 & 0.175 [0.162, 0.187] & 0.024 & 0.257 [0.212, 0.299] & 0.487 \\
Falcon+CC (QC 20\,\%) & 0.986 & 0.812 & 0.786 & 0.174 [0.163, 0.186] & 0.027 & 0.264 [0.218, 0.308] & 0.500 \\
Dolma1.6++ & 0.981 & 0.808 & 0.786 & 0.173 [0.162, 0.185] & 0.022 & 0.278 [0.230, 0.324] & 0.528 \\
FineWeb-Pro & 0.986 & 0.813 & 0.778 & 0.173 [0.162, 0.185] & 0.035 & 0.247 [0.203, 0.288] & 0.468 \\
Dolma1.7 (no math, code) & 0.985 & 0.813 & 0.791 & 0.173 [0.162, 0.184] & 0.021 & 0.273 [0.225, 0.318] & 0.519 \\
DCLM-Baseline (QC 7\,\%, FW2) & 0.982 & 0.812 & 0.786 & 0.170 [0.159, 0.181] & 0.026 & 0.282 [0.233, 0.328] & 0.535 \\
FineWeb-Edu & 0.983 & 0.813 & 0.781 & 0.169 [0.158, 0.180] & 0.032 & 0.288 [0.238, 0.335] & 0.547 \\
DCLM-Baseline (QC 20\,\%) & 0.981 & 0.820 & 0.786 & 0.161 [0.151, 0.173] & 0.034 & 0.269 [0.221, 0.313] & 0.509 \\
Dolma1.7 (no Flan) & 0.989 & 0.829 & 0.796 & 0.160 [0.150, 0.171] & 0.033 & 0.267 [0.221, 0.312] & 0.508 \\
Dolma1.7 & 0.987 & 0.828 & 0.792 & 0.159 [0.149, 0.170] & 0.036 & 0.287 [0.237, 0.335] & 0.545 \\
Falcon+CC (QC Orig 10\,\%) & 0.983 & 0.825 & 0.797 & 0.158 [0.148, 0.169] & 0.028 & 0.261 [0.216, 0.304] & 0.496 \\
Dolma1.7 (no code) & 0.987 & 0.833 & 0.801 & 0.154 [0.143, 0.166] & 0.032 & 0.282 [0.233, 0.328] & 0.536 \\
Falcon+CC & 0.986 & 0.832 & 0.797 & 0.154 [0.143, 0.165] & 0.035 & 0.289 [0.239, 0.337] & 0.549 \\
DCLM-Baseline 25\,\% / Dolma 75\,\% & 0.987 & 0.835 & 0.798 & 0.152 [0.143, 0.163] & 0.037 & 0.276 [0.227, 0.322] & 0.523 \\
DCLM-Baseline 50\,\% / Dolma 50\,\% & 0.984 & 0.839 & 0.795 & 0.145 [0.135, 0.156] & 0.043 & 0.275 [0.227, 0.320] & 0.521 \\
 
\bottomrule
\end{tabular}

\end{table*}

\begin{figure}[h]
  \centering
  \begin{tikzpicture}
\pgfplotstableread[col sep=tab]{figures/data/pair_dist_levels.tsv}\pairlevels
\pgfplotsset{pair row/.style={const plot mark left, mark=none, line width=0.4pt}}
\begin{axis}[
  castor axis,
  width=0.62\textwidth, height=0.42\textwidth,
  xmin=0.44, xmax=1.0, xtick={0.5,0.6,0.7,0.8,0.9,1.0},
  xticklabel style={/pgf/number format/fixed, /pgf/number format/precision=1},
  ymin=0, ymax=3.0, ytick={0.3,1.3,2.3},
  yticklabels={cross-recipe\\pairs, same-recipe\\pairs, MZ\\twins},
  yticklabel style={align=right}, ymajorgrids=false,
  y axis line style={draw=none}, ytick style={draw=none},
  xlabel={corrected per-prompt $r$ of one pair of models},
  stack plots=y, table/col sep=tab,
]
\addplot[pair row, draw=pair-cross, fill=pair-cross]
  table[x=lo, y expr=0.9*\thisrow{h_R}] {figures/data/pair_dist_hist.tsv} \closedcycle;
\addplot[draw=none, forget plot, const plot mark left]
  table[x=lo, y expr=1-0.9*\thisrow{h_R}] {figures/data/pair_dist_hist.tsv};
\addplot[pair row, draw=pair-unrelated, fill=pair-unrelated]
  table[x=lo, y expr=0.9*\thisrow{h_U}] {figures/data/pair_dist_hist.tsv} \closedcycle;
\addplot[draw=none, forget plot, const plot mark left]
  table[x=lo, y expr=1-0.9*\thisrow{h_U}] {figures/data/pair_dist_hist.tsv};
\addplot[pair row, draw=pair-mz, fill=pair-mz]
  table[x=lo, y expr=0.9*\thisrow{h_MZ}] {figures/data/pair_dist_hist.tsv} \closedcycle;
\pgfplotstablegetrowsof{\pairlevels}%
\pgfmathtruncatemacro{\pairlast}{\pgfplotsretval-1}%
\foreach \i in {0,...,\pairlast}{%
  \pgfplotstablegetelem{\i}{row}\of\pairlevels\edef\prow{\pgfplotsretval}%
  \pgfplotstablegetelem{\i}{r}\of\pairlevels\edef\pr{\pgfplotsretval}%
  \pgfplotstablegetelem{\i}{lo}\of\pairlevels\edef\plo{\pgfplotsretval}%
  \pgfplotstablegetelem{\i}{hi}\of\pairlevels\edef\phi{\pgfplotsretval}%
  \pgfplotstablegetelem{\i}{n_pairs}\of\pairlevels\edef\pn{\pgfplotsretval}%
  \pgfplotstablegetelem{\i}{level}\of\pairlevels\edef\plv{\pgfplotsretval}%
  \edef\rowcmd{%
    \noexpand\draw[muted, line width=0.4pt] (axis cs:0.44,\prow) -- (axis cs:1.0,\prow);
    \noexpand\draw[draw=text, line width=0.6pt] (axis cs:\pr,\prow) -- (axis cs:\pr,\prow+0.93);
    \noexpand\draw[draw=text, line width=0.6pt] (axis cs:\plo,\prow+0.93) -- (axis cs:\phi,\prow+0.93);
    \noexpand\node[anchor=south east, font=\noexpand\castortick, text=text, fill=box, inner sep=1pt, xshift=-2.5pt]
      at (axis cs:\pr,\prow+0.5) {\noexpand\pgfmathparse{round(\pr*1000)/1000}$\noexpand\csname r\plv\noexpand\endcsname = \noexpand\pgfmathprintnumber[fixed, fixed zerofill, precision=3]{\noexpand\pgfmathresult}$};
    \noexpand\node[anchor=south west, font=\noexpand\castortick, text=muted, inner sep=1pt]
      at (axis cs:0.44,\prow) {\noexpand\pgfmathprintnumber[fixed, use comma=false]{\pn} pairs};
  }%
  \rowcmd
}
\end{axis}
\end{tikzpicture}
  \caption{\label{fig:pair-dist}\textbf{The correlations of MZ twins and of same-recipe pairs do not overlap, while those of same-recipe and cross-recipe pairs do.}
  The x-axis gives a pair's corrected correlation on the adversarial-harmful subset under WildGuard's refusal label, over the 25-recipe grid.
  Each row is one kind of pair, with 75, 300 and 10,800 pairs in bins of 0.01, and the cross-recipe row's left tail reaches \fiftythreeMinR{}.
  Each row is scaled to its own peak, so the rows compare shapes, not counts.
  A line marks the mean over the row's pairs, and a short bar its 95\,\% interval over prompts from 1,000 draws.}
\end{figure}
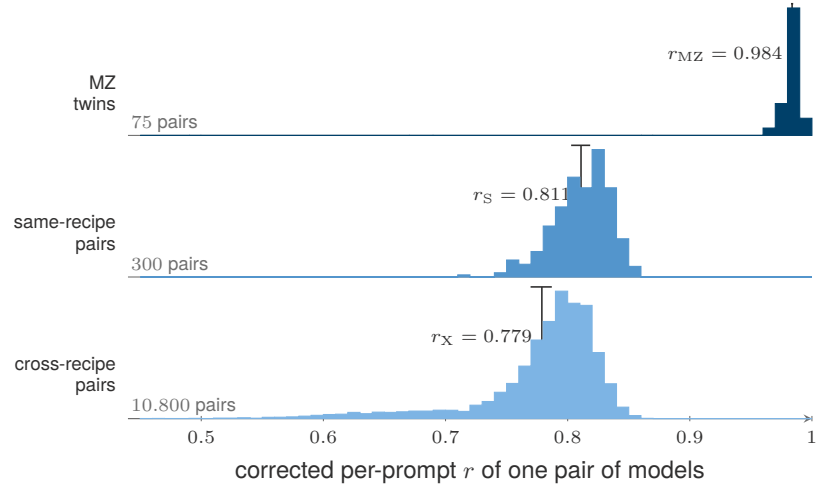

Table~\ref{tab:eight-per-recipe} lists these means per data recipe in order of the gap, and Figure~\ref{fig:grid-heatmap} gives them in DataDecide's Table~1 order, bracketed by dataset group.
Beneath each mean lie the corrected correlations of single pairs, which Figure~\ref{fig:pair-dist} shows by kind of pair.
MZ twins span \fiftythreeMinMZ{}--\fiftythreeMaxMZ{} and same-recipe pairs \fiftythreeMinU{}--\fiftythreeMaxU{}, so these two kinds of pair do not overlap.
Same-recipe pairs and cross-recipe pairs overlap widely.
Cross-recipe pairs span \fiftythreeMinR{}--\fiftythreeMaxR{}.
Below 0.70 lie \fiftythreeNbelow{} of the 10,800 cross-recipe pairs and no same-recipe pair.
Of those \fiftythreeNbelow{} pairs, \fiftythreeNbelowOutlier{} involve DCLM-Baseline (QC FW 3\,\%), the data recipe described below.
Without that data recipe, the lowest cross-recipe pair lies at \fiftythreeMinNoOutlier{}.

\begin{figure}[h]
  \centering
  \providecommand{\heatcell}{25pt}%
\providecommand{\heatrow}{9pt}%
\begin{tikzpicture}
\begin{axis}[
  castor axis,
  name=heat, scale only axis,
  width=4*\heatcell, height=25*\heatrow,
  axis lines=none, ymajorgrids=false,
  enlargelimits=false,
  xmin=-0.5, xmax=3.5, ymin=-0.5, ymax=24.5, y dir=reverse,
  table/col sep=tab,
  colormap name=castor, point meta min=0.6, point meta max=1.0,
]
\addplot[matrix plot*, mesh/cols=4, unbounded coords=jump, draw=box, line width=0.4pt,
  point meta={\thisrow{col}>2 ? nan : \thisrow{value}},
  visualization depends on={\thisrow{value} \as \cellv},
  nodes near coords={\pgfmathprintnumber[fixed, fixed zerofill, precision=3, assume math mode=true]{\cellv}},
  every node near coord/.append style={font=\castortick, anchor=center},
  every node near coord/.append code={%
    \pgfmathparse{\cellv>0.85 ? 1 : 0}%
    \ifnum\pgfmathresult=1 \pgfkeysalso{text=box}\else\pgfkeysalso{text=text}\fi},
] table[x=col, y=row] {figures/data/recipes_heatmap.tsv};
\draw[grid, line width=0.2pt] (axis cs:2.5,-0.5) rectangle (axis cs:3.5,24.5);
\foreach \j/\lab in {0/$\rMZ$, 1/$\rU$, 2/$\rR$, 3/$\rMZ-\rU$}{
  \edef\headcmd{\noexpand\node[anchor=south, font=\noexpand\castortick, text=text]
    at (axis cs:\j,-0.5) {\noexpand\strut\unexpanded\expandafter{\lab}};}
  \headcmd
}
\end{axis}
\pgfplotstableread[col sep=tab]{figures/data/recipes_heatmap_rows.tsv}\heatrows%
\pgfplotstablegetrowsof{\heatrows}%
\pgfmathtruncatemacro{\heatlast}{\pgfplotsretval-1}%
\foreach \i in {0,...,\heatlast}{%
  \pgfplotstablegetelem{\i}{name}\of\heatrows%
  \node[anchor=east, font=\castortick, text=text, inner sep=2pt]
    at ([yshift={-(\i+0.5)*\heatrow}]heat.north west) {\pgfplotsretval};%
}%
\coordinate (heatleft) at ([xshift=-3pt]current bounding box.west);%
\pgfplotstableread[col sep=tab]{figures/data/recipes_groups.tsv}\heatgroups%
\pgfplotstablegetrowsof{\heatgroups}%
\pgfmathtruncatemacro{\heatglast}{\pgfplotsretval-1}%
\foreach \i in {0,...,\heatglast}{%
  \pgfplotstablegetelem{\i}{first}\of\heatgroups\edef\gfirst{\pgfplotsretval}%
  \pgfplotstablegetelem{\i}{last}\of\heatgroups\edef\glast{\pgfplotsretval}%
  \pgfplotstablegetelem{\i}{group}\of\heatgroups\edef\gname{\pgfplotsretval}%
  \path ([yshift={-(\gfirst+0.15)*\heatrow}]heat.north west) coordinate (gtop)
        ([yshift={-(\glast+0.85)*\heatrow}]heat.north west) coordinate (gbot)
        ([yshift={-(\gfirst+\glast+1)/2*\heatrow}]heat.north west) coordinate (gmid);%
  \draw[muted, line width=0.4pt] (heatleft |- gtop) -- (heatleft |- gbot);%
  \node[anchor=east, font=\castortick, text=muted, inner sep=2pt] at (heatleft |- gmid) {\gname};%
}%
\end{tikzpicture}
  \caption{\label{fig:grid-heatmap}\textbf{In each of the 25 data recipes, MZ twins lie above same-recipe pairs and same-recipe pairs above cross-recipe pairs, on point estimates.}
  Values are mean corrected correlations on the adversarial-harmful subset over the 25-recipe grid, with rows in DataDecide's Table~1 order \citep{magnusson2025datadecide}.
  Brackets mark our dataset groups, which merge Table~1's nine sources and are ours, not DataDecide's.
  The columns \rMZ{}, \rU{} and \rR{} give the mean corrected correlations of MZ twins, same-recipe pairs and cross-recipe pairs on one color scale from 0.6 to 1.0.
  A data recipe's \rR{} pairs its six models with the 144 models of the other 24 data recipes, \eightNrPairsRecipe{} pairs, and the gap column is printed without color.}
\end{figure}

\begin{figure}[h]
  \centering
  \input{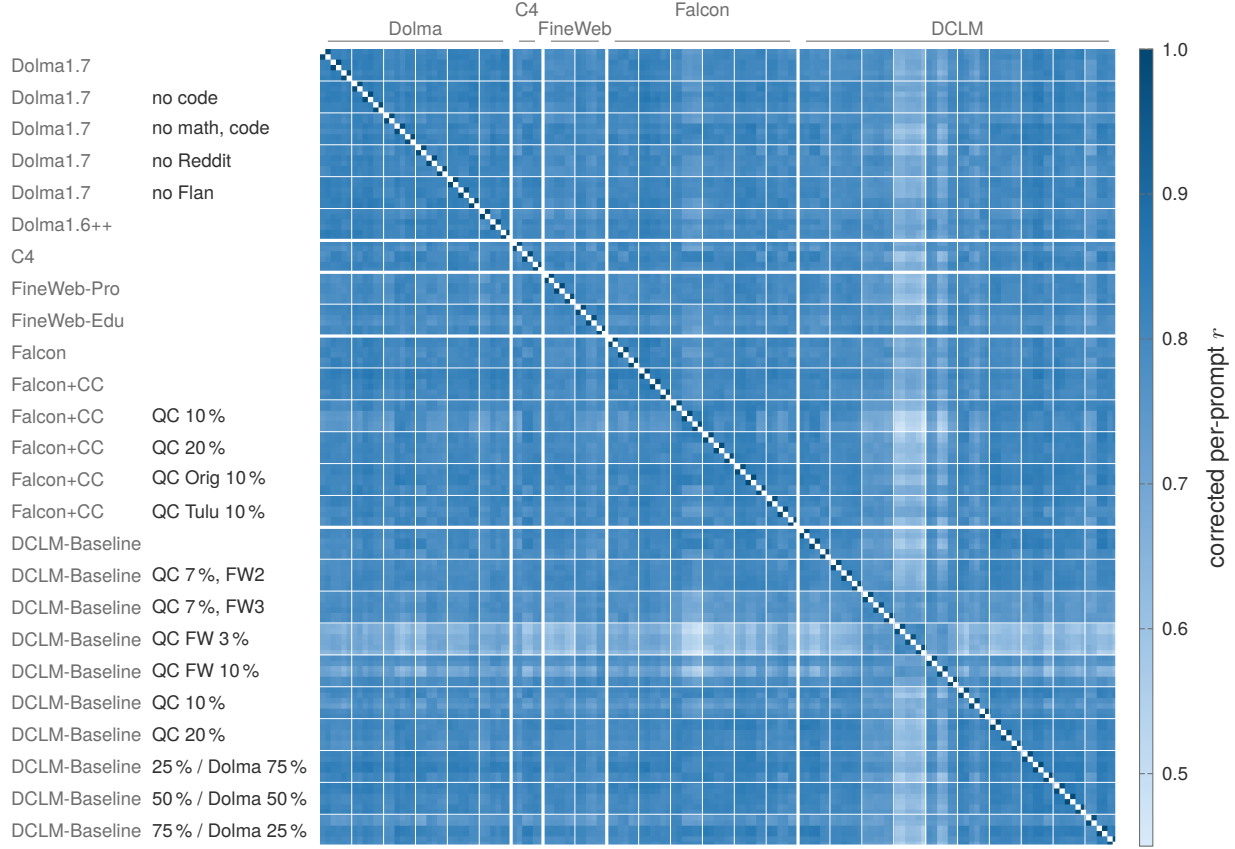}
  \caption{\label{fig:pair-matrix}\textbf{The MZ squares on the diagonal are the visible structure of the pairwise correlations.}
  Each square is one pair's corrected correlation on the adversarial-harmful subset under WildGuard's refusal label, for all 11,175 pairs of the 150 models, each drawn twice around a blank diagonal.
  Rows and columns follow data recipe in DataDecide's Table~1 order, then base model, then post-training seed.
  Each data recipe thus forms a 6 × 6 block, whose three 2 × 2 diagonal squares are MZ twins and whose other squares are same-recipe pairs.
  Every square outside these blocks is a cross-recipe pair.
  The color scale runs from 0.45 to 1.0, and no pair is clipped.
  Thin rules separate data recipes, and wide rules separate dataset groups.
  DCLM-Baseline (QC FW 3\,\%) shows as a pale row and column, and DCLM-Baseline (QC FW 10\,\%) and Falcon+CC (QC 10\,\%) are paler in places.
  Rows and columns are ordered by nested factors, as \citet[Fig.~3a]{mehrer2020individualdiff} order their matrix by layer and, within a layer, by network instance.
  Before plotting, we predicted MZ squares above the other pairs of their data recipe, and those pairs above the cross-recipe field, in every data recipe.
  We also predicted one pale row and column for the outlier data recipe.
  The MZ squares and the pale row and column appear, but the same-recipe pairs are not visibly above the cross-recipe field.}
\end{figure}

Figure~\ref{fig:pair-matrix} shows every pair of the 150 models at once.
The 75 MZ squares on the diagonal are its visible structure.
Within a data recipe, same-recipe pairs are not visibly above the cross-recipe field, which they exceed on average by \eightUminusR{}, about \eightUminusRColourPct{}\% of the color range.
On point estimates, MZ twins still lie above same-recipe pairs, and same-recipe pairs above cross-recipe pairs, in all \eightNrecipesOrdered{} data recipes.
Per data recipe, the same-recipe-minus-cross-recipe difference has no interval.
It runs from \eightMinUminusR{} for DCLM-Baseline (QC 10\,\%) to \eightOutlierUminusR{}.

DataDecide builds DCLM-Baseline (QC FW 3\,\%) from the top 3\,\% of documents by the FineWeb-Edu classifier \citep{magnusson2025datadecide}.
Among the 25 data recipes, it stands apart on the cross-recipe mean, at \eightOutlierRr{}.
It shows as the pale row and column of Figure~\ref{fig:pair-matrix}.
Its same-recipe-minus-cross-recipe difference of \eightOutlierUminusR{} therefore comes near its gap of \eightOutlierGap{}, and no other data recipe has this difference within 0.10 of its gap.
Its same-recipe-pair mean, \eightOutlierRu{}, lies inside the other data recipes' range, while its MZ twins' mean, \eightOutlierRmz{}, is the lowest of the 25, which span \eightRmzRecipeMin{}--\eightRmzRecipeMax{}.
By the judge, it has the highest mean vulnerability, \eightVulnRecipeMax{} against \eightVulnRecipeSecond{} for the next.
Corrected for the judge's precision, its mean vulnerability is \eightLevelCorrRecipeMax{} \eightLevelCorrRecipeMaxCI{}, which overlaps those of the next data recipes and shares their per-source factor from the audit.

\begin{table*}[t]
  \centering\footnotesize
  \caption{\label{tab:eight-secondary}\textbf{The ancestry gap of the 25-recipe grid lies above zero under every outcome and in every prompt subset.}
  For each prompt subset and outcome, the columns give the mean corrected correlations of MZ twins, same-recipe pairs and cross-recipe pairs, then the gap and the same-recipe-minus-cross-recipe difference with 95\,\% intervals over prompts.
  The last column splits the variance of per-model mean vulnerability into the shares of the data recipe, the base model within it and the post-training seed, computed on the judge's rates.
  Incoherent counted is omitted, since it matches WildGuard's refusal label to three decimals.}
  \setlength{\tabcolsep}{3pt}
\begin{tabular}{>{\raggedright\arraybackslash}p{2.6cm}>{\raggedright\arraybackslash}p{2.5cm}llllll}
\toprule
Subset & Outcome & \shortstack{MZ\\twins} & \shortstack{Same-\\recipe\\pairs} & \shortstack{Cross-\\recipe\\pairs} & \shortstack{Ancestry\\gap, 95\,\%\\interval} & \shortstack{Same-recipe\\minus cross-\\recipe, 95\,\%\\interval} & \shortstack{Variance shares:\\data recipe /\\base model /\\post-training seed} \\
\midrule
adversarial-harmful & WildGuard's refusal label & 0.984 & 0.811 & 0.779 & 0.173 [0.166, 0.180] & 0.032 [0.030, 0.034] & 0.51 / 0.31 / 0.18 \\
adversarial-harmful & WildGuard's harmfulness label & 0.986 & 0.836 & 0.803 & 0.150 [0.143, 0.159] & 0.033 [0.031, 0.035] & 0.64 / 0.24 / 0.13 \\
adversarial-harmful & Qwen3Guard as the judge & 0.987 & 0.842 & 0.815 & 0.145 [0.139, 0.151] & 0.026 [0.025, 0.028] & 0.23 / 0.49 / 0.28 \\
adversarial-harmful & judges agree & 0.985 & 0.822 & 0.795 & 0.163 [0.156, 0.170] & 0.028 [0.026, 0.029] & 0.38 / 0.39 / 0.23 \\
plain-harmful & WildGuard's refusal label & 0.990 & 0.890 & 0.861 & 0.100 [0.092, 0.109] & 0.029 [0.026, 0.032] & 0.62 / 0.29 / 0.09 \\
all-harmful & WildGuard's refusal label & 0.986 & 0.846 & 0.817 & 0.140 [0.134, 0.146] & 0.029 [0.027, 0.030] & 0.59 / 0.26 / 0.15 \\
benign & WildGuard's refusal label & 0.987 & 0.867 & 0.844 & 0.120 [0.112, 0.128] & 0.023 [0.020, 0.025] & 0.12 / 0.72 / 0.16 \\
 
\bottomrule
\end{tabular}

\end{table*}

Table~\ref{tab:eight-secondary} repeats the 25-recipe grid's mean corrected correlations under the other subsets and outcomes.
In every subset and outcome, MZ twins lie above same-recipe pairs and same-recipe pairs above cross-recipe pairs.
The gap's interval over prompts lies above zero in each.
The gap is smallest on plain-harmful prompts, at \eightGapPlain{}, and largest on the adversarial-harmful subset under WildGuard's refusal label, at \eightGapMean{}.
Under incoherent counted, the gap is \eightGapInclFour{} against \eightGapFour{} under WildGuard's refusal label.

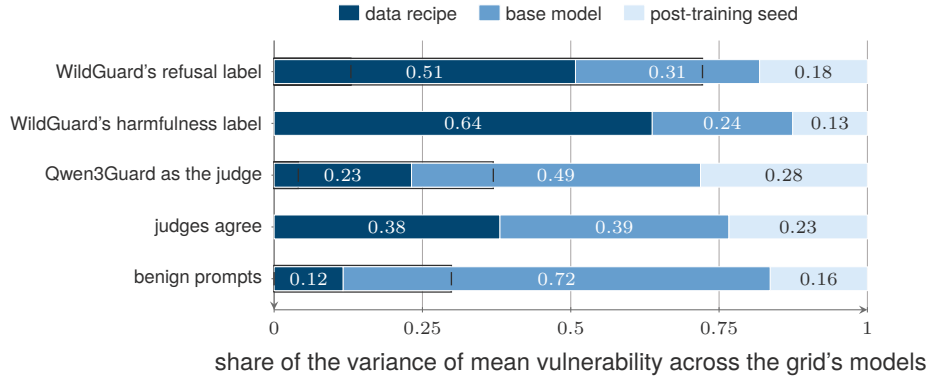
\begin{figure}[h]
  \centering
  \begin{tikzpicture}
\pgfplotsset{share seg/.style={xbar stacked, draw=box, line width=0.3pt, bar width=9pt,
  nodes near coords={\pgfmathprintnumber[fixed, fixed zerofill, precision=2]{\pgfplotspointmeta}},
  point meta=rawx, nodes near coords align={center},
  every node near coord/.append style={font=\castortick, inner sep=0pt}}}
\begin{axis}[
  castor axis,
  width=0.55\textwidth, height=0.3\textwidth,
  xbar stacked,
  xmin=0, xmax=1, xtick={0,0.25,0.5,0.75,1},
  xticklabel style={/pgf/number format/fixed},
  y dir=reverse, ymin=-0.6, ymax=4.6,
  ytick={0,...,4}, yticklabels from table={figures/data/var_shares.tsv}{label},
  ymajorgrids=false, xmajorgrids,
  xlabel={share of the variance of mean vulnerability across the grid's models},
  table/col sep=tab,
  legend style={at={(0.5,1.02)}, anchor=south, legend columns=3, /tikz/every even column/.append style={column sep=6pt}},
  legend image code/.code={\draw[#1] (0cm,-0.08cm) rectangle (0.25cm,0.08cm);},
]
\addplot[share seg, fill=seq-5, every node near coord/.append style={text=box}]
  table[x=recipe, y=y] {figures/data/var_shares.tsv};
\addlegendentry{data recipe}
\addplot[share seg, fill=seq-3, every node near coord/.append style={text=box}]
  table[x=pretrain, y=y] {figures/data/var_shares.tsv};
\addlegendentry{base model}
\addplot[share seg, fill=seq-1, every node near coord/.append style={text=text}]
  table[x=posttrain, y=y] {figures/data/var_shares.tsv};
\addlegendentry{post-training seed}
\addplot[stack plots=false, forget plot, draw=text, line width=0.5pt, unbounded coords=jump,
  mark=|, mark size=2.5pt, mark options={draw=text, line width=0.5pt}]
  table[x=x, y=y] {figures/data/var_shares_ci.tsv};
\end{axis}
\end{tikzpicture}
  \caption{\label{fig:var-shares}\textbf{Which factor has the largest share of variance depends on the outcome and the subset, on point estimates.}
  Each bar splits the variance of per-model mean vulnerability over the 150 models into the shares of the data recipe, the base model within its data recipe and the post-training seed.
  The top four rows are WildGuard's refusal label, WildGuard's harmfulness label, Qwen3Guard as the judge and judges agree on the adversarial-harmful subset, and the bottom row is benign prompts.
  Shares are computed on the judge's rates, and Section~\ref{sec:grid} gives them corrected for the judge's precision.
  Where an interval exists, a bar marks the 95\,\% interval of the data recipe's share from a bootstrap over data recipes, 1,000 draws.
  Every interval is wide, since only 25 data recipes are resampled.}
\end{figure}

Figure~\ref{fig:var-shares} and the last column of Table~\ref{tab:eight-secondary} split the variance of per-model mean vulnerability by factor.
Under WildGuard's labels on harmful prompts, the data recipe's share is the largest point estimate: \eightShareRecipe{} under the refusal label, \eightShareRecipeHarm{} under the harmfulness label, \eightShareRecipePlain{} on plain-harmful and \eightShareRecipeHarmful{} on all-harmful prompts.
Under Qwen3Guard as the judge, the base model's share is larger, \eightSharePretrainQguard{} against \eightShareRecipeQguard{}.
When the judges agree, the two shares are about equal, at \eightSharePretrainAgree{} and \eightShareRecipeAgree{}.
On benign prompts the base model's share is larger again, at \eightSharePretrainBenign{} against \eightShareRecipeBenign{}.
For the data recipe's share, bootstrap intervals over data recipes exist in three of these rows.
They are \fortyfiveShareRecipeCI{} under WildGuard's refusal label, \fortyfiveShareRecipeQguardCI{} under Qwen3Guard as the judge and \fortyfiveShareRecipeBenignCI{} on benign prompts.
The first interval contains the base model's share of \eightSharePretrain{}, whose own interval is \fortyfiveSharePretrainCI{}.
On benign prompts the two intervals do not overlap, the base model's being \fortyfiveSharePretrainBenignCI{}, though no interval for their difference was computed.
Every interval is wide over 25 data recipes, and none is for a difference of shares.
Outside benign prompts, we therefore claim no ordering of shares beyond point estimates.

\begin{table*}[t]
  \centering\footnotesize
  \caption{\label{tab:eight-dfortytwo}\textbf{The template residual and the gap on each adversarial source.}
  Rows give the gap on WildJailbreak, the gap on HarmBench's human jailbreaks and the template-residual gap.
  The first five numeric columns give the 25-recipe grid's three mean corrected correlations, the gap with its 95\,\% interval over prompts and the same-recipe-minus-cross-recipe difference.
  For the template residual, that interval is conditional on the fit.
  The last two columns give the mean gap over the 24 data recipes other than the main one, with its range, and the main data recipe's gap.
  Registration is as for Table~\ref{tab:eight-further-per-recipe}.}
  \setlength{\tabcolsep}{3pt}
\begin{tabular}{>{\raggedright\arraybackslash}p{2.8cm}lllllll}
\toprule
Statistic & \shortstack{MZ\\twins} & \shortstack{Same-recipe\\pairs} & \shortstack{Cross-\\recipe\\pairs} & \shortstack{Ancestry\\gap, 95\,\%\\interval} & \shortstack{Same-recipe\\minus cross-\\recipe} & \shortstack{Mean and range\\over the 24 other\\data recipes} & \shortstack{Main data\\recipe} \\
\midrule
WildJailbreak & 0.986 & 0.847 & 0.823 & 0.139 [0.132, 0.147] & 0.023 & 0.139 (0.121--0.177) & 0.134 \\
HarmBench human jailbreaks & 0.983 & 0.772 & 0.741 & 0.210 [0.198, 0.223] & 0.032 & 0.210 (0.169--0.291) & 0.214 \\
template residual (HarmBench human jailbreaks) & 0.946 & 0.421 & 0.365 & 0.525 [0.507, 0.545] & 0.057 & 0.524 (0.472--0.598) & 0.553 \\
 
\bottomrule
\end{tabular}

\end{table*}

\begin{table*}[t]
  \centering\footnotesize
  \caption{\label{tab:eight-further-per-recipe}\textbf{Further statistics by data recipe.}
  Rows follow DataDecide's Table~1 order, and $^{\ast}$ marks the main data recipe.
  The transfer ratio is given for MZ twins, same-recipe pairs and cross-recipe pairs, without intervals.
  The template-residual gap carries a 95\,\% interval over prompts, conditional on the fit.
  The gaps on WildJailbreak and on HarmBench's human jailbreaks carry 95\,\% intervals over prompts.
  Each statistic was registered for the 24 data recipes other than the main one and computed post hoc for the main one.}
  \setlength{\tabcolsep}{3pt}
\begin{tabular}{lllllll}
\toprule
 & \multicolumn{3}{c}{Transfer ratio} & & & \\
\cmidrule(lr){2-4}
Data recipe & \shortstack{MZ\\twins} & \shortstack{Same-recipe\\pairs} & \shortstack{Cross-\\recipe\\pairs} & \shortstack{Template-\\residual gap,\\95\,\% interval} & \shortstack{Gap on\\WildJailbreak,\\95\,\% interval} & \shortstack{Gap on HarmBench's\\human jailbreaks,\\95\,\% interval} \\
\midrule
Dolma1.7 & 0.998 & 0.943 & 0.921 & 0.509 [0.466, 0.553] & 0.130 [0.119, 0.142] & 0.199 [0.180, 0.219] \\
Dolma1.7 (no code) & 0.996 & 0.944 & 0.920 & 0.518 [0.475, 0.563] & 0.123 [0.112, 0.135] & 0.199 [0.179, 0.220] \\
Dolma1.7 (no math, code) & 0.997 & 0.935 & 0.942 & 0.504 [0.459, 0.553] & 0.142 [0.130, 0.156] & 0.200 [0.179, 0.220] \\
Dolma1.7 (no Reddit) & 0.998 & 0.927 & 0.911 & 0.535 [0.488, 0.583] & 0.154 [0.142, 0.167] & 0.217 [0.196, 0.238] \\
Dolma1.7 (no Flan) & 0.998 & 0.938 & 0.936 & 0.540 [0.497, 0.584] & 0.128 [0.115, 0.140] & 0.196 [0.178, 0.216] \\
Dolma1.6++ & 0.996 & 0.931 & 0.921 & 0.498 [0.453, 0.540] & 0.152 [0.139, 0.165] & 0.200 [0.179, 0.222] \\
C4 & 0.996 & 0.928 & 0.937 & 0.507 [0.459, 0.552] & 0.132 [0.121, 0.145] & 0.208 [0.189, 0.228] \\
FineWeb-Pro & 0.994 & 0.920 & 0.944 & 0.497 [0.461, 0.540] & 0.154 [0.142, 0.167] & 0.195 [0.176, 0.217] \\
FineWeb-Edu & 0.994 & 0.932 & 0.905 & 0.514 [0.471, 0.562] & 0.141 [0.129, 0.153] & 0.194 [0.176, 0.214] \\
Falcon & 0.998 & 0.937 & 0.897 & 0.554 [0.509, 0.599] & 0.132 [0.121, 0.144] & 0.231 [0.207, 0.253] \\
Falcon+CC & 0.992 & 0.940 & 0.916 & 0.578 [0.530, 0.629] & 0.121 [0.110, 0.133] & 0.198 [0.177, 0.220] \\
Falcon+CC (QC 10\,\%) & 0.995 & 0.914 & 0.950 & 0.598 [0.552, 0.649] & 0.129 [0.119, 0.141] & 0.211 [0.192, 0.230] \\
Falcon+CC (QC 20\,\%) & 0.997 & 0.928 & 0.937 & 0.569 [0.525, 0.617] & 0.132 [0.119, 0.144] & 0.220 [0.200, 0.242] \\
Falcon+CC (QC Orig 10\,\%) & 0.992 & 0.926 & 0.938 & 0.493 [0.446, 0.540] & 0.131 [0.119, 0.143] & 0.169 [0.152, 0.185] \\
Falcon+CC (QC Tulu 10\,\%) & 0.996 & 0.936 & 0.905 & 0.549 [0.503, 0.597] & 0.136 [0.124, 0.149] & 0.209 [0.188, 0.230] \\
DCLM-Baseline$^{\ast}$ & 0.994 & 0.932 & 0.933 & 0.553 [0.511, 0.599] & 0.134 [0.122, 0.146] & 0.214 [0.194, 0.236] \\
DCLM-Baseline (QC 7\,\%, FW2) & 0.991 & 0.935 & 0.916 & 0.477 [0.436, 0.523] & 0.137 [0.125, 0.149] & 0.213 [0.193, 0.235] \\
DCLM-Baseline (QC 7\,\%, FW3) & 0.996 & 0.925 & 0.904 & 0.533 [0.491, 0.580] & 0.157 [0.143, 0.171] & 0.227 [0.206, 0.249] \\
DCLM-Baseline (QC FW 3\,\%) & 0.990 & 0.943 & 0.823 & 0.517 [0.458, 0.577] & 0.177 [0.162, 0.192] & 0.291 [0.259, 0.326] \\
DCLM-Baseline (QC FW 10\,\%) & 0.994 & 0.918 & 0.907 & 0.581 [0.536, 0.627] & 0.149 [0.136, 0.162] & 0.268 [0.246, 0.293] \\
DCLM-Baseline (QC 10\,\%) & 0.990 & 0.908 & 0.924 & 0.541 [0.497, 0.587] & 0.157 [0.144, 0.169] & 0.240 [0.219, 0.264] \\
DCLM-Baseline (QC 20\,\%) & 0.992 & 0.933 & 0.921 & 0.483 [0.442, 0.524] & 0.145 [0.133, 0.156] & 0.187 [0.168, 0.208] \\
DCLM-Baseline 25\,\% / Dolma 75\,\% & 0.994 & 0.941 & 0.921 & 0.489 [0.445, 0.536] & 0.125 [0.113, 0.136] & 0.180 [0.162, 0.200] \\
DCLM-Baseline 50\,\% / Dolma 50\,\% & 0.992 & 0.938 & 0.928 & 0.472 [0.428, 0.520] & 0.127 [0.115, 0.139] & 0.172 [0.155, 0.192] \\
DCLM-Baseline 75\,\% / Dolma 25\,\% & 0.990 & 0.926 & 0.916 & 0.525 [0.479, 0.572] & 0.134 [0.123, 0.147] & 0.220 [0.199, 0.242] \\
 
\bottomrule
\end{tabular}

\end{table*}

Beyond the mean corrected correlations, Section~\ref{sec:grid} reports two further statistics, the template residual and the transfer ratio. The transfer ratio takes the tenth of prompts on which one model of a pair is most vulnerable in one half of its samples. There, it divides the other model's mean vulnerability by the first model's own on its other half.
These statistics and the gap on each adversarial source were registered for the 24 data recipes other than the main one.
For the main data recipe, they were computed after the data, as post hoc item~\ref{prereg-posthoc:main-recipe} of Appendix~\ref{app:prereg} records.
Table~\ref{tab:eight-dfortytwo} gives the template residual beside the gap on each adversarial source.
Over the 24 other data recipes, the template-residual gap averages \eightGapResidNonMain{}, ranging \eightGapResidNonMainMin{}--\eightGapResidNonMainMax{}, against \eightGapResidMain{} for the main data recipe.
Once template and behavior effects are removed, the same-recipe-pair mean falls to \eightRuResid{} and the cross-recipe mean to \eightRrResid{}.
On the two adversarial sources, the gap over the 24 averages \eightGapWjNonMain{} on WildJailbreak and \eightGapHjbNonMain{} on HarmBench's human jailbreaks, against \eightGapWjMain{} and \eightGapHjbMain{} for the main data recipe.
Over the 24, the gap ranges \eightGapHjbNonMainMin{}--\eightGapHjbNonMainMax{} on HarmBench's human jailbreaks and \eightGapWjNonMainMin{}--\eightGapWjNonMainMax{} on WildJailbreak.

The transfer ratio has means over the 24 of \eightTransferMzNonMain{} for MZ twins, \eightTransferUNonMain{} for same-recipe pairs and \eightTransferRNonMain{} for cross-recipe pairs.
In the same order, it ranges over the 24 across \eightTransferMzNonMainMin{}--\eightTransferMzNonMainMax{}, \eightTransferUNonMainMin{}--\eightTransferUNonMainMax{} and \eightTransferRNonMainMin{}--\eightTransferRNonMainMax{}.
Over all 25 data recipes, the three means are \eightTransferMz{}, \eightTransferU{} and \eightTransferR{}.
Table~\ref{tab:eight-further-per-recipe} gives every data recipe's value of each statistic.

Two descriptive comparisons place the 25-recipe grid against outside references: Ai2's released OLMo~2 1B DPO model and DataDecide's benchmark results.
The first falls under the same registration as the further statistics.
Each model's corrected correlation with the released model averages \eightRefAll{} over the 150 models, ranging \eightRefMin{}--\eightRefMax{}.
Over the main data recipe's six models, it averages \eightRefMain{}.
Over the 25 data recipes, the judge's mean vulnerability correlates with DataDecide's macro average of its ten OLMES benchmarks \citep{gu2024olmes,magnusson2025datadecide} at a Pearson correlation of \eightOlmesPearsonRecipes{}.
The Spearman correlation is \eightOlmesSpearmanRecipes{}.
Over the 75 base models, the Pearson correlation is \eightOlmesPearsonBases{}.
No interval or test was computed, and DCLM-Baseline (QC FW 3\,\%) is the outlier described above.

\begin{figure}[h]
  \centering
  \input{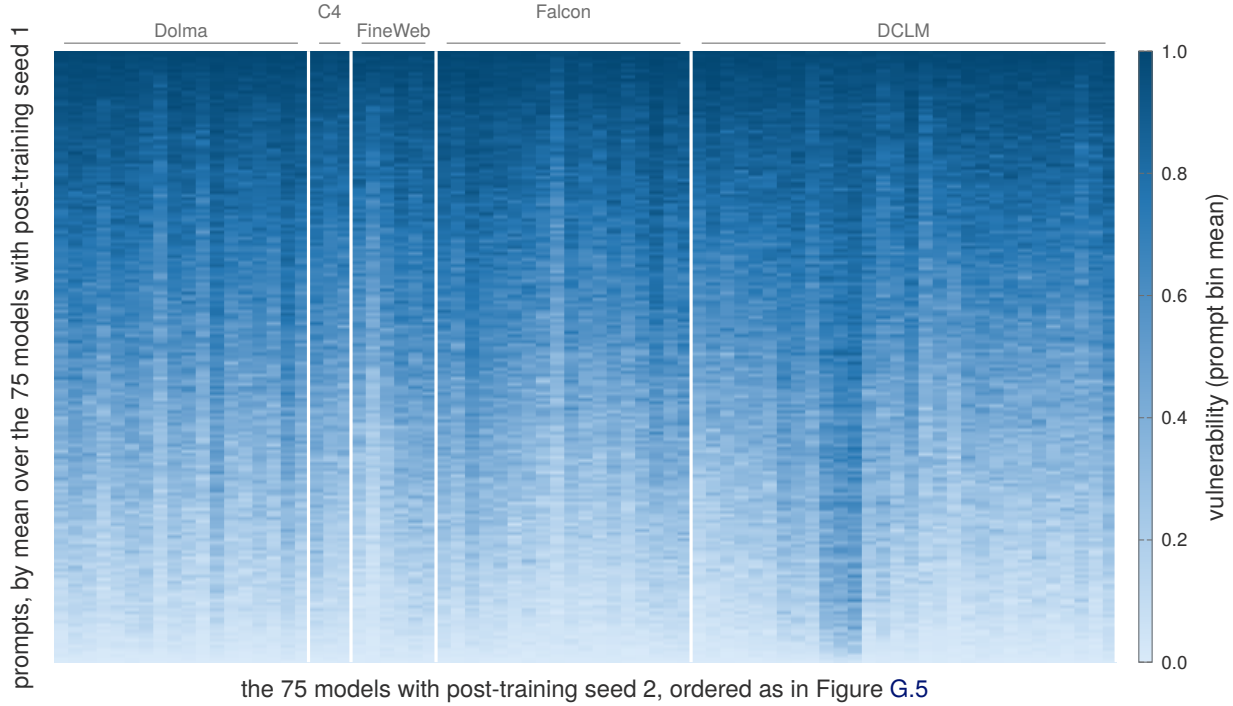}
  \caption{\label{fig:prompt-raster}\textbf{Every column shows the same top-to-bottom gradient, each with its own offset.}
  Rows are the \eightNprompts{} adversarial-harmful prompts, sorted by their mean vulnerability over the 75 models with post-training seed 1, most vulnerable on top.
  Rows are binned into 201 bins of 16 consecutive prompts, the last bin with 12, and color gives the judge's vulnerability from 0 to 1.
  Columns are the 75 models with post-training seed 2, in the order of Figure~\ref{fig:pair-matrix}, with dataset groups named on top, so the sort comes from models other than those shown.
  Before plotting, we predicted horizontal bands for prompts sorted by their mean vulnerability over all 150 models, not the sort shown here.
  At print size it shows no distinct bands, but a gradient shared by every column, plus offsets that span whole columns.}
\end{figure}

At the level of single prompts, Figure~\ref{fig:prompt-raster} sorts the prompts by their mean vulnerability over the 75 models with post-training seed 1, and shows the other 75 models.
Every column shows the same top-to-bottom gradient, each model with its own column-wide offset.
The columns do not repeat one order prompt by prompt, since same-recipe pairs correlate at \eightRu{}.
The columns of DCLM-Baseline (QC FW 3\,\%) stay darker than the other data recipes' into the bottom rows.
\subsection{Model sizes}\label{app:robustness-sizes}

Section~\ref{sec:grid} reports the ancestry gap at three sizes of the main data recipe.
At 150M and 300M, the models come from checkpoints of the main data recipe's three runs near the full budget, which Appendix~\ref{app:pipeline} defines.
Each checkpoint was post-trained with four post-training seeds under the second sharding setting.
At 1B, the twelve models are those of Section~\ref{sec:ladder}.
The 25-recipe grid's gap for the main data recipe, \eightGapMain{}, uses six of these twelve, with post-training seeds 1 and 2, and is not compared here.

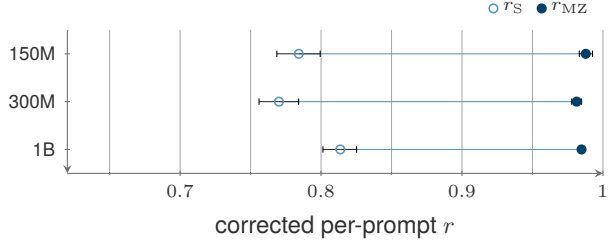
\begin{figure}[h]
  \centering
  \providecommand{\sizesunit}{288pt}%
\providecommand{\sizesrow}{18pt}%
\begin{tikzpicture}
\begin{axis}[
  castor axis,
  name=sizes, scale only axis,
  width=0.7*\sizesunit, height=3*\sizesrow,
  y dir=reverse, ymin=0.5, ymax=3.5,
  table/col sep=tab,
  ytick={1,2,3}, yticklabels from table={figures/data/size_ladder.tsv}{size},
  ymajorgrids=false, xmajorgrids,
  xmin=0.62, xmax=1.0, xtick={0.7,0.8,0.9,1.0}, minor xtick={0.65,0.75,0.85,0.95}, xminorgrids,
  xticklabel style={/pgf/number format/fixed, /pgf/number format/precision=2},
  xlabel={corrected per-prompt $r$},
  legend style={at={(1,1.03)}, anchor=south east, legend columns=2,
    /tikz/every even column/.append style={column sep=3pt}},
]
\addplot[quiver={u=\thisrow{rmz}-\thisrow{ru}, v=0}, draw=pair-unrelated, line width=0.4pt,
  forget plot] table[x=ru, y=x] {figures/data/size_ladder.tsv};
\addplot[only marks, mark=o, draw=pair-unrelated, mark options={fill=box},
  error bars/x dir=both, error bars/x explicit, error bars/error bar style={draw=text}]
  table[x=ru, y=x, x error plus expr=\thisrow{ru_hi}-\thisrow{ru},
        x error minus expr=\thisrow{ru}-\thisrow{ru_lo}] {figures/data/size_ladder.tsv};
\addlegendentry{$\rU$}
\addplot[only marks, mark=*, draw=pair-mz, fill=pair-mz,
  error bars/x dir=both, error bars/x explicit, error bars/error bar style={draw=text}]
  table[x=rmz, y=x, x error plus expr=\thisrow{rmz_hi}-\thisrow{rmz},
        x error minus expr=\thisrow{rmz}-\thisrow{rmz_lo}] {figures/data/size_ladder.tsv};
\addlegendentry{$\rMZ$}
\end{axis}
\end{tikzpicture}
  \caption{\label{fig:grid-sizes}\textbf{At 150M, 300M and 1B, MZ twins of the main data recipe correlate more closely than its same-recipe pairs.}
  Each row is one size, with the main data recipe's twelve models.
  A filled circle marks the mean corrected correlation of MZ twins and an open circle that of same-recipe pairs, on the axis of Figure~\ref{fig:grid}, and the line joining them is the ancestry gap.
  At 1B these are the twelve models of Section~\ref{sec:ladder}, with four post-training seeds.
  Their ancestry gap of \sixteenGapOneB{} therefore differs from that of the main data recipe's row in Figure~\ref{fig:grid}, \eightGapMain{}, which uses post-training seeds 1 and 2.
  Bars are 95\,\% intervals from a bootstrap over prompts.}
\end{figure}

\begin{table}[h]
  \centering\footnotesize
  \caption{\label{tab:sixteen-gate}\textbf{The models at each size.}
  Rows are sizes of the main data recipe, all post-trained under the second sharding setting.
  At 150M and 300M, the twelve models come from checkpoints of three runs near the full budget, and at 1B they are the twelve models of Section~\ref{sec:ladder}.
  Columns give the number of models, the range and mean of their mean vulnerability, and the range of their reliability, all on the adversarial-harmful subset and by the judge.
  Per-prompt variance is the variance of a model's vulnerability over its prompts.
  The last two columns give the shares of a model's prompts always refused and never refused.
  A prompt is always refused when the model refuses it in every sample, and never refused when the model refuses it in no sample.
  These three statistics are computed per model, then averaged over the twelve models of each size.}
  \begin{tabular}{lrllllll}
\toprule
Size & Models & \shortstack{Mean vulnerability,\\range over models} & \shortstack{Mean\\vulnerability} & \shortstack{Reliability,\\range over models} & \shortstack{Per-prompt\\variance} & \shortstack{Prompts always\\refused} & \shortstack{Prompts never\\refused} \\
\midrule
150M & 12 & 0.746--0.805 & 0.784 & 0.822--0.858 & 0.042 & 0.000 & 0.154 \\
300M & 12 & 0.635--0.775 & 0.703 & 0.872--0.909 & 0.069 & 0.007 & 0.134 \\
1B & 12 & 0.462--0.519 & 0.491 & 0.936--0.951 & 0.122 & 0.116 & 0.085 \\
 
\bottomrule
\end{tabular}

\end{table}

Table~\ref{tab:sixteen-gate} shows that every model at every size has mean vulnerability within the bounds of Appendix~\ref{app:prereg} and reliability of at least 0.7.
Its last three columns were registered as a secondary analysis, because a size near floor or ceiling restricts the range of vulnerability.
The smaller models sit near the ceiling, with mean vulnerability of \sixteenVulnMeanOneFifty{} at 150M, \sixteenVulnMeanThreeHundred{} at 300M and \sixteenVulnMeanOneB{} at 1B.
Per-prompt variance is \sixteenVarPOneFifty{}, \sixteenVarPThreeHundred{} and \sixteenVarPOneB{}, so at 150M it is a third of 1B's.
The share of prompts never refused is \sixteenNeverRefusedOneFifty{}, \sixteenNeverRefusedThreeHundred{} and \sixteenNeverRefusedOneB{}.
The share always refused is \sixteenAlwaysRefusedOneFifty{}, \sixteenAlwaysRefusedThreeHundred{} and \sixteenAlwaysRefusedOneB{}.
The reliability correction removes sampling noise, but not the narrower range of the true per-prompt rates.
That narrower range also lowers the same-recipe-pair mean at the small sizes.

Across sizes, the MZ twins' mean stays within \sixteenRmzThreeHundred{}--\sixteenRmzOneFifty{}, so the gap moves with the same-recipe-pair mean.
With intervals over prompts, the gaps of Section~\ref{sec:grid} from 150M to 1B are \sixteenGapOneFifty{} \sixteenGapOneFiftyCI{}, \sixteenGapThreeHundred{} \sixteenGapThreeHundredCI{} and \sixteenGapOneB{} \sixteenGapOneBCI{}.
Over the three runs, the gaps' intervals are \sixteenGapOneFiftyCIbases{}, \sixteenGapThreeHundredCIbases{} and \sixteenGapOneBCIbases{}.
Under one resample of prompts, 150M minus 300M is \sixteenDiffOneFiftyThreeHundred{} \sixteenDiffOneFiftyThreeHundredCI{}.
In the same resample, 150M minus 1B is \sixteenDiffOneFiftyOneB{} \sixteenDiffOneFiftyOneBCI{} and 300M minus 1B is \sixteenDiffThreeHundredOneB{} \sixteenDiffThreeHundredOneBCI{}.
Over prompts, the gaps at both smaller sizes thus lie above the gap at 1B, and 150M and 300M are not resolved.
Over runs, the 150M and 1B intervals overlap while the 300M and 1B intervals do not, and no paired difference over runs was computed.

\begin{table}[h]
  \centering\footnotesize
  \caption{\label{tab:sixteen-secondary}\textbf{The ancestry gap lies above zero at every size under every outcome and in every prompt subset.}
  Rows give a prompt subset and an outcome, and columns the three sizes of the main data recipe.
  Each entry is the gap with its 95\,\% interval over prompts, from \sixteenNboot{} draws, and the 1B models are the twelve of Section~\ref{sec:ladder}.}
  \begin{tabular}{llll}
\toprule
Subset and outcome & 150M & 300M & 1B \\
\midrule
adversarial-harmful, WildGuard's refusal label & 0.204 [0.189, 0.218] & 0.211 [0.198, 0.224] & 0.171 [0.160, 0.183] \\
adversarial-harmful, WildGuard's harmfulness label & 0.064 [0.059, 0.069] & 0.114 [0.106, 0.121] & 0.165 [0.152, 0.180] \\
adversarial-harmful, Qwen3Guard as the judge & 0.112 [0.104, 0.120] & 0.140 [0.131, 0.149] & 0.147 [0.137, 0.157] \\
adversarial-harmful, judges agree & 0.169 [0.157, 0.182] & 0.201 [0.188, 0.213] & 0.161 [0.150, 0.172] \\
plain-harmful, WildGuard's refusal label & 0.130 [0.117, 0.145] & 0.094 [0.084, 0.105] & 0.100 [0.087, 0.114] \\
all-harmful, WildGuard's refusal label & 0.159 [0.149, 0.170] & 0.154 [0.146, 0.162] & 0.140 [0.131, 0.148] \\
benign, WildGuard's refusal label & 0.132 [0.119, 0.147] & 0.127 [0.113, 0.140] & 0.138 [0.126, 0.150] \\
 
\bottomrule
\end{tabular}

\end{table}

Table~\ref{tab:sixteen-secondary} repeats the gap under the other outcomes and subsets, and every interval lies above zero at every size.
Only WildGuard's refusal label has a paired interval between sizes, so the other rows are read from their per-size intervals alone.
Under WildGuard's harmfulness label, the gap is \sixteenGapHarmOneFifty{} \sixteenGapHarmOneFiftyCI{} at 150M against \sixteenGapHarmOneB{} \sixteenGapHarmOneBCI{} at 1B.
Under Qwen3Guard as the judge, it is \sixteenGapQguardOneFifty{} \sixteenGapQguardOneFiftyCI{} at 150M against \sixteenGapQguardOneB{} \sixteenGapQguardOneBCI{} at 1B.
Its 300M and 1B intervals overlap.
In both outcomes, 150M lies below 1B with the two intervals apart, the reverse of the refusal label's order for these two sizes.
At 150M, most adversarial responses are not refusals, with mean vulnerability at \sixteenVulnMeanOneFifty{}.
This is consistent with little variation between models in whether a non-refusal is harmful, which leaves the harm-based outcomes little structure to show, though we have not tested this reading.
When the judges agree, the gap is highest at 300M, at \sixteenGapAgreeThreeHundred{} \sixteenGapAgreeThreeHundredCI{}, against \sixteenGapAgreeOneFifty{} \sixteenGapAgreeOneFiftyCI{} at 150M and \sixteenGapAgreeOneB{} \sixteenGapAgreeOneBCI{} at 1B.
On plain-harmful prompts it is highest at 150M, at \sixteenGapPlainOneFifty{} \sixteenGapPlainOneFiftyCI{}, against \sixteenGapPlainThreeHundred{} and \sixteenGapPlainOneB{}.
From 150M to 1B, the gap is \sixteenGapHarmfulOneFifty{}, \sixteenGapHarmfulThreeHundred{} and \sixteenGapHarmfulOneB{} on all-harmful prompts, and on benign prompts \sixteenGapBenignOneFifty{}, \sixteenGapBenignThreeHundred{} and \sixteenGapBenignOneB{}.

\begin{table}[h]
  \centering\footnotesize
  \caption{\label{tab:sixteen-dfortytwo}\textbf{Further statistics by size.}
  Rows are sizes of the main data recipe, with the 1B models those of Section~\ref{sec:ladder}.
  The first two columns give the mean corrected correlations of MZ twins and same-recipe pairs on WildJailbreak and on HarmBench's human jailbreaks.
  The next two give the template-residual gap and the transfer ratio for MZ twins and same-recipe pairs.
  The last column gives the mean corrected correlation with Ai2's released OLMo~2 1B DPO model.
  No entry carries an interval, and these secondary analyses were stated before the run and are descriptive.}
  \setlength{\tabcolsep}{4pt}
\begin{tabular}{llllll}
\toprule
Size & \shortstack{WildJailbreak:\\MZ twins /\\same-recipe pairs} & \shortstack{HarmBench human\\jailbreaks: MZ twins\\/ same-recipe pairs} & \shortstack{Template-\\residual gap} & \shortstack{Transfer ratio:\\MZ twins /\\same-recipe pairs} & \shortstack{Correlation with\\OLMo~2 1B DPO} \\
\midrule
150M & 0.988 / 0.769 & 0.984 / 0.664 & 0.515 & 0.999 / 0.968 & 0.318 \\
300M & 0.981 / 0.790 & 0.984 / 0.711 & 0.571 & 0.996 / 0.947 & 0.438 \\
1B & 0.987 / 0.854 & 0.983 / 0.773 & 0.545 & 0.993 / 0.932 & 0.672 \\
 
\bottomrule
\end{tabular}

\end{table}

Table~\ref{tab:sixteen-dfortytwo} gives, by size, the mean corrected correlations on each adversarial source, the template-residual gap and the transfer ratio.
Its last column gives each size's correlation with Ai2's released OLMo~2 1B DPO model.
This reference is itself a 1B model.
From 150M to 1B, without intervals, the correlation is \sixteenRefOneFifty{}, \sixteenRefThreeHundred{} and \sixteenRefOneB{}.

One caveat applies at 150M and 300M, where some of the main data recipe's other runs trained past the budget.
We assume that their checkpoints follow the first run's learning-rate schedule, which we read from the code and could not check, since no optimizer state is released.
At 150M, one other run matches the first run's held-out loss at that step to \sixteenLossDiffAuxTwo{} nats.
This match is consistent with the same schedule but does not prove it, and no such check exists at 300M.
Same-recipe pairs show no outlier pair of runs at any size.
If a run were not annealed, the same-recipe-pair mean at that size could fall for a reason other than size.
Such a fall would enlarge the gap at that size, in the direction found above.
Appendix~\ref{app:prereg} gives the registered size prediction and its reading.
\subsection{Which jailbreaks are stable}\label{app:robustness-stable}

\begin{figure}[h]
  \centering
  \includegraphics{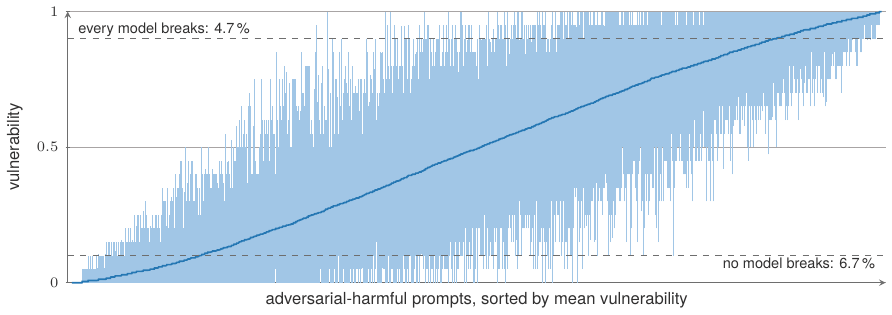}
  \caption{\label{fig:seed-stable}\textbf{Some prompts are vulnerable in all twelve models at least nine times in ten, and some in no model more than one time in ten.}
  The twelve models are the main data recipe's, post-trained under the second sharding setting.
  For each of the \fortyoneStableN{} adversarial-harmful prompts, a thin segment spans the range of vulnerability over the twelve, and a line marks the mean.
  Prompts are sorted by that mean, and dashed lines mark 0.9 and 0.1.}
\end{figure}

The ancestry gap summarizes all prompts at once, and a narrower question asks which single prompts stay stable across models.
Figure~\ref{fig:seed-stable} asks it of the twelve models of Section~\ref{sec:ladder}.
For each prompt, it shows the range of vulnerability over the twelve and the mean.
Of \fortyoneStableN{} prompts, \fortyoneStableNevery{}, or \fortyoneStableEvery{}, are vulnerable in every model at least nine times in ten.
Another \fortyoneStableNnone{}, or \fortyoneStableNone{}, are vulnerable in no model more than one time in ten.

To compare kinds of pair, the same counts run over the 25-recipe grid, on \fortyfourNprompts{} prompts, after dropping one prompt that is undefined in some model.
Recomputed on the twelve models, the grid analysis gives the same two counts.
These counts rest on two thresholds, which also define two statistics of a pair.
The \emph{carry-over rate} of a kind of pair is the share of prompts on which the second model of a pair is vulnerable at least nine times in ten, among those where the first is.
It pools every pair of the type and every prompt, taking each model of a pair as the first in turn.
The \emph{flip share} of a kind of pair is the share of prompts on which exactly one of the two models is vulnerable at least nine times in ten.

\begin{table}[h]
  \centering\footnotesize
  \caption{\label{tab:fortyfour-pairs}\textbf{The carry-over rate is highest between MZ twins.}
  Rows are kinds of pair over the 25-recipe grid, on \fortyfourNprompts{} adversarial-harmful prompts, all but one that is undefined in some model.
  Columns give the number of pairs, the carry-over rate, the flip share, the co-vulnerable share and the share at opposite extremes.
  The co-vulnerable share is the share of prompts on which both models are vulnerable at least nine times in ten.
  The carry-over rate, the flip share and the co-vulnerable share carry 95\,\% intervals over prompts.
  The co-vulnerable share of MZ twins is the same number as the within-base-model share of Table~\ref{tab:fortyfour-levels}.}
  \begin{tabular}{llllll}
\toprule
Kind of pair & Pairs & \shortstack{Carry-over\\rate} & \shortstack{Flip\\share} & \shortstack{Co-vulnerable\\share} & \shortstack{Share at\\opposite extremes} \\
\midrule
MZ twins & 75 & 0.750 [0.740, 0.759] & 0.108 [0.104, 0.112] & 0.162 [0.153, 0.171] & \ensuremath{4.2\times 10^{-6}} \\
same-recipe pairs & 300 & 0.609 [0.594, 0.622] & 0.170 [0.164, 0.176] & 0.132 [0.123, 0.141] & \ensuremath{2.1\times 10^{-3}} \\
cross-recipe pairs & 10{,}800 & 0.587 [0.571, 0.602] & 0.179 [0.172, 0.185] & 0.127 [0.119, 0.136] & \ensuremath{3.9\times 10^{-3}} \\
 
\bottomrule
\end{tabular}

\end{table}

\begin{figure}[h]
  \centering
  \begin{tikzpicture}
\pgfplotsset{stab/.style={castor axis, width=0.33\textwidth, height=0.3\textwidth,
  xmin=0.5, xmax=3.5, xtick={1,2,3},
  xticklabels={MZ\\twins, same-recipe\\pairs, cross-recipe\\pairs},
  xticklabel style={align=center}}}
\newcommand{\stabpoints}[4]{%
  \foreach \pair/\role in {MZ/pair-mz, U/pair-unrelated, R/pair-cross}{
    \edef\plotcmd{\noexpand\addplot[castor level, fill=\role, draw=\role, castor keep={pair}{\pair}, forget plot]}
    \plotcmd table[x=x, y=#2, col sep=tab,
      y error plus expr=\thisrow{#4}-\thisrow{#2},
      y error minus expr=\thisrow{#2}-\thisrow{#3}] {#1};
  }}
\begin{axis}[stab, name=a, ymin=0.5, ymax=0.8, ytick={0.5,0.6,0.7,0.8},
  ylabel={carry-over rate}, title={(a) carry-over rate}, title style={font=\castorlabel, text=text}]
\stabpoints{figures/data/stability_pairs.tsv}{transfer}{transfer_lo}{transfer_hi}
\end{axis}
\begin{axis}[stab, name=b, at={(a.east)}, anchor=west, xshift=0.06\textwidth,
  ymin=0, ymax=0.2, ytick={0,0.1,0.2}, yticklabel style={/pgf/number format/fixed},
  ylabel={flip share}, title={(b) flip share}, title style={font=\castorlabel, text=text}]
\stabpoints{figures/data/stability_pairs.tsv}{flip}{flip_lo}{flip_hi}
\end{axis}
\begin{axis}[stab, name=c, at={(b.east)}, anchor=west, xshift=0.06\textwidth,
  ymin=0.6, ymax=1.0, ytick={0.6,0.7,0.8,0.9,1.0},
  ylabel={corrected template $\rho$}, title={(c) template ranks},
  title style={font=\castorlabel, text=text}]
\stabpoints{figures/data/stability_templates.tsv}{r}{lo}{hi}
\end{axis}
\end{tikzpicture}
  \caption{\label{fig:stability}\textbf{MZ twins share more of their most vulnerable prompts and more of their template ranking than other pairs.}
  A point is one kind of pair over the 25-recipe grid, on \fortyfourNprompts{} adversarial-harmful prompts, all but one that is undefined in some model, with 75, 300 and 10,800 pairs.
  Panel~a gives the carry-over rate, the probability that the other model is vulnerable at least nine times in ten given that one is.
  Panel~b gives the flip share, and panel~c the corrected template rank correlation over \fortyfourNtemplates{} templates.
  Intervals are 95\,\% bootstrap intervals over prompts in panels a and b and over templates in panel c, from 1,000 draws, and no threshold lines are drawn.
  Table~\ref{tab:fortyfour-levels}, not this figure, gives the shares for sets of models.}
\end{figure}
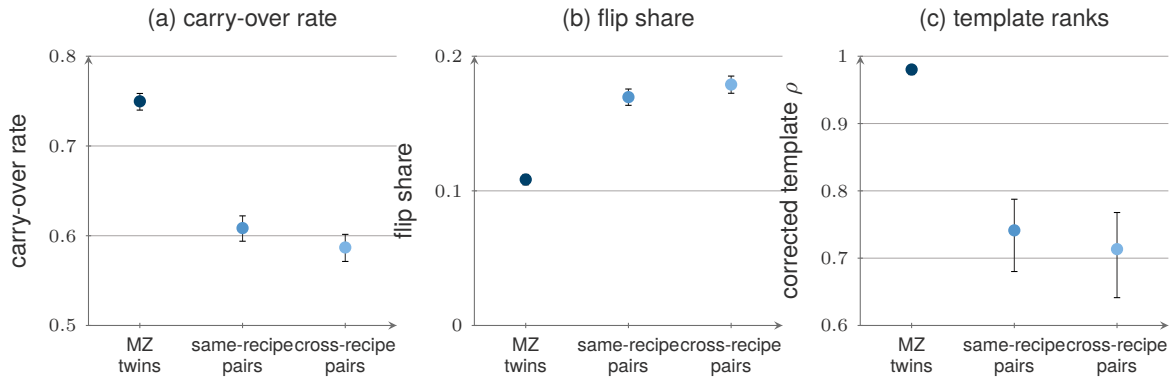

Table~\ref{tab:fortyfour-pairs} and Figure~\ref{fig:stability} give the carry-over rate and the flip share per kind of pair, with intervals over prompts.
The carry-over rate is \fortyfourTransferMZ{} \fortyfourTransferMZCI{} for MZ twins, \fortyfourTransferU{} \fortyfourTransferUCI{} for same-recipe pairs and \fortyfourTransferR{} \fortyfourTransferRCI{} for cross-recipe pairs.
The flip share is \fortyfourFlipMZ{} \fortyfourFlipMZCI{} for MZ twins, against \fortyfourFlipU{} \fortyfourFlipUCI{} and \fortyfourFlipR{} \fortyfourFlipRCI{}.
The carry-over rate falls most with a different base model, and a different data recipe lowers it by a further \fortyfourTransferUminusR{}.
That drop is a difference of point estimates, whose two intervals overlap and which has no paired interval.
This suggests that a jailbreak found reliable on one model carries over more often to models of its base model than to models of another base model.

The share at opposite extremes is the share of prompts at least 0.9 on one model and at most 0.1 on the other.
This share shows that flips between MZ twins are threshold crossings, not reversals.
For MZ twins it is \fortyfourStrongFlipMZ{}, \fortyfourStrongFlipMZCount{} pair-and-prompt combination in \fortyfourPairPromptsMZ{}.
For same-recipe pairs and cross-recipe pairs it is \fortyfourStrongFlipU{} and \fortyfourStrongFlipR{}.
Part of the MZ twins' flip share is sampling noise from \sixK{} samples near the 0.9 line.
That share is therefore an upper bound on flips beyond sampling noise, and we apply no correction.

Of the statistics of Table~\ref{tab:fortyfour-pairs}, only the MZ twins' flip share was registered.
The shares of Table~\ref{tab:fortyfour-levels} and the template correlations were registered too.
The carry-over rate was added at analysis, as change~\ref{prereg-change:carryover} of Appendix~\ref{app:prereg} records.
So were the other flip shares, the co-vulnerable share and the share at opposite extremes.

Templates give a second view of stability, shown in Figure~\ref{fig:stability}c.
The \emph{corrected template rank correlation} of two models is the Spearman correlation of their templates' mean vulnerabilities over \fortyfourNtemplates{} templates, divided by the square root of their two reliabilities of that ranking.
A model's reliability of the ranking is the Spearman correlation between the rankings from its even- and odd-indexed samples, stepped up by the Spearman--Brown formula.
Over models, this reliability runs from \fortyfourTemplateRelMin{} to \fortyfourTemplateRelMax{}.
With intervals over templates, the corrected template rank correlation is \fortyfourTemplateMZ{} \fortyfourTemplateMZCI{} for MZ twins, \fortyfourTemplateU{} \fortyfourTemplateUCI{} for same-recipe pairs and \fortyfourTemplateR{} \fortyfourTemplateRCI{} for cross-recipe pairs.
The paired same-recipe-minus-cross-recipe difference is \fortyfourTemplateUminusR{} \fortyfourTemplateUminusRCI{}.
This difference was chosen after the data, as post hoc item~\ref{prereg-posthoc:template-pair} of Appendix~\ref{app:prereg} records.
A template's mean vulnerability mixes the difficulty of the template with that of its behaviors, which the template residual of Section~\ref{sec:grid} removes.

\begin{table*}[t]
  \centering\footnotesize
  \caption{\label{tab:fortyfour-levels}\textbf{Shares of prompts vulnerable in every model of a set.}
  Rows are sets of models within a base model, within a data recipe and across data recipes, and a fourth row takes all 150 models as one secondary set.
  Columns give what varies within a set and the number of sets times the models in each.
  The two shares count prompts vulnerable at least nine times in ten in every model of the set, and at most one time in ten in every model, both with 95\,\% intervals over prompts.}
  \setlength{\tabcolsep}{3pt}
\begin{tabular}{>{\raggedright\arraybackslash}p{2.6cm}>{\raggedright\arraybackslash}p{2.8cm}lll}
\toprule
Set of models & What varies & \shortstack{Sets $\times$\\models} & \shortstack{Vulnerable at least\\nine times in ten\\in every model} & \shortstack{Vulnerable at most\\one time in ten\\in every model} \\
\midrule
within a base model & post-training seeds & 75 $\times$ 2 & 0.162 [0.153, 0.171] & 0.139 [0.131, 0.148] \\
within a data recipe & base models and post-training seeds & 25 $\times$ 6 & 0.068 [0.061, 0.074] & 0.061 [0.055, 0.067] \\
across data recipes & the post-training-seed-1 model of every base model & 1 $\times$ 75 & 0.005 [0.003, 0.007] & 0.006 [0.004, 0.009] \\
across data recipes, all models & every model & 1 $\times$ 150 & 0.004 [0.002, 0.006] & 0.004 [0.002, 0.006] \\
 
\bottomrule
\end{tabular}

\end{table*}

Table~\ref{tab:fortyfour-levels} turns from pairs to sets of models.
In its first three rows, each set contains 2, 6 or 75 models, so the shares fall partly because the sets grow.
The carry-over rate fixes the count at two models, so it is the comparison matched for set size.
These shares are descriptive, as changes~\ref{prereg-change:carryover} and~\ref{prereg-change:stable-group} of Appendix~\ref{app:prereg} record.
\subsection{How many base models and seeds a comparison needs}\label{app:robustness-seeds}

The shares of Appendix~\ref{app:robustness-grid} split per-model variance by factor.
This split asks how precisely a few models estimate a data recipe's mean, for the practical reading of Section~\ref{sec:limitations}.
We split the variance of a data recipe's mean vulnerability into nested parts: the data recipe, the base model within it and the post-training seed within the base model.
The split uses the 25 × 3 × 2 models of the 25-recipe grid and adds sampling noise at \sixK{} samples per prompt.
We estimate the parts by a method-of-moments nested analysis of variance.
A point is 0.01 of the judge's rate, not corrected for the judge's precision.

The standard deviations are \fortyfiveSDRecipe{} points for the data recipe, \fortyfiveSDPretrain{} for the base model, \fortyfiveSDPosttrain{} for the post-training seed and \fortyfiveSDSampling{} for sampling.
The base model's variance minus the post-training seed's is \fortyfivePreMinusPostPtSq{} \fortyfivePreMinusPostPtSqCI{} squared points.
Its interval includes zero, so we claim no ordering of the two.

\begin{table}[h]
  \centering\footnotesize
  \caption{\label{tab:fortyfive-se}\textbf{Three base models with one post-training seed each reach the 2-point threshold on the point estimate, and no design we measured reaches 1 point.}
  Each entry is the standard error of a data recipe's mean vulnerability from the given numbers of base models and post-training seeds per base model.
  It is in points, one point being 0.01 of the judge's vulnerability.
  The standard errors come from the variance parts over 25 data recipes, with 95\,\% intervals from resampling data recipes.}
  \begin{tabular}{llll}
\toprule
 & \multicolumn{3}{c}{Post-training seeds per base model} \\
\cmidrule(l){2-4}
Base models & 1 & 2 & 4 \\
\midrule
1 & 3.44 [2.96, 3.88] & 3.11 [2.53, 3.60] & 2.93 [2.31, 3.45] \\
2 & 2.43 [2.09, 2.75] & 2.20 [1.79, 2.55] & 2.07 [1.63, 2.44] \\
3 & 1.99 [1.71, 2.24] & 1.79 [1.46, 2.08] & 1.69 [1.33, 1.99] \\
 
\bottomrule
\end{tabular}

\end{table}

\begin{figure}[h]
  \centering
  \begin{tikzpicture}
\begin{axis}[castor axis, name=a, width=0.48\textwidth, height=0.3\textwidth,
  xmin=-0.7, xmax=13.2, ymin=0, ymax=4, ytick={0,1,2,3,4},
  xtick={0,1,2,4,5,6,8,9,10,12.5},
  xticklabels={1$\times$1, 1$\times$2, 1$\times$4, 2$\times$1, 2$\times$2, 2$\times$4,
    3$\times$1, 3$\times$2, 3$\times$4, 12$\times$1},
  xticklabel style={rotate=45, anchor=north east, inner sep=1pt},
  xlabel={design: pretraining $\times$ post-training seeds},
  ylabel={SE of the mean (points)},
  title={(a) standard error by design}, title style={font=\castorlabel, text=text}]
\foreach \t in {1, 2}{
  \edef\drawcmd{\noexpand\draw[muted, dashed, line width=0.4pt]
    ({rel axis cs:0,0}|-{axis cs:0,\t}) -- ({rel axis cs:1,0}|-{axis cs:0,\t});}
  \drawcmd
}
\addplot[castor level, fill=series-main, draw=series-main]
  table[x=x, y expr=100*\thisrow{se}, col sep=tab,
    y error plus expr=100*(\thisrow{se_hi}-\thisrow{se}),
    y error minus expr=100*(\thisrow{se}-\thisrow{se_lo})] {figures/data/seedpower_se.tsv};
\addplot[only marks, mark=o, mark size=2pt, draw=series-main, mark options={fill=box}]
  table[x=x, y expr=100*\thisrow{se}, col sep=tab] {figures/data/seedpower_extrap.tsv};
\draw[muted, line width=0.4pt, dotted] (axis cs:11.25,0) -- (axis cs:11.25,4);
\node[anchor=north, font=\castortick, text=muted, align=center] at (axis cs:12.5,3.9)
  {extra-\\polated};
\end{axis}
\begin{axis}[castor axis, name=b, at={(a.east)}, anchor=west, xshift=0.07\textwidth,
  width=0.4\textwidth, height=0.3\textwidth,
  xmin=-0.5, xmax=4.5, ymin=0, ymax=0.5, ytick={0,0.1,0.2,0.3,0.4,0.5},
  yticklabel style={/pgf/number format/fixed},
  xtick={0,1,2,3,4}, xticklabels={0--1, 1--2, 2--4, 4--8, $\geq$8},
  xlabel={$|$all-seed difference$|$ (points)},
  ylabel={wrong-sign rate},
  legend style={at={(0.97,0.97)}, anchor=north east},
  title={(b) single-seed comparisons}, title style={font=\castorlabel, text=text}]
\addplot[series-main, mark=*, mark size=1.8pt]
  table[x=x, y=flip_empirical, col sep=tab] {figures/data/seedpower_flip.tsv};
\addlegendentry{empirical}
\addplot[series-aux, dashed, mark=o, mark size=1.8pt, mark options={solid, fill=box}]
  table[x=x, y=flip_model, col sep=tab] {figures/data/seedpower_flip.tsv};
\addlegendentry{model-based}
\end{axis}
\end{tikzpicture}
  \caption{\label{fig:seedpower}\textbf{One model per data recipe leaves a standard error of about \fortyfiveSEOneOneRounded{} points and often puts close data recipes in the opposite order to their six-model means.}
  In panel~a, a point is the standard error of a data recipe's mean vulnerability from a number of base models times post-training seeds, in points of 0.01 of the judge's vulnerability.
  Standard errors come from the variance parts over 25 data recipes, with 95\,\% bootstrap intervals over data recipes from 1,000 draws.
  Dashed lines mark 2 and 1 points, and the open mark is \fortyfiveLeOnePtNpre{} × \fortyfiveLeOnePtNpost{}, from the same formula without an interval.
  Panel~b gives the opposite-order share of single-model comparisons, 300 pairs of data recipes × 36, against the six-model difference.
  Its bins are 0--1, 1--2, 2--4, 4--8 and at least 8 points.
  Filled marks are observed shares and open dashed marks the normal model $\Phi(-|\Delta|/\sqrt{2\sigma^2})$, with no interval per bin.}
\end{figure}
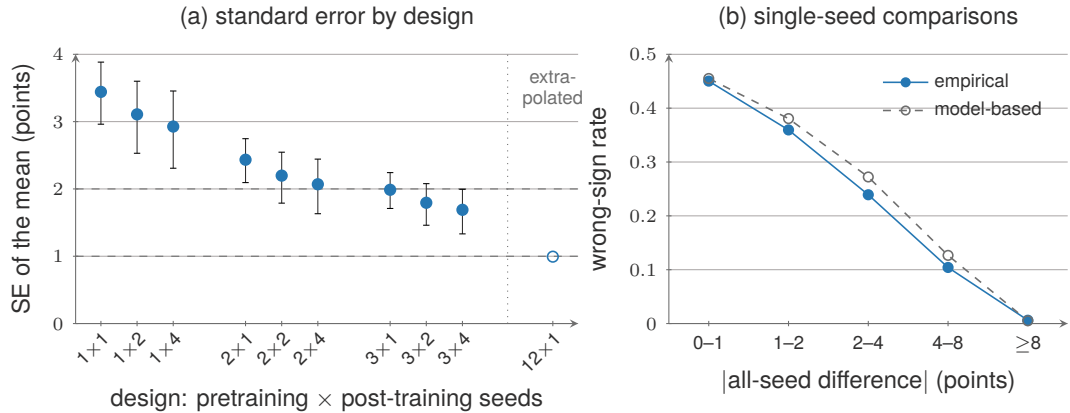

From these parts, the squared standard error of a data recipe's mean is $\sigma^2_{\mathrm{B}}/n_{\mathrm{B}} + \sigma^2_{\mathrm{P}}/(n_{\mathrm{B}} n_{\mathrm{P}})$.
Here $n_{\mathrm{B}}$ base models each get $n_{\mathrm{P}}$ post-training seeds, and $\sigma^2_{\mathrm{B}}$ and $\sigma^2_{\mathrm{P}}$ are the variance parts of the base model and of the post-training seed, sampling included in the second.
Table~\ref{tab:fortyfive-se} and Figure~\ref{fig:seedpower}a give this standard error for the designs measured.
One model leaves \fortyfiveSEOneOne{} \fortyfiveSEOneOneCI{} points.
On the point estimate, the standard error first reaches 2 points at three base models with one post-training seed each, at \fortyfiveSEThreeOne{} \fortyfiveSEThreeOneCI{}.
None of the designs measured, one to three base models with one, two or four post-training seeds each, reaches 1 point.
By the same formula beyond them, \fortyfiveLeOnePtNpre{} base models with one seed each reach it, and no number of post-training seeds does below \fortyfiveFloorNpre{} base models.
On point estimates, one base model with four post-training seeds gives a larger standard error, \fortyfiveSEOneFour{}, than two base models with one each, \fortyfiveSETwoOne{}.
Their intervals overlap, and no paired interval exists.

Figure~\ref{fig:seedpower}b turns from the standard error to single-model comparisons, which take one model per data recipe.
Each of 300 pairs of data recipes gives 36 such comparisons.
In these comparisons, two data recipes come out in the opposite order to their six-model means \fortyfiveFlipOverall{} \fortyfiveFlipOverallCI{} of the time.
For the \fortyfiveNrecipePairsGeTwo{} pairs at least 2 points apart, this opposite-order share is \fortyfiveFlipGeTwo{} \fortyfiveFlipGeTwoCI{}.
From the closest bin to the farthest, it falls from \fortyfiveFlipBinZeroOne{} to \fortyfiveFlipBinEightUp{}.
The normal model takes the difference of two independent one-model means, each with variance $\sigma^2_{\mathrm{B}} + \sigma^2_{\mathrm{P}}$.
We derive this model from first principles.
Two choices in this comparison were made after seeing the data: the bins of Figure~\ref{fig:seedpower}b and its normal model.
The normal model sits slightly above the observed shares, at \fortyfiveFlipModelOverall{} overall.
This is consistent with the six-model mean containing the comparison's own models, a reading we have not tested.

A data recipe can also be compared by its per-prompt profile rather than by its mean.
We estimate how well the average of several models' per-prompt vulnerabilities stands for its data recipe's profile.
The reliability of that average is $p_{\mathrm{S}} / (p_{\mathrm{S}} + (p_{\mathrm{MZ}} - p_{\mathrm{S}})/n_{\mathrm{B}} + (1 - p_{\mathrm{MZ}})/(n_{\mathrm{B}} n_{\mathrm{P}}))$, where $p_{\mathrm{MZ}}$ and $p_{\mathrm{S}}$ are the uncorrected Pearson correlations of a data recipe's MZ twins and same-recipe pairs.
These correlations stand in for intraclass correlations.
With intervals over data recipes, the reliability is \fortyfiveRelOne{} \fortyfiveRelOneCI{} for one model and \fortyfiveRelTwo{} \fortyfiveRelTwoCI{} for two base models.
It is \fortyfiveRelThree{} \fortyfiveRelThreeCI{} for three base models, and \fortyfiveRelSix{} \fortyfiveRelSixCI{} for three base models with two post-training seeds each.
Of a single model's per-prompt noise, \fortyfiveNoiseTotal{}, the base model contributes \fortyfiveNoiseBase{}, sampling \fortyfiveNoiseSampling{} and the post-training seed \fortyfiveNoisePost{}.
The mapping of 1, 2, 3 and 6 averaged models to 1 × 1, 2 × 1, 3 × 1 and 3 × 2 designs was also chosen after seeing the data.

\begin{table*}[t]
  \centering\footnotesize
  \caption{\label{tab:fortyfive-cells}\textbf{The variance analysis on benign prompts and under Qwen3Guard as the judge, beside WildGuard's refusal label on adversarial-harmful prompts.}
  For each prompt subset and outcome, columns give the standard deviations of the data recipe, the base model and the post-training seed in points.
  Next comes the base model's variance minus the post-training seed's in squared points, with its 95\,\% interval over data recipes.
  Then come the one-model standard error and the smallest designs whose standard error reaches 2 and 1 points, as base models × post-training seeds.
  The designs measured are one to three base models × one, two or four post-training seeds.
  Every design in the 1-point column, and the benign 6 × 1, lies beyond them and comes from the same formula.
  The opposite-order share is given over all pairs of data recipes and over pairs at least 2 points apart.
  The last column gives the reliability of a data recipe's per-prompt profile averaged over 1, 3 and 6 models.}
  \setlength{\tabcolsep}{4pt}
\begin{tabular*}{\textwidth}{@{\extracolsep{\fill}}>{\raggedright\arraybackslash}p{3.2cm}lll}
\toprule
Subset and outcome & \shortstack{SD in points: data recipe /\\base model / post-training seed} & \shortstack{Base-model minus post-training-seed\\variance in points$^2$, 95\,\% interval} & \shortstack{One-model SE\\in points} \\
\midrule
adversarial-harmful, WildGuard's refusal label & 3.5 / 2.7 / 2.1 & 3.1 [\ensuremath{-}0.6, 6.9] & 3.44 \\
benign prompts, WildGuard's refusal label & 1.6 / 4.1 / 1.9 & 12.8 [5.1, 22.4] & 4.51 \\
adversarial-harmful, Qwen3Guard as the judge & 1.7 / 2.4 / 1.8 & 2.4 [\ensuremath{-}0.4, 5.1] & 3.02 \\
 
\bottomrule
\end{tabular*}

\medskip
\begin{tabular*}{\textwidth}{@{\extracolsep{\fill}}>{\raggedright\arraybackslash}p{3.2cm}llll}
\toprule
Subset and outcome & \shortstack{Smallest design,\\SE $\le$ 2 points} & \shortstack{Smallest design,\\SE $\le$ 1 point} & \shortstack{Opposite-order share:\\all / $\ge$ 2 points apart} & \shortstack{Profile reliability:\\1 / 3 / 6 models} \\
\midrule
adversarial-harmful, WildGuard's refusal label & 3 $\times$ 1 & 12 $\times$ 1 & 0.234 / 0.148 & 0.759 / 0.904 / 0.919 \\
benign prompts, WildGuard's refusal label & 6 $\times$ 1 (beyond the registered designs) & 19 $\times$ 2 & 0.283 / 0.211 & 0.820 / 0.932 / 0.944 \\
adversarial-harmful, Qwen3Guard as the judge & 3 $\times$ 1 & 10 $\times$ 1 & 0.265 / 0.165 & 0.784 / 0.916 / 0.931 \\
 
\bottomrule
\end{tabular*}

\end{table*}

Table~\ref{tab:fortyfive-cells} repeats the analysis on benign prompts and under Qwen3Guard as the judge.
On benign prompts, the base model's variance minus the post-training seed's is \fortyfivePreMinusPostBenignPtSq{} \fortyfivePreMinusPostBenignPtSqCI{} squared points, an interval above zero.
On the same prompts, one model leaves a standard error of \fortyfiveSEOneOneBenign{} points.
Under Qwen3Guard as the judge, the base model's part is again larger on the point estimate, but the interval of the difference, \fortyfivePreMinusPostQguardPtSqCI{}, includes zero.
Under this judge, the standard deviations of the data recipe, the base model and the post-training seed are \fortyfiveSDRecipeQguard{}, \fortyfiveSDPretrainQguard{} and \fortyfiveSDPosttrainQguard{} points.
The data recipe's part is the smallest, as in the variance shares of the 25-recipe grid.
Appendix~\ref{app:prereg} gives each registered prediction of this appendix with its reading.
\subsection{Other public safety data}\label{app:robustness-safety-data}

Every model above learned to refuse from Tülu 3's safety data, 110,800 of the SFT mixture's 866,138 rows: 50,000 from WildGuardMix, 49,965 from WildJailbreak and 10,835 from CoCoNot.
We replace these rows by 110,800 rows of other public safety data, the second safety data for short, with the same number of refusals of harmful prompts, 52,123.
They come from Nemotron-SFT-Safety-v2, the safety data used in NVIDIA's post-training \citep{nvidia2025nemotron3nano}, with 38,194 rows, Llama-Nemotron's safety subset \citep{bercovich2025llamanemotron} with 9,064, PKU-SafeRLHF \citep{ji2024pkusaferlhf} with 44,621, BeaverTails \citep{ji2023beavertails} with 10,187 and the harmless split of HH-RLHF \citep{bai2022hhrlhf} with 8,734.
The other 755,338 rows, the DPO stage and every other setting are unchanged.
Beyond their sources, the two safety data differ in three declared ways.
The second safety data refuses no benign prompt, where Tülu 3's refuses 10,087, contains no harmful response, where Tülu 3's contains 225, and refuses in 0.470 of its rows against 0.565.
The unchanged DPO stage still meets some of the removed prompts, since 4.35\,\% of its 378,301 rows match a user turn of the removed rows exactly or by 8-gram overlap, 3.89\,\% exactly.
The two sets of models therefore differ in their SFT safety data only.

Under the second safety data we post-train the main data recipe's three base models with post-training seeds 1 and 2.
With seed 1 we also post-train the three base models of each of the next two data recipes in the 25-recipe grid's order, DCLM-Baseline 25\,\% / Dolma 75\,\% and 50\,\% / 50\,\%.
Under Tülu 3's safety data the comparison takes the main data recipe's twelve models, three base models with post-training seeds 1 to 4.
By a \emph{pair of one run across safety data} we mean a model of one pretraining run under Tülu 3's safety data with a model of the same run under the second safety data.
A \emph{pair of two runs across safety data} takes two different runs of the main data recipe, one model under each safety data.
The main data recipe gives \eightyAdoptedNpairs{} pairs of one run and \eightyCrossNpairs{} of two runs across safety data, and \eightySwapMZNpairs{} MZ twins and \eightySwapUNpairs{} same-recipe pairs under the second safety data, against 18 and 48 under Tülu 3's.
Prompts, judge and correction are those of Section~\ref{sec:setup}: the adversarial-harmful prompts, WildGuard's refusal label and the corrected correlation, with prompt-bootstrap intervals from 1,000 draws and, beside the main numbers, a base-model jackknife standard error that is coarse over three base models.

Within the second safety data, the ancestry gap is about the same size as under Tülu 3's.
Table~\ref{tab:other-safety-data} gives every mean corrected correlation of this subsection, over all adversarial-harmful prompts and per prompt source.
Under the second safety data, MZ twins correlate at \eightySwapMZ{} \eightySwapMZCI{} and same-recipe pairs at \eightySwapU{} \eightySwapUCI{}.
Their gap, itself registered, is \eightySwapGap{} \eightySwapGapCI{}, with a jackknife standard error of \eightySwapGapJackSE{}, against \eightGapMain{} for the same three runs at post-training seeds 1 and 2 under Tülu 3's safety data, and against the registered comparator, the 25-recipe grid's gap of \eightGapMean{} \eightGapCIprompts{}.

Under the second safety data, the run also outweighs the data recipe.
In a comparison added after seeing the data, we set the run's difference, MZ twins minus same-recipe pairs, against the data recipe's difference, same-recipe pairs minus cross-recipe pairs, on the three data recipes the second safety data covers.
Under the second safety data the two are \eightyPosthocLadderSwapMZMinusU{} \eightyPosthocLadderSwapMZMinusUCI{} and \eightyPosthocLadderSwapUMinusR{} \eightyPosthocLadderSwapUMinusRCI{}, a ratio of \eightyPosthocLadderSwapRatio{} \eightyPosthocLadderSwapRatioCI{}.
At the same base models and seeds under Tülu 3's safety data they are \eightyPosthocLadderTuluMZMinusU{} \eightyPosthocLadderTuluMZMinusUCI{} and \eightyPosthocLadderTuluUMinusR{} \eightyPosthocLadderTuluUMinusRCI{}, a ratio of \eightyPosthocLadderTuluRatio{} \eightyPosthocLadderTuluRatioCI{}, so the data recipe's difference is larger under the second safety data by \eightyPosthocLadderUMinusRDiff{} \eightyPosthocLadderUMinusRDiffCI{}.
Averaged over the three data recipes, MZ twins thus correlate above same-recipe pairs and same-recipe pairs above cross-recipe pairs under both safety data, and the run's difference is several times the data recipe's.
These three data recipes are DCLM-Baseline and two of its mixes with Dolma rather than all 25 data recipes, their MZ twins come from the main data recipe alone, and the intervals resample prompts, not data recipes.

\begin{table}[h]
  \centering\footnotesize
  \caption{\label{tab:other-safety-data}\textbf{Under each safety data MZ twins correlate above same-recipe pairs, and across the two safety data a pair of one run correlates above a pair of two runs, on both prompt sources.}
  Each entry is the mean corrected correlation over the pairs of its row, with a 95\,\% prompt-bootstrap interval over 1,000 draws, taken within each source in the source columns.
  The registered values are in the first column: the four rows under the second safety data and across safety data, and the difference of MZ twins and same-recipe pairs under the second safety data.
  $^\dagger$: added after seeing the data.}
  \scriptsize
\setlength{\tabcolsep}{4pt}
\begin{tabular}{@{}llccc@{}}
  \toprule
  Kind of pair & Pairs & \shortstack{All adversarial-\\harmful (\eightyNprompts{})} & \shortstack{WildJailbreak$^\dagger$\\(\eightyPosthocWjN{})} & \shortstack{HarmBench human\\jailbreaks$^\dagger$ (\eightyPosthocHjN{})} \\
  \midrule
  \multicolumn{5}{@{}l}{\emph{T\"ulu~3's safety data}} \\
\quad MZ twins & 18 & 0.985 [0.983, 0.987] & 0.987 [0.985, 0.990] & 0.983 [0.980, 0.986] \\
\quad Same-recipe pairs & 48 & 0.814 [0.801, 0.825] & 0.854 [0.842, 0.867] & 0.773 [0.752, 0.793] \\
\midrule\multicolumn{5}{@{}l}{\emph{Other public safety data}} \\
\quad MZ twins & 3 & 0.966 [0.962, 0.970] & 0.960 [0.953, 0.966] & 0.970 [0.965, 0.976] \\
\quad Same-recipe pairs & 12 & 0.788 [0.771, 0.802] & 0.778 [0.753, 0.801] & 0.772 [0.750, 0.793] \\
\midrule\multicolumn{5}{@{}l}{\emph{Across safety data}} \\
\quad Pairs of one run & 24 & 0.558 [0.537, 0.577] & 0.390 [0.362, 0.418] & 0.821 [0.806, 0.837] \\
\quad Pairs of two runs & 48 & 0.470 [0.446, 0.490] & 0.339 [0.309, 0.369] & 0.684 [0.659, 0.707] \\
\midrule\multicolumn{5}{@{}l}{\emph{Differences}} \\
\quad MZ twins $-$ same-recipe pairs, other data &  & +0.178 [0.165, 0.194] & +0.182 [0.161, 0.204] & +0.198 [0.177, 0.220] \\
\quad One run $-$ two runs, across$^\dagger$ &  & +0.088 [0.081, 0.096] & +0.051 [0.043, 0.058] & +0.137 [0.124, 0.153] \\
 
  \bottomrule
\end{tabular}

\end{table}

Across the two safety data, the models of one run correlate above the models of two runs.
Pairs of one run across safety data correlate at \eightyAdopted{} \eightyAdoptedCI{}, with a jackknife standard error of \eightyAdoptedJackSE{}, against pairs of two runs across safety data at \eightyCross{} \eightyCrossCI{}.
This correlation is the registered statistic, read against its prediction below.
The difference of the two, chosen after seeing the data, is \eightyAdoptedMinusCross{} \eightyAdoptedMinusCrossCI{}, with a jackknife standard error of \eightyAdoptedMinusCrossJackSE{}.
On the two other data recipes, with six pairs of one run across safety data per data recipe and outside the registration, these pairs correlate at \eightyExploreTwentyFiveAdopted{} and \eightyExploreFiftyAdopted{}, against \eightyExploreTwentyFiveU{} and \eightyExploreFiftyU{} for their same-recipe pairs under Tülu 3's safety data.

A model's vulnerability profile can also name its run across the change of safety data.
The registration predicted that, with all prompts, at least 0.80 of the twelve models under the second safety data would be matched to their own run, each by its profile to the nearest of the 75 runs' models at post-training seed 1, as in Section~\ref{sec:witness}.
On \eightyFpNprompts{} prompts, \eightyFpCorrect{} of \eightyFpN{} models match their own run, with an exact interval of \eightyFpAccCI{}.
The margins are thin: in uncorrected Pearson correlation the own run averages \eightyFpTrueMean{} against \eightyFpImpostorMean{} for the best other run, and the smallest margin is \eightyFpMarginMin{}.
Fewer prompts name the run less often, \eightyFpThousand{} \eightyFpThousandRange{} with 1,000 random prompts and \eightyFpFifty{} \eightyFpFiftyRange{} with 50, each range from 2.5 to 97.5\,\% over 200 subsets, where a model's MZ twin names its run at \seventysixRunFifty{} with 50 prompts (Section~\ref{sec:witness}).
In the registered descriptive check, the best other run never comes from the model's own data recipe, \eightyFpImpostorSameRecipe{} of \eightyFpN{}, and is Falcon+CC (QC Orig 10\,\%) for \eightyFpImpostorFalconOrigTen{} models, Dolma1.7 (no Flan) for \eightyFpImpostorDolmaNoFlan{} and Dolma1.6++ for \eightyFpImpostorDolmaSixPlus{}.
The profile thus names its run across a change of SFT safety data when all prompts are used, by a small margin.

The registered test compares pairs of one run across safety data with a stricter reference than pairs of two runs across safety data.
The registration predicted that pairs of one run across safety data would correlate above the mean of same-recipe pairs over the 25-recipe grid, \eightyRecordU{}, with an interval excluding it.
It fixed three bands in advance: above that mean, the run outweighs the safety data; between it and the mean of same-recipe pairs under the second safety data, part of the run's share carries over; at or below that second mean, the claim is scoped to fixed safety data.
The observed \eightyAdopted{} \eightyAdoptedCI{} lies below both \eightyRecordU{} and \eightySwapU{}, in the third band.
The difference of pairs of one run across safety data from same-recipe pairs under Tülu 3's safety data is \eightyAdoptedMinusU{} \eightyAdoptedMinusUCI{}.
This contrast comes from the registered analysis but was not itself registered.
With the run fixed, changing the SFT safety data moved per-prompt vulnerability more than changing the run under one safety data, so the paper's comparison of runs is scoped to fixed post-training.
This test rests on one data recipe, three base models and one second safety data, with the SFT safety data changed and the DPO stage unchanged.
Appendix~\ref{app:prereg} gives this prediction's row.
A split by prompt source, added after seeing the data, shows where pairs of one run across safety data fall below same-recipe pairs under Tülu 3's safety data.
It keeps the same pairs and takes reliabilities and the bootstrap within each source.
Pairs of one run across safety data correlate at \eightyPosthocWjAdopted{} \eightyPosthocWjAdoptedCI{} on the \eightyPosthocWjN{} WildJailbreak prompts and at \eightyPosthocHjAdopted{} \eightyPosthocHjAdoptedCI{} on the \eightyPosthocHjN{} human jailbreaks of HarmBench, a difference of \eightyPosthocAdoptedHjMinusWj{} \eightyPosthocAdoptedHjMinusWjCI{}.
Same-recipe pairs under Tülu 3's safety data correlate at \eightyPosthocWjTuluU{} and \eightyPosthocHjTuluU{} on the two sources, so pairs of one run across safety data sit above them by \eightyPosthocHjAdoptedMinusU{} \eightyPosthocHjAdoptedMinusUCI{} on the human jailbreaks and far below them on WildJailbreak, by \eightyPosthocWjAdoptedMinusU{}.
Same-recipe pairs under the second safety data barely differ by source (\eightyPosthocWjSwapU{} against \eightyPosthocHjSwapU{}).
The shortfall across safety data thus sits almost wholly on WildJailbreak.
This is consistent with the removed rows including WildJailbreak's training prompts.

The change of safety data also moves how vulnerable the models are, against the registered expectation.
The registration expected more vulnerability on these prompts under the second safety data, and the main data recipe's models are less vulnerable, at a mean vulnerability of \eightyMeanVulnSwap{} against \eightyMeanVulnTulu{} in the judge's rates.
The two other data recipes fall the same way, \eightyMeanVulnSwapTwentyFive{} against \eightyMeanVulnTuluTwentyFive{} and \eightyMeanVulnSwapFifty{} against \eightyMeanVulnTuluFifty{}.
On the main data recipe, reliabilities are \eightyRelSwap{} under the second safety data and \eightyRelTulu{} under Tülu 3's.
No model's mean vulnerability exceeds 0.9, so the lower correlations of this subsection are not compression at the ceiling.
These rates rest on the judge, which the registration checks on the second safety data with no prediction attached.
The same two authors labeled 200 responses of the main data recipe's six models, blind to the judge, with the rubric and draw design of the 25-recipe grid's audit in Appendix~\ref{app:judge-grid}.
That audit gives the judge's vulnerable calls a precision of \eightAuditPrecision{} \eightAuditPrecisionCI{} and a recall of \eightAuditRecall{} over 25 data recipes, and a precision of \eightAuditPrecDclm{} \eightAuditPrecDclmCI{} on this data recipe alone, from \eightAuditNjvDclm{} judge-vulnerable items.
The design-weighted precision here is \eightyAuditPrecision{} \eightyAuditPrecisionCI{} on \eightyAuditNjv{} judge-vulnerable items, with a recall of \eightyAuditRecall{} \eightyAuditRecallCI{}.
The judge's precision here cannot be told apart from the audit's value on this data recipe, which rests on ten items.
The judge errs both ways, more often by over-calling, with a false-positive mass of \eightyAuditFpMass{} \eightyAuditFpMassCI{} against a false-negative mass of \eightyAuditFnMass{} \eightyAuditFnMassCI{}.
The two annotators agree at Cohen's $\kappa$ \eightyAuditKappaRefusal{}, against \eightAuditKappaRefusal{} in that audit.
The correlations stay on the judge's labels, as registered, and no mean in this subsection is corrected for the judge's precision.%

\FloatBarrier
\section{Pre-registration and provenance}
\label{app:prereg}
Each prediction was pre-registered before its run, together with its primary metric and the rule for interpreting its result.
Table~\ref{tab:prereg} gives one row to each registered prediction whose result the main text reports, and Table~\ref{tab:prereg-appendix} one row to each registered check reported only in the appendices whose result fell outside its rule, whose rule did not apply, or whose value a main-text section cites; the registered checks that passed and are reported only in an appendix are recorded there.
Predictions registered together may share a row, and the register below labels their parts.
The only registered analyses that ran without a row are those registered without a prediction.
Each row gives the prediction, the observed value with its 95\,\% interval where the result is an estimate, and a \emph{reading}: what the observed value means against the prediction.
The rows whose results the body reports are in Table~\ref{tab:prereg}, and those reported only in the appendices in Table~\ref{tab:prereg-appendix}.
In both tables, rows follow the order in which their results appear in the paper, and a final column names the section that reports each result.
Appendix~\ref{app:prereg-deviations} lists each later change to a registered rule, prediction or design that touches a reported number, and Appendix~\ref{app:prereg-posthoc} the analyses chosen after the data.
\begin{table}[h]
  \centering\footnotesize
  \caption{\label{tab:prereg}\textbf{Pre-registered predictions and checks whose results the main text reports.} Intervals in the table are 95\,\% intervals. The register below gives each row's full rule and every interval.}
  \newcolumntype{P}[1]{>{\raggedright\arraybackslash}p{\dimexpr #1\textwidth-2\tabcolsep\relax}}
\newcolumntype{Q}{P{0.17}P{0.30}P{0.28}P{0.17}P{0.08}}
  \setlength{\tabcolsep}{4pt}\begin{tabular}{Q}
    \toprule
    experiment & prediction & observed & reading & \S \\
    \midrule
    partial reruns, seed-spread check & both partial reruns within $z \leq 2$ & largest $z$ 1.60 & inside the seed spread & \S\ref{sec:setup}, App.~\ref{app:factorial} \\
model sizes & the ancestry gap shrinks with size & gap(150M) $-$ gap(300M) \ensuremath{-}0.008 \mbox{[\ensuremath{-}0.024, 0.010]}; gap(150M) $-$ gap(1B) +0.032 \mbox{[0.014, 0.050]} & positive everywhere, not ordered & \S\ref{sec:grid}, App.~\ref{app:robustness} \\
model sizes, per-size check & each size: vulnerability in $[0.1, 0.9]$, reliability $\geq 0.7$ & 36~of~36 models & every model inside both bounds & \S\ref{sec:grid}, App.~\ref{app:robustness} \\
25-recipe grid, ordering & $\rMZ > \rU > \rR$ in most data recipes & 25~of~25 data recipes & ordered in every data recipe & \S\ref{sec:grid} \\
25-recipe grid, data recipe against seed & effect of the data recipe at least the seed's & \ensuremath{-}0.141 \mbox{[\ensuremath{-}0.152, \ensuremath{-}0.128]} & reverse: seed outweighs data recipe & \S\ref{sec:grid} \\
25-recipe grid, variance shares & data recipes' variance share exceeds the seeds' & 0.51 \mbox{[0.13, 0.72]} against 0.31 & above pretraining seed; interval wide & \S\ref{sec:grid} \\
human labels & human-labeled gap positive; judge minus human interval includes zero & 0.156 \mbox{[0.036, 0.294]}; judge $-$ human 0.047 \mbox{[\ensuremath{-}0.060, 0.157]} & not a judge artifact & \S\ref{sec:grid}, App.~\ref{app:judge} \\
partial reruns, ordering & same-seed rerun above partial reruns above same-recipe pairs & 0.822 / 0.822 against 0.823 & not above pairs sharing neither & \S\ref{sec:ladder}, App.~\ref{app:factorial} \\
second same-seed rerun, same hardware and software & the two reruns within 0.02 of the first rerun's pair & +0.0110 \mbox{[\ensuremath{-}0.0009, 0.0227]} & within 0.02, covers 0 & \S\ref{sec:ladder} \\
same-seed reruns, mean of three pairs & between $\rU$ and $\rMZ$, nearer $\rU$ & 0.8441 \mbox{[0.8347, 0.8530]} & between, nearer $\rU$ & \S\ref{sec:ladder} \\
seed-stable jailbreaks & fewer stable prompts at each wider level of ancestry & shares 0.162 \mbox{[0.153, 0.171]}, 0.068 \mbox{[0.061, 0.074]}, 0.005 \mbox{[0.003, 0.007]}; template ranks $\rU$ 0.741, $\rR$ 0.713 & descending; ranks ordered, close & \S\ref{sec:witness}, App.~\ref{app:robustness} \\
seed counts, standard error & one seed of each kind: standard error above one point & 3.4 \mbox{[3.0, 3.9]} points & above one point & \S\ref{sec:witness}, \S\ref{sec:limitations}, App.~\ref{app:robustness} \\
seed counts, sign flips & single-seed comparisons of data recipes flip sign non-negligibly & 0.234 \mbox{[0.176, 0.293]} & about one in four & \S\ref{sec:witness}, \S\ref{sec:limitations}, App.~\ref{app:robustness} \\
attack-order reversals (1) & 2--5-point template pairs reverse across runs, twice MZ twins & 0.585 \mbox{[0.578, 0.593]}; MZ 0.186 & above; the binning inflates it & \S\ref{sec:witness} \\
attack-order reversals (2) & the top template changes between runs of a data recipe & 0.733 \mbox{[0.650, 0.813]} & above a third & \S\ref{sec:witness}, \S\ref{sec:limitations} \\
fingerprints (1) & a model's profile identifies its pretraining run among 75 & 1.000 \mbox{[0.952, 1.000]}; 50 prompts 0.981 & above both registered thresholds & \S\ref{sec:witness}, \S\ref{sec:limitations} \\
fingerprints (2) & data recipe from an unseen run: below 0.5 & 0.48 \mbox{[0.34, 0.63]} & below 0.5, above chance & \S\ref{sec:witness} \\
 
    \bottomrule
  \end{tabular}

\end{table}

\begin{table}[h]
  \centering\footnotesize
  \caption{\label{tab:prereg-appendix}\textbf{Pre-registered predictions and checks reported only in the appendices.} Intervals in the table are 95\,\% intervals.}
  \newcolumntype{P}[1]{>{\raggedright\arraybackslash}p{\dimexpr #1\textwidth-2\tabcolsep\relax}}
\newcolumntype{Q}{P{0.17}P{0.30}P{0.28}P{0.17}P{0.08}}
  \setlength{\tabcolsep}{4pt}\begin{tabular}{Q}
    \toprule
    experiment & prediction & observed & reading & \S \\
    \midrule
    GPU sharing & time-sliced $\leq 1.1\times$; MPS 1.5--1.9$\times$; copies inside loss check & 0.81 / 1.91 / 1.44; 4~of~6 copies outside & most copies outside; none eligible & App.~\ref{app:compute} \\
pretraining reproduction, perplexity & perplexity within 1\,\% of Ai2's at every evaluated step & largest difference 4.7\,\% & outside the registered 1\,\% & App.~\ref{app:nondet} \\
first same-seed rerun & between $\rU$ and $\rMZ$, nearer $\rU$ & 0.8395 \mbox{[0.8273, 0.8505]} & between, nearer $\rU$ & App.~\ref{app:nondet} \\
post-training reproduction, first check & our SFT and DPO match Ai2's within seed noise & 0.811 against 0.933; 7~of~7 sources & correlation below the rule & App.~\ref{app:pipeline} \\
post-training reproduction, second check & our SFT and DPO match Ai2's within seed noise & 0.949 against 0.922; 6~of~7 sources & one source outside 3 points & App.~\ref{app:pipeline} \\
twelve models, first sharding setting: one-seed rule & one post-training seed if all MZ twins within 0.02 & 16~of~18 pairs within 0.02 & two post-training seeds kept & App.~\ref{app:pipeline} \\
judge audit, 25-recipe grid & precision near the first audit's, no data recipe dependence & 0.553 / 0.484; $p$ 0.321 & WildJailbreak close, human jailbreaks lower & App.~\ref{app:judge} \\
judge audit, twelve models of the first sharding setting & refusal accuracy $\geq 0.90$, $\kappa \geq 0.8$ & 0.880 \mbox{[0.853, 0.905]}; $\kappa$ 0.517 & above the rule's 0.85 & App.~\ref{app:judge} \\
projection transplant, closure & another run's projection closes part of the gap & 0.483 \mbox{[0.451, 0.514]} & about half the gap & App.~\ref{app:mechanism} \\
projection transplant, replication & nine more pairs within 0.10~of~0.483 & 0.495 \mbox{[0.472, 0.518]} & the closure replicates & App.~\ref{app:mechanism} \\
refusal direction, projection gap & projection gap positive, interval excluding zero; below behavioral gap & 0.401 \mbox{[0.384, 0.418]} against 0.173 & positive; larger, not smaller & App.~\ref{app:mechanism} \\
refusal direction, ablation at layer \sevenSiteLayer{} & ablation at layer \sevenSiteLayer{}, position \sevenSitePos{} doubles vulnerability everywhere & 10~of~12 models & not in every model & App.~\ref{app:mechanism} \\
seed counts, seed components & pretraining-seed variance exceeds post-training-seed variance & 3.1 \mbox{[\ensuremath{-}0.6, 6.9]} pt$^2$ & point above, interval spans zero & App.~\ref{app:robustness} \\
other safety data, same pretraining run & pairs of one run across safety data exceed same-recipe runs & 0.558 \mbox{[0.537, 0.577]} against 0.811 & scoped to fixed safety data & App.~\ref{app:robustness} \\
forecast from the base model & base model readouts predict the run's residual & 0.014 \mbox{[\ensuremath{-}0.005, 0.032]}; within data recipe 0.460 & below the registered thresholds & here \\
refuse-if-any & same-recipe pair cuts reliably vulnerable share 1.5$\times$ MZ twins' & 1.28 \mbox{[1.21, 1.35]}; over-refusal +0.002 \mbox{[\ensuremath{-}0.014, 0.018]} & above 1, below 1.5$\times$ & here \\
 
    \bottomrule
  \end{tabular}

\end{table}
\FloatBarrier
\subsection{Register of pre-registered predictions}\label{app:prereg-register}
The register below repeats each row of Table~\ref{tab:prereg} in full, in the terms of this paper: the registered prediction and rule, every observed value and interval, the reading, and the section that reports the result.

{\small
\textbf{Partial reruns, seed-spread check.} Predicted: both partial reruns lie inside the spread of DataDecide's released 1B runs; rule: for each rerun and each of two metrics, final loss on DataDecide's validation sets and the OLMES-10 macro average, $z \leq 2$ against the pretraining-seed standard deviation of the released 1B runs pooled over the 25 data recipes, our reruns and the released runs evaluated by the same code. Observed: $z$ 0.06 (loss), 0.81 (OLMES-10); $z$ 0.50 (loss), 1.60 (OLMES-10). Reading: both runs inside the pretraining-seed spread. Reported in \S\ref{sec:setup}, App.~\ref{app:factorial}.\par

\textbf{Model sizes.} Predicted: the ancestry gap on adversarial harmful prompts shrinks with size, ordered 150M $>$ 300M $>$ 1B, with 0.173 at 1B; rule: the three point estimates are ordered as stated and the 95\,\% prompt-bootstrap interval of gap(150M) $-$ gap(1B) excludes 0. Observed: gap 150M 0.204 [0.189, 0.218]; 300M 0.211 [0.198, 0.224]; 1B 0.171 [0.160, 0.183]; gap(150M) $-$ gap(300M) \ensuremath{-}0.008 [\ensuremath{-}0.024, 0.010], gap(150M) $-$ gap(1B) +0.032 [0.014, 0.050], gap(300M) $-$ gap(1B) +0.040 [0.024, 0.055] (prompt bootstrap). Reading: positive at every size; 1B below both smaller sizes; 150M and 300M not ordered. Reported in \S\ref{sec:grid}, App.~\ref{app:robustness}.\par

\textbf{Model sizes, per-size check.} Predicted: mean vulnerability on adversarial harmful prompts in $[0.1, 0.9]$ and reliability $\geq 0.7$ for every one of the twelve models of each size. Observed: mean vulnerability (raw judge rates) and reliability by size: 150M 0.746--0.805, reliability 0.822--0.858; 300M 0.635--0.775, reliability 0.872--0.909; 1B 0.462--0.519, reliability 0.936--0.951. Reading: every model of every size inside both bounds; every size enters the comparison. Reported in \S\ref{sec:grid}, App.~\ref{app:robustness}.\par

\textbf{25-recipe grid, ordering.} Predicted: on adversarial harmful prompts, (1) $\rMZ > \rU > \rR$ in most data recipes; (2) the data recipe's effect, $\rU - \rR$, at least as large as the pretraining seed's, $\rMZ - \rU$; (3) the data recipes' mean vulnerabilities vary more than those of the pretraining seeds within a data recipe. Observed: ordered in 25 of 25 data recipes; primary $\rMZ - \rU$ (mean over data recipes) 0.173 [0.167, 0.179]. Reading: the ordering in every data recipe. Reported in \S\ref{sec:grid}.\par

\textbf{25-recipe grid, data recipe against seed.} Predicted: on adversarial harmful prompts, (1) $\rMZ > \rU > \rR$ in most data recipes; (2) the data recipe's effect, $\rU - \rR$, at least as large as the pretraining seed's, $\rMZ - \rU$; (3) the data recipes' mean vulnerabilities vary more than those of the pretraining seeds within a data recipe. Observed: $\rU - \rR$ 0.032 against $\rMZ - \rU$ 0.173; difference \ensuremath{-}0.141 [\ensuremath{-}0.152, \ensuremath{-}0.128]. Reading: the reverse: the seed outweighs the data recipe. Reported in \S\ref{sec:grid}.\par

\textbf{25-recipe grid, variance shares.} Predicted: on adversarial harmful prompts, (1) $\rMZ > \rU > \rR$ in most data recipes; (2) the data recipe's effect, $\rU - \rR$, at least as large as the pretraining seed's, $\rMZ - \rU$; (3) the data recipes' mean vulnerabilities vary more than those of the pretraining seeds within a data recipe. Observed: share of the data recipe 0.51, pretraining seed 0.31, post-training seed 0.18; share of the data recipe [0.13, 0.72] (bootstrap over data recipes); on the mean vulnerabilities corrected for judge precision (App.~\ref{app:judge}) 0.49 [0.12, 0.70], 0.32, 0.19. Reading: point shares in the predicted order; the data recipe's share interval [0.13, 0.72] contains the base model's 0.31; the two seed factors together 0.49 (0.51 corrected). Reported in \S\ref{sec:grid}.\par

\textbf{Human labels.} Predicted: on human labels, the ancestry gap is positive with an interval excluding 0, and on the same items the gap under the judge's labels exceeds it by no more than the interval allows: the interval of the paired difference, judge $-$ human, includes 0. Observed: human gap 0.156 [0.036, 0.294], judge gap 0.203 [0.101, 0.341] on the same items, judge $-$ human 0.047 [\ensuremath{-}0.060, 0.157]; each annotator alone 0.144 and 0.124 (raw $r$, $K = 3$; prompt bootstrap). Reading: the gap survives human labels; the judge does not inflate it beyond the interval. Reported in \S\ref{sec:grid}, App.~\ref{app:judge}.\par

\textbf{Partial reruns, ordering.} Predicted: MZ twins ($\rMZ$) above the first same-seed rerun's pairs, above shared-initialization and shared-order pairs, above same-recipe pairs ($\rU$), above cross-recipe pairs ($\rR$). Observed: shared-initialization pairs 0.822 [0.811, 0.833], shared-order pairs 0.822 [0.810, 0.833], pairs sharing neither seed 0.823 [0.811, 0.834] (prompt bootstrap; first sharding setting). Reading: the partial reruns' means are not above the mean of pairs sharing neither seed; the step was set aside after the first same-seed rerun's data and before the partial reruns'. Reported in \S\ref{sec:ladder}, App.~\ref{app:factorial}.\par

\textbf{Second same-seed rerun, same hardware and software.} Predicted: the corrected correlation between the models post-trained from the two same-seed reruns (16 model pairs) lies within $\pm 0.02$ of 0.8395, the value between the first same-seed rerun's models and those of base model 1, the run it copies; and the 95\,\% prompt-bootstrap interval of the difference between the two covers 0. Observed: the two reruns 0.8505 [0.8389, 0.8616] against the first rerun with base model 1 0.8395; difference +0.0110 [\ensuremath{-}0.0009, 0.0227] (prompt bootstrap), jackknife standard error 0.0047. Reading: within 0.02 of the first rerun's pair with base model 1 and the interval covers 0, so moving to our hardware and software costs no correlation we can detect on the primary outcome; so, as registered, the three pairs are reported as one mean, with each pair's value shown. Reported in \S\ref{sec:ladder}.\par

\textbf{Same-seed reruns, mean of three pairs.} Predicted: the mean corrected correlation over the three same-seed pairs of base models (the first rerun with base model 1, the two reruns, the second rerun with base model 1; 48 model pairs), read by the first same-seed rerun's rule: strictly between $\rU$ and $\rMZ$, nearer $\rU$, with an interval that covers neither. Observed: 0.8441 [0.8347, 0.8530] (prompt bootstrap, 48 model pairs) against $\rU$ 0.8136 and $\rMZ$ 0.9848. Reading: between and nearer $\rU$, the interval clear of both; the same-seed mean that \S\ref{sec:ladder} reports. Reported in \S\ref{sec:ladder}.\par

\textbf{Seed-stable jailbreaks.} Predicted: prompts stable across the post-training seeds of one base model outnumber prompts stable across the base models of a data recipe, which outnumber prompts stable across all 25 data recipes; template success ranks agree more between the base models of one data recipe than between data recipes. Observed: shares of prompts stable within a base model, within a data recipe and across data recipes (every model of the group has vulnerability at least 0.9) 0.162 [0.153, 0.171], 0.068 [0.061, 0.074], 0.005 [0.003, 0.007]; corrected Spearman correlation of template success ranks, MZ twins 0.980 [0.973, 0.983]; same-recipe pairs 0.741 [0.680, 0.788]; cross-recipe pairs 0.713 [0.641, 0.768] (template bootstrap). Reading: the shares descend as predicted, partly because more models must agree at each wider level; template ranks agree more within a data recipe than between data recipes on the point estimates, with overlapping intervals. Reported in \S\ref{sec:witness}, App.~\ref{app:robustness}.\par

\textbf{Seed counts, standard error.} Predicted: (1) the pretraining seed's variance component of a data recipe's mean vulnerability on adversarial harmful prompts exceeds the post-training seed's; with one seed of each kind, (2) the standard error of that mean is above 1 point and (3) the share of comparisons between two data recipes whose sign reverses is not negligible, a share stated without a threshold. Observed: 3.4 [3.0, 3.9] points with one pretraining and one post-training seed (bootstrap over data recipes); the smallest measured design at or below 2 points on the point estimate, 3 pretraining seeds with 1 post-training seed each, gives 1.99 [1.71, 2.24] points. Reading: above 1 point; no measured design reaches 1 point, which would take 12 pretraining seeds (extrapolated). Reported in \S\ref{sec:witness}, \S\ref{sec:limitations}, App.~\ref{app:robustness}.\par

\textbf{Seed counts, sign flips.} Predicted: (1) the pretraining seed's variance component of a data recipe's mean vulnerability on adversarial harmful prompts exceeds the post-training seed's; with one seed of each kind, (2) the standard error of that mean is above 1 point and (3) the share of comparisons between two data recipes whose sign reverses is not negligible, a share stated without a threshold. Observed: 0.234 [0.176, 0.293]; 0.148 [0.099, 0.197] for data recipes at least 2 points apart. Reading: about one single-seed comparison in four points the wrong way. Reported in \S\ref{sec:witness}, \S\ref{sec:limitations}, App.~\ref{app:robustness}.\par

\textbf{Attack-order reversals (1).} Predicted: among pairs of templates whose success rates, averaged over six models, differ by 2--5 points, the share whose order reverses between the two models of a same-recipe pair is at least 0.25 and at least twice the share for MZ twins. Observed: same-recipe pairs 0.585 [0.578, 0.593], MZ twins 0.186 [0.177, 0.196], ratio 3.14 [2.96, 3.35] (bootstrap over data recipes); binned instead by the gap seen on one model (a secondary set after the data), 0.434 [0.430, 0.438] against 0.293 [0.286, 0.300], ratio 1.48 [1.44, 1.52]. Reading: above both registered values; the registered bins condition on a mean that contains both compared models, which raises the same-recipe pairs' rate at small gaps by construction; by the gap seen on one model the rate stays above 0.25 and the ratio falls below two. Reported in \S\ref{sec:witness}.\par

\textbf{Attack-order reversals (2).} Predicted: the most successful template differs between two base models of one data recipe in at least a third of such pairs. Observed: the most successful template differs between two base models of one data recipe in 0.733 [0.650, 0.813] of their 300 pairs, and between MZ twins in 0.333 [0.213, 0.453] (bootstrap over data recipes). Reading: above a third of pairs of base models; MZ twins at a third. Reported in \S\ref{sec:witness}, \S\ref{sec:limitations}.\par

\textbf{Fingerprints (1).} Predicted: a model's vulnerability profile identifies its pretraining run among the 75 with accuracy $\geq 0.95$ on all prompts and $\geq 0.80$ on 50 random prompts. Observed: 75 of 75 queries (1.000, exact interval [0.952, 1.000]; the registered bootstrap over queries is degenerate at [1.000, 1.000]), the same in the reverse direction; with 50 random prompts 0.981 (2.5--97.5\,\% range over 200 subsets [0.933, 1.000]). Reading: every query is assigned to its own run with all prompts, and 98 in 100 with 50 random prompts. Reported in \S\ref{sec:witness}, \S\ref{sec:limitations}.\par

\textbf{Fingerprints (2).} Predicted: the data recipe of a pretraining run left out of the reference is identified with accuracy below 0.5: above chance, and far below the identification of the run. Observed: 24 of 50 queries (0.48, bootstrap over queries [0.34, 0.60], exact [0.34, 0.63]) against chance 0.04. Reading: the point estimate is below 0.5 and the interval spans it; above chance throughout. Reported in \S\ref{sec:witness}.\par
 
}
\subsection{Changes after registration}
\label{app:prereg-deviations}
Each item below records one change to a registered rule, prediction or design, made after its registration.
Every change listed touches a reported number.
Its item gives the change's effect on that number and ends on the section that reports it.

\begin{enumerate}
\item \label{prereg-change:perplexity}Pretraining reproduction: perplexity was registered to stay within 1\,\% of Ai2's in-loop values at every evaluated step; the largest difference was 4.7\,\%, and the reproduction was accepted on three grounds: its step-0 weights are bit-exact, its data order is exact, and a second run of ours with the same seeds deviates from it by an amount of the same order. Effect: the reproduction's perplexity row above (App.~\ref{app:nondet}).
\item \label{prereg-change:dpo-rule}Our DPO against Ai2's: the rule was registered on harmful prompts' refusal rates and was run on all prompts' rates. Effect: within on 7 of 7 sources on all prompts and on 6 of 6 on harmful prompts, as registered (App.~\ref{app:pipeline}).
\item \label{prereg-change:stop}Post-training reproduction, second check: a miss was registered to stop the work for a diagnosis that included a step-by-step comparison of Ai2's intermediate SFT checkpoints; after the miss, post-training of the MZ twins started before the diagnosis had located the gap, and the step-by-step comparison was not run. Effect: every post-trained model comes from the pipeline that check examined; the check's row reports the miss (App.~\ref{app:pipeline}).
\item \label{prereg-change:same-seed-level}Same-seed mean: redefined as the mean over three same-seed pairs after the first pair was measured. Effect: the same-seed mean of \S\ref{sec:ladder} is the three pairs' mean, 0.8441 [0.8347, 0.8530]; the first pair's value, 0.8395, is reported in App.~\ref{app:nondet}.
\item \label{prereg-change:ladder-step}Partial reruns, ordering: the step placing the partial reruns between the same-seed rerun and same-recipe pairs was set aside after the first same-seed rerun's data and before the partial reruns'. Effect: no ordering is claimed; its row prints 0.822 / 0.822 against 0.823 (\S\ref{sec:ladder}, App.~\ref{app:factorial}).
\item \label{prereg-change:carryover}Seed-stable jailbreaks: the registered shares by ancestry became descriptive, and the carry-over rate (count-matched), added at analysis, became the stability statistic. Effect: the stability numbers of \S\ref{sec:witness} and App.~\ref{app:robustness}.
\item \label{prereg-change:stable-group}Seed-stable jailbreaks: the registered group of every data recipe's models with post-training seed 1 was read as the 75 such models, and the interval's resampling units were set at analysis. Effect: the share of prompts stable across data recipes (App.~\ref{app:robustness}).
\item \label{prereg-change:shares-corrected}25-recipe grid, variance shares: shares on the mean vulnerabilities corrected for judge precision are printed beside the registered raw ones. Effect: the observed column of the variance-share row (\S\ref{sec:grid}, this appendix).
\item \label{prereg-change:fingerprints}Fingerprints: the registered 3{,}212 prompts were read as all prompts, 3{,}211 after one prompt undefined for a model of the 25-recipe grid was dropped; and the registered bootstrap over queries, degenerate at [1.000, 1.000] when every query is identified, is printed beside an exact interval, which the text quotes. Effect: the fingerprint numbers and the interval of run identification (\S\ref{sec:witness}, \S\ref{sec:limitations}).
\item \label{prereg-change:forecast}Forecast from the base model: the registered target averaged the main data recipe's four post-training seeds; the 25-recipe grid has two for every run, so two were used throughout and the four-seed value is printed beside it (0.014 against 0.014). Effect: none at three decimals (this appendix only).
\item \label{prereg-change:refuse-ratio}Refuse-if-any: the pooled ratio was set at analysis as the ratio of the mean reductions over data recipes, not a mean of each data recipe's ratio, its interval resampling data recipes and recomputing the ratio in each draw. Effect: the ratio of the refuse-if-any row (this appendix only).
 
\end{enumerate}
\subsection{Analyses chosen after the data}\label{app:prereg-posthoc}

We chose the analyses below after seeing the data, so their results are exploratory.
Each item names an analysis, the section that reports it and why it counts as chosen after the data.

\begin{enumerate}
\item\label{prereg-posthoc:run} The interpretation that shared vulnerability is carried by the pretraining run, not by the pretraining seed, reported in Section~\ref{sec:ladder}.
We chose this interpretation after the first same-seed rerun's data.

\item\label{prereg-posthoc:main-recipe} The transfer ratio, the template residual, the ancestry gap per prompt source and each model's correlation with Ai2's released OLMo~2 1B DPO model, reported for the main data recipe in Section~\ref{sec:grid} and Appendix~\ref{app:robustness}.
The correlation with OLMo~2 is a descriptive reference.
Their pre-registration covered only the 24 other data recipes.

\item\label{prereg-posthoc:site-map} A map over layers and token positions, reported in Appendix~\ref{app:mechanism}.
The within-pair link on new same-recipe pairs then pre-registered its layer and position from the map, so that later result is confirmatory.

\item\label{prereg-posthoc:harm-split} The harm-direction control split by pretraining run, reported in Appendix~\ref{app:mechanism}.
We wrote the rule that sets the runs apart after seeing the data.

\item\label{prereg-posthoc:template-pair} The paired template comparison of same-recipe pairs minus cross-recipe pairs, reported in Appendix~\ref{app:robustness}.
We added this comparison beyond the registered list of analyses.

\item\label{prereg-posthoc:residue} The \emph{same-seed residue} and the mean corrected correlation for each of the six pairs of base models, reported in Appendix~\ref{app:nondet}.
The same-seed residue is a descriptive quantity: the same-seed rerun pair's corrected correlation minus the mean corrected correlation of the rerun's pairs with the other released runs.
We added the same-seed residue, and extended the means to all six pairs, after the first same-seed rerun's data.

\item\label{prereg-posthoc:data-order} The check that the pretraining reproduction's data order is base model 1's, reported in Appendix~\ref{app:factorial}.
We ran this check while diagnosing the pretraining reproduction's difference in perplexity.
The registration of the pretraining reproduction named no check of data order.

\item\label{prereg-posthoc:binning} The reversal rate of two templates binned by the gap seen on one model, with its contrasts between kinds of pair, reported in Section~\ref{sec:witness} and beside the registered bins in the register above.
We set this binning after the registered numbers showed same-recipe pairs reversing more than half of small leads.
The registered bins condition on a mean containing both compared models, which raises that rate by construction.

\item\label{prereg-posthoc:companion} The companion statistic of the human-label check, one minus the ratio of the same-recipe pairs' mean correlation to that of MZ twins, reported in Appendix~\ref{app:judge}.
We fixed it after the judge-only dry run and before any human label; it was not registered.

\item\label{prereg-posthoc:rotation} The fingerprints' identification of the data recipe with each base model held out in turn, reported only here: holding out base models 1, 2 and 3 gives \seventysixRecipeHeldOne{}, \seventysixRecipeHeldTwo{} and \seventysixRecipeHeldThree{}.
The registered test holds out base model 3; we added the rotation after the data.

\item\label{prereg-posthoc:recipe-count} The count of data recipes in which the refuse-if-any ratio is above 1, reported in Table~\ref{tab:prereg-appendix}.
We added this count beyond the registered statistics.
\end{enumerate}
\subsection{Provenance of reported numbers}\label{app:prereg-provenance}
One script writes each measured value as a macro, read from an artifact, and records the value's provenance: the artifact, the entry, the commit and the producing script.
The same artifacts stop the build when one is missing or contradicts a statement the prose makes in words.
Tests check the macros, recomputing the mean corrected correlations and the ancestry gap from the artifacts' per-pair entries and comparing each with its macro.
These tests and the macro-writing script are released with our code.

\FloatBarrier

\end{document}